%% file: main_new.tex
\documentclass[14pt]{article}
\usepackage{xcolor}
\usepackage{amsmath}
\usepackage{PRIMEarxiv}
\usepackage[utf8]{inputenc} 
\usepackage[T1]{fontenc}    
\usepackage{hyperref}       
\usepackage{url}            
\usepackage{booktabs}       
\usepackage{amsfonts}       
\usepackage{nicefrac}       
\usepackage{microtype}      
\usepackage{lipsum}
\usepackage{graphicx}
\graphicspath{{media/}}     
\newtheorem{proposition}{Proposition}
\newtheorem{lemma}{Lemma}

\newtheorem{definition}{Definition}
\newtheorem{remark}{Remark}
\newtheorem{theorem}{Theorem}

\usepackage{cancel}

\usepackage{color}
\usepackage{soul,xargs}
\newenvironment{proof}{\paragraph{\textit{Proof:}}}{}
\usepackage{amsfonts} 
\input{Macros.tex}

\input{krish_macros}

\usepackage{tikz}
\usepackage{comment}
\usepackage[colorinlistoftodos,prependcaption,textsize=tiny]{todonotes}
\usepackage{xfrac}  
\usepackage{lipsum}
\usepackage{amsfonts,dsfont}
\usepackage{graphicx}
\usepackage{epstopdf}
\usepackage{algorithmic}
\usepackage{tikz}
\usepackage{graphicx}
\usepackage[normalem]{ulem}
\usepackage{cleveref}

\crefname{figure}{figure}{figures}
\Crefname{figure}{Figure}{Figures}

\usepackage{enumitem,comment}
\usepackage{subcaption}
\usepackage{graphicx}
\usepackage{hyperref}
\usepackage{multirow}
\usepackage{booktabs}
\usepackage{tabularray}
\usepackage{amssymb}

\usepackage{setspace}
\def\sX{{\mathcal X}}

\def\sG{{\mathcal G}}
\def\sW{{\mathcal W}}
\def\sP{{\mathcal P}}
\def\sN{{\mathcal N}}
\usepackage[utf8]{inputenc}
\usepackage{pgfplots}
\DeclareUnicodeCharacter{2212}{−}
\usepgfplotslibrary{groupplots,dateplot}
\usetikzlibrary{patterns,shapes.arrows}
\pgfplotsset{compat=newest}

\usetikzlibrary{arrows.meta, positioning, chains, ext.paths.ortho, quotes, shapes.geometric, shadows}
\ExplSyntaxOn\makeatletter\tl_replace_once:Nnn\tikz@do@matrix@cont{\tikz@node@finish}{\tikz@alias\tikz@node@finish}\makeatother\ExplSyntaxOff
\tikzset{tight matrix/.style={every outer matrix/.append style={inner sep=+0pt, outer sep=+0pt}}}

\definecolor{cinereous}{rgb}{0.6, 0.51, 0.48}

\definecolor{aliceblue}{rgb}{0.94, 0.97, 1.0}

\definecolor{almond}{rgb}{0.94, 0.87, 0.8}

\definecolor{antiquebrass}{rgb}{0.8, 0.58, 0.46}

\definecolor{apricot}{rgb}{0.98, 0.81, 0.69}

\definecolor{babyblueeyes}{rgb}{0.63, 0.79, 0.95}

\definecolor{celadon}{rgb}{0.67, 0.88, 0.69}

\definecolor{corn}{rgb}{0.98, 0.93, 0.36}

\definecolor{applegreen}{rgb}{0.55, 0.71, 0.0}

\definecolor{brilliantlavender}{rgb}{0.96, 0.73, 1.0}

\definecolor{lavenderblue}{rgb}{0.8, 0.8, 1.0}

\definecolor{candypink}{rgb}{0.89, 0.44, 0.48}

\tikzset{
    bigbox/.style={draw, rounded corners, minimum width=2.5cm, minimum height=1.5cm, fill=celadon, drop shadow},
    bigboxstacked/.style={draw, rounded corners, minimum width=2.5cm, minimum height=1.5cm, fill=celadon, drop shadow, double copy shadow},
    verybigbox/.style={draw, rounded corners, minimum width=13.5cm, minimum height=4.5cm, fill=almond, drop shadow},
    verybigbox2/.style={draw, rounded corners, minimum width=14.5cm, minimum height=4.5cm, fill=almond, drop shadow},
    verybigsmallbox/.style={draw, rounded corners, minimum width=10.3cm, minimum height=4.0cm, fill=almond, drop shadow},
    verybigboxstacked/.style={draw, rounded corners, minimum width=13.3cm, minimum height=4.0cm, fill=almond, drop shadow, double copy shadow},
    verybigsmallboxstacked/.style={draw, rounded corners, minimum width=10.3cm, minimum height=4.0cm, fill=almond, drop shadow, double copy shadow},
    smallbox/.style={draw, rounded corners, minimum width=2cm, minimum height=1cm, fill=candypink, drop shadow},
    smallboxstacked/.style={draw, rounded corners, minimum width=2cm, minimum height=1cm, fill=candypink, drop shadow, double copy shadow},
    smallboxdyn/.style={draw, rounded corners, minimum width=2cm, minimum height=1cm, fill=corn, drop shadow},
    smallboxdynstacked/.style={draw, rounded corners, minimum width=2cm, minimum height=1cm, fill=corn, drop shadow, double copy shadow},
    smallboxinit/.style={draw, rounded corners, minimum width=2cm, minimum height=1cm, fill=babyblueeyes, drop shadow},
    smallboxinitstacked/.style={draw, rounded corners, minimum width=2cm, minimum height=1cm, fill=babyblueeyes, drop shadow, double copy shadow},
    smallboxtime/.style={draw, rounded corners, minimum width=2cm, minimum height=1cm, fill=cinereous, drop shadow},
    smallboxtimestacked/.style={draw, rounded corners, minimum width=2cm, minimum height=1cm, fill=cinereous, drop shadow, double copy shadow},
    smallboxdec/.style={draw, rounded corners, minimum width=2cm, minimum height=1cm, fill=brilliantlavender, drop shadow},
    smallboxdecstacked/.style={draw, rounded corners, minimum width=2cm, minimum height=1cm, fill=brilliantlavender, drop shadow, double copy shadow},
    bigcircle/.style={draw, circle, minimum size=2cm, fill=white, drop shadow},
    bigellipse/.style={draw, ellipse, minimum width=3cm, minimum height=2.5cm, fill=white, drop shadow},
    place/.style={inner sep=0pt, outer sep=0pt},
    fork/.style={decorate, decoration={show path construction, lineto code={
        \draw[rounded corners, ->](\tikzinputsegmentfirst)-|($(\tikzinputsegmentfirst)!.5!(\tikzinputsegmentlast)$)|-(\tikzinputsegmentlast);}
    }}
}

\title{\lilan: A linear latent Network approach for real-time solutions to stiff nonlinear ordinary differential equations

}

\author{
  William Cole Nockolds \\ 
  Oden Institute for Computational
  Engineering and Sciences, \\
  University of Texas at Austin \\
  Austin, TX 78712\\
  \texttt{cole.nockolds@utexas.edu} \\
   \And
  C G Krishnanunni \\
  Dept of Aerospace Engineering $\&$ Engineering Mechanics, \\
   University of Texas at Austin \\
  Austin, TX 78712\\
  \texttt{krishnanunni@utexas.edu} \\
   \And
  Tan Bui-Thanh \\ 
  Oden Institute for Computational
  Engineering and Sciences, \\
  Dept of Aerospace Engineering $\&$ Engineering Mechanics, \\
  University of Texas at Austin \\
  Austin, TX 78712\\
  \texttt{tanbui@oden.utexas.edu} \\
   \And
  Xianzhu Tang \\ 
  Theoretical Division, Los Alamos National Laboratory\\
  Los Alamos, NM 87545, United States of America \\
  \texttt{xtang@lanl.gov} 
}

\begin{document}
\maketitle

\begin{abstract}
Solving stiff ordinary differential equations (StODEs) requires sophisticated numerical solvers, which are often computationally expensive. In general, traditional explicit time integration
schemes with restricted time step sizes are not suitable for StODEs, and one must resort to costly implicit methods  to compute solutions. On the other hand, state-of-the-art machine learning (ML) based methods such as Neural ODE (NODE) poorly handle the timescale separation of various elements of the solutions to StODEs, while still requiring expensive implicit/explicit  integration at inference time. In this work, we propose a linear latent network (\lilan) approach in which the dynamics in the latent space can be integrated analytically, and thus numerical integration is completely avoided. At the heart of \lilans are the following key ideas:
i) two encoder networks to encode initial condition together with parameters of the ODE to the slope and the initial condition for the latent dynamics, respectively. Since the latent dynamics, by design, are linear, the solution can be evaluated analytically; ii) a neural network to map the initial condition, parameters, and the physical time to latent times, one for each latent variable. Intuitively, this allows for the "stretching/squeezing" of time in the latent space, thereby allowing for varying levels of attention to different temporal scales in the solution. Finally, iii) a decoder network to decode the latent solutions into the physical solution at the corresponding physical time. {\em \lilans is thus a solution operator approach that instantly provides an approximate flow map solution of a system of nonlinear ODEs at any time.}
We provide a universal approximation theorem for the proposed \lilans approach, showing that it can approximate the solution of any stiff nonlinear system on a compact set to any degree of accuracy $\epsilon$.  We also show an interesting fact that the dimension of the latent dynamical system in \lilans is independent of $\epsilon$. Numerical results on  "Robertson Stiff Chemical Kinetics Model" \cite{anantharaman2021acceleratingsimulationstiffnonlinear} and "Plasma Collisional-Radiative Model" \cite{flychk} suggest that \lilans outperformed state-of-the-art  machine learning approaches for handling StODEs. Numerically, we also show that \lilans outperformed other machine learning methods on multiple partial differential equations (PDEs) with known stiff behaviors, such as the "Allen-Cahn" and "Cahn-Hilliard" PDEs \cite{montanelli2020solvingperiodicsemilinearstiff}.

\end{abstract}


\section{Introduction}
\label{sect:intro}


\input{intro}

\section{Mathematical framework}

In this work, all the vectors are represented in boldface small, and matrices are represented in boldface capital.
We consider  the following autonomous parametrized StODE:
\begin{equation}
\dd{\xb}{t}=\bff\LRp{\xb,\ \mathbf{p}},\quad \xb(0)=\xb_0,
    \label{original_dynamics}
\end{equation}
where $\bff\LRp{\xb,\ \mathbf{p}}:  \sG \times \sX_p \rightarrow \real^{n_x}$, with $ \sG\subseteq \real^{n_x}$ 
denoting a non-empty open set, and $\sX_p\subset \real^{n_p}$ denoting a non-empty, compact set, and  $\xb_0\in  \sG$. Here, $\mathbf{p}$ is the $n_p$ dimensional parameter vector which does not vary with time and $\xb(t)$ is the $n_x$ dimensional state vector.    

\begin{definition}[Flow map]
\label{flow}
By $\sW$ we denote the following
\[\sW=\bigcup_{\xb_0\in \sG,\ \mathbf{p} \in \sX_p} I_{\xb_0; \mathbf{p}} \times \{\xb_0 \} \times \{\mathbf{p} \}\subseteq \real \times \sG \times \sX_p, \]
where  $I_{\xb_0; \mathbf{p}}$ is the maximal interval of existence of the solution\footnote{Refer to \cite{logemann2014ordinary} for the definition of maximal interval of existence of the solution.}  of \eqref{original_dynamics} for a given $\xb_0,\ \mathbf{p}$. We define the flow map (if it exists) to be the function  $\bpsi: \sW \rightarrow \sG$  such that $\bpsi\LRp{t,\xb_0,\mathbf{p}}=\xb(t)$, where $\xb(t)$ is a solution of \eqref{original_dynamics} at time $t$ for a given $\{\xb_0,\ \mathbf{p}\}$. 
\end{definition}

The objective in this work is to develop a surrogate model for approximating the solution $\xb(t)$, and thus the flow map, on some closed and bounded interval $I$, for any given $\mathbf{p}\in \sX_p,\ \xb_0\in \sG$ (assuming that $I\subseteq I_{\xb_0; \mathbf{p}}$ for all $\mathbf{p},\ \xb_0$). 



\subsection{Description of the proposed \lilans approach}
\label{method}
In this section, we present the key components of \lilan. \lilans deploys four neural networks:


\input{methodology_new}

\section{Numerical Experiments}
\label{num_re}
In this section, we numerically demonstrate the effectiveness of \lilans on a variety of prototype problems such as:
\begin{enumerate}
    \item Robertson Stiff Chemical Kinetics Model \cite{anantharaman2021acceleratingsimulationstiffnonlinear}
    \item Plasma Collisional-Radiative Model \cite{flychk}
    \item Allen-Cahn Phase Separation PDE \cite{montanelli2020solvingperiodicsemilinearstiff}
    \item Cahn-Hilliard Phase Separation PDE \cite{montanelli2020solvingperiodicsemilinearstiff}
\end{enumerate}
General experimental settings for all the problems and
descriptions of methods adopted for comparison are detailed in \Cref{hyper_parameter}.

\input{ROBER}
\input{CR}

\input{PDEs}

\input{DataReduction}

\section{Summary of results}
\Cref{fig:err_table} shows the prediction error 
computed using \eqref{eq:R2} for each machine learning method, on each problem. We see that \lilans produced the lowest relative error for all the problems except the CR charge state problem, where the Sulzer and Buck II approach, a modified form of the method in \cite{sulzer2023speedingastrochemicalreactionnetworks}, slightly outperformed \lilan, though the errors were on the same order of magnitude. \Cref{fig:err_table} also shows that for the Robertson ODE Model \cref{sec:Robertson}, \lilans outperformed  NODE  by approximately two order of magnitude and outperformed DeepONet and Sulzer and Buck I by one order of magnitude in terms of the relative error achieved. For the CR charge state model, \lilans and Sulzer and Buck I and II achieved errors on the same order of magnitude which were three and one orders of magnitude less than the errors of NODE and DeepONet respectively. For the full CR model, DeepONet and Sulzer and Buck I exceeded the error of \lilans by an order of magnitude. On the same problem, Sulzer and Buck II struggled significantly compared to \lilans, having two orders of magnitude greater relative error. On the two PDE problems, all methods had the same order of magnitude error, on the test dataset; however, \lilans's was quantitatively the lowest. The one notable exception was DeepONet, which had one order of magnitude greater error on the Cahn-Hilliard problem compared to the other methods.

\Cref{fig:speed_table} shows the speedup of \lilans in comparison to the traditional stiff numerical solver used to generate the data in each problem. To measure the speedup achieved,  $1000$ initial conditions were chosen to run the simulations for each problem. \Cref{fig:speed_table} shows the total time taken by both the numerical solver and \lilans to compute the solutions for all the $1000$ initial conditions.  We see that \lilans achieved  $\mathcal{O}(10^1)$-$\mathcal{O}(10^3)$ speedup on all problems, in terms of wall clock computation time, with especially remarkable speedup displayed on the three ODE problems (\cref{sec:Robertson,,sec:CR-model}).

\begin{table}[hbt!]
\caption{Relative error on test data for the three machine learning approaches}
\begin{tabular}{ |p{3cm}||p{2cm}|p{2.5cm}|p{2.5cm}|p{2cm}|p{2cm}|  }
 \hline
 \multicolumn{6}{|c|}{Relative Error} \\
 \hline
 Problem & Robertson & CR charge state  & Full CR  model & Allen-Cahn & Cahn-Hilliard\\
 \hline
 NODE   & \tiny $1.112\cdot10^{-2}$ & \tiny $2.209\cdot10^{0}$ & \tiny $\mathbf{OOM}$ & \tiny $6.070\cdot10^{-2}$ & \tiny $8.810\cdot10^{-2}$ \\
 \hline
 DeepONet & \tiny $1.697\cdot10^{-3}$ & \tiny $2.003\cdot10^{-2}$ & \tiny $3.968\cdot10^{-2}$ & \tiny $8.589\cdot10^{-2}$ & \tiny $2.492\cdot10^{-1}$ \\
 \hline
 Sulzer and Buck I & \tiny $1.475\cdot10^{-3}$ & \tiny $1.970\cdot10^{-3}$ & \tiny $2.132\cdot10^{-2}$ & \tiny $3.633\cdot10^{-2}$ & \tiny $5.627\cdot10^{-2}$ \\
 \hline
 Sulzer and Buck II & \tiny $7.645\cdot10^{-4}$ & \tiny $\mathbf{1.598\cdot10^{-3}}$ & \tiny $1.797\cdot10^{-1}$ & \tiny $3.391\cdot10^{-2}$ & \tiny $5.517\cdot10^{-2}$ \\
 \hline
 \lilans & \tiny $\mathbf{2.80\cdot10^{-4}}$ & \tiny $1.981\cdot10^{-3}$ & \tiny $\mathbf{7.486\cdot10^{-3}}$ & \tiny $\mathbf{1.072\cdot10^{-2}}$ & \tiny $\mathbf{4.741\cdot10^{-2}}$ \\
 \hline
\end{tabular}
\label{fig:err_table}

\end{table}

\begin{table}[hbt!]

\centering
\caption{Wall clock time comparison of \lilans with  traditional numerical solver}
\begin{tabular}{ |p{3cm}||p{2cm}|p{2.5cm}|p{2.5cm}|p{2cm}|p{2cm}|  }
 \hline
 \multicolumn{6}{|c|}{Speed tests in wall clock time} \\
 \hline
 Problem & Robertson & CR charge state  & Full CR model & Allen-Cahn & Cahn-Hilliard\\
 \hline
 Numerical Solver   & \tiny $6232.8$s & \tiny $11034.1$s & \tiny $10955.5$s & \tiny $53.25$s & \tiny $212.1$s \\
\lilans & \tiny $7.648$s & \tiny $7.366$s & \tiny $13.604$s & \tiny $3.729$s & \tiny $15.485$s \\
 \hline
 Speedup & \tiny $\mathbf{815.0x}$ & \tiny $\mathbf{1498.0x}$ & \tiny $\mathbf{805.3x}$ & \tiny $\mathbf{14.3x}$ & \tiny $\mathbf{13.7x}$ \\
 \hline
\end{tabular}
\label{fig:speed_table}

\end{table}

\section{Conclusion}
\label{conc}
In this paper, we have presented a novel, machine learning approach, called \lilan, for fast and accurate simulation of stiff, nonlinear, ordinary and partial differential equations. \lilans is inspired by kernel methods in which the original problem under consideration is featurized/lifted to (much) higher dimensional feature spaces where the problem is now easier. In particular, \lilan, via nonlinear neural networks, lifts the original stiff problems into higher dimensions where the problem is no longer stiff.  Unlike kernel methods which rely on particular feature maps to organize the latent dynamics so that hopefully it is easier to simulate, \lilans forces the latent dynamics to be trivially linear, and then learns the nonlinear feature maps using neural networks\textemdash the encoders.
With linear latent dynamics that admits analytical solutions, no numerical integration, but instead evaluations of encoders/decoders, are needed during the inference stage, and this is key to \lilan's real-time efficiency. \lilans is thus a solution operator approach that learns the flow maps of nonlinear systems of ODEs.
Indeed, theoretical results show that \lilans can approximate an arbitrary continuous flow map to any desired accuracy and that the latent dimension is independent of the accuracy. Various numerical results with varying dimensionality,  complexity, and stiffness have been presented to demonstrate the speed and accuracy of \lilans against contemporary machine learning and numerical methods. Numerical results also show that the temporal nonlinear mapping into the latent space  plays a vital role in achieving better generalization performance especially when using coarse time grids for training the networks.
In particular, \lilans architectures show better generalization capabilities compared to a direct learning method for approximating the flow map of a system of ODEs. 
It should be pointed out that, in this paper, though the motivation and numerical examples for \lilans are stiff ODEs, it is equally applicable to non-stiff ODEs.
Part of our future work will focus on additional theoretical analysis (non-asymptotic analysis) and numerical results  to address the question of why \lilans outperformed the direct learning approach for learning flow maps.
For many problems, data generation may be costly, and part of our future work is to develop algorithms for adaptive data sampling (active learning) to reduce the amount of data required to train each model, while also answering the question of finding the optimal neural network architecture (number of layers, number of neurons in each layer) for prediction such as in \cite{krishnanunni2025topological}.

\section{Acknowledgments}

This research is partially funded by the National Science Foundation award NSF-OAC-2212442 and by the Department
of Energy awards DE-SC0024724 and DE-SC0024633. The authors would like to thank Wesley Lao and Hai Nguyen for productive conversations. The authors also acknowledge the Texas Advanced Computing Center (TACC)
at The University of Texas at Austin for providing HPC
resources that have been vital to the research results reported within this paper. URL:
\url{http://www.tacc.utexas.edu}

\bibliographystyle{unsrt}  
\bibliography{references} 

\appendix

\begin{section}{General setting for numerical experiments}{}
   \label{hyper_parameter}

\input{Appendix}
\end{section}

\end{document}

%% file: Macros.tex
\newcommand{\bs}[1]{\boldsymbol{#1}}

\newcommand{\bx}{\boldsymbol{x}}
\newcommand{\bW}{{\boldsymbol{W}}}
\newcommand{\bpsi}{\boldsymbol{\psi}}

\newcommand{\by}{\boldsymbol{y}}

\newcommand{\bb}{\boldsymbol{b}}
\newcommand{\bcc}{\boldsymbol{c}}
\newcommand{\bphi}{\boldsymbol{\phi}}

\newcommand{\bz}{\boldsymbol{z}}

\newcommand{\nor}[1]{\left\| #1 \right\|}

\newcommand{\norm}[1]{\left \Vert #1 \right \Vert}

\newcommand{\mc}[1]{\mathcal{#1}}

\newcommand{\xb}{\bs{x}}
\newcommand{\zb}{\bs{z}}
\newcommand{\yb}{\bs{y}}

\newcommand{\pp}[2]{\frac{\partial #1}{\partial #2}} %

\newcommand{\E}{\mc{E}}

\newcommand{\lilan}{\texttt{LiLaN}}
\newcommand{\lilans}{\lilan\,}

%% file: krish_macros.tex
\newcommand{\LRp}[1]{\left( #1 \right)} 
\newcommand{\LRs}[1]{\left[ #1 \right]} 

                             \newcommand{\dd}[2]{\frac{d #1}{d #2}} 

\newcommand{\mb}[1]{\mathbf{#1}} 

\newcommand{\p}{p}

\newcommand{\real}{\mathbb{R}}

\newcommand{\w}{{w}}

\newcommand{\wb}{\mb{\w}}

\newcommand{\gb}{\mb{g}}

\newcommand{\bff}{\boldsymbol{f}}

\renewcommand{\epsilon}{\varepsilon}
\newcommand{\eval}[2][\right]{\relax
  \ifx#1\right\relax \left.\fi#2#1\rvert}

\def\sD{{\bs{d}}}
\def\sE{{\bs{e}}}



%% file: intro.tex
Stiff ordinary differential equations (StODEs), arising in many scientific and engineering disciplines, pose unique challenges for numerical integration. While a precise definition of stiffness has eluded mathematicians for decades, systems characterized by numerical stiffness typically have important behaviors on timescales that differ by several orders of magnitude. In order to realize these features and ensure that the solution remains bounded during numerical integration with explicit methods, extremely small time steps must be taken \cite{miranker1976computational}, making explicit integration prohibitively expensive.
It has been established that stiff problems are generally solved more efficiently using implicit methods, which are more costly at each integration step, than with explicit methods \cite{MDchem}. Stiff solvers often require solving a nonlinear system of equations at each integration step \cite{MDchem} which drastically increases the computational requirements for large systems. For certain applications, specifically for control, design, optimization, and uncertainty quantification scenarios, one often needs to run an expensive forward model many times \cite{wilcox}, which renders implicit approaches computationally infeasible. The expensive nature of solving StODEs, whether by explicit or implicit numerical methods, necessitates the development of novel machine learning (ML) techniques for accelerating simulation of such systems.

In recent years, scientific machine learning techniques have been deployed as a faster alternative 
to traditional numerical solvers \cite{anantharaman2021acceleratingsimulationstiffnonlinear, galaris2021numericalsolutionstiffodes, van2024model, sulzer2023speedingastrochemicalreactionnetworks, DHZHYS}.
One popular approach for modeling dynamical systems is the Neural ODE (NODE) approach, which approximates the right-hand side of an ordinary differential equation (ODE) using a neural network \cite{chen2019neuralordinarydifferentialequations,nguyen2022model}. However, as discussed in \cite{Kim_2021}, NODE  struggles to learn solutions of stiff systems of ODEs. The authors propose scaling factors, derived from the data, which can be used to normalize the NODE dynamics such that learning stiff dynamical systems becomes feasible \cite{Kim_2021}. Specifically, the scaling factors help balance the scales of different components in the loss function to make training less challenging for the optimizer. Furthermore, it was shown that, for stiff problems, the implementation of explicit conservation of mass constraints in the loss function during NODE training led to significantly better performance than an unconstrained NODE approach \cite{KTKAPP}. Another NODE-based approach considers encoding the ODE system to a latent space and implements an explicit constraint in the loss function to minimize an estimate of the stiffness ratio for the latent dynamics, showing promising results on two reaction equations \cite{DHZHYS}. These modifications to vanilla NODE exhibited great success in the simulation of stiff ODEs. However, though NODE-based approaches were able to achieve reasonable accuracy on stiff problems, all of the methods discussed still require the use of an expensive, implicit integrator at inference time.

Another approach, the physics-informed neural network (PINN), described in \cite{raissi2017physicsinformeddeeplearning}, has received significant attention for its ability to speed up the simulation of ODEs and PDEs. Training a PINN results in a function that can be evaluated to approximate the solution of an ODE/PDE, and thus avoids expensive numerical integration entirely.
However, it was shown in \cite{wang2020understandingmitigatinggradientpathologies} that the gradient flow dynamics for training PINNs are typically stiff themselves, placing strict limitations on the gradient descent step size required for stable training. Stiff gradient dynamics may mean that gradient descent cannot effectively optimize network parameters during training and the optimization may diverge \cite{wang2020understandingmitigatinggradientpathologies}.
One solution to this problem is proposed in  \cite{Ji_2021}, which illustrates a method utilizing quasi-steady state assumptions (QSSA) to convert a stiff system into one that is non-stiff and thus easier for PINNs to handle. The work in \cite{baty2023solvingstiffordinarydifferential} proposed a different, but effective solution to the problem of learning stiff ODEs, using PINNs, by augmenting the loss function with conservation of energy constraints and implementing a sequential training process which prioritizes learning the early temporal evolution of the ODE during early epochs and gradually increasing the time interval seen in training. Additionally, in \cite{nasiri2022reducedpinnintegrationbasedphysicsinformedneural}, the authors were able to accurately simulate stiff ODEs using a modified PINN architecture based on solving the corresponding integral form of the ODE of interest. The aforementioned approaches showed significant improvements in accuracy over the basic PINN architecture, which typically struggles to learn solutions for stiff ODEs. While the PINN approach is efficient due to avoiding numerical integration, and the modifications to the PINN approach described in \cite{Ji_2021, baty2023solvingstiffordinarydifferential, nasiri2022reducedpinnintegrationbasedphysicsinformedneural} have allowed PINNs to model stiff problems more accurately than before, PINNs have their own issues, such as not being generalizable to unseen initial and boundary conditions, unless retraining is carried out.

Several approaches to simulate dynamical systems involve the use of autoencoders to learn a latent dynamical system, typically of lower dimensionality than the original system, that is faster to solve than the original model. The latent solutions are then decoded to approximate solutions using trained decoders \cite{sulzer2023speedingastrochemicalreactionnetworks, DHZHYS, Xie_2024, Lee_2021, Lusch_2018, azencot2020forecastingsequentialdatausing, goswami2023learningstiffchemicalkinetics}. The encoders and decoders can be trained to handle a wide range of initial conditions, boundary conditions, and input parameters so that the latent dynamical system holds significant information about the true dynamics under many different circumstances. 
However, without a good estimate of the intrinsic dimensionality of the system, important information could be lost in the encoding process \cite{Levina2004MaximumLE}. On the other hand, some new approaches propose increasing the dimension of the state through nonlinear featurization. In the area of reservoir computing, methods that utilize high-dimensional "reservoir" representations of data have shown promise in the integration of stiff and even chaotic systems \cite{anantharaman2021acceleratingsimulationstiffnonlinear, galaris2021numericalsolutionstiffodes, chaos}. Additionally, in the reinforcement learning (RL) community,  augmenting the state vector with nonlinear featurization has been shown to improve the training of RL algorithms for certain tasks \cite{ota2020increasinginputdimensionalityimprove}. Even though these reservoir computing approaches increase the dimensionality of the original problem, they achieve speedup at inference time because they do not require classical numerical integration at inference time (such as in the case for NODE). Instead, the result of training is a neural network that can be evaluated to obtain the solution to the stiff system at the desired future time.  The approach we will propose in this paper follows this philosophy, i.e., expanding the dimension of the problem through nonlinear featurization to approximate solution operators\textemdash i.e., the flow map\textemdash of the stiff ODE without requiring numerical integration.


While the approaches we have discussed above have shown some speedup in integration, while still accurately predicting the solutions to StODEs, major challenges remain. In the NODE approaches, for example, while \cite{DHZHYS, Kim_2021, KTKAPP} demonstrate the ability of NODE to learn stiff dynamics, each approach still necessitates the use of an expensive
implicit solver at inference time. On the other hand, while the PINN approaches \cite{Ji_2021, baty2023solvingstiffordinarydifferential, nasiri2022reducedpinnintegrationbasedphysicsinformedneural} do not require expensive implicit integration, they must be retrained for new boundary and/or initial conditions. The QSSA approach \cite{Ji_2021} works by removing the fastest evolving components of the solution, meaning that the information about those components will be lost, though that information is important for many applications such as those in this paper. 
Similarly, the approach in \cite{anantharaman2021acceleratingsimulationstiffnonlinear}, though it obtained significant speedup results, can be inaccurate due to exhibiting oscillatory behavior, resembling Gibb's phenomenon, in prediction. Other work that presented speedup results showed only mild speedup and often only under specific test conditions \cite{galaris2021numericalsolutionstiffodes, DHZHYS}. In \cite{sulzer2023speedingastrochemicalreactionnetworks}, multiple orders of magnitude of speedup were achieved; however, the approach presented lacks any theoretical guarantees. In this work, we will propose an approach which not only provides an approximate solution operator
to achieve a remarkable speedup, but also maintains a high level of accuracy when evaluated across a wide range of new initial conditions and parameter inputs, even for the fastest evolving components of the solution to stiff systems. 

In particular, we developed an autoencoder-based approach, called \lilans (Linear Latent Network), to accelerate the simulation of StODEs by approximating the solution operator (i.e., the flow map in the context of ODEs). \lilans is inspired by kernel methods, in which the original problem under consideration is featurized/lifted to (much) higher-dimensional feature spaces where the problem is now easier. In particular, a similar intuition is that, given a stiff dynamical system, it may be possible to learn a non-stiff, higher-dimensional latent dynamical system with trivial solutions. Unlike kernel methods, which rely on particular feature maps to organize the latent dynamics so that hopefully it is easier to simulate, \lilans forces the latent dynamics to be trivially linear, and then learns the nonlinear feature maps using neural networks. To achieve this goal, we deploy an encoder that takes in the initial condition of the original dynamical system as input and outputs the initial condition for the latent dynamics. Furthermore, by implicitly imposing a constant velocity latent dynamics structure, we completely evade numerical integration in the latent space since the solution is a sequence of fully linear trajectories and can therefore be computed analytically. The trained decoder can then retrieve, from the latent representation, the solution to the original dynamics, at any desirable time. Our theoretical analysis reveals that under mild assumptions, one can approximate the flow map of a stiff, nonlinear system on a compact set to any degree of accuracy, $\epsilon$, using \lilan, and that the required dimension of the latent space is independent of $\epsilon$. Additionally, our theoretical results mandate that employing a nonlinear transformation on the time variable is necessary to guarantee that a \lilans architecture exists which can approximate a given flow map. Intuitively, this transformation allows for "stretching/squeezing" time in the latent space, thereby allowing for varying levels of attention to different temporal regions of the solution.  Numerical experiments on the "Robertson Stiff Chemical Kinetics Model" \cite{anantharaman2021acceleratingsimulationstiffnonlinear} and the "Plasma Collisional-Radiative Model" \cite{flychk} suggest that \lilans can outperform state-of-the-art ML approaches for calculating solutions to StODEs. Additionally, \lilans outperformed other machine learning methods on multiple PDEs with known stiff behaviors, such as the "Allen-Cahn" and "Cahn-Hilliard" PDEs \cite{montanelli2020solvingperiodicsemilinearstiff}. It should be pointed out that though we focus on stiff ODEs, \lilans is equally applicable to non-stiff ODEs.

%% file: methodology_new.tex

\begin{itemize}
    \item {\bf{Encoders} $\boldsymbol{\sE}\LRp{\xb_0,\ \mathbf{p}; \ \boldsymbol{\beta}}$ and $\boldsymbol{\bs{c}}\LRp{\xb_0,\ \mathbf{p}; \ \boldsymbol{\alpha}}$}: Note that both $\boldsymbol{\sE}$ and $\boldsymbol{\bs{c}}$ are neural networks with parameters $\boldsymbol{\beta}$ and $\boldsymbol{\alpha}$ respectively. The network $\boldsymbol{\sE}\LRp{.,.,\boldsymbol{\beta}}: \sG \times \sX_p\mapsto \real^m$ takes in $\xb_0,\ \mathbf{p}$, as defined in \eqref{original_dynamics}, as inputs and outputs an $m$ dimensional initial condition for the latent dynamical system given below:
\begin{equation}
        \begin{aligned}
&diag\LRp{\dd{\yb}{\boldsymbol{\tau}}}=\boldsymbol{\bs{c}}(\xb_0, \mathbf{p};\ \boldsymbol{\alpha}),\quad \ \ \yb({\bf{0}})=\boldsymbol{\sE}\LRp{\xb_0, \mathbf{p}; \ \boldsymbol{\beta}} = \yb_0,
        \end{aligned}
        \label{latent_dynamics}
    \end{equation}    
where $diag\LRp{\dd{\yb}{\boldsymbol{\tau}}}$ extracts the diagonal elements of matrix $\LRs{\dd{\yb}{\boldsymbol{\tau}}}$ into a column vector, $ \yb(\boldsymbol{\tau})\in\real^m$ denotes the solution to \eqref{latent_dynamics} given by \eqref{solution_latent},
  $\boldsymbol{\tau} \in \real^m$ is a nonlinear transformation of time as described by a neural network below. The network $\boldsymbol{\bs{c}}\LRp{.,.,\boldsymbol{\alpha}}: \sG \times \sX_p\mapsto \real^m$ takes in $\xb_0,\ \mathbf{p}$, as defined in \eqref{original_dynamics}, as inputs and outputs an $m$ dimensional slope vector (constant velocity) for the latent dynamics as described in \eqref{latent_dynamics}.

    \item {\bf{Nonlinear Time Transformation $\boldsymbol{\tau}\LRp{t,\ \xb_0,\ \mathbf{p}; \ \boldsymbol{\nu}}$}}: In \eqref{latent_dynamics}, $\boldsymbol{\tau}$ is a neural network with parameters $\boldsymbol{\nu}$. The network $\boldsymbol{\tau}\LRp{.,.,.;\boldsymbol{\nu}}: I \times \sG \times \sX_p \mapsto \real^m$ performs a nonlinear transformation on the time variable which "stretches/squeezes" time in the latent space, thereby allowing for varying levels of attention to different regions/dynamics in the solution depending on the inputs $\xb_0,\ \mathbf{p}$. Note that the solution of \eqref{latent_dynamics} is a sequence of linear trajectories, 
\begin{equation}
  \yb(\boldsymbol{\tau})=\boldsymbol{\sE}(\xb_0, \mathbf{p};\ \boldsymbol{\beta}) +  \boldsymbol{\tau}\ \circ\ \boldsymbol{\bs{c}}(\xb_0, \mathbf{p};\ \boldsymbol{\alpha}),
    \label{solution_latent}
    \end{equation}
    where $\circ$ denotes the Hadamard product for vectors.


    \item  {\bf{Decoder}}  $\boldsymbol{\sD}\LRp{\yb;\ \boldsymbol{\theta}}:$ The decoder network $\boldsymbol{\sD}(.; \ \boldsymbol{\theta}): \real^m\mapsto \real^{n_x} $    
with parameters, $\boldsymbol{\theta}$, decodes the solution, $\yb$, in \eqref{solution_latent}, back to the solution of original dynamics, $\xb$, in \eqref{original_dynamics}.

\end{itemize}
Based on the above networks, our procedure can be summarized as follows. For a given initial condition, $\xb_0$, and parameters, $\mathbf{p}$, the encoders, $\boldsymbol{\sE}$ and $\boldsymbol{\bs{c}}$
generate the initial condition and the constant velocity vector for the latent dynamics described in \eqref{latent_dynamics}. The solution of this latent dynamics at a latent time $\bf{\tau}$ is given in \eqref{solution_latent}. The decoder then retrieves the solution $\xb(t)$ to the original dynamics \eqref{original_dynamics}. The networks are trained, in tandem, to find the optimal parameters $\boldsymbol{\alpha},\ \boldsymbol{\beta},\ \boldsymbol{\theta},\ \boldsymbol{\nu}$ that minimize the loss function (details in \Cref{sec:loss}) that compares the generated solution $\hat{\xb}(t)$ to the true solution $\xb(t)$. 
The training data for \lilans take the form:
\begin{equation}
    \{\{\xb_0,\ \mathbf{p}\}_i,\ \{\xb(t_0),\ \xb(t_1),\dots \xb(t_M)\}_i \}_{i=1}^N,
    \label{train_form}
    \end{equation}
    where  $N$ is the total number of training sample trajectories, and $t_0,\dots, t_M$ are the temporal collocation points for each trajectory, which are used to construct the loss function (see  \Cref{sec:loss} for more details).

It should be noted that an independent approach\footnote{We were not aware of their work during the development of \lilan. Our inspiration is kernel methods as discussed at the end of \cref{sect:intro}.} by Sulzer and Buck \cite{sulzer2023speedingastrochemicalreactionnetworks} considers a fully linear latent space to simulate stiff dynamics. Even though the method lacks theoretical guarantees, it achieves multiple orders of magnitude of speedup versus a traditional numerical solver. Here we highlight key differences between \lilans and the method proposed by Sulzer and Buck \cite{sulzer2023speedingastrochemicalreactionnetworks}. Notably, their method does not consider a learned nonlinear transformation on the time variable, denoted by $\boldsymbol{\tau}$ in this work. The authors also emphasize the use of a lower-dimensional latent space (i.e., $m<n_x$) in their approach. Additionally, they assumed that the parameters $\bold{p}$ are fixed-i.e., $\bold{p}$ in \eqref{original_dynamics} remains constant across all training samples. The objective then is to train a surrogate model to simulate the trajectory $\xb(t)$ for new unseen initial conditions $\xb_0$ in \eqref{original_dynamics}.
The latent NODE in their approach, analogous to our $\boldsymbol{\bs{c}}$, takes as input the latent initial condition $\mathbf{y}_0$ rather than the true initial condition $\xb_0$. In \cref{num_re}, we provide extensive numerical experiments comparing the performance of \lilans with the approach by Sulzer and Buck \cite{sulzer2023speedingastrochemicalreactionnetworks}. Specifically, we evaluate two variants of their method: i) Sulzer and Buck I: Their original approach  with a reduced latent dimension (i.e. $m\leq n_x$); and ii) Sulzer and Buck II: Their approach with an expanded latent dimension (i.e. $m \ge n_x$).


\subsection{The \lilans approach}
\label{sec:motive}

\subsubsection{"Direct learning" is a special case of \lilans}
\label{special_case}

When dealing with stiff systems, integration using both implicit and explicit methods can be computationally expensive. If the flow map is continuous on a compact set, then by the universal approximation theorem for neural networks \cite{kidger2020universal, Bui-Thanh2022}, theoretically, one could use a single hidden layer neural network to learn the flow map to any degree of accuracy, $\epsilon$. That is, there exists a one-hidden layer neural network that can predict the future state of the system for any given initial condition, parameters, and future time. In particular, denoting the flow map in \cref{flow} as $\bpsi\LRp{t,\xb_0,\mathbf{p}}=\bpsi\LRp{\bz}$ (where $\bz=\LRp{t,\xb_0,\mathbf{p}}$), 
and the universal neural network as
\begin{equation}
\boldsymbol{\sN}_{\mathrm{direct}}\LRp{\bz}=\bW_1\phi\LRp{\bW \bz+\bb}+ \bcc_1,
\label{direct_net}
\end{equation}
where $\bW_1\in \real^{n_x\times m},\ \bW\in \real^{m\times (1+n_x+n_p)},\ \bb\in \real^{m},\ \bcc_1\in \real^{n_x}$ denotes the weights and biases of the network $\boldsymbol{\sN}_{\mathrm{direct}}$, $\phi$ is a nonlinear activation function, and $m$ denotes the number of neurons in the hidden layer. Note that in this approach, the neural network \eqref{direct_net} tries to learn the ground truth flow map $\bpsi\LRp{\bz}$ based on the training data \ref{train_form}.
This procedure completely avoids numerical integration at inference time and provides an extremely fast surrogate model. We call this approach the "{\bf direct learning}" approach.

We now show that "{\bf direct learning}"  is a special case of our proposed \lilans  approach in \cref{method}. To that end, note that the direct neural network in \eqref{direct_net} can be equivalently written as:
 \begin{equation}
 \begin{aligned}
\boldsymbol{\sN}_{\mathrm{direct}}\LRp{t,\xb_0,\mathbf{p}}&=\bW_1\bphi\LRp{\bW_t \ t+\bW_{\xb_0}\xb_0+\bW_{\mathbf{p}}\mathbf{p}+\bb}+ \bcc_1,\\
     \end{aligned}
     \label{simm}
 \end{equation}
where $\bW_t\in \real^{m\times 1}$ denotes the first column of $\bW$, $\bW_{\xb_0}\in \real^{m\times n_x}$ denotes the $2^{nd}$ to $(n_x+1)^{th}$ columns of $\bW$,  $\bW_{\mathbf{p}}\in \real^{m\times n_p}$ denotes the $(n_x+2)^{th}$ to $(n_x+n_p+1)^{th}$ columns of $\bW$.  Next, by choosing
\begin{equation}
    \begin{aligned}
        \boldsymbol{\sE}\LRp{\xb_0,\ \mathbf{p}; \ \{ \bW_{\xb_0},\bW_{\mathbf{p}},\bb\}}&=\bW_{\xb_0}\xb_0+\bW_{\mathbf{p}}\mathbf{p}+\bb, \ \ \boldsymbol{\bs{c}}\LRp{\xb_0,\ \mathbf{p}}=\mathbf{1}_m,\ \ \boldsymbol{\tau}\LRp{t,\ \xb_0,\ \mathbf{p}; \ \bW_t}=\bW_t\ t,\ \,\\
         \boldsymbol{\sD}\LRp{\yb;\ \{ \bW_1, \bcc_1 \}}&=\bW_1\bphi\LRp{\yb}+\bcc_1,
    \end{aligned}
    \label{dl_l}
\end{equation}
then the \lilans  approach in \cref{method} reduces to \cref{simm}.

\begin{remark}$ $
\begin{enumerate}
    \item 
    Note that instead of employing simple linear encoders and linear time transformation as in \cref{dl_l}, \lilans deploys deep nonlinear encoders  $\boldsymbol{\sE}\LRp{\xb_0,\ \mathbf{p}; \ \boldsymbol{\beta}}$ and $\boldsymbol{\bs{c}}\LRp{\xb_0,\ \mathbf{p}; \ \boldsymbol{\alpha}}$, and a deep nonlinear time transformation network $\boldsymbol{\tau}\LRp{t,\ \xb_0,\ \mathbf{p}; \ \boldsymbol{\nu}}$.  This is expected to improve the expressiveness of networks, possibly requiring significantly fewer parameters to learn the flow map \cite{augustine2024survey, guhring2020expressivitydeepneuralnetworks}. 
    In \Cref{fig:loss_curve}, we demonstrate how employing nonlinear encoders and time transformation improves the generalization performance of the architecture.
    \Cref{fig:loss_curve} (left) depicts the average point-wise relative error of predictions (see \eqref{eq:R1}) over time for the test dataset from \cref{sec:full-CR}, with \lilan, shown in blue, and the "{\bf direct learning}" approach  shown in orange. \lilans greatly outperforms the "{\bf direct learning}" approach, having at least an order of magnitude lower error than the direct learning approach for the majority of the simulation time interval. Both methods become less accurate in the middle portion of the trajectory, where the dynamics for the collisional-radiative cooling model (\cref{sec:full-CR}) begin evolving on very small timescales. However, the "{\bf direct learning}" approach, struggles more in this region, shown by the larger increase in error. The error of the direct learning approach decreases to a similar value to that of \lilans as the dynamics approach steady state near $t=10^{-3}s$. Viewing \cref{fig:loss_curve} (right), we observe that while both approaches reached very low final losses on the training data, the "{\bf direct learning}" approach failed to generalize to unseen cases as well, signified by the larger validation loss, which did not converge. Note that, for a fair comparison, both approaches were given approximately the same number of learnable parameters and the same data-points for training.
\item In \cref{special_case} we will present a universal approximation theorem for \lilan, which also provides a new proof of the universal approximation theorem for the direct approach as a by-product.
    
\begin{figure}[h!t!b!]
 \begin{minipage}{0.49\textwidth}
  \centering
    \resizebox{1\linewidth}{!}{\input{figs/motivation_rel_errors}}
  \end{minipage}
  \begin{minipage}{0.49\textwidth}
  \centering
    \resizebox{1\linewidth}{!}{\input{figs/loss_curves}}
  \end{minipage}
  \caption{Learning flow map by \lilan, 
  vs. "{\bf direct learning}" approach. Left to Right: Average point-wise relative error \eqref{eq:R1} in predictions for the full CR model (\cref{sec:CR-model}) on the test dataset, depicting \lilans having at least an order of magnitude lower error for the majority of the simulation; Training and validation loss curves for the two approaches. These curves demonstrate that validation loss of the "{\bf direct learning}" approach fails to converge, while \lilans validation loss remains very close to the training loss.}
  \label{fig:loss_curve}
  \end{figure}
\end{enumerate}
\end{remark}

\subsubsection{Universal approximation for \lilans}
\label{special_case}
In the following sections, we will prove a universal approximation theorem for \lilans presented in \cref{method}.
We begin with a definition. 
\begin{definition}[Class of functions $\sN^{\phi}_{a,b}$]
\label{class}
    Let $\phi:\real\rightarrow \real$ be any non-affine, Lipschitz continuous function which is continuously differentiable at least at one point, having a nonzero derivative at that point.  We denote  by $\sN^{\phi}_{a,b}$ the class of functions described by feedforward neural networks with `a' neurons in the input layer, `b' neurons in the output layer, and an arbitrary number of hidden layers, each with `$a+b+2$' neurons and activation function $\phi$. Every neuron in the output layer has the identity activation function.
\end{definition}

\begin{remark}
    Note that commonly employed activation functions in deep learning such as `Tanh', `Sigmoid', 'ReLU', and 'ELU' satisfy the conditions in \cref{class}.
\end{remark}

We will start by examining the conditions on $\bff\LRp{\xb(t),\ {\mathbf{p}}}$ such that for any given $\xb_0,\ \mathbf{p}$ the solution exists on the entire real line and the flow map is continuous with respect to its arguments. The result is summarized in \cref{lem}.



\begin{lemma}
\label{lem}
        Consider the autonomous ordinary differential equation in \eqref{original_dynamics}. Assume that $\bff\LRp{\xb(t),\ {\mathbf{p}}}$ is globally Lipschitz continuous with respect to both  $\xb\LRp{t}$ and $\mathbf{p}$\footnote{  $\bff\LRp{\xb,\ \mathbf{p}}$ is globally Lipschitz continuous if there exists a non-negative constant $M$ such that $\nor{\bff(\xb_1,\ \mathbf{p}_1)-\bff(\xb_2,\ \mathbf{p}_2)}_2\leq M\nor{\mathbf{r}_1-\mathbf{r_2}}_2,\forall \mathbf{r}_1,\mathbf{r}_2\in \real^{n_x+n_p}$, where $\mathbf{r}_1=[\xb_1,\ \mathbf{p}_1]$ and $\mathbf{r}_2=[\xb_2,\ \mathbf{p}_2]$.}.
 Then, the flow map in \cref{flow}, $\bpsi(t,\ \xb_0,\ \mathbf{p})$, is  continuous on $\sW$ with $I_{\xb_0; \mathbf{p}}=\real$.
\end{lemma}
\begin{proof}
    Rewriting the ODE \eqref{original_dynamics} in terms of a new variable $\tilde{\xb}(t)$ as:
    \begin{equation}
        \dd{\tilde{\xb}}{t}=\tilde{\bff}\LRp{\tilde{\xb}(t)},\quad \tilde{\xb}(t)= \begin{bmatrix}
\xb(t) \\
\mathbf{p}
\end{bmatrix}, \quad \tilde{\xb}(0)= \tilde{\xb}_0= \begin{bmatrix}
\xb_0 \\
\mathbf{p}
\end{bmatrix},
\label{new_ode}
    \end{equation}
where
\[\tilde{\bff}\LRp{\tilde{\xb}(t)}=\begin{bmatrix}
{\bff}\LRp{I_1 \tilde{\xb}(t),\ {I_2 \tilde{\xb}(t)}} \\
\mathbf{0}
\end{bmatrix}.
\]
where, $I_1$ projects $\tilde{\xb}(t)$ to $\xb(t)$, and $I_2$ projects $\tilde{\xb}(t)$ to $\mathbf{p}$, and they are  linear and differentiable. Since $\bff\LRp{\xb,\ {\mathbf{p}}}$ is globally Lipschitz continuous, by the global existence theorem 
(Theorem B, Chapter 13 in \cite{simmons2016differential}), we note that $\real$ is the interval of existence of the unique maximal solution of \eqref{new_ode} for any given $\tilde{\xb}_0$. Also note that
$\tilde{\boldsymbol{f}}$ is  continuous   on $\sP$,
where $\sP=\sG\times \sX_p$, due to the assumption that $\bff\LRp{\bx(t),\ \mathbf{p}}$ is  globally Lipschitz continuous with respect to both $\mathbf{p}$ and $\bx(t)$. Now introduce the set:
\[ \tilde{\sW}=\bigcup_{\tilde{\xb}_0\in \sP} \real \times \{\tilde{\xb}_0 \}  \subseteq \real \times \sP, \]
  and define the corresponding flow map for \eqref{new_ode} as $\bpsi_{\tilde{\xb}}:\tilde{\sW} \rightarrow \sP$ such that $\bpsi_{\tilde{\xb}}(t,\tilde{\xb}_0)=\tilde{\xb}(t)$, where $\tilde{\xb}(t)$ solves \eqref{new_ode} for a given $\tilde{\xb}_0$. Since $\tilde{\boldsymbol{f}}$ is  continuous on $\sP$, by Theorem 6.1 in \cite{teschl2024ordinary}, we have $\bpsi_{\tilde{\xb}}(t,\tilde{\xb}_0)$ is continuous on $\tilde{\sW}$.
Now, defining $\bpsi(t,\ \xb_0,\ \mathbf{p})=\bpsi_{\tilde{\xb}}(t,\tilde{\xb_0})$  concludes the proof.
\end{proof}

\begin{proposition}
    There exists a continuous function ${\boldsymbol{l}}\LRp{\yb}: \real^{n_x}\mapsto \real^m$ ($m\geq n_x$) with a  continuous left-inverse    ${\boldsymbol{l}}^{-1}\LRp{\zb}: \mathrm{Range}({\boldsymbol{l}})\mapsto \real^{n_x}$.
    \label{prop:leftInverse}
\end{proposition}
\begin{proof}
    We prove the assertion by constructing a function with the desirable properties. In this proof, the exponential and logarithm are Hadamard exponential and logarithm, that is, with a given vector input, both exponential and logarithm act in a componentwise fashion. Now, consider the functions ${\boldsymbol{l}}\LRp{\yb}: \real^{n_x}\mapsto \real^m$ and  $\bs{g}\LRp{\zb}:\mathrm{Range}({\boldsymbol{l}})\mapsto \real^{n_x}$ as follows:
\begin{equation}
    {\boldsymbol{l}}\LRp{\yb}=\log \LRp{\mathbf{1}+e^{{\bf{A}}\yb}},\quad  {\boldsymbol{l}}^{-1}\LRp{\zb}={\bf{A}}^{\dagger}\LRp{\log(e^{\zb}-\mathbf{1})},
    \label{construct}
\end{equation}
where ${\bf{A}}\in \real^{m\times n_x}$ ($m\geq n_x$) is an arbitrary matrix with linearly independent columns, and
${\bf{A}}^{\dagger}\in \real^{n_x\times m}$ is the pseudo-inverse of ${\bf{A}}$. 
Clearly $\bs{g}\LRp{\zb}$ is a left-inverse of ${\boldsymbol{l}}\LRp{\yb}$ and both are continuous functions on $\real^{n_x}$ and $\mathrm{Range}(\boldsymbol{l})$, respectively.
\end{proof}

Next, we will determine the conditions under which the flow map admits the following decomposition:
\begin{equation}
\bpsi\LRp{t,\ \xb_0, \mathbf{p}}={\boldsymbol{l}}^{-1}\LRp{\gb(\xb_0,\mathbf{p})+\boldsymbol{h}(t,\xb_0,\mathbf{p})\ \circ\ \boldsymbol{b}(\xb_0,\mathbf{p})},
\end{equation}
where a continuous  function ${\boldsymbol{l}}\LRp{\yb}:\real^{n_x}\mapsto \real^m$  and its left-inverse ${\boldsymbol{l}}^{-1}$ are given in \cref{prop:leftInverse}, $\gb\LRp{\xb_0,\ \mathbf{p}}: \sG_0\times \sX_p\mapsto \real^m$  continuous with respect to both arguments, $ \boldsymbol{b}\LRp{\xb_0,\ \mathbf{p}}: \sG_0\times \sX_p\mapsto \real^m$ continuous with respect to both arguments, and $\boldsymbol{h}(t,\xb_0,\mathbf{p}): I\times \sG_0\times \sX_p \mapsto \real^m$ continuous with respect to all arguments. Here, $I$ is a closed and bounded interval, $\xb_0\in \sG_0\subset \sG$ is a given non-empty compact set. The result is summarized in \cref{lemma_w}.

\begin{lemma}
\label{lemma_w}
Consider the autonomous ordinary differential equation in \eqref{original_dynamics} and assume the following:
\begin{enumerate}
\label{cond_1_l}
    \item The conditions in \cref{lem} are satisfied. 
    \label{cond_33}
    \item Assume that the initial condition $\xb_0\in \sG_0\subset \sG$, where $\sG_0$ is a non-empty compact set, and consider a set $E=I\times \sG_0\times \sX_p$, where $I=[0,\ t^u]$, with $t^u\in \real^{+}$, is a closed and bounded interval.
    \label{cond_333}
\end{enumerate}
 Then, there exists a (non-unique\footnote{Non-uniqueness here implies that there are infinitely many choices for functions  ${\boldsymbol{l}}$, $\boldsymbol{b}$,   $\gb$, $\boldsymbol{h}$ for which \eqref{non_uniq} holds true.}) decomposition for a given flow map $\bpsi$ as follows:
 \begin{equation}
\bpsi\LRp{t,\ \xb_0, \mathbf{p}}={\boldsymbol{l}}^{-1}\LRp{\gb(\xb_0,\mathbf{p})+\boldsymbol{h}(t,\xb_0,\mathbf{p})\ \circ \ \boldsymbol{b}(\xb_0,\mathbf{p})},
\label{non_uniq}
\end{equation}
 where the functions  ${\boldsymbol{l}}$, $\boldsymbol{b}$,   $\gb$, $\boldsymbol{h}$ have the following properties:
 \begin{itemize}
     \item  ${\boldsymbol{l}}\LRp{\yb}: \real^{n_x}\mapsto \real^m$ ($m\geq n_x$) belongs to the space of 
continuous functions with a  continuous left-inverse denoted as  ${\boldsymbol{l}}^{-1}\LRp{\zb}:\mathrm{Range}({\boldsymbol{l}})\mapsto \real^{n_x}$\footnote{When ${\boldsymbol{l}}$ satisfies this condition we say that ${\boldsymbol{l}}$ is a topological embedding of $\real^{n_x}$ into $\real^m$\cite{lee2000introduction}.};
\item   $ \boldsymbol{b}\LRp{\xb_0,\ \mathbf{p}}: \sG_0\times \sX_p\mapsto \real^m$ belongs to the space of  strictly non-zero continuous functions on $\sG_0\times \sX_p$;
\item $\gb\LRp{\xb_0,\ \mathbf{p}}: \sG_0\times \sX_p\mapsto \real^m$ belongs to the space of continuous functions on $\sG_0\times \sX_p$, and $\boldsymbol{h}(t,\xb_0,\mathbf{p}): I\times \sG_0\times \sX_p \mapsto \real^m$  belong to the space of continuous functions on $I\times \sG_0\times \sX_p$.
 \end{itemize}


\end{lemma}
\begin{proof}

Given the flow map $\bpsi$ and a function $\boldsymbol{l}$
let us decompose the action of $\boldsymbol{l}$ on the flow map $\bpsi\LRp{t,\ \xb_0, \mathbf{p}}$ as follows:
\begin{equation}
{\boldsymbol{l}}\LRp {\bpsi\LRp{t,\ \xb_0, \mathbf{p}}}= {\boldsymbol{l}}\LRp {\bpsi\LRp{t^*,\ \xb_0, \mathbf{p}}}+{\boldsymbol{u}(t,\xb_0,\mathbf{p})},
\label{action}
\end{equation}
for some  $t^*>0$ in the interval $I$, and the function $\boldsymbol{u}(t,\ \xb_0,\ \mathbf{p})$ is defined as $\boldsymbol{u}(t,\ \xb_0,\ \mathbf{p})={\boldsymbol{l}}\LRp {\bpsi\LRp{t,\ \xb_0, \mathbf{p}}}-{\boldsymbol{l}}\LRp {\bpsi\LRp{t^*,\ \xb_0, \mathbf{p}}}$.  Note that $\bpsi\LRp{t^*,\ \xb_0, \mathbf{p}}$ exists since the solution exists on the entire real line  for any $(\xb_0,\ \mathbf{p})\in \sG_0\times \sX_p$ (see \cref{lem}). Further, 
 $\boldsymbol{u}(t,\ \xb_0,\ \mathbf{p})$ is continuous on $E$ due to the fact the continuity of  $\bpsi\LRp{t,\ \xb_0, \mathbf{p}}$ and  $\boldsymbol{l}$. Now define the following functions:
\[\gb(\xb_0,\mathbf{p})={\boldsymbol{l}}\LRp {\bpsi\LRp{t^*,\ \xb_0, \mathbf{p}}}, \]
\[ \boldsymbol{h}_i(t,\xb_0,\mathbf{p})\  \boldsymbol{b}_i(\xb_0,\mathbf{p})=\boldsymbol{u}_i(t,\xb_0,\mathbf{p})={\boldsymbol{r}_i\LRp{\xb_0,\ \mathbf{p}}} \frac{\boldsymbol{u}_i(t,\xb_0,\mathbf{p})}{{\boldsymbol{r}_i\LRp{\xb_0,\ \mathbf{p}}}},\]
where  $\boldsymbol{r}_i\LRp{\xb_0,\ \mathbf{p}}:\sG_0\times \sX_p\mapsto \real^m$ is a strictly non-zero continuous function. Finally, by taking $\boldsymbol{b}_i\LRp{\xb_0,\ \mathbf{p}}=\boldsymbol{r}_i\LRp{\xb_0,\ \mathbf{p}}$ and $\boldsymbol{h}_i(t,\xb_0,\mathbf{p})=\frac{\boldsymbol{u}_i(t,\xb_0,\mathbf{p})}{{\boldsymbol{r}_i\LRp{\xb_0,\ \mathbf{p}}}}$, \eqref{action} can be rewritten as:
\begin{equation}
{\boldsymbol{l}}\LRp {\bpsi\LRp{t,\ \xb_0, \mathbf{p}}}= \gb(\xb_0,\mathbf{p})+ \boldsymbol{h}(t,\xb_0,\mathbf{p})\ \circ\  \boldsymbol{b}(\xb_0,\mathbf{p}).
\label{pr_left}
\end{equation}
 Now applying the left inverse ${\boldsymbol{l}}^{-1}$ on both sides of \eqref{pr_left} we have:
\[{\boldsymbol{l}}^{-1}\LRp{ {\boldsymbol{l}}\LRp{\bpsi\LRp{t,\ \xb_0, \mathbf{p}}}}={\boldsymbol{l}}^{-1}\LRp{\gb(\xb_0,\mathbf{p})+\boldsymbol{h}(t,\xb_0,\mathbf{p})\ \circ \ \boldsymbol{b}(\xb_0,\mathbf{p})}\]
\begin{equation}
\implies \bpsi\LRp{t,\ \xb_0, \mathbf{p}}={\boldsymbol{l}}^{-1}\LRp{\gb(\xb_0,\mathbf{p})+\boldsymbol{h}(t,\xb_0,\mathbf{p})\ \circ \ \boldsymbol{b}(\xb_0,\mathbf{p})},
\label{subs}
\end{equation}
where we used the fact that ${\boldsymbol{l}}^{-1}\LRp{ {\boldsymbol{l}}(\yb)}=\yb$ since the left-inverse exists by assumption, and this concludes the proof.
\end{proof}

We are now in the position to present the universal approximation theorem for \lilan. 

\begin{theorem}
\label{theo_im}
Consider the autonomous ordinary differential equation in \eqref{original_dynamics} and assume that the conditions in \cref{lemma_w} are satisfied.

Then, $\forall m\geq n_x$, $m$ being the latent dimension, and $\forall \epsilon >0$,  there exist four neural networks $\boldsymbol{\sD}\LRp{.;\ \boldsymbol{\theta}}:\real^m\mapsto \real^{n_x}$, 
$\boldsymbol{\sE}\LRp{.,.;\boldsymbol{\beta}}: \sG_0 \times \sX_p\mapsto \real^m$, $\boldsymbol{\bs{c}}\LRp{.,.;\boldsymbol{\alpha}}: \sG_0 \times \sX_p\mapsto \real^m$,   and $\boldsymbol{\tau}\LRp{.,.,.;\boldsymbol{\nu}}: \real \times \sG_0 \times \sX_p \mapsto \real^m$ with the associated  latent dynamics:
\begin{equation}
        \begin{aligned}
&diag\LRp{\dd{\yb}{\boldsymbol{\tau}}}=\boldsymbol{\bs{c}}(\xb_0, \mathbf{p};\ \boldsymbol{\alpha}),\quad \ \ \yb(\bf{{0}})=\boldsymbol{\sE}\LRp{\xb_0, \mathbf{p}; \ \boldsymbol{\beta}} = \yb_0,
        \end{aligned}
        \label{latent_dynamic}
    \end{equation}  
 such that:
 \begin{equation}
      \sup_{(t,\ \xb_0, \, \mathbf{p})\in E}\LRp{\nor{\bpsi\LRp{t,\ \xb_0, \mathbf{p}}-{\boldsymbol{\sD}}\LRp{\boldsymbol{\sE}(\xb_0, \mathbf{p};\ \boldsymbol{\beta}) +  \boldsymbol{\tau}\LRp{t,\xb_0,\mathbf{p};\ \boldsymbol{\nu}}\ \circ \  \boldsymbol{\bs{c}}(\xb_0, \mathbf{p};\ \boldsymbol{\alpha}); \ \boldsymbol{\theta}}}_2}\leq \epsilon,
      \label{univ_appr}
 \end{equation}
where each network $\boldsymbol{\sD},\ \boldsymbol{\E},\ \boldsymbol{\bs{c}},$ and $\ \boldsymbol{\tau}$ belong to the appropriate class  functions defined in \cref{class} and satisfies:

\[\sup_{(t,\ \xb_0, \, \mathbf{p})\in E}\nor{{\boldsymbol{l}}^{-1}\LRp{\gb(\xb_0,\mathbf{p})+\boldsymbol{h}(t,\xb_0,\mathbf{p})\ \circ \ \boldsymbol{b}(\xb_0,\mathbf{p})}-\boldsymbol{\sD}\LRp{\gb(\xb_0,\mathbf{p})+\boldsymbol{h}(t,\xb_0,\mathbf{p})\ \circ \ \boldsymbol{f}(\xb_0,\mathbf{p});\ \boldsymbol{\theta}}}_2\leq \frac{\epsilon}{4},\]

\[\sup_{(t,\ \xb_0, \, \mathbf{p})\in E}\nor{\gb(\xb_0,\mathbf{p})-\boldsymbol{\sE}\LRp{ \xb_0, \, \mathbf{p}\ ; \boldsymbol{\beta}}  }_2\leq \frac{\epsilon}{4L},\quad \sup_{(t,\ \xb_0, \, \mathbf{p})\in E}\nor{\boldsymbol{b}(\xb_0,\mathbf{p})-\boldsymbol{\bs{c}}\LRp{ \xb_0, \, \mathbf{p}; \boldsymbol{\alpha}} }_2\leq \frac{\epsilon}{4L c_2}, \]
\[\sup_{(t,\ \xb_0, \, \mathbf{p})\in E}\nor{\boldsymbol{h}(t,\xb_0,\mathbf{p})-\boldsymbol{\tau}\LRp{ t,\ \xb_0, \, \mathbf{p};\ \boldsymbol{\nu}}}_2\leq \frac{\epsilon}{4Lc_1}, \]
where $\boldsymbol{l}(\yb),\ \gb\LRp{\xb_0,\ \mathbf{p}},\ \boldsymbol{b}\LRp{\xb_0,\ \mathbf{p}},\ \boldsymbol{h}(t,\xb_0,\mathbf{p})$ are functions defined in \cref{lemma_w}, $L$ is the Lipschitz constant of $\boldsymbol{\sD}\LRp{.;\ \boldsymbol{\theta}}$ with respect to the first argument,  $c_1=\sup_{(\xb_0, \, \mathbf{p})\in E} \nor{\boldsymbol{b}(\xb_0,\mathbf{p})}_{\infty}$, $c_2=\sup_{(t,\ \xb_0, \, \mathbf{p})\in E}\nor{\boldsymbol{\tau}\LRp{t,\xb_0,\mathbf{p};\ \boldsymbol{\nu}}}_\infty$.


\end{theorem}
\begin{proof}
  Note that the analytical solution for the constant velocity latent dynamics \eqref{latent_dynamic} can be written as:
    \[\yb(\boldsymbol{\tau})=\boldsymbol{\sE}(\xb_0, \mathbf{p};\ \boldsymbol{\beta}) +  \boldsymbol{\tau}\ \circ \ \boldsymbol{\bs{c}}(\xb_0, \mathbf{p};\ \boldsymbol{\alpha}). \]
For any $\LRp{t,\xb_0,\mathbf{p}}\in E$,  define the error $\rho$ between the ground truth flow map $\bpsi\LRp{t,\ \xb_0, \mathbf{p}}$ and the one predicted by \lilans (for arbitrary choice of networks and it's parameters $\boldsymbol{\theta},\ \boldsymbol{\beta},\ \boldsymbol{\alpha},\ \boldsymbol{\nu}$)  as follows: 
\begin{equation}
    \rho=\nor{\bpsi\LRp{t,\ \xb_0, \mathbf{p}}-{\boldsymbol{\sD}}\LRp{\boldsymbol{\sE}(\xb_0, \mathbf{p};\ \boldsymbol{\beta}) +  \boldsymbol{\tau}\ \circ \ \boldsymbol{\bs{c}}(\xb_0, \mathbf{p};\ \boldsymbol{\alpha});\ \boldsymbol{\theta} }   }_2.
    \label{error}
\end{equation}
Now, using \eqref{subs} in \cref{lemma_w}, the error \eqref{error} can be rewritten as:
\[\rho=\nor{{\boldsymbol{l}}^{-1}\LRp{\gb(\xb_0,\mathbf{p})+\boldsymbol{h}(t,\xb_0,\mathbf{p})\ \circ\ \boldsymbol{b}(\xb_0,\mathbf{p})}-{\boldsymbol{\sD}}\LRp{\boldsymbol{\sE}(\xb_0, \mathbf{p};\ \boldsymbol{\beta}) +  \boldsymbol{\tau}\ \circ\ \boldsymbol{\bs{c}}(\xb_0, \mathbf{p};\ \boldsymbol{\alpha});\ \boldsymbol{\theta}}}_2\]
Adding and subtracting term ${\boldsymbol{\sD}}\LRp{\gb(\xb_0,\mathbf{p})+\boldsymbol{h}(t,\xb_0,\mathbf{p})\ \circ\ \boldsymbol{b}(\xb_0,\mathbf{p});\ \boldsymbol{\theta}}$ above and applying a triangle inequality we have:
\begin{equation}
\begin{aligned}
  & \rho=\nor{{\boldsymbol{l}}^{-1}\LRp{\gb(\xb_0,\mathbf{p})+\boldsymbol{h}(t,\xb_0,\mathbf{p})\ \circ \ \boldsymbol{b}(\xb_0,\mathbf{p})}-{\boldsymbol{\sD}}\LRp{\boldsymbol{\sE}(\xb_0, \mathbf{p};\ \boldsymbol{\beta}) +  \boldsymbol{\tau}\ \circ \ \boldsymbol{\bs{c}}(\xb_0, \mathbf{p};\ \boldsymbol{\alpha});\ \boldsymbol{\theta}}}_2  \\
    &\leq \underbrace{\nor{{\boldsymbol{l}^{-1}}\LRp{\gb(\xb_0,\mathbf{p})+\boldsymbol{h}(t,\xb_0,\mathbf{p})\ \circ \ \boldsymbol{b}(\xb_0,\mathbf{p})}-{\boldsymbol{\sD}}\LRp{\gb(\xb_0,\mathbf{p})+\boldsymbol{h}(t,\xb_0,\mathbf{p})\ \circ\ \boldsymbol{b}(\xb_0,\mathbf{p});\ \boldsymbol{\theta}}    }_2}_{I}\\
&+\underbrace{\nor{{\boldsymbol{\sD}}\LRp{\gb(\xb_0,\mathbf{p})+\boldsymbol{h}(t,\xb_0,\mathbf{p})\ \circ \ \boldsymbol{b}(\xb_0,\mathbf{p});\ \boldsymbol{\theta}}-
{\boldsymbol{\sD}}\LRp{\boldsymbol{\sE}(\xb_0, \mathbf{p};\ \boldsymbol{\beta}) +  \boldsymbol{\tau}\ \circ \ \boldsymbol{\bs{c}}(\xb_0, \mathbf{p};\ \boldsymbol{\alpha});\ \boldsymbol{\theta}}}_2}_{II}
\end{aligned}
\label{finl_decom}
\end{equation}
Now let us consider the term (I) in  \eqref{finl_decom}. Given the continuous functions ${\boldsymbol{l}}^{-1}$, ${\gb},\ \boldsymbol{h},\ \boldsymbol{b}$, $\forall \epsilon_1$ there exists a neural network $\sD\LRp{.;\ \boldsymbol{\theta}(\epsilon_1)}\in \sN^{\phi}_{m,n_x}$ with parameters $\boldsymbol{\theta}(\epsilon_1)$ such that the following holds (the universal approximation result  in \cite[Theorem 3.2]{kidger2020universal}):
\begin{equation}
\sup_{(t,\ \xb_0, \, \mathbf{p})\in E}\nor{{\boldsymbol{l}}^{-1}\LRp{\gb(\xb_0,\mathbf{p})+\boldsymbol{h}(t,\xb_0,\mathbf{p})\ \circ \ \boldsymbol{b}(\xb_0,\mathbf{p})}-\boldsymbol{\sD}\LRp{\gb(\xb_0,\mathbf{p})+\boldsymbol{h}(t,\xb_0,\mathbf{p})\ \circ \ \boldsymbol{b}(\xb_0,\mathbf{p});\boldsymbol{\theta}(\epsilon_1)}}_2\leq \epsilon_1,
\label{second_network}
\end{equation}
Now let us consider the term (II) in  \eqref{finl_decom}.  We have:
\begin{equation}
    \begin{aligned}
      &  \nor{{\boldsymbol{\sD}}\LRp{\gb(\xb_0,\mathbf{p})+\boldsymbol{h}(t,\xb_0,\mathbf{p})\ \circ \ \boldsymbol{b}(\xb_0,\mathbf{p});\ \boldsymbol{\theta}}-
{\boldsymbol{\sD}}\LRp{\boldsymbol{\sE}(\xb_0, \mathbf{p};\ \boldsymbol{\beta}) +  \boldsymbol{\tau}\ \circ \ \boldsymbol{\bs{c}}(\xb_0, \mathbf{p};\ \boldsymbol{\alpha});\ \boldsymbol{\theta}}}_2\\
        & \leq L \LRp{\underbrace{\nor{\gb(\xb_0,\mathbf{p})+\boldsymbol{h}(t,\xb_0,\mathbf{p})\ \circ \  \boldsymbol{b}(\xb_0,\mathbf{p})-\boldsymbol{\sE}(\xb_0, \mathbf{p};\ \boldsymbol{\beta}) +  \boldsymbol{\tau}\ \circ \ \boldsymbol{\bs{c}}(\xb_0, \mathbf{p},\ \boldsymbol{\alpha})}_2}_{III}}
    \end{aligned}
    \label{term_5}
\end{equation}
where we have used the fact that $\sD\LRp{.;\ \boldsymbol{\theta}}\in \sN_{m,n_x}^{\phi}$ is Lipschitz continuous (with constant $L$). Note that the Lipschitz continuity of $\sD$ is easily obtained by employing Lipschitz continuous activation functions.
We can bound (III) in \eqref{term_5} triangle inequality as
\begin{equation}
\begin{aligned}
  &  \nor{\gb(\xb_0,\mathbf{p})+\boldsymbol{h}(t,\xb_0,\mathbf{p})\ \circ \ \boldsymbol{b}(\xb_0,\mathbf{p})-\boldsymbol{\sE}(\xb_0, \mathbf{p};\ \boldsymbol{\beta}) +  \boldsymbol{\tau}\ \circ \ \boldsymbol{\bs{c}}(\xb_0, \mathbf{p},\ \boldsymbol{\alpha})}_2\\
    & \leq \underbrace{\nor{\gb(\xb_0,\mathbf{p})-\boldsymbol{\sE}(\xb_0, \mathbf{p};\boldsymbol{\beta})  }_2}_{IV}+\underbrace{\nor{\boldsymbol{h}(t,\xb_0,\mathbf{p})\ \circ \ \boldsymbol{b}(\xb_0,\mathbf{p})-  \boldsymbol{\tau}\ \circ \ \boldsymbol{\bs{c}}(\xb_0, \mathbf{p},\boldsymbol{\alpha})}_2}_{V}.
\end{aligned}
\label{6_7}
\end{equation}
For (IV) in \eqref{6_7}, again by invoking \cite[Theorem 3.2]{kidger2020universal}, we have $\forall \epsilon_2$  there exists a neural network  $\boldsymbol{\sE}\LRp{ \xb_0, \, \mathbf{p}\ ; \boldsymbol{\beta}(\epsilon_2)} \in \sN^{\phi}_{n_x+n_p,m}$ with parameters $ \boldsymbol{\beta}(\epsilon_2)$  such that, 
\begin{equation}
\sup_{(t,\ \xb_0, \, \mathbf{p})\in E}\nor{\gb(\xb_0,\mathbf{p})-\boldsymbol{\sE}\LRp{ \xb_0, \, \mathbf{p}\ ; \boldsymbol{\beta}(\epsilon_2)}  }_2\leq \epsilon_2.
\label{third_network}
\end{equation}
Analyzing term (V) in  \eqref{6_7}, we have:
\begin{equation}
\begin{aligned}
    &\nor{\boldsymbol{h}(t,\xb_0,\mathbf{p})\ \circ \ \boldsymbol{b}(\xb_0,\mathbf{p})-  \boldsymbol{\tau}\circ \boldsymbol{\bs{c}}(\xb_0, \mathbf{p},\boldsymbol{\alpha})}_2\\
    & = \nor{\boldsymbol{h}(t,\xb_0,\mathbf{p})\ \circ \ \boldsymbol{b}(\xb_0,\mathbf{p}) -  \boldsymbol{\tau}\ \circ \ \boldsymbol{b}(\xb_0, \mathbf{p}) + \boldsymbol{\tau}\ \circ \ \boldsymbol{b}(\xb_0, \mathbf{p})-  \boldsymbol{\tau}\ \circ \ \boldsymbol{\bs{c}}(\xb_0, \mathbf{p},\boldsymbol{\alpha})}_2\\
    & \leq \nor{\LRp{\boldsymbol{h}(t,\xb_0,\mathbf{p})-\boldsymbol{\tau}}\ \circ \ \boldsymbol{b}(\xb_0,\mathbf{p})}_2+\nor{\boldsymbol{\tau}\ \circ \ \LRp{\boldsymbol{b}(\xb_0,\mathbf{p})-\boldsymbol{\bs{c}}(\xb_0, \mathbf{p},\boldsymbol{\alpha})}}_2 \\
    & \leq \nor{{\mathbf{F}}_{f}\LRp{\xb_0,\ \mathbf{p}}\LRp{\boldsymbol{h}(t,\xb_0,\mathbf{p})-\boldsymbol{\tau}}}_2+\nor{{\mathbf{T}}_{\tau}\LRp{t,\xb_0,\mathbf{p}} \LRp{\boldsymbol{b}(\xb_0,\mathbf{p})-\boldsymbol{\bs{c}}(\xb_0, \mathbf{p},\boldsymbol{\alpha})}}_2\\
    &\leq  \nor{{\mathbf{F}}_f\LRp{\xb_0,\ \mathbf{p}}}_2\nor{\boldsymbol{h}(t,\xb_0,\mathbf{p})-\boldsymbol{\tau}}_2+\nor{{\mathbf{T}}_{\tau}\LRp{t,\xb_0,\mathbf{p}}}_2\nor{\boldsymbol{b}(\xb_0,\mathbf{p})-\boldsymbol{\bs{c}}(\xb_0, \mathbf{p},\boldsymbol{\alpha})}_2
\end{aligned}   
\label{last_two}
\end{equation}
where ${\mathbf{F}}_f\LRp{\xb_0,\ \mathbf{p}}$ is a  diagonal matrix whose diagonal elements are $\boldsymbol{b}_i(\xb_0,\mathbf{p})$ and ${\mathbf{T}}_{\tau}\LRp{t,\xb_0,\mathbf{p}}$ is a diagonal matrix whose diagonal elements are $\boldsymbol{\tau}_i\LRp{t,\xb_0,\mathbf{p};\ \boldsymbol{\nu}}$. From \cite[Theorem 3.2]{kidger2020universal},  $\forall \epsilon_3$ there exists a neural network  $\boldsymbol{\bs{c}}\LRp{ \xb_0, \, \mathbf{p}; \boldsymbol{\alpha}(\epsilon_3)} \in \sN^{\phi}_{n_x+n_p,m}$ such that,
\begin{equation}
\sup_{(t,\ \xb_0, \, \mathbf{p})\in E}\nor{\boldsymbol{b}(\xb_0,\mathbf{p})-\boldsymbol{\bs{c}}\LRp{ \xb_0, \, \mathbf{p}; \boldsymbol{\alpha}(\epsilon_3)} }_2\leq \epsilon_3,
\label{fourth_network}
\end{equation}
and $\forall \epsilon_4$  there exists a neural network  $\boldsymbol{\tau}\LRp{ t,\ \xb_0, \, \mathbf{p};\ \boldsymbol{\nu}(\epsilon_4)}\in \sN^{\phi}_{n_x+n_p+1,m}$ such that,
\begin{equation}
\sup_{(t,\ \xb_0, \, \mathbf{p})\in E}\nor{\boldsymbol{h}(t,\xb_0,\mathbf{p})-\boldsymbol{\tau}\LRp{ t,\ \xb_0, \, \mathbf{p};\ \boldsymbol{\nu}(\epsilon_4)}}_2\leq \epsilon_4.
\label{fifth_network}
\end{equation}
Plugging the results  \eqref{term_5}, \eqref{6_7}, \eqref{last_two} back in \eqref{finl_decom}, the error $\rho$ in \eqref{error} can now be bounded as follows:
\begin{equation}
\begin{aligned}
   &\rho= \nor{\bpsi\LRp{t,\ \xb_0, \mathbf{p}}-{\boldsymbol{\sD}}\LRp{\boldsymbol{\sE}(\xb_0, \mathbf{p};\ \boldsymbol{\beta}) +  \boldsymbol{\tau}\ \circ \ \boldsymbol{\bs{c}}(\xb_0, \mathbf{p};\ \boldsymbol{\alpha});\ \boldsymbol{\theta} }   }_2\\
   &\leq \nor{{\boldsymbol{l}^{-1}}\LRp{\gb(\xb_0,\mathbf{p})+\boldsymbol{h}(t,\xb_0,\mathbf{p})\ \circ \ \boldsymbol{b}(\xb_0,\mathbf{p})}-{\boldsymbol{\sD}}\LRp{\gb(\xb_0,\mathbf{p})+\boldsymbol{h}(t,\xb_0,\mathbf{p})\ \circ \ \boldsymbol{b}(\xb_0,\mathbf{p});\ \boldsymbol{\theta}}    }_2\\
    &+ L \LRp{\nor{\gb(\xb_0,\mathbf{p})-\boldsymbol{\sE}(\xb_0, \mathbf{p}; \boldsymbol{\beta})  }_2+\nor{{\mathbf{F}}_f\LRp{\xb_0,\ \mathbf{p}}}_2\nor{\boldsymbol{h}(t,\xb_0,\mathbf{p})-\boldsymbol{\tau}}_2+\nor{{\mathbf{T}}_{\tau}\LRp{t,\xb_0,\mathbf{p}}}_2\nor{\boldsymbol{b}(\xb_0,\mathbf{p})-\boldsymbol{\bs{c}}(\xb_0, \mathbf{p};\boldsymbol{\alpha})}_2}\\
    &\leq \nor{{\boldsymbol{l}^{-1}}\LRp{\gb(\xb_0,\mathbf{p})+\boldsymbol{h}(t,\xb_0,\mathbf{p})\ \circ \ \boldsymbol{b}(\xb_0,\mathbf{p})}-{\boldsymbol{\sD}}\LRp{\gb(\xb_0,\mathbf{p})+\boldsymbol{h}(t,\xb_0,\mathbf{p})\ \circ  \ \boldsymbol{b}(\xb_0,\mathbf{p});\ \boldsymbol{\theta}}    }_2\\
    & + L \sup_{(t,\ \xb_0, \, \mathbf{p})\in E} \LRp{\nor{\gb(\xb_0,\mathbf{p})-\boldsymbol{\sE}(\xb_0, \mathbf{p}; \boldsymbol{\beta})  }_2+\nor{{\mathbf{F}}_f\LRp{\xb_0,\ \mathbf{p}}}_2\nor{\boldsymbol{h}(t,\xb_0,\mathbf{p})-\boldsymbol{\tau}}_2}\\
    &+L \sup_{(t,\ \xb_0, \, \mathbf{p})\in E}\LRp{\nor{{\mathbf{T}}_{\tau}\LRp{t,\xb_0,\mathbf{p}}}_2\nor{\boldsymbol{b}(\xb_0,\mathbf{p})-\boldsymbol{\bs{c}}(\xb_0, \mathbf{p};\boldsymbol{\alpha})}_2}.
\end{aligned}    
\label{fi_l}
\end{equation}
Next, substituting \eqref{second_network},  \eqref{third_network}, \eqref{fourth_network} and \eqref{fifth_network} in \eqref{fi_l}, and by setting $\sD\LRp{.;\ \boldsymbol{\theta}}=\sD\LRp{.;\ \boldsymbol{\theta}(\epsilon_1)}$, $\boldsymbol{\tau}=\boldsymbol{\tau}\LRp{ t,\ \xb_0, \, \mathbf{p};\ \boldsymbol{\nu}(\epsilon_4)}$, $\boldsymbol{\bs{c}}(\xb_0, \mathbf{p};\boldsymbol{\alpha})=\boldsymbol{\bs{c}}(\xb_0, \mathbf{p};\boldsymbol{\alpha}(\epsilon_3))$, $\boldsymbol{\sE}\LRp{ \xb_0, \, \mathbf{p}\ ; \boldsymbol{\beta}}  =\boldsymbol{\sE}\LRp{ \xb_0, \, \mathbf{p}\ ; \boldsymbol{\beta}(\epsilon_2)} $,   we have the following estimate for $\rho$:
\[ \rho \leq \epsilon_1+L \LRp{\epsilon_2+\epsilon_4\times \sup_{(t,\ \xb_0, \, \mathbf{p})\in E} \nor{{\mathbf{F}}_f\LRp{\xb_0,\ \mathbf{p}}}_2+\epsilon_3 \times \sup_{(t,\ \xb_0, \, \mathbf{p})\in E}\nor{{\mathbf{T}}_{\tau}\LRp{t,\xb_0,\mathbf{p}}}_2}\]
Now, by Weierstrauss extreme value theorem \cite{rudin2021principles} we have $\sup_{( \xb_0, \, \mathbf{p})\in E} \nor{{\mathbf{F}}_f\LRp{\xb_0,\ \mathbf{p}}}_2\le \sqrt{m}\sup_{(\xb_0, \, \mathbf{p})\in E} \nor{\boldsymbol{b}(\xb_0,\mathbf{p})}_{\infty}=c_1$ and $\sup_{(t,\ \xb_0, \, \mathbf{p})\in E}\nor{{\mathbf{T}}_{\tau}\LRp{t,\xb_0,\mathbf{p}}}_2\le \sqrt{m}\sup_{(t,\ \xb_0, \, \mathbf{p})\in E}\nor{\boldsymbol{\tau}\LRp{t,\xb_0,\mathbf{p};\ \boldsymbol{\nu}(\epsilon_4)}}_\infty=c_2(\epsilon_4)$.
 Also note that the Lipschitz constant $L$ now depends on $\epsilon_1$ due to the choice of parameter $\boldsymbol{\theta}=\boldsymbol{\theta}(\epsilon_1)$ in \eqref{term_5} and we denote this dependence as $L(\epsilon_1)$.  Therefore, we have:
\[\rho \leq \epsilon_1+L(\epsilon_1) \times \LRp{\epsilon_2+\epsilon_4\times c_1+\epsilon_3 \times c_2(\epsilon_4)}. \]
Finally, setting $\epsilon_1=\frac{\epsilon}{4}$,  $\epsilon_2=\frac{\epsilon}{4L(\epsilon_1)}$,  $\epsilon_4=\frac{\epsilon}{4L(\epsilon_1) c_1}$, $\epsilon_3=\frac{\epsilon}{4L(\epsilon_1) c_2(\epsilon_4)}$ concludes the proof. 


\end{proof}

\begin{remark}[Independence of the latent dimension on the accuracy $\epsilon$]

\Cref{theo_im} shows that \lilans  can approximate the flow map, $\bpsi\LRp{t,\ \xb_0, \mathbf{p}}$, on a compact set to any degree of accuracy, $\epsilon$, as long as the dimension of the latent dynamics is greater than or equal to the 
dimension the original dynamics \eqref{original_dynamics}.
 Further, the latent dimension does not depend on the accuracy $\epsilon$.

\end{remark}

\begin{remark}[Relaxing assumption in \cref{lem}]
\label{relaxed_rem}
Note that the assumption of  $\bff\LRp{\xb(t),\ {\mathbf{p}}}$ to be globally Lipschitz, is necessary to guarantee the existence of solution on the entire real line, for any $(\xb_0,\ \mathbf{p})\in \sG_0\times \sX_p$. However, if it is known that the solution exists on some closed and bounded interval, $\tilde{I}=[0,\ \tilde{t}]$, for any $(\xb_0,\ \mathbf{p})\in \sG_0\times \sX_p$, then the requirement for $\bff\LRp{\xb(t),\ {\mathbf{p}}}$ to be globally Lipschitz can be relaxed. In such cases case, 
the set $E=I\times  \sG_0\times \sX_p$ in \cref{lemma_w} needs to be such that $I\subseteq \tilde{I}$. With these changes, the result in \cref{theo_im} holds. 

For example, consider the Robertson Stiff Chemical Kinetics Model in \cref{sec:Robertson}. In this case the solution  $\xb(t)$ of  remains bounded in $[0,\ 1]$ for any time $t\in \real^{+}$ and $\forall \mathbf{p}\in \sX_p$ (species concentration is always positive, and the sum of species concentration is 1 at any time). Based on the existence theorem in \cite{hirsch2013differential} (page 147), given that $\bff\LRp{\xb(t),\ {\mathbf{p}}}$ is  continuously differentiable with respect to both arguments and the solution does not leave the closed and bounded set $[0,\ 1]$, solution exists on $\real^{+}$ for any $\mathbf{p\in \sX_p}$. Therefore, in this case we do not need $\bff\LRp{\xb(t),\ {\mathbf{p}}}$ to be globally Lipschitz.

\end{remark}


\subsection{Architectural Variants}
\label{sec:variations}

This section presents a few variants of \lilans presented in \cref{method}.  There are several, simple modifications that can be made to create "stacked" variants of the architecture, similar to the DeepONet implementation in \cite{goswami2023learningstiffchemicalkinetics}. 
We tested the following four variants:
\begin{enumerate}
    \item {\em The "Full learning" approach}. This approach is exactly the same as the one described in \cref{method}. 
    In particular, we have three encoders, ($\boldsymbol{\sE},\ \boldsymbol{\bs{c}},\ \boldsymbol{\tau}$), and one decoder, $\boldsymbol{\sD}$. The description of each network is provided in \cref{method}. A schematic of the approach is provided in \cref{fig:varF}.
    \item {\em The "Independent learning" approach}. In this case, we have $n_x$ triples of encoders, $\LRp{\boldsymbol{\sE}_i,\ \boldsymbol{\bs{c}}_i,\ \boldsymbol{\tau}_i}$, where each triple encodes to a unique set of latent dynamics. Additionally, there are separate decoders, $\boldsymbol{\sD}_i$,
    for each $i^{th}$ latent dynamics, which recovers the associated $i^{th}$ component of the solution vector $\xb(t)$ in \eqref{original_dynamics}, denoted as $\xb_{i}(t)$. 
    A schematic of the approach is provided in \cref{fig:varI}.       
    \item   {\em The "Common encoder learning" approach}. In this case, we have three encoders, ($\boldsymbol{\sE},\ \boldsymbol{\bs{c}},\ \boldsymbol{\tau}$), as described in \cref{method}. However, we have $n_x$ separate decoders, $\boldsymbol{\sD}_i$, which decode the same latent dynamics to recover the associated $i^{th}$ component of the solution vector $\xb(t)$   in \eqref{original_dynamics}. A schematic of the approach is provided in \cref{fig:varCE}.  
    \item {\em The "Common decoder learning" approach}. In this case, we have $n_x$ triples of encoders, $\LRp{\boldsymbol{\sE}_i,\ \boldsymbol{\bs{c}}_i,\ \boldsymbol{\tau}_i}$, where each triple encodes to a unique set of latent dynamics. However, we have a single decoder, $\boldsymbol{\sD}$, which separately decodes each $i^{th}$ latent dynamics to recover the associated $i^{th}$ component of the solution vector $\xb(t)$   in \eqref{original_dynamics}. A schematic of the approach is provided in \cref{fig:varCD}.      
    \end{enumerate}
\Cref{fig:variants} shows the average point-wise relative error \eqref{eq:R1} (for a test data set), on the collisional-radiative charge state model in \cref{sec:CR-model}, achieved by each approach presented above. Each approach was tested with approximately the same number of trainable parameters and the same training data. Based on the numerical result in \cref{fig:variants}, the "independent learning" approach shows the best performance, for the given problem, illustrated by i) the error being the lowest on average for the full time series and ii) its error spike near $t=10^{-8}s$, the region where the dynamics evolve quickly, being lowest. The "common decoder" approach struggled, never outperforming any other method for any period of time. The "common encoder" and "full" approaches, on the other hand, had competitive performance with one another. Though they were not more accurate than "independent learning," except near $t=10^{-3}s$, where steady state is usually reached. However, the rapid increase of error at $t=10^{-8}s$ for the "common encoder" approach was uniquely large, leading to a two order of magnitude increase in error.  For all the ODE numerical results (sections \cref{sec:Robertson}, \cref{sec:CR-model}), we only present results for the "independent learning" approach, while for the PDE problems (sections \cref{sec:AC}, \cref{sec:CH}), we present results only for the "full" approach. Note that for PDE problems such as \eqref{sys_al}, a fine spatial discretization leads to a high-dimensional state space. As a result, the "independent learning" approach incurs a heavy computational cost during training, since using a separate network for each dimension results in a very large number of parameters (weights and biases). Hence, the results for the "independent learning" approach are not presented for PDE problems.

\begin{figure}[h!t!b!] 
  \centering
    \input{figs/arch_FULL_1}
  \caption{Schematic of the "full learning" approach.  
  }
  \label{fig:varF}
\end{figure}

\begin{figure}[h!t!b!] 
  \centering
    \input{figs/arch_IND_stack_2}
  \caption{Schematic of "independent learning" approach. 
  }
  \label{fig:varI}
\end{figure} 

\begin{figure}[h!t!b!] 
  \centering
    \input{figs/arch_CE_2}
  \caption{Schematic of "common encoder learning" approach: Note the single triple of encoder networks which map the initial conditions and parameters to the latent dynamics and $n_x$ decoder networks which map from the latent space to the predicted solution.   
  }
  \label{fig:varCE}
\end{figure} 

\begin{figure}[h!t!b!] 
  \centering
    \input{figs/arch_CD_2}
  \caption{Schematic of "common decoder learning" approach: Note the $n_x$ triples of encoder networks which map the initial conditions and parameters to the latent dynamics and single decoder network which maps from the latent space to the solution.  
  }
  \label{fig:varCD}
\end{figure}

\begin{figure}[h!t!b!]      
  \centering
  \input{figs/methods_compare}
  \caption{Average point-wise relative error \eqref{eq:R1}, achieved by different approaches presented in \cref{sec:variations}, for the full CR model (\cref{sec:CR-model}) on the test dataset. The "independent" approach maintains the lowest error on average, and, in particular, has the least error near $t=10^{-7}s$, in the fastest evolving temporal interval in the dynamics.}
  \label{fig:variants}
\end{figure}

%% file: figs/motivation_rel_errors.tex
\begin{tikzpicture}

\definecolor{darkgray176}{RGB}{176,176,176}
\definecolor{darkorange25512714}{RGB}{255,127,14}
\definecolor{lightgray204}{RGB}{204,204,204}
\definecolor{steelblue31119180}{RGB}{31,119,180}

\begin{axis}[
legend cell align={left},
legend style={
  fill opacity=0.8,
  draw opacity=1,
  text opacity=1,
  at={(1,0)},
  anchor=south east,
  draw=lightgray204
},
log basis x={10},
log basis y={10},
tick align=outside,
tick pos=left,
title={Relative Error on Test Dataset},
x grid style={darkgray176},
xlabel={Time (s)},
xmin=1e-16, xmax=1,
xmode=log,
xtick style={color=black},
y grid style={darkgray176},
ylabel={Error},
ymin=9.85104505966717e-06, ymax=1.77151826208149,
ymode=log,
ytick style={color=black}
]
\addplot [ultra thick, steelblue31119180]
table {%
1.09673133155882e-16 3.15820871037431e-05
1.20281961362278e-16 3.10439936583862e-05
1.31916995647357e-16 3.04477380268509e-05
1.44677502291565e-16 2.98916038445896e-05
1.58672349734831e-16 2.93771972792456e-05
1.74020937406248e-16 2.88327573798597e-05
1.90854214400669e-16 2.83691824733978e-05
2.09315796693257e-16 2.78924926533364e-05
2.2956319242369e-16 2.74343456112547e-05
2.51769145703728e-16 2.69929969363147e-05
2.76123110413076e-16 2.65633880189853e-05
3.02832866557495e-16 2.6139026886085e-05
3.32126292979374e-16 2.57197189057479e-05
3.64253311544964e-16 2.53170364885591e-05
3.99488019395418e-16 2.4924034732976e-05
4.38131027453331e-16 2.46000054175965e-05
4.80512025136123e-16 2.42593723669415e-05
5.26992593157566e-16 2.39969667745754e-05
5.77969288415332e-16 2.36957384913694e-05
6.33877027283849e-16 2.34070412261644e-05
6.95192796177559e-16 2.31427311518928e-05
7.62439721041914e-16 2.29325905820588e-05
8.36191530491633e-16 2.27113232540432e-05
9.17077450674293e-16 2.25050080189249e-05
1.00578757362058e-15 2.23153674596688e-05
1.10307874488222e-15 2.21524460357614e-05
1.2097810206889e-15 2.19785233639413e-05
1.32680474971472e-15 2.18525765376398e-05
1.45514833987319e-15 2.17179403989576e-05
1.59590677640473e-15 2.16153712244704e-05
1.7502809639301e-15 2.15187410503859e-05
1.91958797217311e-15 2.14471638173563e-05
2.1052722727657e-15 2.14042247534962e-05
2.30891806300419e-15 2.1371810362325e-05
2.5322627816988e-15 2.13389448617818e-05
2.77721193242936e-15 2.13284911296796e-05
3.04585534067428e-15 2.13268012885237e-05
3.34048498351324e-15 2.13000366784399e-05
3.66361454402072e-15 2.13410185097018e-05
4.01800085718209e-15 2.13562707358506e-05
4.40666743030178e-15 2.1419598851935e-05
4.83293023857175e-15 2.14860410778783e-05
5.30042601587968e-15 2.15452837437624e-05
5.81314328222471e-15 2.16174230445176e-05
6.37545637245649e-15 2.17225951928413e-05
6.99216275665938e-15 2.18285840674071e-05
7.66852397058703e-15 2.19478697545128e-05
8.4103105053526e-15 2.20587426156271e-05
9.22385103935846e-15 2.22008366108639e-05
1.01160864324958e-14 2.23335664486513e-05
1.10946289432752e-14 2.24581526708789e-05
1.21678271741092e-14 2.2593932953896e-05
1.33448372988384e-14 2.27465243369807e-05
1.46357011801908e-14 2.29023044084897e-05
1.60514320436476e-14 2.30700643442106e-05
1.76041084386555e-14 2.32316124311183e-05
1.93069772888325e-14 2.341440449527e-05
2.11745669103571e-14 2.35785082622897e-05
2.32228109627773e-14 2.37743261095602e-05
2.54691843897454e-14 2.39880755543709e-05
2.79328525094825e-14 2.41998004639754e-05
3.06348345269608e-14 2.43999002123019e-05
3.35981828628379e-14 2.45715673372615e-05
3.68481798291168e-14 2.47881234827219e-05
4.04125533295059e-14 2.5031944460352e-05
4.43217134247609e-14 2.52907921094447e-05
4.86090117813064e-14 2.55035956797656e-05
5.33110262166703e-14 2.57428346230881e-05
5.84678727693758e-14 2.5960096536437e-05
6.41235479557692e-14 2.62009652942652e-05
7.03263041338066e-14 2.64484242507024e-05
7.71290611762799e-14 2.66602091869572e-05
8.45898579657428e-14 2.6871561203734e-05
9.27723475631406e-14 2.70148357230937e-05
1.01746340274761e-13 2.71599110419629e-05
1.11588399250775e-13 2.71713724941947e-05
1.22382493696819e-13 2.71382796199759e-05
1.34220715271601e-13 2.69856573140714e-05
1.472040637826e-13 2.67471641564043e-05
1.6144330888316e-13 2.62809808191378e-05
1.77059935122689e-13 2.56843359238701e-05
1.94187178412825e-13 2.4866572857718e-05
2.12971162752348e-13 2.39566998061491e-05
2.33572146909012e-13 2.28505195991602e-05
2.56165891694572e-13 2.16900116356555e-05
2.80945159498141e-13 2.0507191948127e-05
3.08121358871401e-13 1.9414335838519e-05
3.37926348196743e-13 1.8494301912142e-05
3.70614413826622e-13 1.77648962562671e-05
4.06464439570963e-13 1.7245269191335e-05
4.45782286041972e-13 1.70762632478727e-05
4.88903400156144e-13 1.70741750480374e-05
5.36195677056881e-13 1.72300824488048e-05
5.88062598874676e-13 1.7509608369437e-05
6.44946677103763e-13 1.78720074472949e-05
7.07333227964444e-13 1.82552357728127e-05
7.7575451296124e-13 1.86243341886438e-05
8.50794279962744e-13 1.88714584510308e-05
9.33092743546168e-13 1.9106706531602e-05
1.02335204709726e-12 1.91821800399339e-05
1.12234225326641e-12 1.91520575754112e-05
1.2309079138896e-12 1.91362742043566e-05
1.34997527542643e-12 1.9004764908459e-05
1.48056018138991e-12 1.88832636922598e-05
1.62377673918872e-12 1.85874487215187e-05
1.78084682532468e-12 1.83528118213871e-05
1.95311051004064e-12 1.81987961695995e-05
2.14203749035838e-12 1.8171658666688e-05
2.34923962904965e-12 1.83026240847539e-05
2.57648470651837e-12 1.84588552656351e-05
2.82571150292083e-12 1.87297519005369e-05
3.09904633919943e-12 1.91418803296983e-05
3.39882121815266e-12 1.96259825315792e-05
3.72759372031494e-12 2.01497168745846e-05
4.0881688243913e-12 2.05005308089312e-05
4.48362283841191e-12 2.09854933928e-05
4.91732964577901e-12 2.15857307921397e-05
5.39298949012887e-12 2.20000983972568e-05
5.91466054459176e-12 2.23183760681422e-05
6.48679353478851e-12 2.26362553803483e-05
7.11426971095572e-12 2.30271307373187e-05
7.80244249316504e-12 2.3308739400818e-05
8.55718314494002e-12 2.35378047364065e-05
9.38493086494272e-12 2.3755226720823e-05
1.02927477240961e-11 2.39836208493216e-05
1.12883789168469e-11 2.41025300056208e-05
1.2380318840614e-11 2.42923306359444e-05
1.35778835671893e-11 2.43323520408012e-05
1.48912903243941e-11 2.4406959710177e-05
1.63317446661017e-11 2.4491484509781e-05
1.79115360743323e-11 2.45149658439914e-05
1.96441428090663e-11 2.44852126343176e-05
2.15443469003188e-11 2.45912196987774e-05
2.36283602635518e-11 2.45650844590273e-05
2.59139630143966e-11 2.4561553800595e-05
2.84206551627451e-11 2.46148847509176e-05
3.11698229804114e-11 2.46429681283189e-05
3.41849214617593e-11 2.47825100814225e-05
3.74916744339889e-11 2.4938932256191e-05
4.11182940243583e-11 2.51368637691485e-05
4.50957213567614e-11 2.55622107943054e-05
4.94578905312064e-11 2.61400036833948e-05
5.42420181383804e-11 2.69443444267381e-05
5.94889207793433e-11 2.80378208117327e-05
6.52433632993261e-11 2.94753372145351e-05
7.15544407066457e-11 3.12735719489865e-05
7.84759970351462e-11 3.34576616296545e-05
8.60670847237616e-11 3.61398851964623e-05
9.43924684324764e-11 3.90954883187078e-05
1.03523177593074e-10 4.25576654379256e-05
1.13537112408852e-10 4.64539843960665e-05
1.24519708473503e-10 5.08250159327872e-05
1.36564665679461e-10 5.55619226361159e-05
1.4977474763452e-10 6.07247275183909e-05
1.64262658407093e-10 6.63685787003487e-05
1.80152004080202e-10 7.24376805010252e-05
1.9757834731787e-10 7.89349578553811e-05
2.16690363941118e-10 8.59010615386069e-05
2.37651111381108e-10 9.34268900891766e-05
2.60639419831435e-10 0.00010144199768547
2.85851417968447e-10 0.000110045359178912
3.13502206256511e-10 0.000119228585390374
3.43827692114331e-10 0.00012909880024381
3.77086602599345e-10 0.00013962852244731
4.1356269178177e-10 0.000150975654833019
4.5356716164087e-10 0.000163198870723136
4.97441317137746e-10 0.000176329776877537
5.45559478116851e-10 0.000190559687325731
5.98332172879627e-10 0.000206045413506217
6.56209640676755e-10 0.000223166338400915
7.19685673001153e-10 0.000242362642893568
7.89301826454357e-10 0.000264161615632474
8.65652043129092e-10 0.000289028888801113
9.49387717927582e-10 0.000317704601911828
1.04122325604831e-09 0.000351099617546424
1.14194216805586e-09 0.000390443805372342
1.25240375453507e-09 0.00043658321374096
1.37355043736051e-09 0.000490788777824491
1.50641580012959e-09 0.000554135476704687
1.65213340635737e-09 0.000627916073426604
1.81194647066712e-09 0.000713845016434789
1.98721846548805e-09 0.000814126920886338
2.17944475375299e-09 0.000931252841837704
2.39026534684239e-09 0.00106791488360614
2.62147889662135e-09 0.00122789142187685
2.87505804094488e-09 0.00141516723670065
3.15316623355437e-09 0.00163421430625021
3.45817620195238e-09 0.0018905628239736
3.79269019073224e-09 0.00219075451605022
4.15956216307184e-09 0.0025426943320781
4.56192214980747e-09 0.00295597827062011
5.003202953826e-09 0.00344208278693259
5.48716943760859e-09 0.00401532836258411
6.01795064379733e-09 0.00469278963282704
6.6000750228271e-09 0.0054959156550467
7.23850906817325e-09 0.006452277302742
7.93869968883821e-09 0.00759767647832632
8.70662068058512e-09 0.00897664483636618
9.54882369239568e-09 0.0106453541666269
1.04724941229815e-08 0.0126766907051206
1.14855124242393e-08 0.0151619594544172
1.25965213346714e-08 0.0182091519236565
1.38149996163832e-08 0.021930618211627
1.51513429247605e-08 0.0264076814055443
1.66169525007768e-08 0.031597550958395
1.82243324426266e-08 0.0371334701776505
1.99871963865725e-08 0.04196497797966
2.19205845071732e-08 0.0440999493002892
2.40409918350996e-08 0.0423492528498173
2.63665089873036e-08 0.0417103581130505
2.8916976510203e-08 0.0481493808329105
3.171415415269e-08 0.0611492283642292
3.47819065131412e-08 0.0720182806253433
3.81464066443117e-08 0.0746234208345413
4.18363593532002e-08 0.0690167099237442
4.58832461010084e-08 0.0623771287500858
5.03215935925998e-08 0.0585776306688786
5.51892683469737e-08 0.0575045645236969
6.05277997619336e-08 0.0559521242976189
6.63827344292308e-08 0.0559639446437359
7.28040247230855e-08 0.0532440841197968
7.98464549773908e-08 0.0522648543119431
8.75701088876052e-08 0.0521911419928074
9.60408821250537e-08 0.0514758042991161
1.05331044537093e-07 0.0519404336810112
1.15519856729648e-07 0.0518623217940331
1.26694246292591e-07 0.0505538545548916
1.38949549437314e-07 0.0497408770024776
1.52390324373882e-07 0.0486968196928501
1.67131243367249e-07 0.0471349805593491
1.83298071083244e-07 0.0460622683167458
2.01028737571288e-07 0.0443795807659626
2.20474515038147e-07 0.0434581264853477
2.41801308452572e-07 0.0393419899046421
2.65191070991854e-07 0.0363948158919811
2.90843356406404e-07 0.0356920994818211
3.18977021546631e-07 0.0323785804212093
3.49832093577503e-07 0.0277203060686588
3.83671817811265e-07 0.0242135301232338
4.2078490362974e-07 0.0222091674804688
4.61487987657692e-07 0.0194013752043247
5.06128335202221e-07 0.0186189115047455
5.55086803005981e-07 0.019156564027071
6.08781088591476e-07 0.0194654203951359
6.67669293918754e-07 0.020978594198823
7.32253833760453e-07 0.0234153717756271
8.03085722139152e-07 0.0262921676039696
8.80769273397545e-07 0.029324883595109
9.6596725800938e-07 0.0322069190442562
1.05940655711885e-06 0.0356640629470348
1.1618843640511e-06 0.0395436137914658
1.27427498570313e-06 0.0436905696988106
1.39753730184229e-06 0.0480465553700924
1.53272294595261e-06 0.0524374507367611
1.68098527742537e-06 0.0566563047468662
1.84358922164149e-06 0.0604816526174545
2.02192206189835e-06 0.0636458918452263
2.21750527525393e-06 0.0659881606698036
2.43200751326795e-06 0.0674224495887756
2.6672588383874e-06 0.0679723620414734
2.92526633743664e-06 0.0677721276879311
3.20823124542107e-06 0.0670038238167763
3.51856772573927e-06 0.0658484846353531
3.8589234670299e-06 0.0644423440098763
4.23220227237926e-06 0.0628721341490746
4.64158883361277e-06 0.0611853078007698
5.09057590203669e-06 0.0594304911792278
5.58299408744191e-06 0.0575720556080341
6.12304453960516e-06 0.0555830970406532
6.71533479111514e-06 0.0534726046025753
7.36491806732297e-06 0.0512414425611496
8.0773363987967e-06 0.0488105267286301
8.8586679041008e-06 0.0463224537670612
9.71557864630185e-06 0.0439042299985886
1.06553795056231e-05 0.0417992174625397
1.16860885534665e-05 0.0399054288864136
1.28164994599575e-05 0.038221962749958
1.40562565186421e-05 0.0366381108760834
1.54159369284227e-05 0.0351518467068672
1.69071410347358e-05 0.033795453608036
1.85425912998785e-05 0.03262098133564
2.03362408468667e-05 0.0315694771707058
2.2303392502885e-05 0.03061961568892
2.4460829357968e-05 0.0295935422182083
2.68269579527972e-05 0.0286933798342943
2.94219653172437e-05 0.0279560051858425
3.22679911994581e-05 0.0274596121162176
3.53893169549098e-05 0.0270348507910967
3.88125727069152e-05 0.0265140682458878
4.25669645460787e-05 0.025933688506484
4.66845237070379e-05 0.0253655146807432
5.12003798484088e-05 0.0247939266264439
5.61530607674626e-05 0.0243142396211624
6.15848211066026e-05 0.0238758288323879
6.75420028560559e-05 0.0234855860471725
7.40754307284716e-05 0.0231951158493757
8.12408457786294e-05 0.0229710228741169
8.9099380967761e-05 0.022739851847291
9.77180827298391e-05 0.0223994310945272
0.000107170482989671 0.0219654217362404
0.000117537226513063 0.0214254725724459
0.000128906758941402 0.0207354798913002
0.000141376081380736 0.0199677143245935
0.000155051577983262 0.0190465580672026
0.000170049923581879 0.018054474145174
0.000186499079121429 0.0170645136386156
0.000204539383379339 0.0161520503461361
0.000224324750289842 0.0153808603063226
0.000246023982086977 0.014739416539669
0.000269822209469653 0.0141493985429406
0.000295922471075796 0.0135749066248536
0.000324547445741133 0.0130979577079415
0.000355941352321685 0.0126865403726697
0.000390372033288608 0.0123626431450248
0.00042813323987194 0.0121826408430934
0.000469547138249342 0.0121106207370758
0.000514967058161832 0.0120767718181014
0.000564780507406755 0.0121627617627382
0.000619412477926676 0.0123211499303579
0.000679329071700669 0.0124909887090325
0.000745041477372889 0.0127265099436045
0.000817110331545719 0.0130078867077827
0.000896150501946605 0.0133596295490861
0.000982836333277001 0.0136933894827962
0.00107790740049927 0.0140213835984468
0.00118217481864667 0.01433549169451
0.00129652816298967 0.0146381435915828
0.00142194305859916 0.0149343712255359
0.00155948950405828 0.0152160888537765
0.00171034100033784 0.015492370352149
0.00187578456272016 0.015756206586957
0.00205723170118955 0.0159995257854462
0.00225623046297063 0.0162347555160522
0.00247447863995735 0.0164357740432024
0.00271383825371428 0.016604408621788
0.00297635144163131 0.0167393330484629
0.00326425787976731 0.0168260484933853
0.00358001389102857 0.0168897081166506
0.00392631340170684 0.0169249400496483
0.00430611092517116 0.0169584099203348
0.00472264676880293 0.0169855188578367
0.0051794746792312 0.0170077942311764
0.00568049216172843 0.0170307476073503
0.00622997373244183 0.0170624200254679
0.00683260738715738 0.0171011444181204
0.00749353459773574 0.0171419698745012
0.00821839417745681 0.0171879641711712
0.00901337038951742 0.0172427631914616
0.00988524570912824 0.017309982329607
0.0108414586893583 0.0173941273242235
0.0118901674244199 0.0174993760883808
0.0130403191518413 0.0176309254020452
0.0143017265873508 0.0177955776453018
0.0156851516437354 0.0180056914687157
0.017202397247936 0.0182546488940716
0.0188664080397325 0.0185507796704769
0.0206913808111478 0.0189063847064972
0.0226928856288008 0.0193393528461456
0.0248879986725867 0.0198285784572363
0.02729544792402 0.0203692242503166
0.0299357729472048 0.0209615956991911
0.0328315001256304 0.0216076988726854
0.0360073348498562 0.0223100390285254
0.0394903722957669 0.0230710227042437
0.0433103285916898 0.0238919984549284
0.047499794346614 0.0247757229954004
0.0520945127025321 0.0257226824760437
0.0571336842831557 0.0267326831817627
0.0626603016407263 0.0278053637593985
0.0687215160543111 0.0289386156946421
0.0753690398089854 0.0301292557269335
0.082659587388018 0.0313725583255291
0.0906553593421632 0.0326610393822193
0.099424572964274 0.0339863561093807
0.109042044296775 0.0353374257683754
0.119589826437498 0.0367030911147594
0.131157909589684 0.0380707047879696
0.143844988828766 0.0394275411963463
0.157759306136236 0.0407617092132568
0.173019573884589 0.0420630760490894
0.189755987652184 0.0433211736381054
0.208111337009039 0.044525932520628
0.22824222375041 0.045669462531805
0.250320397971732 0.046745877712965
0.27453422338387 0.0477497018873692
0.301090284370259 0.048677321523428
0.330215148496818 0.0495250672101974
0.362157299511807 0.0502917729318142
0.397189257327329 0.0509795434772968
0.43560990306946 0.0515910014510155
0.477747029033577 0.0521306023001671
0.523960135300263 0.0525999516248703
0.574643496871595 0.0530034489929676
0.630229527495601 0.0533471927046776
0.691192468877937 0.0536379851400852
0.758052436755925 0.0538858883082867
0.83137985835473 0.0540928803384304
0.911800339084566 0.0542650148272514
1 0.0544041320681572
};
\addlegendentry{\lilan}
\addplot [ultra thick, darkorange25512714]
table {%
1.09673133155882e-16 0.000150910927914083
1.20281961362278e-16 0.000150043240864761
1.31916995647357e-16 0.000149213345139287
1.44677502291565e-16 0.000148416234878823
1.58672349734831e-16 0.000147616447065957
1.74020937406248e-16 0.000146840466186404
1.90854214400669e-16 0.000146076534292661
2.09315796693257e-16 0.000145347701618448
2.2956319242369e-16 0.00014461092359852
2.51769145703728e-16 0.000143883575219661
2.76123110413076e-16 0.000143170851515606
3.02832866557495e-16 0.00014244629710447
3.32126292979374e-16 0.000141754760988988
3.64253311544964e-16 0.00014106735761743
3.99488019395418e-16 0.000140447227749974
4.38131027453331e-16 0.000139827956445515
4.80512025136123e-16 0.000139223455335014
5.26992593157566e-16 0.000138615985633805
5.77969288415332e-16 0.000138011993840337
6.33877027283849e-16 0.000137445284053683
6.95192796177559e-16 0.00013688464241568
7.62439721041914e-16 0.000136368340463378
8.36191530491633e-16 0.000135833586682566
9.17077450674293e-16 0.00013536965707317
1.00578757362058e-15 0.000134873320348561
1.10307874488222e-15 0.000134433954372071
1.2097810206889e-15 0.00013396107533481
1.32680474971472e-15 0.000133548775920644
1.45514833987319e-15 0.000133233406813815
1.59590677640473e-15 0.000132864399347454
1.7502809639301e-15 0.000132538087200373
1.91958797217311e-15 0.000132260684040375
2.1052722727657e-15 0.000131967186462134
2.30891806300419e-15 0.000131719541968778
2.5322627816988e-15 0.00013144631520845
2.77721193242936e-15 0.000131253473227844
3.04585534067428e-15 0.000131039632833563
3.34048498351324e-15 0.000130844666273333
3.66361454402072e-15 0.000130699278088287
4.01800085718209e-15 0.00013062241487205
4.40666743030178e-15 0.000130540123791434
4.83293023857175e-15 0.00013045959349256
5.30042601587968e-15 0.000130435219034553
5.81314328222471e-15 0.000130426895339042
6.37545637245649e-15 0.000130400323541835
6.99216275665938e-15 0.000130417945911177
7.66852397058703e-15 0.000130488537251949
8.4103105053526e-15 0.000130550470203161
9.22385103935846e-15 0.000130679298308678
1.01160864324958e-14 0.000130856584291905
1.10946289432752e-14 0.000131028791656718
1.21678271741092e-14 0.000131254317238927
1.33448372988384e-14 0.000131520631839521
1.46357011801908e-14 0.000131814900669269
1.60514320436476e-14 0.000132159068016335
1.76041084386555e-14 0.000132561312057078
1.93069772888325e-14 0.000132993751321919
2.11745669103571e-14 0.000133498004288413
2.32228109627773e-14 0.000134079076815397
2.54691843897454e-14 0.000134681846247986
2.79328525094825e-14 0.000135380119900219
3.06348345269608e-14 0.000136129441671073
3.35981828628379e-14 0.000136975591885857
3.68481798291168e-14 0.000137954164529219
4.04125533295059e-14 0.000139063849928789
4.43217134247609e-14 0.000140243239002302
4.86090117813064e-14 0.000141595955938101
5.33110262166703e-14 0.000143135068356059
5.84678727693758e-14 0.000144881123560481
6.41235479557692e-14 0.000146883437992074
7.03263041338066e-14 0.000149197352584451
7.71290611762799e-14 0.000151875006849878
8.45898579657428e-14 0.000155004381667823
9.27723475631406e-14 0.000158700466272421
1.01746340274761e-13 0.000163111646543257
1.11588399250775e-13 0.000168377562658861
1.22382493696819e-13 0.000174870961927809
1.34220715271601e-13 0.000182856150786392
1.472040637826e-13 0.000192709994735196
1.6144330888316e-13 0.000205010452191345
1.77059935122689e-13 0.000220454880036414
1.94187178412825e-13 0.000240029097767547
2.12971162752348e-13 0.000264771253569052
2.33572146909012e-13 0.000295890000415966
2.56165891694572e-13 0.000334845826728269
2.80945159498141e-13 0.000382870814064518
3.08121358871401e-13 0.000440954783698544
3.37926348196743e-13 0.000509509758558124
3.70614413826622e-13 0.000587792601436377
4.06464439570963e-13 0.000673982198350132
4.45782286041972e-13 0.00076520413858816
4.88903400156144e-13 0.000857689068652689
5.36195677056881e-13 0.000947522348724306
5.88062598874676e-13 0.00103142624720931
6.44946677103763e-13 0.00110693532042205
7.07333227964444e-13 0.00117287074681371
7.7575451296124e-13 0.0012291488237679
8.50794279962744e-13 0.00127630319911987
9.33092743546168e-13 0.00131553679239005
1.02335204709726e-12 0.00134798290673643
1.12234225326641e-12 0.00137470115441829
1.2309079138896e-12 0.0013968147104606
1.34997527542643e-12 0.00141502951737493
1.48056018138991e-12 0.00143009447492659
1.62377673918872e-12 0.00144251855090261
1.78084682532468e-12 0.00145268184132874
1.95311051004064e-12 0.00146101717837155
2.14203749035838e-12 0.00146780640352517
2.34923962904965e-12 0.00147336244117469
2.57648470651837e-12 0.00147783756256104
2.82571150292083e-12 0.00148147461004555
3.09904633919943e-12 0.00148447533138096
3.39882121815266e-12 0.00148710515350103
3.72759372031494e-12 0.00148927292320877
4.0881688243913e-12 0.001491145696491
4.48362283841191e-12 0.00149301392957568
4.91732964577901e-12 0.00149483664426953
5.39298949012887e-12 0.00149675516877323
5.91466054459176e-12 0.00149877066724002
6.48679353478851e-12 0.00150088884402066
7.11426971095572e-12 0.00150344462599605
7.80244249316504e-12 0.00150626781396568
8.55718314494002e-12 0.00150948402006179
9.38493086494272e-12 0.00151308404747397
1.02927477240961e-11 0.00151719397399575
1.12883789168469e-11 0.00152171240188181
1.2380318840614e-11 0.00152687763329595
1.35778835671893e-11 0.00153264158871025
1.48912903243941e-11 0.0015390410553664
1.63317446661017e-11 0.00154614122584462
1.79115360743323e-11 0.00155403872486204
1.96441428090663e-11 0.00156281853560358
2.15443469003188e-11 0.00157281139399856
2.36283602635518e-11 0.00158399099018425
2.59139630143966e-11 0.00159651215653867
2.84206551627451e-11 0.00161071424372494
3.11698229804114e-11 0.00162663857918233
3.41849214617593e-11 0.00164447096176445
3.74916744339889e-11 0.00166398868896067
4.11182940243583e-11 0.00168524752371013
4.50957213567614e-11 0.00170787621755153
4.94578905312064e-11 0.00173160620033741
5.42420181383804e-11 0.0017560813575983
5.94889207793433e-11 0.00178096140734851
6.52433632993261e-11 0.00180599757004529
7.15544407066457e-11 0.00183094129897654
7.84759970351462e-11 0.00185556011274457
8.60670847237616e-11 0.00187981373164803
9.43924684324764e-11 0.00190341507550329
1.03523177593074e-10 0.00192654016427696
1.13537112408852e-10 0.00194914999883622
1.24519708473503e-10 0.00197116238996387
1.36564665679461e-10 0.00199249805882573
1.4977474763452e-10 0.00201307376846671
1.64262658407093e-10 0.0020322254858911
1.80152004080202e-10 0.00204741908237338
1.9757834731787e-10 0.00205045007169247
2.16690363941118e-10 0.00202116882428527
2.37651111381108e-10 0.0019395430572331
2.60639419831435e-10 0.00181638333015144
2.85851417968447e-10 0.00169591174926609
3.13502206256511e-10 0.00161793758161366
3.43827692114331e-10 0.00158408761490136
3.77086602599345e-10 0.00157988304272294
4.1356269178177e-10 0.00159496255218983
4.5356716164087e-10 0.00162569235544652
4.97441317137746e-10 0.00167492346372455
5.45559478116851e-10 0.001756597077474
5.98332172879627e-10 0.00187711464241147
6.56209640676755e-10 0.00203277892433107
7.19685673001153e-10 0.00222387607209384
7.89301826454357e-10 0.00245177652686834
8.65652043129092e-10 0.00271760066971183
9.49387717927582e-10 0.00302380975335836
1.04122325604831e-09 0.00337305222637951
1.14194216805586e-09 0.0037659858353436
1.25240375453507e-09 0.00419982057064772
1.37355043736051e-09 0.00467023067176342
1.50641580012959e-09 0.00517004216089845
1.65213340635737e-09 0.00568542489781976
1.81194647066712e-09 0.00618060864508152
1.98721846548805e-09 0.00658025592565536
2.17944475375299e-09 0.00691891089081764
2.39026534684239e-09 0.00729075167328119
2.62147889662135e-09 0.00765281263738871
2.87505804094488e-09 0.00795859005302191
3.15316623355437e-09 0.0081876777112484
3.45817620195238e-09 0.00833945907652378
3.79269019073224e-09 0.00844325870275497
4.15956216307184e-09 0.00858390610665083
4.56192214980747e-09 0.00888563320040703
5.003202953826e-09 0.00944881327450275
5.48716943760859e-09 0.0103257866576314
6.01795064379733e-09 0.011551127769053
6.6000750228271e-09 0.0131673365831375
7.23850906817325e-09 0.0152252605184913
7.93869968883821e-09 0.0177991352975368
8.70662068058512e-09 0.0209885947406292
9.54882369239568e-09 0.0248820073902607
1.04724941229815e-08 0.029596971347928
1.14855124242393e-08 0.0352786779403687
1.25965213346714e-08 0.0422567762434483
1.38149996163832e-08 0.0510791130363941
1.51513429247605e-08 0.0622238032519817
1.66169525007768e-08 0.0763204321265221
1.82243324426266e-08 0.0941807627677917
1.99871963865725e-08 0.116731807589531
2.19205845071732e-08 0.14481483399868
2.40409918350996e-08 0.178680762648582
2.63665089873036e-08 0.215934902429581
2.8916976510203e-08 0.247654736042023
3.171415415269e-08 0.266585320234299
3.47819065131412e-08 0.277540594339371
3.81464066443117e-08 0.284025251865387
4.18363593532002e-08 0.291593641042709
4.58832461010084e-08 0.301233381032944
5.03215935925998e-08 0.311712265014648
5.51892683469737e-08 0.324217110872269
6.05277997619336e-08 0.341522872447968
6.63827344292308e-08 0.36507385969162
7.28040247230855e-08 0.392265945672989
7.98464549773908e-08 0.424319177865982
8.75701088876052e-08 0.459659069776535
9.60408821250537e-08 0.497711032629013
1.05331044537093e-07 0.534437119960785
1.15519856729648e-07 0.568914949893951
1.26694246292591e-07 0.603348851203918
1.38949549437314e-07 0.637966513633728
1.52390324373882e-07 0.67642205953598
1.67131243367249e-07 0.718699276447296
1.83298071083244e-07 0.758776962757111
2.01028737571288e-07 0.793781399726868
2.20474515038147e-07 0.832733690738678
2.41801308452572e-07 0.873442471027374
2.65191070991854e-07 0.904076337814331
2.90843356406404e-07 0.936331033706665
3.18977021546631e-07 0.966639697551727
3.49832093577503e-07 0.982735395431519
3.83671817811265e-07 0.988418161869049
4.2078490362974e-07 1.00101232528687
4.61487987657692e-07 1.01525950431824
5.06128335202221e-07 1.02185356616974
5.55086803005981e-07 1.02208781242371
6.08781088591476e-07 1.01721012592316
6.67669293918754e-07 1.00957560539246
7.32253833760453e-07 1.00110292434692
8.03085722139152e-07 0.992368042469025
8.80769273397545e-07 0.983836472034454
9.6596725800938e-07 0.975637853145599
1.05940655711885e-06 0.967885732650757
1.1618843640511e-06 0.960917353630066
1.27427498570313e-06 0.954669535160065
1.39753730184229e-06 0.948937356472015
1.53272294595261e-06 0.943393111228943
1.68098527742537e-06 0.937636137008667
1.84358922164149e-06 0.931245803833008
2.02192206189835e-06 0.923825263977051
2.21750527525393e-06 0.915020704269409
2.43200751326795e-06 0.904524564743042
2.6672588383874e-06 0.892091572284698
2.92526633743664e-06 0.877557218074799
3.20823124542107e-06 0.860872268676758
3.51856772573927e-06 0.842118144035339
3.8589234670299e-06 0.821504950523376
4.23220227237926e-06 0.799341022968292
4.64158883361277e-06 0.775991678237915
5.09057590203669e-06 0.751833915710449
5.58299408744191e-06 0.727215170860291
6.12304453960516e-06 0.702422857284546
6.71533479111514e-06 0.677679300308228
7.36491806732297e-06 0.653154134750366
8.0773363987967e-06 0.628970265388489
8.8586679041008e-06 0.605187714099884
9.71557864630185e-06 0.581842541694641
1.06553795056231e-05 0.558969616889954
1.16860885534665e-05 0.536591291427612
1.28164994599575e-05 0.51474392414093
1.40562565186421e-05 0.493458926677704
1.54159369284227e-05 0.472764045000076
1.69071410347358e-05 0.45263084769249
1.85425912998785e-05 0.433044046163559
2.03362408468667e-05 0.413946241140366
2.2303392502885e-05 0.395269900560379
2.4460829357968e-05 0.377051889896393
2.68269579527972e-05 0.359359622001648
2.94219653172437e-05 0.342311322689056
3.22679911994581e-05 0.325922191143036
3.53893169549098e-05 0.310371339321136
3.88125727069152e-05 0.295628011226654
4.25669645460787e-05 0.281643003225327
4.66845237070379e-05 0.268456399440765
5.12003798484088e-05 0.256059199571609
5.61530607674626e-05 0.244421303272247
6.15848211066026e-05 0.233440279960632
6.75420028560559e-05 0.223064497113228
7.40754307284716e-05 0.213169917464256
8.12408457786294e-05 0.20367731153965
8.9099380967761e-05 0.194445252418518
9.77180827298391e-05 0.185427024960518
0.000107170482989671 0.176670864224434
0.000117537226513063 0.168188005685806
0.000128906758941402 0.159938246011734
0.000141376081380736 0.151952609419823
0.000155051577983262 0.144292950630188
0.000170049923581879 0.136939406394958
0.000186499079121429 0.129925757646561
0.000204539383379339 0.123302146792412
0.000224324750289842 0.11716278642416
0.000246023982086977 0.111424535512924
0.000269822209469653 0.106016501784325
0.000295922471075796 0.100885286927223
0.000324547445741133 0.0959863439202309
0.000355941352321685 0.091311402618885
0.000390372033288608 0.086864560842514
0.00042813323987194 0.0826418474316597
0.000469547138249342 0.078671395778656
0.000514967058161832 0.0750387832522392
0.000564780507406755 0.071882389485836
0.000619412477926676 0.069305881857872
0.000679329071700669 0.0672386288642883
0.000745041477372889 0.0656015500426292
0.000817110331545719 0.0643022134900093
0.000896150501946605 0.0633503496646881
0.000982836333277001 0.0627677515149117
0.00107790740049927 0.0625130757689476
0.00118217481864667 0.0626129806041718
0.00129652816298967 0.0630026608705521
0.00142194305859916 0.0636170655488968
0.00155948950405828 0.0644124150276184
0.00171034100033784 0.0653323605656624
0.00187578456272016 0.0663365125656128
0.00205723170118955 0.0673953965306282
0.00225623046297063 0.0684713870286942
0.00247447863995735 0.0695393234491348
0.00271383825371428 0.0705837234854698
0.00297635144163131 0.0715953782200813
0.00326425787976731 0.0725666731595993
0.00358001389102857 0.0734899416565895
0.00392631340170684 0.074360728263855
0.00430611092517116 0.0751753002405167
0.00472264676880293 0.0759309902787209
0.0051794746792312 0.0766256302595139
0.00568049216172843 0.0772586613893509
0.00622997373244183 0.0778300687670708
0.00683260738715738 0.0783416926860809
0.00749353459773574 0.0787955820560455
0.00821839417745681 0.0791952461004257
0.00901337038951742 0.0795441493391991
0.00988524570912824 0.0798466131091118
0.0108414586893583 0.0801066383719444
0.0118901674244199 0.0803285986185074
0.0130403191518413 0.0805160403251648
0.0143017265873508 0.0806721523404121
0.0156851516437354 0.0807995274662971
0.017202397247936 0.0809004828333855
0.0188664080397325 0.0809770971536636
0.0206913808111478 0.0810305252671242
0.0226928856288008 0.0810624212026596
0.0248879986725867 0.0810739099979401
0.02729544792402 0.0810669437050819
0.0299357729472048 0.0810429751873016
0.0328315001256304 0.0810041725635529
0.0360073348498562 0.080952525138855
0.0394903722957669 0.0808896943926811
0.0433103285916898 0.080816887319088
0.047499794346614 0.0807356908917427
0.0520945127025321 0.0806480571627617
0.0571336842831557 0.0805557444691658
0.0626603016407263 0.0804596990346909
0.0687215160543111 0.0803608223795891
0.0753690398089854 0.0802591443061829
0.082659587388018 0.0801557824015617
0.0906553593421632 0.080051489174366
0.099424572964274 0.0799470618367195
0.109042044296775 0.0798433944582939
0.119589826437498 0.0797406509518623
0.131157909589684 0.0796390175819397
0.143844988828766 0.0795384868979454
0.157759306136236 0.0794410482048988
0.173019573884589 0.0793468505144119
0.189755987652184 0.0792561918497086
0.208111337009039 0.0791687369346619
0.22824222375041 0.0790851563215256
0.250320397971732 0.079005628824234
0.27453422338387 0.078931912779808
0.301090284370259 0.0788635015487671
0.330215148496818 0.0788007900118828
0.362157299511807 0.0787435844540596
0.397189257327329 0.0786927118897438
0.43560990306946 0.078648529946804
0.477747029033577 0.0786112546920776
0.523960135300263 0.0785808116197586
0.574643496871595 0.0785582736134529
0.630229527495601 0.0785440057516098
0.691192468877937 0.0785371214151382
0.758052436755925 0.0785384476184845
0.83137985835473 0.0785480961203575
0.911800339084566 0.0785668045282364
1 0.0785945653915405
};
\addlegendentry{Direct Learning}
\end{axis}

\end{tikzpicture}

%% file: figs/arch_FULL_1.tex
\begin{tikzpicture}[scale=.8, transform shape, node distance=1cm, >=Latex]
  \node[place](W){};
  \node(Z)[verybigbox, right=1.2cm of W]{};
  \node(A)[smallbox]{$\xb_0$, $\mathbf{p}, t$};
  \node[place](X)[right=2.0cm of A]{};
  \node(D)[smallboxtime, right=of A]{Encoder $\boldsymbol{\tau}$};
  \node(C)[smallboxinit, above=2.5mm of D]{Encoder $\boldsymbol{\sE}$};
  \node(E)[smallboxdyn, below=2.5mm of D]{Encoder $\bs{c}$};
  \node(H)[smallboxinit, right=of C]{$\mathbf{y}_0$};
  \node(HH)[smallboxtime, right=0.9cm of D]{$\boldsymbol{\tau}$};
  \node(I)[smallboxdyn, right=0.95cm of E]{$\frac{d\mathbf{y}}{d\boldsymbol{\tau}}$};
  \node[place](Y)[right=6.0cm of A]{};
  \node(J)[bigbox, right=of Y]{$\mathbf{y}({\boldsymbol{\tau}}) = \mathbf{y}_0 + \boldsymbol{\tau}
  \circ \frac{d\mathbf{y}}{d\boldsymbol{\tau}}$};
  \node(K)[smallboxdec, right=of J]{Decoder $\boldsymbol{\sD}$};
  \node(L)[smallbox, right=of K]{ $\xb(t)$};
  \draw[fork](A.east)--(C.west);
  \draw[fork](A.east)--(E.west);
  \draw[->](A)--(D);
  \draw[->](C)--(H);
  \draw[->](D)--(HH);
  \draw[->](E)--(I);
  \draw[->](HH)--(J);
  \draw[fork](H.east)--(J.west);
  \draw[fork](I.east)--(J.west);
  \draw[->](J)--(K);
  \draw[->](K)--(L);
\end{tikzpicture}

%% file: figs/arch_IND_stack_2.tex
\begin{tikzpicture}[scale=.8, transform shape, node distance=1cm, >=Latex]
  \node[place](W){};
  \node(Z)[verybigbox2, right=1.2cm of W]{};
  \node(A)[smallbox]{$\xb_0$, $\mathbf{p}, t$};
  \node[place](X)[right=2.0cm of A]{};
  \node(D)[smallboxtimestacked, right=of A]{Encoder $\boldsymbol{\tau}_i$};
  \node(C)[smallboxinitstacked, above=3.5mm of D]{Encoder $\boldsymbol{\sE}_i$};
  \node(E)[smallboxdynstacked, below=3.5mm of D]{Encoder $\bs{c}_i$};
  \node(H)[smallboxinitstacked, right=of C]{$[\mathbf{y}_0]_i$};
  \node(HH)[smallboxtimestacked, right=0.9cm of D]{$\boldsymbol{\tau}_i$};
  \node(I)[smallboxdynstacked, right=0.95cm of E]{$\frac{d\mathbf{y}_i}{d\boldsymbol{\tau}}$};
  \node[place](Y)[right=6.0cm of A]{};
  \node(J)[bigboxstacked, right=of Y]{$\mathbf{y}_i({\boldsymbol{\tau}}) = [\mathbf{y}_0]_i + \boldsymbol{\tau}_i
  \circ \frac{d\mathbf{y}_i}{d\boldsymbol{\tau}}$};
  \node(K)[smallboxdecstacked, right=of J]{Decoder $\boldsymbol{\sD}_i$};
  \node(L)[smallboxstacked, right=of K]{ $\xb_i(t)$};
  \draw[fork](A.east)--(C.west);
  \draw[fork](A.east)--(E.west);
  \draw[->](A)--(D);
  \draw[->](C)--(H);
  \draw[->](D)--(HH);
  \draw[->](E)--(I);
  \draw[->](HH)--(J);
  \draw[fork](H.east)--(J.west);
  \draw[fork](I.east)--(J.west);
  \draw[->](J)--(K);
  \draw[->](K)--(L);
\end{tikzpicture}

%% file: figs/arch_CE_2.tex
\begin{tikzpicture}[scale=.8, transform shape, node distance=1cm, >=Latex]
  \node[place](W){};
  \node(Z)[verybigbox, right=1.2cm of W]{};
  \node(A)[smallbox]{$\xb_0$, $\mathbf{p}, t$};
  \node[place](X)[right=2.0cm of A]{};
  \node(D)[smallboxtime, right=of A]{Encoder $\boldsymbol{\tau}$};
  \node(C)[smallboxinit, above=2.5mm of D]{Encoder $\boldsymbol{\sE}$};
  \node(E)[smallboxdyn, below=2.5mm of D]{Encoder $\bs{c}$};
  \node(H)[smallboxinit, right=of C]{$\mathbf{y}_0$};
  \node(HH)[smallboxtime, right=0.9cm of D]{$\boldsymbol{\tau}$};
  \node(I)[smallboxdyn, right=0.95cm of E]{$\frac{d\mathbf{y}}{d\boldsymbol{\tau}}$};
  \node[place](Y)[right=6.0cm of A]{};
  \node(J)[bigbox, right=of Y]{$\mathbf{y}({\boldsymbol{\tau}}) = \mathbf{y}_0 + \boldsymbol{\tau}
  \circ \frac{d\mathbf{y}}{d\boldsymbol{\tau}}$};
  \node(K)[smallboxdecstacked, right=of J]{Decoder $\boldsymbol{\sD}_i$};
  \node(L)[smallboxstacked, right=of K]{ $\xb_i(t)$};
  \draw[fork](A.east)--(C.west);
  \draw[fork](A.east)--(E.west);
  \draw[->](A)--(D);
  \draw[->](C)--(H);
  \draw[->](D)--(HH);
  \draw[->](E)--(I);
  \draw[->](HH)--(J);
  \draw[fork](H.east)--(J.west);
  \draw[fork](I.east)--(J.west);
  \draw[->](J)--(K);
  \draw[->](K)--(L);
\end{tikzpicture}

%% file: figs/arch_CD_2.tex
\begin{tikzpicture}[scale=.8, transform shape, node distance=1cm, >=Latex]
  \node[place](W){};
  \node(Z)[verybigbox2, right=1.2cm of W]{};
  \node(A)[smallbox]{$\xb_0$, $\mathbf{p}, t$};
  \node[place](X)[right=2.0cm of A]{};
  \node(D)[smallboxtimestacked, right=of A]{Encoder $\boldsymbol{\tau}_i$};
  \node(C)[smallboxinitstacked, above=3.5mm of D]{Encoder $\boldsymbol{\sE}_i$};
  \node(E)[smallboxdynstacked, below=3.5mm of D]{Encoder $\bs{c}_i$};
  \node(H)[smallboxinitstacked, right=of C]{$[\mathbf{y}_0]_i$};
  \node(HH)[smallboxtimestacked, right=0.9cm of D]{$\boldsymbol{\tau}_i$};
  \node(I)[smallboxdynstacked, right=0.95cm of E]{$\frac{d\mathbf{y}_i}{d\boldsymbol{\tau}}$};
  \node[place](Y)[right=6.0cm of A]{};
  \node(J)[bigboxstacked, right=of Y]{$\mathbf{y}_i({\boldsymbol{\tau}}) = [\mathbf{y}_0]_i + \boldsymbol{\tau}_i
  \circ \frac{d\mathbf{y}_i}{d\boldsymbol{\tau}}$};
  \node(K)[smallboxdec, right=of J]{Decoder $\boldsymbol{\sD}$};
  \node(L)[smallbox, right=of K]{ $\xb(t)$};
  \draw[fork](A.east)--(C.west);
  \draw[fork](A.east)--(E.west);
  \draw[->](A)--(D);
  \draw[->](C)--(H);
  \draw[->](D)--(HH);
  \draw[->](E)--(I);
  \draw[->](HH)--(J);
  \draw[fork](H.east)--(J.west);
  \draw[fork](I.east)--(J.west);
  \draw[->](J)--(K);
  \draw[->](K)--(L);
\end{tikzpicture}

%% file: ROBER.tex
\subsection{Robertson Stiff Chemical Kinetics Model}
\label{sec:Robertson}

The Robertson chemical kinetics problem is a prototype stiff system of ODEs that describes the concentration of three species of reactants in a chemical reaction \cite{anantharaman2021acceleratingsimulationstiffnonlinear}. 
The model has been widely used to evaluate the performance of stiff integrators in traditional numerical analysis. The reaction is characterized by the following system of ODEs:
\begin{equation}
\begin{aligned}
  & \dd{x_1}{t}  = -p_1 x_1 + p_3 x_2 x_3, \\
   & \frac{dx_2}{dt} = p_1 x_1 - p_3 x_2 x_3 - p_2 x_2^2, \\
    &\frac{dx_3}{dt} = p_2 x_2^2,
\end{aligned}    
\label{rob}
\end{equation}
 where the second species, $x_2$, is the fastest evolving component and leads to numerical stiffness and difficulties in integration by explicit numerical solvers. In \eqref{rob}, the three reaction rates are typically chosen as  $p_1 = 4 \cdot 10^{-2}$, $p_2 = 3 \cdot 10^7$, and $p_3 = 10^4$. Note that in \eqref{original_dynamics}, we have the parameters of the StODE as $\mathbf{p}=[p_1,\ p_2,\ p_3]$ and states as $\xb=[x_1,\ x_2,\ x_3]$.
\subsubsection{Data Generation}
\label{sec:Robertson_data}
For a given initial condition, $\xb_0$, and parameters, $\mathbf{p}$, we integrate the system \eqref{rob} with the \texttt{Kvaerno5} stiff solver \cite{kvaerno2004singly}, from the \texttt{Diffrax} library \cite{kidger2021on}, on a 50-point (i.e. $M=50$ in \eqref{train_form}) logarithmically scaled time grid, spanning from $t=10^{-5}s$ to $t=10^{5}s$. In this problem, the initial condition is fixed  as $\xb_0= [1, 0, 0]$ and the reaction rates $\mathbf{p}$ vary across training samples.  Similar to the data generation approach in \cite{anantharaman2021acceleratingsimulationstiffnonlinear}, training input samples $\LRp{p_1,\ \p_2,\ p_3}$ were sampled from $[0.2\cdot10^{-2}, 0.6\cdot10^{-2}]\times [1.5 \cdot 10^{7}, 3.5 \cdot 10^{7}] \times [5 \cdot 10^{3},\ 1.5 \cdot 10^{4}]$. The parameters of the training set, $\LRp{p_1,\ \p_2,\ p_3}$, were sampled on a uniformly discretized grid of 16 linearly spaced collocation points in each domain. Thus, we consider a total of  $4096$ training samples. An additional $512$ validation samples and $1000$ test samples were generated using a uniformly discretized grid of 8 and 10 points in each domain respectively.

\subsubsection{Results}
\label{sec:Robertson_results}

\lilans was tested and compared with the NODE \cite{Lee_2021}, DeepONet \cite{goswami2023learningstiffchemicalkinetics}, and the Sulzer and Buck I and II approaches. The choice of latent dimension $m$ for each method is provided in  \cref{late}, \Cref{appen_late}. \Cref{fig:ROBerr} shows a comparison of the average point-wise relative error, calculated via \eqref{eq:R1}, for each method when applied to the test dataset. From \cref{fig:ROBerr}, we see that \lilans outperformed DeepONet by approximately one order of magnitude over the entire time, and also outperformed NODE by several orders of magnitude. Compared to Sulzer and Buck I and II, the performance of \lilans is similar until the middle of the figure, where \lilans takes the lead in accuracy. Note that the error of Sulzer and Buck I, the direct implementation of the methodology in \cite{sulzer2023speedingastrochemicalreactionnetworks}, becomes greater than that of the modified Sulzer and Buck II with latent dimension expansion, which reveals the benefit of increasing the problem dimensionality. The computation is done using single-precision floats, which means the error is close to machine precision for prediction on the early steps of the time series. The average error achieved by different machine learning methods has been tabulated in \cref{fig:err_table} and shows that \lilans achieved the lowest error out of  all the approaches, when evaluated on the test dataset. \Cref{fig:err_table} also shows that the method by Sulzer and Buck \cite{sulzer2023speedingastrochemicalreactionnetworks} produced a competitive relative error compared to \lilan. \lilans outperformed NODE by two orders of magnitude and DeepONet by an order of magnitude. \Cref{spec0} depicts a simulated trajectory using \lilans for a randomly chosen set of parameters, $\mathbf{p}$, from the test dataset, where we observe the predicted solution matches closely to the one calculated by the numerical solver. In \cref{fig:speed_table}, results on the speedup achieved over the chosen traditional numerical solver (\texttt{Kvaerno5} stiff solver) have been tabulated, where it is presented that the \lilans approach achieved over 800x speedup in the task of simulating trajectories from 1000 initial conditions. 




\begin{figure}[h!t!b!]      
  \centering
  \input{figs/ROBER_rel_errors_new}
  \caption{Average point-wise relative error over time \eqref{eq:R1} when evaluating Robertson chemical kinetics model on test data. All methods, besides NODE, performed well on this problem, with \lilans having a slight performance edge due to greater accuracy at the end of the time series $t\in[10^0s,10^5s]$.}
  \label{fig:ROBerr}
\end{figure} 

\begin{figure}[h!t!b!] 
\begin{minipage}{0.32\textwidth}
    \centering
    \resizebox{1\linewidth}{!}{\input{figs/dofs0}}
\end{minipage}
\begin{minipage}{0.32\textwidth}
    \centering
    \resizebox{1\linewidth}{!}{\input{figs/dofs1}}
\end{minipage}
\begin{minipage}{0.32\textwidth}
    \centering
    \resizebox{1\linewidth}{!}{\input{figs/dofs2}}
\end{minipage}
\caption{Simulated trajectory using \lilans for a randomly chosen set of parameters $\mathbf{p}$, from the test dataset. Left to Right: Evolution of species $x_1$ with time; Evolution of species $x_2$ with time; Evolution of species $x_3$ with time. For each species, the \lilans prediction is a near perfect visual match to the true dynamics. }
 \label{spec0}
\end{figure}

%% file: figs/ROBER_rel_errors_new.tex
\begin{tikzpicture}

\definecolor{crimson2143940}{RGB}{214,39,40}
\definecolor{darkgray176}{RGB}{176,176,176}
\definecolor{darkorange25512714}{RGB}{255,127,14}
\definecolor{forestgreen4416044}{RGB}{44,160,44}
\definecolor{lightgray204}{RGB}{204,204,204}
\definecolor{mediumpurple148103189}{RGB}{148,103,189}
\definecolor{steelblue31119180}{RGB}{31,119,180}

\begin{axis}[
legend cell align={left},
legend style={
  fill opacity=0.8,
  draw opacity=1,
  text opacity=1,
  at={(1,0)},
  anchor=south east,
  draw=lightgray204
},
log basis x={10},
log basis y={10},
tick align=outside,
tick pos=left,
title={Relative Error on Test Dataset},
x grid style={darkgray176},
xlabel={Time (s)},
xmin=1.59985871960606e-05, xmax=100000,
xmode=log,
xtick style={color=black},
xtick={1e-07,1e-05,0.001,0.1,10,1000,100000,10000000},
xticklabels={
  \(\displaystyle {10^{-7}}\),
  \(\displaystyle {10^{-5}}\),
  \(\displaystyle {10^{-3}}\),
  \(\displaystyle {10^{-1}}\),
  \(\displaystyle {10^{1}}\),
  \(\displaystyle {10^{3}}\),
  \(\displaystyle {10^{5}}\),
  \(\displaystyle {10^{7}}\)
},
y grid style={darkgray176},
ylabel={Error},
ymin=9.67972521797489e-09, ymax=0.0456406829012049,
ymode=log,
ytick style={color=black},
ytick={1e-10,1e-09,1e-08,1e-07,1e-06,1e-05,0.0001,0.001,0.01,0.1,1},
yticklabels={
  \(\displaystyle {10^{-10}}\),
  \(\displaystyle {10^{-9}}\),
  \(\displaystyle {10^{-8}}\),
  \(\displaystyle {10^{-7}}\),
  \(\displaystyle {10^{-6}}\),
  \(\displaystyle {10^{-5}}\),
  \(\displaystyle {10^{-4}}\),
  \(\displaystyle {10^{-3}}\),
  \(\displaystyle {10^{-2}}\),
  \(\displaystyle {10^{-1}}\),
  \(\displaystyle {10^{0}}\)
}
]
\addplot [semithick, steelblue31119180]
table {%
1.59985871960606e-05 3.08307583907208e-08
2.55954792269953e-05 1.94627158833782e-08
4.09491506238043e-05 2.02269667681776e-08
6.55128556859551e-05 2.06827586168856e-08
0.000104811313415469 4.86586735348737e-08
0.000167683293681101 4.21710772968709e-08
0.000268269579527973 5.89156918806566e-08
0.000429193426012878 6.87071377569737e-08
0.0006866488450043 9.48121794408507e-08
0.00109854114198756 1.12682130293251e-07
0.00175751062485479 1.53097289512516e-07
0.00281176869797423 2.87100505147464e-07
0.00449843266896944 5.43273017683532e-07
0.00719685673001152 8.00580892246217e-07
0.0115139539932645 1.31707463424391e-06
0.0184206996932672 2.07976154342759e-06
0.0294705170255181 3.08467110698984e-06
0.0471486636345739 5.32354806637159e-06
0.0754312006335462 8.91883701115148e-06
0.120679264063933 1.49962561408756e-05
0.193069772888325 2.81488901237026e-05
0.308884359647748 5.25902287336066e-05
0.494171336132384 8.35168830235489e-05
0.79060432109077 0.00011217802966712
1.2648552168553 0.000138055751449428
2.02358964772516 0.000174243279616348
3.23745754281765 0.000220706715481356
5.17947467923121 0.00031784106977284
8.28642772854684 0.000428244937211275
13.2571136559011 0.000519196211826056
21.2095088792019 0.000616912730038166
33.9322177189533 0.000734582601580769
54.2867543932386 0.000850565731525421
86.8511373751352 0.000882657070178539
138.949549437314 0.000803728646133095
222.29964825262 0.000682168116327375
355.648030622313 0.000640174432191998
568.98660290183 0.000666823179926723
910.298177991523 0.000728213461115956
1456.34847750124 0.000785965239629149
2329.95181051537 0.000800231820903718
3727.59372031494 0.0007526318076998
5963.62331659464 0.000663000799249858
9540.95476349996 0.000550709315575659
15264.1796717524 0.000435171707067639
24420.5309454865 0.000339795224135742
39069.3993705462 0.000267217867076397
62505.5192527398 0.000219254739931785
100000 0.000191685729078017
};
\addlegendentry{\lilans}
\addplot [semithick, darkorange25512714]
table {%
1.59985871960606e-05 3.90101142500043e-08
2.55954792269953e-05 4.51655672861762e-08
4.09491506238043e-05 5.51528884784602e-08
6.55128556859551e-05 3.73849786683422e-08
0.000104811313415469 7.49627133700415e-08
0.000167683293681101 8.64904023956115e-08
0.000268269579527973 1.11946519609774e-07
0.000429193426012878 2.3297407381051e-07
0.0006866488450043 2.65321773440519e-07
0.00109854114198756 4.50399880946861e-07
0.00175751062485479 8.88673355348146e-07
0.00281176869797423 9.50775017827254e-07
0.00449843266896944 9.50128139720618e-07
0.00719685673001152 1.42830538152339e-06
0.0115139539932645 2.48186302087561e-06
0.0184206996932672 3.66188805855927e-06
0.0294705170255181 5.56696522835409e-06
0.0471486636345739 1.00770821518381e-05
0.0754312006335462 1.73190837813308e-05
0.120679264063933 2.66785082203569e-05
0.193069772888325 4.17153969465289e-05
0.308884359647748 7.6775970228482e-05
0.494171336132384 0.000137608920340426
0.79060432109077 0.000252023834036663
1.2648552168553 0.000426866376074031
2.02358964772516 0.00059743324527517
3.23745754281765 0.000719504256267101
5.17947467923121 0.000918330915737897
8.28642772854684 0.00130781508050859
13.2571136559011 0.00192522641737014
21.2095088792019 0.00275308429263532
33.9322177189533 0.00361457653343678
54.2867543932386 0.00440101837739348
86.8511373751352 0.00487308716401458
138.949549437314 0.00514222076162696
222.29964825262 0.0051712142303586
355.648030622313 0.005132463760674
568.98660290183 0.00493653444573283
910.298177991523 0.00467363279312849
1456.34847750124 0.00430101016536355
2329.95181051537 0.00393386743962765
3727.59372031494 0.00349000957794487
5963.62331659464 0.00307405623607337
9540.95476349996 0.00269166636280715
15264.1796717524 0.00226583168841898
24420.5309454865 0.00186562293674797
39069.3993705462 0.00146901037078351
62505.5192527398 0.00115217140410095
100000 0.000866740709170699
};
\addlegendentry{Sulzer and Buck I}
\addplot [semithick, forestgreen4416044]
table {%
1.59985871960606e-05 3.67379335841633e-08
2.55954792269953e-05 3.23809281610465e-08
4.09491506238043e-05 2.74620095552791e-08
6.55128556859551e-05 3.23975619664907e-08
0.000104811313415469 4.58458266905382e-08
0.000167683293681101 8.27076860332454e-08
0.000268269579527973 1.25701959063917e-07
0.000429193426012878 1.34323173028861e-07
0.0006866488450043 1.10801970265584e-07
0.00109854114198756 2.95900690616691e-07
0.00175751062485479 5.15896033448371e-07
0.00281176869797423 4.96169718644524e-07
0.00449843266896944 5.65515961170604e-07
0.00719685673001152 6.67897097628156e-07
0.0115139539932645 9.00719896890223e-07
0.0184206996932672 1.82966289230535e-06
0.0294705170255181 3.80998039872793e-06
0.0471486636345739 5.87462000112282e-06
0.0754312006335462 7.944215212774e-06
0.120679264063933 9.50564935919829e-06
0.193069772888325 1.69487320818007e-05
0.308884359647748 4.25904181611259e-05
0.494171336132384 9.24187334021553e-05
0.79060432109077 0.000172213680343702
1.2648552168553 0.000272475619567558
2.02358964772516 0.000359496858436614
3.23745754281765 0.000427382998168468
5.17947467923121 0.00051911681657657
8.28642772854684 0.000668516324367374
13.2571136559011 0.000914492993615568
21.2095088792019 0.00120237690862268
33.9322177189533 0.0014207442291081
54.2867543932386 0.00152255059219897
86.8511373751352 0.0016133802710101
138.949549437314 0.00169882806949317
222.29964825262 0.00183014804497361
355.648030622313 0.00205006008036435
568.98660290183 0.00235544145107269
910.298177991523 0.0026820432394743
1456.34847750124 0.00289811799302697
2329.95181051537 0.00291926926001906
3727.59372031494 0.00273310532793403
5963.62331659464 0.00236114813014865
9540.95476349996 0.00188327301293612
15264.1796717524 0.00144277338404208
24420.5309454865 0.0011059797834605
39069.3993705462 0.000874875928275287
62505.5192527398 0.000734174565877765
100000 0.000613291398622096
};
\addlegendentry{Sulzer and Buck II}
\addplot [semithick, crimson2143940]
table {%
1.59985871960606e-05 3.14492005770717e-08
2.55954792269953e-05 2.1248878212532e-08
4.09491506238043e-05 2.42789521820441e-08
6.55128556859551e-05 2.860958758788e-08
0.000104811313415469 5.16421678753431e-08
0.000167683293681101 7.3781151854746e-08
0.000268269579527973 1.36467221523162e-07
0.000429193426012878 2.20873943135302e-07
0.0006866488450043 5.17254363785469e-07
0.00109854114198756 1.09289896954579e-06
0.00175751062485479 2.44149327954801e-06
0.00281176869797423 4.84966585645452e-06
0.00449843266896944 1.0104669854627e-05
0.00719685673001152 1.89065449376358e-05
0.0115139539932645 2.74102876574034e-05
0.0184206996932672 2.96469461318338e-05
0.0294705170255181 2.42636433540611e-05
0.0471486636345739 2.1496456611203e-05
0.0754312006335462 4.92692888656165e-05
0.120679264063933 0.000107481304439716
0.193069772888325 0.000202276176423766
0.308884359647748 0.000372211041394621
0.494171336132384 0.000654281640890986
0.79060432109077 0.00104156241286546
1.2648552168553 0.00146118376869708
2.02358964772516 0.00179097300861031
3.23745754281765 0.00192260730545968
5.17947467923121 0.00183489813935012
8.28642772854684 0.00197020918130875
13.2571136559011 0.00276852189563215
21.2095088792019 0.00363445677794516
33.9322177189533 0.00422926293686032
54.2867543932386 0.00468315603211522
86.8511373751352 0.00507776765152812
138.949549437314 0.0053832228295505
222.29964825262 0.00570202711969614
355.648030622313 0.00620419904589653
568.98660290183 0.00646328879520297
910.298177991523 0.00613196846097708
1456.34847750124 0.00527374306693673
2329.95181051537 0.00400962913408875
3727.59372031494 0.00266844732686877
5963.62331659464 0.00170015555340797
9540.95476349996 0.00134886766318232
15264.1796717524 0.00151570921298116
24420.5309454865 0.00154564424883574
39069.3993705462 0.00136577489320189
62505.5192527398 0.00108994438778609
100000 0.00080842844909057
};
\addlegendentry{DeepONet}
\addplot [semithick, mediumpurple148103189]
table {%
1.59985871960606e-05 0.00281916814856231
2.55954792269953e-05 0.00492784241214395
4.09491506238043e-05 0.00468146521598101
6.55128556859551e-05 0.00504846265539527
0.000104811313415469 0.0050610164180398
0.000167683293681101 0.0050414577126503
0.000268269579527973 0.00487425643950701
0.000429193426012878 0.00461791409179568
0.0006866488450043 0.00430777901783586
0.00109854114198756 0.00437276205047965
0.00175751062485479 0.00524263037368655
0.00281176869797423 0.006375711876899
0.00449843266896944 0.00748551776632667
0.00719685673001152 0.00856632087379694
0.0115139539932645 0.00970785971730947
0.0184206996932672 0.0109161892905831
0.0294705170255181 0.0121821705251932
0.0471486636345739 0.0135056944563985
0.0754312006335462 0.0148898418992758
0.120679264063933 0.0163339860737324
0.193069772888325 0.017824549227953
0.308884359647748 0.019319674000144
0.494171336132384 0.0207301173359156
0.79060432109077 0.0219050645828247
1.2648552168553 0.0226367693394423
2.02358964772516 0.022699261084199
3.23745754281765 0.0219292156398296
5.17947467923121 0.02033331990242
8.28642772854684 0.0183844044804573
13.2571136559011 0.0168889947235584
21.2095088792019 0.0156927853822708
33.9322177189533 0.0145819559693336
54.2867543932386 0.0135725922882557
86.8511373751352 0.0127351265400648
138.949549437314 0.0123075684532523
222.29964825262 0.0120465038344264
355.648030622313 0.0115802027285099
568.98660290183 0.0108378333970904
910.298177991523 0.0098959356546402
1456.34847750124 0.00897479709237814
2329.95181051537 0.00825905613601208
3727.59372031494 0.00771280098706484
5963.62331659464 0.00731573440134525
9540.95476349996 0.00711006578058004
15264.1796717524 0.00708374753594398
24420.5309454865 0.00717187346890569
39069.3993705462 0.00740335276350379
62505.5192527398 0.00801475532352924
100000 0.0090134609490633
};
\addlegendentry{NODE}
\end{axis}

\end{tikzpicture}

%% file: figs/dofs0.tex
\begin{tikzpicture}

\definecolor{darkgray176}{RGB}{176,176,176}
\definecolor{steelblue31119180}{RGB}{0,127,0}
\definecolor{lightgray204}{RGB}{204,204,204}

\begin{axis}[
legend cell align={left},
legend style={
  fill opacity=0.8,
  draw opacity=1,
  text opacity=1,
  at={(0.03,0.03)},
  anchor=south west,
  draw=lightgray204
},
log basis x={10},
log basis y={10},
tick align=outside,
tick pos=left,
title={Species 1 Concentration},
x grid style={darkgray176},
xlabel={Time (s)},
xmin=1.59985871960606e-05, xmax=100000,
xmode=log,
xtick style={color=black},
xtick={1e-07,1e-05,0.001,0.1,10,1000,100000,10000000},
xticklabels={
  \(\displaystyle {10^{-7}}\),
  \(\displaystyle {10^{-5}}\),
  \(\displaystyle {10^{-3}}\),
  \(\displaystyle {10^{-1}}\),
  \(\displaystyle {10^{1}}\),
  \(\displaystyle {10^{3}}\),
  \(\displaystyle {10^{5}}\),
  \(\displaystyle {10^{7}}\)
},
y grid style={darkgray176},
ylabel={$x_1$},
ymin=0.0106302065859966, ymax=1.24217103449876,
ymode=log,
ytick style={color=black},
ytick={0.001,0.01,0.1,1,10,100},
yticklabels={
  \(\displaystyle {10^{-3}}\),
  \(\displaystyle {10^{-2}}\),
  \(\displaystyle {10^{-1}}\),
  \(\displaystyle {10^{0}}\),
  \(\displaystyle {10^{1}}\),
  \(\displaystyle {10^{2}}\)
}
]
\addplot [ultra thick, steelblue31119180]
table {%
1.59985871960606e-05 0.999384880065918
2.55954792269953e-05 0.999460637569427
4.09491506238043e-05 0.999543190002441
6.55128556859551e-05 0.999633967876434
0.000104811313415469 0.999731838703156
0.000167683293681101 0.999837517738342
0.000268269579527973 0.999948620796204
0.000429193426012878 1.0000638961792
0.0006866488450043 1.00017917156219
0.00109854114198756 1.00028848648071
0.00175751062485479 1.00038194656372
0.00281176869797423 1.0004448890686
0.00449843266896944 1.00045549869537
0.00719685673001152 1.0003821849823
0.0115139539932645 1.00017893314362
0.0184206996932672 0.999777138233185
0.0294705170255181 0.999080300331116
0.0471486636345739 0.997947990894318
0.0754312006335462 0.996177852153778
0.120679264063933 0.993482708930969
0.193069772888325 0.989459574222565
0.308884359647748 0.983557939529419
0.494171336132384 0.9750657081604
0.79060432109077 0.9631307721138
1.2648552168553 0.946861624717712
2.02358964772516 0.92551326751709
3.23745754281765 0.898705303668976
5.17947467923121 0.866556882858276
8.28642772854684 0.829631626605988
13.2571136559011 0.788718402385712
21.2095088792019 0.744550168514252
33.9322177189533 0.697582483291626
54.2867543932386 0.64790666103363
86.8511373751352 0.595387876033783
138.949549437314 0.54003918170929
222.29964825262 0.482391089200974
355.648030622313 0.423533350229263
568.98660290183 0.364847987890244
910.298177991523 0.307731330394745
1456.34847750124 0.253508120775223
2329.95181051537 0.203449249267578
3727.59372031494 0.158727467060089
5963.62331659464 0.120251007378101
9540.95476349996 0.0884723886847496
15264.1796717524 0.06329595297575
24420.5309454865 0.0441364422440529
39069.3993705462 0.030087199062109
62505.5192527398 0.0201192926615477
100000 0.0132440254092216
};
\addlegendentry{\lilan}
\addplot [ultra thick, black, dashed]
table {%
1e-05 1
1.59985871960606e-05 0.999999664079173
2.55954792269953e-05 0.999999126653543
4.09491506238043e-05 0.999998266849063
6.55128556859551e-05 0.999996891284913
0.000104811313415469 0.999994690580595
0.000167683293681101 0.999991169775186
0.000268269579527973 0.99998553701544
0.000429193426012878 0.999976525516558
0.0006866488450043 0.99996210899159
0.00109854114198756 0.99993904744937
0.00175751062485479 0.999902162512447
0.00281176869797423 0.999843180449486
0.00449843266896944 0.999748890823459
0.00719685673001152 0.999598227873412
0.0115139539932645 0.99935766589078
0.0184206996932672 0.998974016689849
0.0294705170255181 0.998363319878149
0.0471486636345739 0.997394095200447
0.0754312006335462 0.995863049376045
0.120679264063933 0.993462136222248
0.193069772888325 0.989739227467331
0.308884359647748 0.984063156012763
0.494171336132384 0.975619215589551
0.79060432109077 0.963477094083476
1.2648552168553 0.946764341638235
2.02358964772516 0.924909607140196
3.23745754281765 0.897815055076302
5.17947467923121 0.865812732576529
8.28642772854684 0.829431518940081
13.2571136559011 0.789159888105977
21.2095088792019 0.745342274814842
33.9322177189533 0.698204334198272
54.2867543932386 0.647938806648361
86.8511373751352 0.594800261364194
138.949549437314 0.539186980310618
222.29964825262 0.481702805139949
355.648030622313 0.42319393638103
568.98660290183 0.364753050671007
910.298177991523 0.30768123878187
1456.34847750124 0.2534003312656
2329.95181051537 0.203316452152825
3727.59372031494 0.158650506283357
5963.62331659464 0.120269023641839
9540.95476349996 0.0885607186940407
15264.1796717524 0.0633996317636168
24420.5309454865 0.0442102900743589
39069.3993705462 0.0301133753227995
62505.5192527398 0.0201023612362476
100000 0.0131985228019457
};
\addlegendentry{Numerical Solver}
\end{axis}

\end{tikzpicture}

%% file: figs/dofs1.tex
\begin{tikzpicture}

\definecolor{darkgray176}{RGB}{176,176,176}
\definecolor{steelblue31119180}{RGB}{0,127,0}
\definecolor{lightgray204}{RGB}{204,204,204}

\begin{axis}[
legend cell align={left},
legend style={
  fill opacity=0.8,
  draw opacity=1,
  text opacity=1,
  at={(0.03,0.03)},
  anchor=south west,
  draw=lightgray204
},
log basis x={10},
log basis y={10},
tick align=outside,
tick pos=left,
title={Species 2 Concentration},
x grid style={darkgray176},
xlabel={Time (s)},
xmin=1.59985871960606e-05, xmax=100000,
xmode=log,
xtick style={color=black},
xtick={1e-07,1e-05,0.001,0.1,10,1000,100000,10000000},
xticklabels={
  \(\displaystyle {10^{-7}}\),
  \(\displaystyle {10^{-5}}\),
  \(\displaystyle {10^{-3}}\),
  \(\displaystyle {10^{-1}}\),
  \(\displaystyle {10^{1}}\),
  \(\displaystyle {10^{3}}\),
  \(\displaystyle {10^{5}}\),
  \(\displaystyle {10^{7}}\)
},
y grid style={darkgray176},
ylabel={$x_2$},
ymin=3.8517671955313e-08, ymax=5.05603231456631e-05,
ymode=log,
ytick style={color=black},
ytick={1e-09,1e-08,1e-07,1e-06,1e-05,0.0001,0.001},
yticklabels={
  \(\displaystyle {10^{-9}}\),
  \(\displaystyle {10^{-8}}\),
  \(\displaystyle {10^{-7}}\),
  \(\displaystyle {10^{-6}}\),
  \(\displaystyle {10^{-5}}\),
  \(\displaystyle {10^{-4}}\),
  \(\displaystyle {10^{-3}}\)
}
]
\addplot [ultra thick, steelblue31119180]
table {%
1.59985871960606e-05 3.36239338594169e-07
2.55954792269953e-05 8.75295938840281e-07
4.09491506238043e-05 1.73658861513104e-06
6.55128556859551e-05 3.10667769554129e-06
0.000104811313415469 5.27001975569874e-06
0.000167683293681101 8.63483364810236e-06
0.000268269579527973 1.36895077957888e-05
0.000429193426012878 2.06543936656089e-05
0.0006866488450043 2.8412243409548e-05
0.00109854114198756 3.3916330721695e-05
0.00175751062485479 3.59928089892492e-05
0.00281176869797423 3.64749794243835e-05
0.00449843266896944 3.64792213076726e-05
0.00719685673001152 3.64151164831128e-05
0.0115139539932645 3.63555518561043e-05
0.0184206996932672 3.62847276846878e-05
0.0294705170255181 3.61820493708365e-05
0.0471486636345739 3.60201302100904e-05
0.0754312006335462 3.57615026587155e-05
0.120679264063933 3.53508185071405e-05
0.193069772888325 3.47060922649689e-05
0.308884359647748 3.37173551088199e-05
0.494171336132384 3.22649102599826e-05
0.79060432109077 3.02624157484388e-05
1.2648552168553 2.77089311566669e-05
2.02358964772516 2.47183616011171e-05
3.23745754281765 2.15008967643371e-05
5.17947467923121 1.82991389010567e-05
8.28642772854684 1.53162363858428e-05
13.2571136559011 1.26738787002978e-05
21.2095088792019 1.0411796210974e-05
33.9322177189533 8.51348886499181e-06
54.2867543932386 6.93500578563544e-06
86.8511373751352 5.62585910302005e-06
138.949549437314 4.53974053016282e-06
222.29964825262 3.63815843229531e-06
355.648030622313 2.89045738099958e-06
568.98660290183 2.27231043936627e-06
910.298177991523 1.76422088316031e-06
1456.34847750124 1.35019308800111e-06
2329.95181051537 1.0167486834689e-06
3727.59372031494 7.52164339701267e-07
5963.62331659464 5.45944203622639e-07
9540.95476349996 3.88489951319571e-07
15264.1796717524 2.70963852244677e-07
24420.5309454865 1.85320061518723e-07
39069.3993705462 1.24410817647913e-07
62505.5192527398 8.21103185444372e-08
100000 5.33819104475697e-08
};
\addlegendentry{\lilan}
\addplot [ultra thick, black, dashed]
table {%
1e-05 0
1.59985871960606e-05 3.35909602228474e-07
2.55954792269953e-05 8.73170287929092e-07
4.09491506238043e-05 1.73182520218405e-06
6.55128556859551e-05 3.10119708324974e-06
0.000104811313415469 5.27230712954176e-06
0.000167683293681101 8.66199285034116e-06
0.000268269579527973 1.37512694401447e-05
0.000429193426012878 2.06984769100587e-05
0.0006866488450043 2.83706212874287e-05
0.00109854114198756 3.40094337025314e-05
0.00175751062485479 3.616425029189e-05
0.00281176869797423 3.64817580440854e-05
0.00449843266896944 3.64777891452135e-05
0.00719685673001152 3.64500254635403e-05
0.0115139539932645 3.64056090167048e-05
0.0184206996932672 3.63348637113597e-05
0.0294705170255181 3.62224793963776e-05
0.0471486636345739 3.60446937071741e-05
0.0754312006335462 3.57652983783278e-05
0.120679264063933 3.53307395407203e-05
0.193069772888325 3.46655523817751e-05
0.308884359647748 3.36717141796748e-05
0.494171336132384 3.22388303355248e-05
0.79060432109077 3.02740815452805e-05
1.2648552168553 2.77531175524258e-05
2.02358964772516 2.47692895247125e-05
3.23745754281765 2.15352397221448e-05
5.17947467923121 1.83138586046373e-05
8.28642772854684 1.53230338036929e-05
13.2571136559011 1.26835836328882e-05
21.2095088792019 1.04270306528241e-05
33.9322177189533 8.53084357881318e-06
54.2867543932386 6.95018959715473e-06
86.8511373751352 5.63653721164234e-06
138.949549437314 4.54582696089523e-06
222.29964825262 3.64087874064147e-06
355.648030622313 2.89127664109535e-06
568.98660290183 2.27238616576141e-06
910.298177991523 1.76420304637266e-06
1456.34847750124 1.35029972504056e-06
2329.95181051537 1.01692129097788e-06
3727.59372031494 7.5224976479035e-07
5963.62331659464 5.45828918438636e-07
9540.95476349996 3.88167383342997e-07
15264.1796717524 2.70530580594087e-07
24420.5309454865 1.84913706516827e-07
39069.3993705462 1.24145726353921e-07
62505.5192527398 8.20384263449537e-08
100000 5.34915191112241e-08
};
\addlegendentry{Numerical Solver}
\end{axis}

\end{tikzpicture}

%% file: figs/dofs2.tex
\begin{tikzpicture}

\definecolor{darkgray176}{RGB}{176,176,176}
\definecolor{steelblue31119180}{RGB}{0,127,0}
\definecolor{lightgray204}{RGB}{204,204,204}

\begin{axis}[
legend cell align={left},
legend style={
  fill opacity=0.8,
  draw opacity=1,
  text opacity=1,
  at={(0.97,0.03)},
  anchor=south east,
  draw=lightgray204
},
log basis x={10},
log basis y={10},
tick align=outside,
tick pos=left,
title={Species 3 Concentration},
x grid style={darkgray176},
xlabel={Time (s)},
xmin=1.59985871960606e-05, xmax=100000,
xmode=log,
xtick style={color=black},
xtick={1e-07,1e-05,0.001,0.1,10,1000,100000,10000000},
xticklabels={
  \(\displaystyle {10^{-7}}\),
  \(\displaystyle {10^{-5}}\),
  \(\displaystyle {10^{-3}}\),
  \(\displaystyle {10^{-1}}\),
  \(\displaystyle {10^{1}}\),
  \(\displaystyle {10^{3}}\),
  \(\displaystyle {10^{5}}\),
  \(\displaystyle {10^{7}}\)
},
y grid style={darkgray176},
ylabel={$x_3$},
ymin=3.18388299976938e-12, ymax=3.48078022087124,
ymode=log,
ytick style={color=black},
ytick={1e-14,1e-12,1e-10,1e-08,1e-06,0.0001,0.01,1,100,10000},
yticklabels={
  \(\displaystyle {10^{-14}}\),
  \(\displaystyle {10^{-12}}\),
  \(\displaystyle {10^{-10}}\),
  \(\displaystyle {10^{-8}}\),
  \(\displaystyle {10^{-6}}\),
  \(\displaystyle {10^{-4}}\),
  \(\displaystyle {10^{-2}}\),
  \(\displaystyle {10^{0}}\),
  \(\displaystyle {10^{2}}\),
  \(\displaystyle {10^{4}}\)
}
]
\addplot [ultra thick, steelblue31119180]
table {%
1.59985871960606e-05 1.1263785910931e-11
2.55954792269953e-05 1.74781522588319e-10
4.09491506238043e-05 1.3230123663277e-09
6.55128556859551e-05 7.5471469074273e-09
0.000104811313415469 3.72654760383284e-08
0.000167683293681101 1.68795452282211e-07
0.000268269579527973 7.13493818693678e-07
0.000429193426012878 2.78970969702641e-06
0.0006866488450043 9.60456327447901e-06
0.00109854114198756 2.70217260549543e-05
0.00175751062485479 6.1456325056497e-05
0.00281176869797423 0.000120425582281314
0.00449843266896944 0.00021554681006819
0.00719685673001152 0.000366692838724703
0.0115139539932645 0.000607001129537821
0.0184206996932672 0.00098944001365453
0.0294705170255181 0.00159795267973095
0.0471486636345739 0.00256452290341258
0.0754312006335462 0.00409358832985163
0.120679264063933 0.00649432558566332
0.193069772888325 0.0102177960798144
0.308884359647748 0.0158884227275848
0.494171336132384 0.024307444691658
0.79060432109077 0.0363947525620461
1.2648552168553 0.0530398450791836
2.02358964772516 0.074876457452774
3.23745754281765 0.102079220116138
5.17947467923121 0.134330481290817
8.28642772854684 0.171016380190849
13.2571136559011 0.211526662111282
21.2095088792019 0.255452543497086
33.9322177189533 0.302591860294342
54.2867543932386 0.352818161249161
86.8511373751352 0.405923634767532
138.949549437314 0.461504518985748
222.29964825262 0.518907189369202
355.648030622313 0.577229142189026
568.98660290183 0.635358512401581
910.298177991523 0.69203919172287
1456.34847750124 0.745960235595703
2329.95181051537 0.795860290527344
3727.59372031494 0.840642035007477
5963.62331659464 0.879482388496399
9540.95476349996 0.91191840171814
15264.1796717524 0.937882363796234
24420.5309454865 0.957688450813293
39069.3993705462 0.971963047981262
62505.5192527398 0.98153680562973
100000 0.987329423427582
};
\addlegendentry{\lilan}
\addplot [ultra thick, black, dashed]
table {%
1e-05 0
1.59985871960606e-05 1.12246193703943e-11
2.55954792269953e-05 1.76169012920251e-10
4.09491506238043e-05 1.32573485317015e-09
6.55128556859551e-05 7.51800427441068e-09
0.000104811313415469 3.71122757972618e-08
0.000167683293681101 1.68231963598991e-07
0.000268269579527973 7.11715119608965e-07
0.000429193426012878 2.7760065317621e-06
0.0006866488450043 9.52038712266293e-06
0.00109854114198756 2.69431169276741e-05
0.00175751062485479 6.16732372613965e-05
0.00281176869797423 0.000120337792470332
0.00449843266896944 0.000214631387396565
0.00719685673001152 0.000365322101125139
0.0115139539932645 0.000605928500204333
0.0184206996932672 0.00098964844644076
0.0294705170255181 0.00160045764245489
0.0471486636345739 0.0025698601058458
0.0754312006335462 0.00410118532557713
0.120679264063933 0.00650253303821138
0.193069772888325 0.010226106980288
0.308884359647748 0.0159031722730583
0.494171336132384 0.024348545580114
0.79060432109077 0.0364926318349797
1.2648552168553 0.0532079052442136
2.02358964772516 0.0750656235702798
3.23745754281765 0.102163409683977
5.17947467923121 0.134168953564867
8.28642772854684 0.170553158026116
13.2571136559011 0.21082742831039
21.2095088792019 0.254647298154506
33.9322177189533 0.30178713495815
54.2867543932386 0.352054243162042
86.8511373751352 0.405194102098594
138.949549437314 0.460808473862421
222.29964825262 0.51829355398131
355.648030622313 0.576803172342328
568.98660290183 0.635244676942826
910.298177991523 0.692316997015083
1456.34847750124 0.746598318434675
2329.95181051537 0.796682530925884
3727.59372031494 0.841348741466879
5963.62331659464 0.879730430529242
9540.95476349996 0.911438893138576
15264.1796717524 0.936600097705803
24420.5309454865 0.955789525011935
39069.3993705462 0.969886500531474
62505.5192527398 0.979897556725326
100000 0.986801423706535
};
\addlegendentry{Numerical Solver}
\end{axis}

\end{tikzpicture}

%% file: CR.tex
\subsection{Plasma Collisional-Radiative Model}
\label{sec:CR-model}
In this section, we consider a collisional-radiative (CR) model, for a lithium plasma, which has more significant multiscale behavior in comparison to the Robertson problem, in addition to being stiff. The CR model is a nonlinear dynamical system written as follows: 
\begin{equation}
    \frac{d\mathbf{n}}{dt} = {\mathbf{R}}(\mathbf{n}) \cdot \mathbf{n}, \quad \mathbf{n}(0)=\mathbf{n}_{initial}.
    \label{eq:1}
\end{equation}
In \cref{eq:1}, $\mathbf{n}\in\real^{n_x}$ and ${\mathbf{R}}$ is the $n_x\times n_x$ rate matrix with nonlinear dependence on $\mathbf{n}$. The rate matrix ${\mathbf{R}}$ contains a set of rules for the up-transitions and down-transitions of electrons between the $n_x$ discrete excited state levels of the state vector $\mathbf{n}$. The rate matrix explicitly depends on the temperature $T_e$ and the electron density $n_e$. The dependence on temperature comes from the fact that the electron distribution is assumed to be a Maxwellian distribution, parameterized by $T_e$ \cite{garland2022efficient}. 
The electron density is defined as
\begin{equation}
    n_e = \sum_{j=0}^{Z}jn_{j},
    \label{electron_density}
\end{equation}
where $Z$ denotes the atomic number of the plasma element (in our simulation we consider lithium, for which $Z=3$), $j$ is the ion charge state, and $n_j$ is the lithium ion density at charge state $j$ given by \eqref{eq:4states} \cite{garland2022efficient}.  We have $L=31$ in \eqref{eq:4states}, implying that  $\mathbf{n}\in \real^{94}$, i.e. there are $94$ discrete excited states, and 
\begin{equation}
    n_0 = \sum_{i=0}^{L-1}\mathbf{n}_{i}, \ \ \ \ n_1 = \sum_{i=L}^{2L-1}\mathbf{n}_{i}, \ \ \ \ n_2 = \sum_{i=2L}^{3L-1}\mathbf{n}_{i}, \ \ \ \ n_3 = \mathbf{n}_{3L}.
    \label{eq:4states}
\end{equation} The traditional approach to solving \eqref{eq:1}  are methods well suited for stiff problems, such as Backward Differentiation Formula (BDF) methods, described in detail in \cite{süli2003introduction}. These approaches are implicit linear multistep methods and require solving a system of $n_x$ equations at each time step.  Solving this system becomes very expensive when $n_x$ is large, as is the case for CR models with high-Z element impurities or when $L$ is large.

\subsubsection{Data Generation}
\label{sec:CR_data}

For data generation, \eqref{eq:1} is integrated using the six-step BDF formula on a 400-step (i.e. $M=400$ in \eqref{train_form}) logarithmic time grid on $t \in [1^{-16},1^0]$. We generate different initial conditions ${\mathbf{n}}_{initial}$  for \eqref{eq:1} through two parameters $n_A$ and $n_{per}$ as follows:
\begin{equation}
\begin{aligned}
   &  \LRp{\mathbf{n}_{initial}}_0 = \LRp{1 - n_{per} - 92 \cdot 10^{-5} \cdot n_{per}} \cdot n_A,\\  
   & \LRp{\mathbf{n}_{initial}}_1, \LRp{\mathbf{n}_{initial}}_2, \dots \LRp{\mathbf{n}_{initial}}_{92} = 10^{-5} \cdot n_{per} \cdot n_A,\\
   & \LRp{\mathbf{n}_{initial}}_{93} = n_{per} \cdot n_A,
\end{aligned}    
\label{f_map}
\end{equation}
where $\LRp{\mathbf{n}_{initial}}_i$ denotes the $i^{th}$ component of $\mathbf{n}_{initial}$. We sample parameters $\LRp{n_A,\ n_{per},\ T_e}$ from $[10^{14}, 10^{15}]\times [1 \cdot 10^{-3}, 2 \cdot 10^{-3}] \times [5,\ 95]$, by considering a uniformly discretized grid ($25$ equally spaced collocation points) in each variable,  to generate the training data. This leads to a total of $15625$ training samples. Here,  
$n_A$ is the total number of lithium ions, of all charge states, in the plasma, $T_e$ is the temperature in KeV, and $n_{per}$ is a constant used to parameterize the initial electron distribution. An additional $1000$ validation samples and $4096$ test samples were also generated using respective grids of $10$ and $16$ linearly discretized collocation points. Note that the sum of the entries in $\mathbf{n}_{initial}$ will be $n_A$. In fact, the total number of electrons is conserved for all time steps of the simulation. Note that in \eqref{original_dynamics}, we have the parameters $\mathbf{p}=[T_e, \ n_A]$ and $\xb_0=\LRp{\mathbf{n}_{initial}}=\mathcal{F}\LRp{\mathbf{n_p}}$, where $\mathbf{n_p}=[n_A, \ n_{per}]$ and $\mathcal{F}$ is a map determined by \eqref{f_map}. In implementation, our approach only acts on the variables $n_A$, $T_e$, $n_{per}$, and $t$, and learning the parameterization \eqref{f_map} becomes part of the learning task. 


\subsubsection{Results}
\label{sec:chargestate}

We demonstrate our approach on two cases:
\begin{enumerate}
    \item CR charge state model: The aim here is to predict only the four charge states $n_0,\dots, n_3$ in \eqref{eq:4states}.
    \item Full CR model: The aim here is to predict the full state $\mathbf{n}\in \real^{94}$ in \eqref{eq:1}.
    \end{enumerate}

{\bf{Results on the CR charge state model}}

\lilans was tested and compared against NODE \cite{Lee_2021},  DeepONet \cite{goswami2023learningstiffchemicalkinetics}, the approach by Sulzer and Buck \cite{sulzer2023speedingastrochemicalreactionnetworks}, as well as the modified Sulzer and Buck II. The choice of latent dimension $m$ for each method is provided in \cref{late}, \Cref{appen_late}.  The right subfigure of \cref{fig:ChargeErr} shows a comparison of the average point-wise relative error, calculated via \eqref{eq:R1}, for each method when applied to the test dataset. For this problem, we see that the error produced by \lilans is lower than that of DeepONet. In particular, DeepONet has a large spike in error up to $\mathcal{O}(10^{-1})$ near $t=10^{-8}s$, which is typically the beginning of the temporal region where rapid changes in the solution take place. The error increase is present in the error of \lilan, but it is not as significant, with the error remaining on $\mathcal{O}(10^{-2})$. The error of DeepONet decreases to nearly the same value as that of \lilans in the temporal region where the steady state is typically reached, $t\in[10^{-3}s, 10^{0}s]$. Additionally, NODE performs very poorly for this problem. Sulzer and Buck I and II produce a very close relative error to that of \lilans throughout the simulation time, appearing marginally less accurate in the initial range $t \in [10^{-16}s, 10^{-10}s]$. The left subfigure of \cref{fig:ChargeErr} depicts a simulated trajectory using \lilans  for a randomly chosen initial condition and set of parameters, $\mathbf{x}_0, \,\mathbf{p}$, from the test dataset. We observe that the predicted solution closely matches the numerical solution. \Cref{fig:err_table} shows that Sulzer and Buck II method produced slightly better relative error than \lilan, when averaged over all time steps for all test samples.  Sulzer and Buck I performed second best, followed very closely by \lilan. The difference in their averaged test errors was approximately $1.1 \cdot 10^{-5}$. \Cref{fig:err_table} also shows that DeepONet was an order of magnitude worse in terms of error than \lilans and Sulzer and Buck I and II, while NODE was notably inaccurate, lagging three orders of magnitude behind. The speedup achieved over the chosen numerical solver (BDF6 method) of \lilans was almost 1500x, when generating full trajectories from 1000 test initial conditions, as appears in \cref{fig:speed_table}.

\begin{figure}[h!t!b!] 
    \begin{minipage}{0.48\textwidth}
    \centering
    \resizebox{1.0\linewidth}{!}{\input{figs/charge_one_sample}}
    \end{minipage}
    \begin{minipage}{0.48\textwidth}
    \centering
    \resizebox{1.0\linewidth}{!}{\input{figs/CHARGE_rel_errors_new}}
    \end{minipage}
    \caption{Left to Right: Comparison between numerical solution and machine learning prediction by \lilans for a randomly chosen set of test initial conditions and parameters for CR charge state model (\cref{sec:CR-model}). All four charge states are predicted with a high degree of accuracy; Average point-wise relative error over time \eqref{eq:R1} when evaluating CR charge state model on test data. \lilans and Sulzer and Buck I and II have similar performance. DeepONet is slightly worse, with a larger error increase near $t=10^{-8s}$. NODE struggles particularly.}
    \label{fig:ChargeErr}
\end{figure}

{\bf{Expanding Latent Space and Hidden Dimension}}

\begin{figure}[h!t!b!]      
  \centering
  \input{figs/Expansion}
  \caption{Training and testing loss achieved by different architectures. The x-axis, in the context of "latent dimension expansion", represents the ratio of the latent space dimension to the original problem dimension, i.e. $\frac{m}{n_x}$ in \cref{method}. The x-axis, in the context of "hidden dimension expansion", represents the number of neurons in each hidden layer. Note that 1 is subtracted from the loss \eqref{eq:J} because the minimum of the loss function is 1. A much greater decrease in loss is achieved when the latent dimension is increased in comparison to increasing the hidden layer width.}
  \label{fig:expansion}
\end{figure} 
\Cref{fig:expansion} shows the change in training and test loss when different architectures are chosen, i.e. the dimension of the latent space and hidden layers were altered. For networks where the latent dimension was expanded (referred to as "latent dimension expansion" in \cref{fig:expansion}), all hidden layers were held at 100 neurons, while the latent dimension was allowed to vary. The x-axis of \cref{fig:expansion}, for the  "latent dimension expansion" case, represents the ratio of the latent space dimension to the original problem dimension, i.e. $\frac{m}{n_x}$ in \cref{method}. For networks where the hidden layer width was expanded (referred to as "hidden dimension expansion" in \cref{fig:expansion}), the latent dimension multiplier was kept at one, indicating that the latent space has the same dimension as the original system. The x-axis of \cref{fig:expansion}, for the "hidden dimension expansion" case, represents the number of neurons in each hidden layer. The result shows that much greater gains in accuracy are achieved by expanding the problem dimension to a higher-dimensional latent space. This agrees with our motivation during the development of \lilan, namely, that lifting a stiff problem to higher-dimensional latent space makes the problem easier to solve with high accuracy.
In contrast, expanding the dimension of hidden layers, while not allowing for latent expansion, resulted in very little gain in accuracy.

{\bf{Results on the full CR model}}

To reduce the memory requirements associated with training the full CR model, we use $4096$ data samples for training while $2500$ samples are reserved for testing. The choice of latent dimension $m$ for each method is provided in  \cref{late}, \Cref{appen_late}.
For this higher-dimensional problem, training a NODE was not computationally feasible on our hardware (NVIDIA Quadro RTX 5000) due to high memory requirements, even with the reduced training data set size. Hence, results for the NODE approach are not presented here. 
\Cref{fig:Full_CR_errs} (right) shows a comparison of the average point-wise relative error, calculated via \eqref{eq:R1}, for each method, when applied to the test dataset. We see that \lilans outperformed  all other methods. Another key observation from \cref{fig:Full_CR_errs} (right) is that, while all of the methods have a spike in error near $t=10^{-8}s$, \lilan's error increase is the smallest by a wide margin, remaining on $\mathcal{O}(10^{-2})$, while Sulzer and Buck I reaches $\mathcal{O}(10^{-1})$ and DeepONet and Sulzer and Buck II increase to $\mathcal{O}(10^{0})$. Recall from \cref{sec:chargestate}, that $t=10^{-8}s$ marks the approximate beginning of the region of the most rapid evolution of the dynamics, indicating that \lilans handles this complex behavior, which is a significant challenge to predict, the best of all methods considered. Moving to the left subfigure of \cref{fig:Full_CR_errs}, there is a depiction of the solution to the CR model using \lilans for a randomly chosen initial condition, $\mathbf{x}_0$, and set of parameters, $\mathbf{p}$, from the test dataset. The figure shows that \lilans produces a good approximation of the numerical solution. This approximation is particularly impressive near $t=10^{-7}s$, where several components of the solution collapse in a nearly discontinuous manner. Again, the average error achieved by different machine learning methods has been tabulated in \cref{fig:err_table}, and additional results on the speedup achieved over the chosen traditional numerical solver (BDF6 method) have been tabulated in \cref{fig:speed_table}. We observe from \cref{fig:err_table} that \lilans produced the lowest relative error in comparison to all other methods. 
\lilans outperformed DeepONet and Sulzer and Buck I by one order of magnitude. Sulzer and Buck II did not perform well on this problem and had two orders of magnitude greater error than \lilan. \Cref{fig:speed_table} reports the speedup achieved over the numerical solver (BDF6 method) by \lilan, which was approximately 800x in the same experiment as described in \cref{sec:Robertson,,sec:chargestate} (CR charge state model).

\label{sec:full-CR}


\begin{figure}[h!t!b!]   
  \begin{minipage}{0.43\textwidth}
  \resizebox{1.0\linewidth}{!}{\input{figs/plot_1159}}
  \end{minipage}
  \begin{minipage}{0.56\textwidth}
  \centering
  \input{figs/CR_rel_errors_new}
  \end{minipage}
  \caption{Left to Right: Comparison between numerical solution and machine learning prediction by \lilans for a random set of test initial conditions and parameters for full CR model (\cref{sec:CR-model}). The \lilan's prediction shows impressive accuracy for this highly-stiff, multiscale physics; Average point-wise relative error over time \eqref{eq:R1} when evaluating the full CR model on test data. Note that the \lilans surrogate has the smallest error growth near $t=10^{-8}s$, where the dynamics are evolving rapidly. }
  \label{fig:Full_CR_errs}
\end{figure}

%% file: figs/Expansion.tex
\begin{tikzpicture}

\definecolor{crimson2143940}{RGB}{214,39,40}
\definecolor{darkgray176}{RGB}{176,176,176}
\definecolor{darkorange25512714}{RGB}{255,127,14}
\definecolor{forestgreen4416044}{RGB}{44,160,44}
\definecolor{lightgray204}{RGB}{204,204,204}
\definecolor{steelblue31119180}{RGB}{31,119,180}

\begin{axis}[
legend cell align={left},
legend style={
  fill opacity=0.8,
  draw opacity=1,
  text opacity=1,
  at={(0.91,0.5)},
  anchor=east,
  draw=lightgray204
},
log basis y={10},
tick align=outside,
tick pos=left,
x grid style={darkgray176},
xlabel={Dimension},
xmin=1, xmax=50,
xtick style={color=black},
y grid style={darkgray176},
ylabel={Loss - 1},
ymin=0.00739155943673751, ymax=1.84913316260932,
ymode=log,
ytick style={color=black}
]
\addplot [ultra thick, steelblue31119180]
table {%
1 0.912515640258789
2 0.269077777862549
3 0.0242962837219238
4 0.0104358196258545
5 0.0122742652893066
6 0.00950050354003906
7 0.010007381439209
8 0.0109317302703857
9 0.0098799467086792
10 0.0103014707565308
12 0.0110207796096802
15 0.0102488994598389
18 0.0121402740478516
20 0.0120888948440552
22 0.0126101970672607
25 0.0111042261123657
28 0.0126948356628418
30 0.0101653337478638
32 0.0103832483291626
35 0.0123393535614014
38 0.0145386457443237
40 0.0110265016555786
42 0.0100284814834595
45 0.0154595375061035
48 0.0123429298400879
50 0.0151213407516479
};
\addlegendentry{\tiny Latent Dimension Expansion (Train)}
\addplot [ultra thick, darkorange25512714]
table {%
1 0.946917653083801
2 0.318706750869751
3 0.0301058292388916
4 0.0160820484161377
5 0.0187867879867554
6 0.0158431529998779
7 0.0811147689819336
8 0.0134366750717163
9 0.0124049186706543
10 0.0156028270721436
12 0.0170485973358154
15 0.014936089515686
18 0.0157601833343506
20 0.0159978866577148
22 0.0164306163787842
25 0.0203858613967896
28 0.0271339416503906
30 0.0138435363769531
32 0.0179600715637207
35 0.0229508876800537
38 0.0526807308197021
40 0.0204046964645386
42 0.0182240009307861
45 0.029407262802124
48 0.0183042287826538
50 0.0275795459747314
};
\addlegendentry{\tiny Latent Dimension Expansion (Test)}
\addplot [ultra thick, forestgreen4416044]
table {%
1 1.36205339431763
2 1.03193306922913
3 0.960177302360535
4 0.951293349266052
5 0.936946749687195
6 0.947170972824097
7 0.939349770545959
8 0.919877529144287
9 0.945632219314575
10 0.92163610458374
20 0.942105174064636
30 0.93574857711792
40 0.921094417572021
50 0.912097454071045
};
\addlegendentry{\tiny Hidden Dimension Expansion (Train)}
\addplot [ultra thick, crimson2143940]
table {%
1 1.43865823745728
2 1.06923317909241
3 0.998565196990967
4 0.984026193618774
5 0.971981644630432
6 0.982934236526489
7 0.977688908576965
8 0.95474100112915
9 0.975375771522522
10 0.952386975288391
20 0.976752042770386
30 0.971861720085144
40 0.956856489181519
50 0.951345920562744
};
\addlegendentry{\tiny Hidden Dimension Expansion (Test)}
\end{axis}

\end{tikzpicture}

%% file: PDEs.tex
\subsection{Allen-Cahn PDE in 1D}
\label{sec:AC}
The governing equation for the Allen-Cahn problem \cite{montanelli2020solvingperiodicsemilinearstiff} is given below:
\begin{equation}
    u_t = \epsilon u_{xx} + u - u^3,
    \label{sys_al}
\end{equation} where the solution is $u(x,t).$ The equation describes phase separation in iron alloys and has a second order diffusive term and a cubic reaction term \cite{montanelli2020solvingperiodicsemilinearstiff}. In order to write \eqref{sys_al} in the form \eqref{original_dynamics}, the spatial domain is discretized and integrated as a system of ODEs by the method of lines \cite{schiesser2012numerical}. The spatial domain, $x \in [0,2\pi]$ is discretized by 201 linearly spaced points leading to dimension $n_x=201$ in \eqref{original_dynamics}. In the below discussions, $\xb$ denotes the vector containing the ordered set of discrete points on the spatial domain.

\subsubsection{Data Generation}

Initial conditions are generated using a sum of three sinusoidal waves as follows:
\begin{equation}
    \label{eq:ac_initialconditions}
    \mathbf{u}_0 = \sum_{i=1}^3 \alpha sin(\beta \xb + \phi) + \alpha cos(\beta \xb + \phi)
\end{equation}
\begin{equation}
    \label{eq:ac_dists}
    \alpha \sim \mathcal{U}(0,1/6), \, \, \beta \sim \mathcal{U}_{discrete}(1,3), \, \, \phi \sim \mathcal{U}(0,2\pi),
\end{equation}
where $\mathcal{U}$ represents the continuous uniform distribution and  $\mathcal{U}_{discrete}$ denotes the discrete uniform distribution. The amplitude $\alpha$ and phase $\phi$ are sampled from continuous uniform distributions, while the frequency $\beta$ of each wave is sampled from a discrete uniform distribution where the frequencies $1$, $2$, or $3$ are the discrete choices, each with equal probability of being selected. The integer frequencies ensure that the solution, $\mathbf{u}$, is periodic on the problem domain. After the construction of the initial conditions, each is evolved in time using the ETDRK4 explicit, exponential integrator proposed in \cite{COX2002430}. The problem is solved over the time interval $t\in[0,10]$ seconds, which is discretized by $101$ linearly spaced points, (i.e. $M=101$ in \eqref{train_form}). The dataset contains $900$ samples for training, $100$ samples for validation, and an additional $200$ test samples, where each sample is a complete, time-integrated trajectory from a random initial condition generated by \eqref{eq:ac_initialconditions}.

\subsubsection{Results}
\label{resu_allen}
The results for this problem are presented in \cref{fig:ACex}, where 
the left subfigure shows the numerical solution, while the middle subfigure shows the solution produced by \lilan, from the same random initial condition, unseen during training. We observe that the solutions are visibly identical. The right subfigure in \cref{fig:ACex} shows the absolute point-wise error between \lilans and the numerical code. The errors are quite low nearly everywhere, except at phase boundaries where the solution quickly switches between $1$ and $-1$, indicated by a quick shift from red to blue in the two leftmost subfigures of \cref{fig:ACex}. This is a naturally challenging region to approximate, considering the sharp boundaries, yet the absolute pointwise error remains on $\mathcal{O}(10^{-2})$. In terms of performance over the entire test set, \cref{fig:ACerr} presents the average error over time \eqref{eq:R1}, for five machine learning approaches: \lilan, Sulzer and Buck I and II, DeepONet, and NODE. Each approach is trained with the same data, hyperparameter settings, and approximately the same total number of trainable parameters. Additionally, the choice of latent dimension $m$ for each method is provided in  \cref{late}, \Cref{appen_late}. In \cref{fig:ACerr}, NODE performs slightly better than \lilan \ during the initial, short time interval. This is because NODE begins integration from a known initial condition, $\xb_0$ \cite{Lee_2021}. With no error accumulation at the starting point, the solution computed by NODE is most accurate in the vicinity of $\xb_0$. \Cref{fig:ACerr} also shows that while \lilan\ is particularly effective (with very low relative error) between 0s-7s,
the approach by Sulzer and Buck \cite{sulzer2023speedingastrochemicalreactionnetworks} produced slightly lower error than \lilan\ between 7s-10s. We also observe that the relative errors of all methods converge to similar values toward the end of the simulation, suggesting a need for further investigation. However, the reason could be that all of the methods are nearly equally good for predicting the steady state of the problem, a phenomenon that was observed between \lilan, DeepONet, and Sulzer and Buck I and II in previous results \cref{sec:Robertson,,sec:chargestate} (\cref{fig:ROBerr,,fig:ChargeErr}). The average error achieved by different machine learning methods, for the Allen-Cahn PDE, are included in \cref{fig:err_table}, and report that the error produced by \lilans is lowest in comparison to Sulzer and Buck I and II, DeepONet, and NODE, though they are all on the same order of magnitude. \lilans achieved 14.3x speedup over the solver (ETDRK4 method), as shown in \cref{fig:speed_table}. The experiment to test the speed of \lilans versus the traditional solver is the same as in \cref{sec:Robertson}. The speedup is less radical than previous results sections because exponential integrators are already extremely efficient for simulating stiff PDEs, meaning any speedup is a significant success.

\begin{figure}[h!t!b!]      
  \centering
  \resizebox{1.0\linewidth}{!}{\input{figs/ACexample}}
  \caption{Left to right: Prediction by numerical method for Allen-Cahn PDE (\cref{sec:AC}); Prediction by \lilan; Point-wise absolute error between the two approaches. The largest errors are concentrated on sharp phase boundaries.}
  \label{fig:ACex}
\end{figure}


\begin{figure}[h!t!b!]      
  \centering
  \resizebox{0.5\linewidth}{!}{\input{figs/AC_errors_new}}
  \caption{Average point-wise relative error over time \eqref{eq:R1} when evaluating Allen-Cahn PDE (\cref{sec:AC}) on test data, for five machine learning approaches. Note that \lilans has the best accuracy in first six seconds where the phase separation dynamics take place, while all methods converge to similar errors in the later time region where steady state is typically reached and there is no more evolution in the solution.}
  \label{fig:ACerr}
\end{figure} 


\subsection{Cahn-Hilliard PDE in 1D}
\label{sec:CH}
The governing equation for the Cahn-Hilliard problem \cite{montanelli2020solvingperiodicsemilinearstiff} is given below: 
\begin{equation}
    u_t = \alpha \left( -u_{xx} - \gamma u_{xxxx} + (u^3)_{xx} \right)
\end{equation} where the solution is $u(x,t).$ This is another equation describing phase separation in alloys, but with much more complex physics due to the presence of the higher-order derivatives in the governing equation. Just as with the Allen-Cahn problem (\cref{sec:AC}), the spatial domain is discretized and integrated as a system of ODEs by the method of lines \cite{schiesser2012numerical}. The spatial domain, $x \in [-1,1]$ is discretized by 201 linearly spaced points, and is denoted by $\xb$ in the below discussions.

\subsubsection{Data Generation}

Initial conditions in the Cahn-Hilliard problem are generated as follows:
\begin{equation}
    \label{eq:ch_initialconditions}
    \mathbf{u}_0 = \alpha sin(\beta \pi (\xb+1) + \phi) + \alpha cos(\beta \pi (\xb+1) + \phi),
\end{equation}
\begin{equation}
    \label{eq:ch_dists}
    \alpha \sim \mathcal{U}(0.1,0.6), \, \, \beta \sim \mathcal{U}_{discrete}(1,3), \, \, \phi \sim \mathcal{U}(0,2\pi),
\end{equation}
where $\mathcal{U}$ represents the continuous uniform distribution and  $\mathcal{U}_{discrete}$ denotes the discrete uniform distribution.
For each initial condition, generated by \eqref{eq:ch_initialconditions}, the amplitude $\alpha$ and phase $\phi$ are sampled from continuous uniform distributions, specified in \eqref{eq:ch_dists},  while the frequency coefficient $\beta$ of the wave is sampled from a discrete uniform distribution where the frequencies $1$, $2$, or $3$ are the discrete choices, each with equal probability of being selected. The integer frequencies ensure that the solution, $\mathbf{u}$, is periodic on the problem domain. After the construction of the initial conditions, each is evolved in time using the ETDRK4 explicit, exponential integrator \cite{COX2002430}. The problem is solved over the time interval $t\in[0,20]$ seconds, which is discretized by $401$ linearly spaced points (i.e. M = 401 in \eqref{train_form}). The dataset contains $900$ samples for training, $100$ samples for validation, and an additional $200$ test samples, where each sample is a complete, time-integrated trajectory from a random initial condition generated by \eqref{eq:ch_initialconditions}.

\subsubsection{Results}
\begin{figure}[h!t!b!]      
  \centering
  \resizebox{1.0\linewidth}{!}{\input{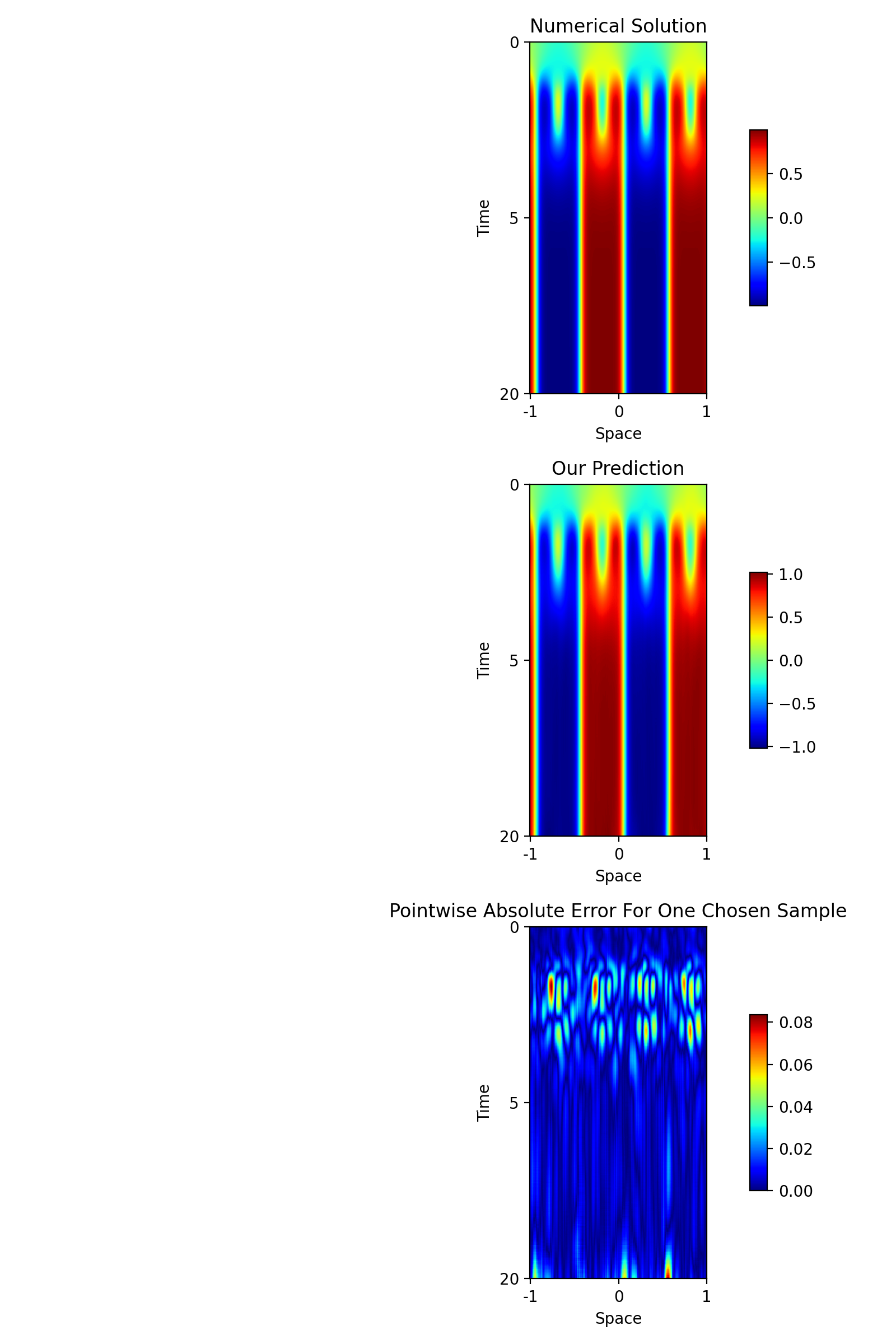}}
  \caption{Left to right: Prediction by numerical method for Cahn-Hilliard PDE (\cref{sec:CH}); Prediction by \lilan; Point-wise absolute error between the two approaches. The error concentrates early in the evolution, near phase boundaries, while steady state phase boundaries do not have as much significant error seen in \cref{sec:AC}.}
    \label{fig:CHex}
\end{figure}

\Cref{fig:CHex} (left) shows a solution generated by numerical code, while \cref{fig:CHex} (middle) shows the solution produced by \lilan, from the same initial condition that was not seen during training. The solutions are strikingly similar visibly. \Cref{fig:CHex} (right) shows the absolute point-wise error between our approach and the numerical code for the example initial condition considered. Notice that on this problem, the largest errors are not at steady state phase boundaries, in contrast to the Allen-Cahn PDE results from \cref{sec:AC}. Instead, high errors are most prevalent in the initial transient evolution before the solution reaches steady state, while the steady state has lower error than in \cref{sec:AC}. In terms of performance over the entire test set, \cref{fig:CHerr} depicts the average error over time \eqref{eq:R1}, for \lilan, Sulzer and Buck I and II, DeepONet, and NODE. Each approach is trained with the same data, hyperparameter settings, and approximately the same total number of trainable parameters. Again, the choice of latent dimension $m$ for each method is provided in  \cref{late}, \Cref{appen_late}. From \cref{fig:CHerr} we see that \lilans outperformed both NODE and DeepONet for prediction on the Cahn-Hilliard PDE. NODE performs slightly better than \lilans during the initial, short time interval. The reason for this has been highlighted in \cref{resu_allen}. We also observe that the approach of Sulzer and Buck I and II produce a very close relative error to that of \lilans towards the end of the simulation time. This, again, could be due to the fact that predicting the steady state is easier that predicting the time dependent evolution. Looking at the example in \cref{fig:CHex}, the evolution is done before $t=3s$, which is typical across the entire dataset, with some exceptions where the evolution is slower. Now, the region of \cref{fig:CHerr} prior to $t=4s$ is where \lilans performs the best, beating all other methods, suggesting that it is uniquely suited for handling the time dependent portion of the solution. \Cref{fig:err_table} illustrates that the average relative error produced by \lilans is lowest of all methods tested. Significantly, DeepONet has an order of magnitude greater error than \lilan, while the Sulzer and Buck I and II and NODE methods, though less accurate than \lilan, achieve errors on the same order of magnitude. The speedup achieved over the chosen traditional numerical solver (ETDRK4 method) is tabulated in \cref{fig:speed_table}, showing that \lilans was 13.7x faster when generating full trajectories from 1000 test initial conditions. Again, even though the speedup is less than that of the ODE problems, it is still impressive due to the competition being conducted against an exponential integrator.


\begin{figure}[h!t!b!]     
  \centering
  \resizebox{0.5\linewidth}{!}{\input{figs/CH_errors_new}}
  \caption{Average point-wise relative error over time \eqref{eq:R1} when evaluating Cahn-Hilliard PDE (\cref{sec:CH}) on test data, for five machine learning approaches.}
  \label{fig:CHerr}
\end{figure} 


%% file: figs/ACexample.tex
\begin{tikzpicture}

\definecolor{darkgray176}{RGB}{176,176,176}

\pgfplotsset{/pgfplots/group/.cd,
    horizontal sep=4.0cm,
    vertical sep=0.5cm
}

\begin{groupplot}[group style={group size=3 by 1}]
\nextgroupplot[
colorbar,
colorbar style={ylabel={Phase}},
colormap={mymap}{[1pt]
  rgb(0pt)=(0,0,0.5);
  rgb(22pt)=(0,0,1);
  rgb(25pt)=(0,0,1);
  rgb(68pt)=(0,0.86,1);
  rgb(70pt)=(0,0.9,0.967741935483871);
  rgb(75pt)=(0.0806451612903226,1,0.887096774193548);
  rgb(128pt)=(0.935483870967742,1,0.0322580645161291);
  rgb(130pt)=(0.967741935483871,0.962962962962963,0);
  rgb(132pt)=(1,0.925925925925926,0);
  rgb(178pt)=(1,0.0740740740740741,0);
  rgb(182pt)=(0.909090909090909,0,0);
  rgb(200pt)=(0.5,0,0)
},
point meta max=0.999999344348907,
point meta min=-0.999999344348907,
tick align=outside,
tick pos=left,
x grid style={darkgray176},
xlabel={(a)},
xmin=-0.5, xmax=201,
xtick style={color=black},
xtick={0,100.5,201},
xticklabels={0,$\pi$,2$\pi$},
y dir=reverse,
y grid style={darkgray176},
ylabel={Time (s)},
ymin=-0.5, ymax=101,
ytick style={color=black},
ytick={0,50.5,101},
yticklabels={0,5,10}
]
\addplot graphics [includegraphics cmd=\pgfimage,xmin=-0.5, xmax=200.5, ymin=100.5, ymax=-0.5] {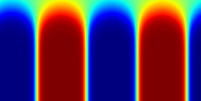};

\nextgroupplot[
colorbar,
colorbar style={ylabel={Phase}},
colormap={mymap}{[1pt]
  rgb(0pt)=(0,0,0.5);
  rgb(22pt)=(0,0,1);
  rgb(25pt)=(0,0,1);
  rgb(68pt)=(0,0.86,1);
  rgb(70pt)=(0,0.9,0.967741935483871);
  rgb(75pt)=(0.0806451612903226,1,0.887096774193548);
  rgb(128pt)=(0.935483870967742,1,0.0322580645161291);
  rgb(130pt)=(0.967741935483871,0.962962962962963,0);
  rgb(132pt)=(1,0.925925925925926,0);
  rgb(178pt)=(1,0.0740740740740741,0);
  rgb(182pt)=(0.909090909090909,0,0);
  rgb(200pt)=(0.5,0,0)
},
point meta max=1.00786602497101,
point meta min=-1.00861203670502,
tick align=outside,
tick pos=left,
x grid style={darkgray176},
xlabel={(b)},
xmin=-0.5, xmax=201,
xtick style={color=black},
xtick={0,100.5,201},
xticklabels={0,$\pi$,2$\pi$},
y dir=reverse,
y grid style={darkgray176},
ylabel={Time (s)},
ymin=-0.5, ymax=101,
ytick style={color=black},
ytick={0,50.5,101},
yticklabels={0,5,10}
]
\addplot graphics [includegraphics cmd=\pgfimage,xmin=-0.5, xmax=200.5, ymin=100.5, ymax=-0.5] {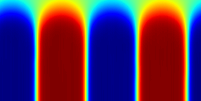};

\nextgroupplot[
colorbar,
colorbar style={ylabel={Absolute Error ($\cdot 10^{-2}$})},
colormap={mymap}{[1pt]
  rgb(0pt)=(0,0,0.5);
  rgb(22pt)=(0,0,1);
  rgb(25pt)=(0,0,1);
  rgb(68pt)=(0,0.86,1);
  rgb(70pt)=(0,0.9,0.967741935483871);
  rgb(75pt)=(0.0806451612903226,1,0.887096774193548);
  rgb(128pt)=(0.935483870967742,1,0.0322580645161291);
  rgb(130pt)=(0.967741935483871,0.962962962962963,0);
  rgb(132pt)=(1,0.925925925925926,0);
  rgb(178pt)=(1,0.0740740740740741,0);
  rgb(182pt)=(0.909090909090909,0,0);
  rgb(200pt)=(0.5,0,0)
},
point meta max=6.79640918970108,
point meta min=-3.64299699249033e-19,
tick align=outside,
tick pos=left,
x grid style={darkgray176},
xlabel={(c)},
xmin=-0.5, xmax=201,
xtick style={color=black},
xtick={0,100.5,201},
xticklabels={0,$\pi$,2$\pi$},
y dir=reverse,
y grid style={darkgray176},
ylabel={Time (s)},
ymin=-0.5, ymax=101,
ytick style={color=black},
ytick={0,50.5,101},
yticklabels={0,5,10}
]
\addplot graphics [includegraphics cmd=\pgfimage,xmin=-0.5, xmax=200.5, ymin=100.5, ymax=-0.5] {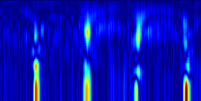};
\end{groupplot}

\end{tikzpicture}

%% file: figs/AC_errors_new.tex
\begin{tikzpicture}

\definecolor{crimson2143940}{RGB}{214,39,40}
\definecolor{darkgray176}{RGB}{176,176,176}
\definecolor{darkorange25512714}{RGB}{255,127,14}
\definecolor{forestgreen4416044}{RGB}{44,160,44}
\definecolor{lightgray204}{RGB}{204,204,204}
\definecolor{mediumpurple148103189}{RGB}{148,103,189}
\definecolor{steelblue31119180}{RGB}{31,119,180}

\begin{axis}[
legend cell align={left},
legend style={fill opacity=0.8, draw opacity=1, text opacity=1, draw=lightgray204},
log basis y={10},
tick align=outside,
tick pos=left,
title={Relative Error on Holdout Dataset},
x grid style={darkgray176},
xlabel={Time (s)},
xmin=0.1, xmax=10,
xtick style={color=black},
y grid style={darkgray176},
ylabel={Relative Error},
ymin=0.00414871499126675, ymax=0.411008569284888,
ymode=log,
ytick style={color=black},
ytick={0.0001,0.001,0.01,0.1,1,10},
yticklabels={
  \(\displaystyle {10^{-4}}\),
  \(\displaystyle {10^{-3}}\),
  \(\displaystyle {10^{-2}}\),
  \(\displaystyle {10^{-1}}\),
  \(\displaystyle {10^{0}}\),
  \(\displaystyle {10^{1}}\)
}
]
\addplot [semithick, steelblue31119180]
table {%
0.1 0.0119883567094803
0.2 0.0110961170867085
0.3 0.0103281373158097
0.4 0.00969468429684639
0.5 0.00909615214914083
0.6 0.00861850380897522
0.7 0.00823754724115133
0.8 0.00794451683759689
0.9 0.00771683035418391
1 0.00750828115269542
1.1 0.00742051843553782
1.2 0.00737359141930938
1.3 0.00738140847533941
1.4 0.0074049006216228
1.5 0.00745340343564749
1.6 0.007512585259974
1.7 0.0075917118228972
1.8 0.00769734429195523
1.9 0.00775890564545989
2 0.00788652990013361
2.1 0.0080210268497467
2.2 0.00816783029586077
2.3 0.00830724090337753
2.4 0.00844394043087959
2.5 0.00857485271990299
2.6 0.00868985988199711
2.7 0.00879298057407141
2.8 0.00889989733695984
2.9 0.00900874007493258
3 0.00912358239293098
3.1 0.00923950131982565
3.2 0.00938905961811543
3.3 0.00956257991492748
3.4 0.00976595003157854
3.5 0.00998338591307402
3.6 0.0102323079481721
3.7 0.0105214053764939
3.8 0.0108797615393996
3.9 0.011237371712923
4 0.0116246724501252
4.1 0.0120896147564054
4.2 0.0125752529129386
4.3 0.0130993407219648
4.4 0.0136653864756227
4.5 0.0143096158280969
4.6 0.0149826603010297
4.7 0.0157134626060724
4.8 0.0164816956967115
4.9 0.0173204820603132
5 0.0182364322245121
5.1 0.0191771388053894
5.2 0.020215256139636
5.3 0.0212675966322422
5.4 0.0223926771432161
5.5 0.0235312189906836
5.6 0.0247865580022335
5.7 0.0259838439524174
5.8 0.027306504547596
5.9 0.0285979956388474
6 0.0299647655338049
6.1 0.0313027240335941
6.2 0.0327143147587776
6.3 0.0340777710080147
6.4 0.0354963019490242
6.5 0.0368154905736446
6.6 0.0382428541779518
6.7 0.0395366288721561
6.8 0.0408989042043686
6.9 0.0421469695866108
7 0.0434409119188786
7.1 0.0446477346122265
7.2 0.0459023527801037
7.3 0.0470349080860615
7.4 0.048207052052021
7.5 0.0492702051997185
7.6 0.0503088571131229
7.7 0.051312044262886
7.8 0.0522263832390308
7.9 0.0531839840114117
8 0.0540042109787464
8.1 0.0548217706382275
8.2 0.0555419251322746
8.3 0.0562832541763783
8.4 0.0570031814277172
8.5 0.0577214322984219
8.6 0.0583183169364929
8.7 0.0589887239038944
8.8 0.0596428960561752
8.9 0.0602511540055275
9 0.0607197657227516
9.1 0.0613710470497608
9.2 0.0618055388331413
9.3 0.0622798055410385
9.4 0.0626786574721336
9.5 0.06307702511549
9.6 0.063531905412674
9.7 0.0639484450221062
9.8 0.0643648356199265
9.9 0.0647820979356766
10 0.0650938004255295
};
\addlegendentry{\tiny \lilans}
\addplot [semithick, darkorange25512714]
table {%
0.1 0.0550904348492622
0.2 0.0415665619075298
0.3 0.0356791019439697
0.4 0.0324432253837585
0.5 0.0300912596285343
0.6 0.0283588878810406
0.7 0.0271893441677094
0.8 0.0265321768820286
0.9 0.0262212697416544
1 0.0260832067579031
1.1 0.0260493028908968
1.2 0.0260095708072186
1.3 0.0259543173015118
1.4 0.0258781425654888
1.5 0.0257774852216244
1.6 0.0256871320307255
1.7 0.0256720781326294
1.8 0.0257739797234535
1.9 0.0260158982127905
2 0.0263919830322266
2.1 0.0268420446664095
2.2 0.0273080132901669
2.3 0.0276980418711901
2.4 0.027979651466012
2.5 0.0281132571399212
2.6 0.0281166359782219
2.7 0.0279883705079556
2.8 0.0277791209518909
2.9 0.0275422763079405
3 0.0273163244128227
3.1 0.0271501243114471
3.2 0.0270875114947557
3.3 0.0271625556051731
3.4 0.0274016615003347
3.5 0.0277782753109932
3.6 0.0283013340085745
3.7 0.0289673041552305
3.8 0.0297139789909124
3.9 0.0305219516158104
4 0.0313617810606956
4.1 0.0322036072611809
4.2 0.0330289974808693
4.3 0.0338142588734627
4.4 0.034558117389679
4.5 0.0352404527366161
4.6 0.0358617566525936
4.7 0.0364175103604794
4.8 0.0369164422154427
4.9 0.0373488143086433
5 0.0377306565642357
5.1 0.0380783900618553
5.2 0.0383727550506592
5.3 0.0386489033699036
5.4 0.0388970226049423
5.5 0.0391130000352859
5.6 0.0393070355057716
5.7 0.0395013056695461
5.8 0.0396689809858799
5.9 0.0398347564041615
6 0.0399877578020096
6.1 0.0401379466056824
6.2 0.0402852818369865
6.3 0.0404122546315193
6.4 0.0405569262802601
6.5 0.0406854674220085
6.6 0.040831446647644
6.7 0.0409612841904163
6.8 0.041109886020422
6.9 0.0412533171474934
7 0.0413921400904655
7.1 0.0415539257228374
7.2 0.0417122766375542
7.3 0.0418812446296215
7.4 0.042060486972332
7.5 0.0422455668449402
7.6 0.0424335673451424
7.7 0.0426282770931721
7.8 0.0428346619009972
7.9 0.0430298671126366
8 0.0432335734367371
8.1 0.0434402078390121
8.2 0.0436405017971992
8.3 0.0438375882804394
8.4 0.0440362170338631
8.5 0.0442371852695942
8.6 0.0444242171943188
8.7 0.0446248613297939
8.8 0.0448205061256886
8.9 0.0450124479830265
9 0.0452070832252502
9.1 0.0453975945711136
9.2 0.045598428696394
9.3 0.0457931235432625
9.4 0.0459833294153214
9.5 0.0461893826723099
9.6 0.0463771633803844
9.7 0.0465787015855312
9.8 0.0467757396399975
9.9 0.0469784401357174
10 0.0471733473241329
};
\addlegendentry{\tiny Sulzer and Buck I}
\addplot [semithick, forestgreen4416044]
table {%
0.1 0.0555080883204937
0.2 0.0396915599703789
0.3 0.0338386259973049
0.4 0.0311508625745773
0.5 0.028923524543643
0.6 0.0268676988780499
0.7 0.025323648005724
0.8 0.0244902223348618
0.9 0.0242363456636667
1 0.0242201518267393
1.1 0.0241562034934759
1.2 0.0239880029112101
1.3 0.0237732548266649
1.4 0.0235981028527021
1.5 0.0235508047044277
1.6 0.0237278714776039
1.7 0.0241423975676298
1.8 0.0248112957924604
1.9 0.0256625507026911
2 0.0266085807234049
2.1 0.0275256335735321
2.2 0.0283418651670218
2.3 0.0289767496287823
2.4 0.0294031091034412
2.5 0.0296053513884544
2.6 0.0296268984675407
2.7 0.0294838063418865
2.8 0.0292285773903131
2.9 0.0289220809936523
3 0.0286028906702995
3.1 0.0283192917704582
3.2 0.0281149055808783
3.3 0.0280286259949207
3.4 0.0280784890055656
3.5 0.0282813105732203
3.6 0.0286342948675156
3.7 0.0291228462010622
3.8 0.0297151897102594
3.9 0.0303842592984438
4 0.0310858283191919
4.1 0.0317831113934517
4.2 0.0324627906084061
4.3 0.0330965518951416
4.4 0.0336760357022285
4.5 0.0341938883066177
4.6 0.0346461497247219
4.7 0.0350352264940739
4.8 0.0353653579950333
4.9 0.0356336757540703
5 0.0358522795140743
5.1 0.0360346548259258
5.2 0.0361784659326077
5.3 0.0362937860190868
5.4 0.0363845862448215
5.5 0.0364543609321117
5.6 0.0365095697343349
5.7 0.0365587212145329
5.8 0.0365923009812832
5.9 0.0366215035319328
6 0.0366544872522354
6.1 0.0366781540215015
6.2 0.0367082878947258
6.3 0.0367268845438957
6.4 0.0367591232061386
6.5 0.0367824546992779
6.6 0.0368180014193058
6.7 0.0368541404604912
6.8 0.0368964709341526
6.9 0.0369562059640884
7 0.0370142310857773
7.1 0.0370865352451801
7.2 0.03717952221632
7.3 0.0372825898230076
7.4 0.0373982675373554
7.5 0.0375330708920956
7.6 0.0376727767288685
7.7 0.0378335416316986
7.8 0.0379947833716869
7.9 0.0381629168987274
8 0.0383347533643246
8.1 0.0385059714317322
8.2 0.0386797450482845
8.3 0.0388490743935108
8.4 0.0390218906104565
8.5 0.0391950830817223
8.6 0.0393632911145687
8.7 0.039532758295536
8.8 0.0396974347531796
8.9 0.0398662090301514
9 0.0400355122983456
9.1 0.0402090027928352
9.2 0.0403712466359138
9.3 0.0405468232929707
9.4 0.0407169759273529
9.5 0.0408913716673851
9.6 0.0410664081573486
9.7 0.0412462167441845
9.8 0.0414186380803585
9.9 0.0415950939059258
10 0.0417812019586563
};
\addlegendentry{\tiny Sulzer and Buck II}
\addplot [semithick, crimson2143940]
table {%
0.1 0.333523511886597
0.2 0.242778956890106
0.3 0.167933076620102
0.4 0.107892423868179
0.5 0.063677042722702
0.6 0.0427313446998596
0.7 0.0552459023892879
0.8 0.0767390206456184
0.9 0.0962200686335564
1 0.112279333174229
1.1 0.124843083322048
1.2 0.134345427155495
1.3 0.141044139862061
1.4 0.145350083708763
1.5 0.147561982274055
1.6 0.147963255643845
1.7 0.146780088543892
1.8 0.144257366657257
1.9 0.140561521053314
2 0.135860919952393
2.1 0.130323231220245
2.2 0.124118074774742
2.3 0.117369621992111
2.4 0.110250473022461
2.5 0.102822929620743
2.6 0.0952680930495262
2.7 0.0877589955925941
2.8 0.0803309828042984
2.9 0.0732227489352226
3 0.0665652006864548
3.1 0.0605833381414413
3.2 0.0554812997579575
3.3 0.051905021071434
3.4 0.0499050430953503
3.5 0.0489998050034046
3.6 0.0491627678275108
3.7 0.0500870421528816
3.8 0.0516633838415146
3.9 0.053661897778511
4 0.0559011436998844
4.1 0.0583132542669773
4.2 0.0607080534100533
4.3 0.063090868294239
4.4 0.0651656091213226
4.5 0.0671262368559837
4.6 0.0687756538391113
4.7 0.0702291280031204
4.8 0.0713486820459366
4.9 0.0722377225756645
5 0.0728081986308098
5.1 0.0732725858688354
5.2 0.0734600350260735
5.3 0.0733638629317284
5.4 0.0731442049145699
5.5 0.0728824138641357
5.6 0.0724643915891647
5.7 0.0720060244202614
5.8 0.071489043533802
5.9 0.0709386095404625
6 0.0705046728253365
6.1 0.0700855702161789
6.2 0.0698046535253525
6.3 0.0696672648191452
6.4 0.0696178898215294
6.5 0.0696892589330673
6.6 0.0698858350515366
6.7 0.0702226981520653
6.8 0.0706159546971321
6.9 0.0711414963006973
7 0.0716958045959473
7.1 0.0722432211041451
7.2 0.0728276148438454
7.3 0.0734265148639679
7.4 0.0738940313458443
7.5 0.0744207873940468
7.6 0.0748018249869347
7.7 0.0750440135598183
7.8 0.0753322616219521
7.9 0.0753835290670395
8 0.0754517316818237
8.1 0.0753529369831085
8.2 0.0751254484057426
8.3 0.0748866498470306
8.4 0.0746099948883057
8.5 0.0741569399833679
8.6 0.0737342610955238
8.7 0.0733181685209274
8.8 0.0729953274130821
8.9 0.0727028101682663
9 0.0725641027092934
9.1 0.0726379379630089
9.2 0.0730279982089996
9.3 0.0737375989556313
9.4 0.0747739896178246
9.5 0.0761970430612564
9.6 0.0780466943979263
9.7 0.0804171711206436
9.8 0.08310866355896
9.9 0.0861125066876411
10 0.0895596742630005
};
\addlegendentry{\tiny DeepONet}
\addplot [semithick, mediumpurple148103189]
table {%
0.1 0.00511255534365773
0.2 0.00961453560739756
0.3 0.013587118126452
0.4 0.0170861519873142
0.5 0.0201549120247364
0.6 0.0228336658328772
0.7 0.02516377158463
0.8 0.0271898247301579
0.9 0.0289546251296997
1 0.0305037293583155
1.1 0.0318797938525677
1.2 0.0331173241138458
1.3 0.0342434272170067
1.4 0.035271804779768
1.5 0.0362094677984715
1.6 0.0370627455413342
1.7 0.0378395840525627
1.8 0.0385514311492443
1.9 0.0392147041857243
2 0.0398473665118217
2.1 0.0404695682227612
2.2 0.0411008819937706
2.3 0.0417587645351887
2.4 0.042457427829504
2.5 0.043207086622715
2.6 0.0440141148865223
2.7 0.0448813959956169
2.8 0.0458085350692272
2.9 0.0467932373285294
3 0.0478318594396114
3.1 0.0489187575876713
3.2 0.0500486679375172
3.3 0.0512135699391365
3.4 0.0524048879742622
3.5 0.0536136999726295
3.6 0.0548287369310856
3.7 0.0560390278697014
3.8 0.0572330467402935
3.9 0.0583985112607479
4 0.0595234483480453
4.1 0.0605971366167068
4.2 0.0616114400327206
4.3 0.0625608041882515
4.4 0.0634422302246094
4.5 0.064256526529789
4.6 0.0650050118565559
4.7 0.0656909197568893
4.8 0.0663192346692085
4.9 0.0668939352035522
5 0.0674204528331757
5.1 0.0679049491882324
5.2 0.068352535367012
5.3 0.0687686800956726
5.4 0.0691580250859261
5.5 0.0695245414972305
5.6 0.0698726624250412
5.7 0.0702055618166924
5.8 0.070525735616684
5.9 0.0708360970020294
6 0.0711389407515526
6.1 0.0714367255568504
6.2 0.0717309713363647
6.3 0.0720232874155045
6.4 0.0723156556487083
6.5 0.0726097896695137
6.6 0.0729070156812668
6.7 0.0732090100646019
6.8 0.0735170766711235
6.9 0.0738331228494644
7 0.0741589367389679
7.1 0.0744960829615593
7.2 0.0748463720083237
7.3 0.0752114504575729
7.4 0.075593613088131
7.5 0.0759943127632141
7.6 0.076415590941906
7.7 0.0768584981560707
7.8 0.0773231461644173
7.9 0.0778092294931412
8 0.0783162638545036
8.1 0.0788429975509644
8.2 0.0793878361582756
8.3 0.0799495130777359
8.4 0.0805262327194214
8.5 0.0811168253421783
8.6 0.0817203000187874
8.7 0.0823360681533813
8.8 0.0829637497663498
8.9 0.0836033821105957
9 0.0842537879943848
9.1 0.0849152579903603
9.2 0.0855879187583923
9.3 0.0862715318799019
9.4 0.0869654268026352
9.5 0.0876700356602669
9.6 0.088384747505188
9.7 0.0891098827123642
9.8 0.0898450836539268
9.9 0.0905904099345207
10 0.0913455784320831
};
\addlegendentry{\tiny NODE}
\end{axis}

\end{tikzpicture}

%% file: figs/CHexample.tex
\begin{tikzpicture}

\definecolor{darkgray176}{RGB}{176,176,176}

\pgfplotsset{/pgfplots/group/.cd,
    horizontal sep=4.0cm,
    vertical sep=0.5cm
}

\begin{groupplot}[group style={group size=3 by 1}]
\nextgroupplot[
colorbar,
colorbar style={ylabel={Phase}},
colormap={mymap}{[1pt]
  rgb(0pt)=(0,0,0.5);
  rgb(22pt)=(0,0,1);
  rgb(25pt)=(0,0,1);
  rgb(68pt)=(0,0.86,1);
  rgb(70pt)=(0,0.9,0.967741935483871);
  rgb(75pt)=(0.0806451612903226,1,0.887096774193548);
  rgb(128pt)=(0.935483870967742,1,0.0322580645161291);
  rgb(130pt)=(0.967741935483871,0.962962962962963,0);
  rgb(132pt)=(1,0.925925925925926,0);
  rgb(178pt)=(1,0.0740740740740741,0);
  rgb(182pt)=(0.909090909090909,0,0);
  rgb(200pt)=(0.5,0,0)
},
point meta max=0.999930322170258,
point meta min=-0.999930322170258,
tick align=outside,
tick pos=left,
x grid style={darkgray176},
xlabel={(a)},
xmin=-0.5, xmax=201,
xtick style={color=black},
xtick={0,100.5,201},
xticklabels={-1,0,1},
y dir=reverse,
y grid style={darkgray176},
ylabel={Time (s)},
ymin=-0.5, ymax=401,
ytick style={color=black},
ytick={0,200.5,401},
yticklabels={0,5,20}
]
\addplot graphics [includegraphics cmd=\pgfimage,xmin=-0.5, xmax=200.5, ymin=400.5, ymax=-0.5] {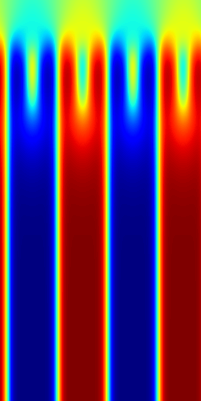};

\nextgroupplot[
colorbar,
colorbar style={ylabel={Phase}},
colormap={mymap}{[1pt]
  rgb(0pt)=(0,0,0.5);
  rgb(22pt)=(0,0,1);
  rgb(25pt)=(0,0,1);
  rgb(68pt)=(0,0.86,1);
  rgb(70pt)=(0,0.9,0.967741935483871);
  rgb(75pt)=(0.0806451612903226,1,0.887096774193548);
  rgb(128pt)=(0.935483870967742,1,0.0322580645161291);
  rgb(130pt)=(0.967741935483871,0.962962962962963,0);
  rgb(132pt)=(1,0.925925925925926,0);
  rgb(178pt)=(1,0.0740740740740741,0);
  rgb(182pt)=(0.909090909090909,0,0);
  rgb(200pt)=(0.5,0,0)
},
point meta max=1.02281224727631,
point meta min=-1.02118098735809,
tick align=outside,
tick pos=left,
x grid style={darkgray176},
xlabel={(b)},
xmin=-0.5, xmax=201,
xtick style={color=black},
xtick={0,100.5,201},
xticklabels={-1,0,1},
y dir=reverse,
y grid style={darkgray176},
ylabel={Time (s)},
ymin=-0.5, ymax=401,
ytick style={color=black},
ytick={0,200.5,401},
yticklabels={0,5,20}
]
\addplot graphics [includegraphics cmd=\pgfimage,xmin=-0.5, xmax=200.5, ymin=400.5, ymax=-0.5] {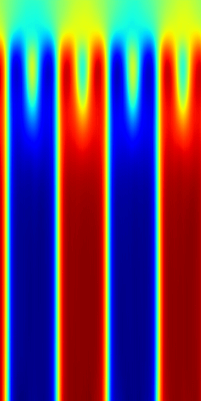};

\nextgroupplot[
colorbar,
colorbar style={ylabel={Absolute Error ($\cdot 10^{-2}$})},
colormap={mymap}{[1pt]
  rgb(0pt)=(0,0,0.5);
  rgb(22pt)=(0,0,1);
  rgb(25pt)=(0,0,1);
  rgb(68pt)=(0,0.86,1);
  rgb(70pt)=(0,0.9,0.967741935483871);
  rgb(75pt)=(0.0806451612903226,1,0.887096774193548);
  rgb(128pt)=(0.935483870967742,1,0.0322580645161291);
  rgb(130pt)=(0.967741935483871,0.962962962962963,0);
  rgb(132pt)=(1,0.925925925925926,0);
  rgb(178pt)=(1,0.0740740740740741,0);
  rgb(182pt)=(0.909090909090909,0,0);
  rgb(200pt)=(0.5,0,0)
},
point meta max=8.36205780506134,
point meta min=-1.16240341400596e-18,
tick align=outside,
tick pos=left,
x grid style={darkgray176},
xlabel={(c)},
xmin=-0.5, xmax=201,
xtick style={color=black},
xtick={0,100.5,201},
xticklabels={-1,0,1},
y dir=reverse,
y grid style={darkgray176},
ylabel={Time (s)},
ymin=-0.5, ymax=401,
ytick style={color=black},
ytick={0,200.5,401},
yticklabels={0,5,20}
]
\addplot graphics [includegraphics cmd=\pgfimage,xmin=-0.5, xmax=200.5, ymin=400.5, ymax=-0.5] {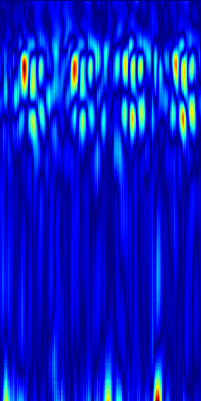};
\end{groupplot}

\end{tikzpicture}

%% file: figs/CH_errors_new.tex
\begin{tikzpicture}

\definecolor{crimson2143940}{RGB}{214,39,40}
\definecolor{darkgray176}{RGB}{176,176,176}
\definecolor{darkorange25512714}{RGB}{255,127,14}
\definecolor{forestgreen4416044}{RGB}{44,160,44}
\definecolor{lightgray204}{RGB}{204,204,204}
\definecolor{mediumpurple148103189}{RGB}{148,103,189}
\definecolor{steelblue31119180}{RGB}{31,119,180}

\begin{axis}[
legend cell align={left},
legend style={
  fill opacity=0.8,
  draw opacity=1,
  text opacity=1,
  at={(0.03,0.03)},
  anchor=south west,
  draw=lightgray204
},
log basis y={10},
tick align=outside,
tick pos=left,
title={Relative Error on Holdout Dataset},
x grid style={darkgray176},
xlabel={Time (s)},
xmin=0.025, xmax=10,
xtick style={color=black},
y grid style={darkgray176},
ylabel={Relative Error},
ymin=0.00818844845578246, ymax=0.443689696362379,
ymode=log,
ytick style={color=black},
ytick={0.0001,0.001,0.01,0.1,1,10},
yticklabels={
  \(\displaystyle {10^{-4}}\),
  \(\displaystyle {10^{-3}}\),
  \(\displaystyle {10^{-2}}\),
  \(\displaystyle {10^{-1}}\),
  \(\displaystyle {10^{0}}\),
  \(\displaystyle {10^{1}}\)
}
]
\addplot [semithick, steelblue31119180]
table {%
0.025 0.0687038600444794
0.05 0.0626419112086296
0.075 0.0573688223958015
0.1 0.0528988875448704
0.125 0.0494628809392452
0.15 0.0468039885163307
0.175 0.044955063611269
0.2 0.0435466170310974
0.225 0.0424666292965412
0.25 0.0416184328496456
0.275 0.0409127473831177
0.3 0.0403225682675838
0.325 0.0397958382964134
0.35 0.0392971970140934
0.375 0.0388238243758678
0.4 0.0383814722299576
0.425 0.0379613973200321
0.45 0.037496380507946
0.475 0.0370298586785793
0.5 0.0365990810096264
0.525 0.0361213348805904
0.55 0.0356381200253963
0.575 0.0351577028632164
0.6 0.0347087122499943
0.625 0.0341841317713261
0.65 0.0337474755942822
0.675 0.0332857631146908
0.7 0.0328336097300053
0.725 0.0323757827281952
0.75 0.0319897644221783
0.775 0.0315455086529255
0.8 0.0312002263963223
0.825 0.0308092795312405
0.85 0.0304934550076723
0.875 0.0302412174642086
0.9 0.0300442352890968
0.925 0.0298896487802267
0.95 0.029727766290307
0.975 0.0296451188623905
1 0.0296130143105984
1.025 0.0295975655317307
1.05 0.0296275578439236
1.075 0.029626689851284
1.1 0.0296586360782385
1.125 0.0296650733798742
1.15 0.0296811424195766
1.175 0.0297101158648729
1.2 0.0296961292624474
1.225 0.0296959280967712
1.25 0.0296752899885178
1.275 0.0296703781932592
1.3 0.0296437405049801
1.325 0.0296302493661642
1.35 0.0296511352062225
1.375 0.0296589583158493
1.4 0.0297534912824631
1.425 0.0298293493688107
1.45 0.0300005171447992
1.475 0.0302034523338079
1.5 0.0305318348109722
1.525 0.0309467054903507
1.55 0.0314377509057522
1.575 0.0319379791617393
1.6 0.0324882790446281
1.625 0.0330222547054291
1.65 0.0336044579744339
1.675 0.0341241434216499
1.7 0.0346709452569485
1.725 0.0351030118763447
1.75 0.0355593897402287
1.775 0.035893902182579
1.8 0.0362443551421165
1.825 0.0364746600389481
1.85 0.0367801934480667
1.875 0.0369482040405273
1.9 0.0371132865548134
1.925 0.0373468957841396
1.95 0.0374584794044495
1.975 0.0376630164682865
2 0.0378159023821354
2.025 0.0381130650639534
2.05 0.0384553521871567
2.075 0.038955956697464
2.1 0.0394401624798775
2.125 0.0400153808295727
2.15 0.0405001491308212
2.175 0.0410382971167564
2.2 0.0414973571896553
2.225 0.0420067869126797
2.25 0.0424358062446117
2.275 0.0429279990494251
2.3 0.0433614738285542
2.325 0.0437903478741646
2.35 0.0441922098398209
2.375 0.0446213223040104
2.4 0.0449686236679554
2.425 0.0453417375683784
2.45 0.0456675216555595
2.475 0.046021718531847
2.5 0.0464594215154648
2.525 0.0469109751284122
2.55 0.0472877509891987
2.575 0.0476707071065903
2.6 0.0479568652808666
2.625 0.0481217429041862
2.65 0.0482341647148132
2.675 0.0483100600540638
2.7 0.0482733808457851
2.725 0.0481638088822365
2.75 0.0480394177138805
2.775 0.0479212217032909
2.8 0.0478220842778683
2.825 0.0477158315479755
2.85 0.0476805679500103
2.875 0.0476129613816738
2.9 0.0476127341389656
2.925 0.0476447269320488
2.95 0.0477442927658558
2.975 0.047823004424572
3 0.0480618886649609
3.025 0.0483517050743103
3.05 0.0487278215587139
3.075 0.0490169413387775
3.1 0.0494117736816406
3.125 0.0498293489217758
3.15 0.0502279065549374
3.175 0.0506756454706192
3.2 0.0510177947580814
3.225 0.0515494719147682
3.25 0.0521084293723106
3.275 0.0526683777570724
3.3 0.0531244650483131
3.325 0.0536750070750713
3.35 0.0542189739644527
3.375 0.0547608472406864
3.4 0.0551984496414661
3.425 0.0557506643235683
3.45 0.0562779791653156
3.475 0.056791178882122
3.5 0.0571602135896683
3.525 0.0575824342668056
3.55 0.0579178035259247
3.575 0.0581739991903305
3.6 0.0583285614848137
3.625 0.0584445372223854
3.65 0.0584853626787663
3.675 0.0584583729505539
3.7 0.0583409667015076
3.725 0.0582337938249111
3.75 0.0580868199467659
3.775 0.0579706951975822
3.8 0.0578490607440472
3.825 0.0576881170272827
3.85 0.0576049610972404
3.875 0.0575300790369511
3.9 0.0574453286826611
3.925 0.0573711134493351
3.95 0.0573077574372292
3.975 0.0572476089000702
4 0.057186096906662
4.025 0.0570431426167488
4.05 0.0569422468543053
4.075 0.0567924380302429
4.1 0.0566544979810715
4.125 0.0564519874751568
4.15 0.0562404617667198
4.175 0.0560508221387863
4.2 0.0558093711733818
4.225 0.0556309223175049
4.25 0.0553821250796318
4.275 0.0551882721483707
4.3 0.0549850352108479
4.325 0.0547609403729439
4.35 0.0545013323426247
4.375 0.0542342849075794
4.4 0.0539338923990726
4.425 0.0536898709833622
4.45 0.0534891225397587
4.475 0.0531792640686035
4.5 0.0529682710766792
4.525 0.0526790283620358
4.55 0.0525480434298515
4.575 0.0524395927786827
4.6 0.0521686412394047
4.625 0.0520720854401588
4.65 0.0519948527216911
4.675 0.0518408566713333
4.7 0.0517451018095016
4.725 0.0516268350183964
4.75 0.0516708567738533
4.775 0.0516181848943233
4.8 0.0516542792320251
4.825 0.0516661927103996
4.85 0.0518411062657833
4.875 0.0519742667675018
4.9 0.0521668381989002
4.925 0.0523592829704285
4.95 0.0526778399944305
4.975 0.0529003515839577
5 0.0530441254377365
5.025 0.053215067833662
5.05 0.0532949194312096
5.075 0.0533630326390266
5.1 0.0533962622284889
5.125 0.0535384453833103
5.15 0.053401030600071
5.175 0.0532803982496262
5.2 0.0530775710940361
5.225 0.0527911856770515
5.25 0.0525639727711678
5.275 0.0522420406341553
5.3 0.0519061833620071
5.325 0.0517906472086906
5.35 0.0513795353472233
5.375 0.05101028829813
5.4 0.0506229773163795
5.425 0.0502271354198456
5.45 0.0498824194073677
5.475 0.0495509505271912
5.5 0.0492118746042252
5.525 0.0490916520357132
5.55 0.048793014138937
5.575 0.0485814362764359
5.6 0.048271119594574
5.625 0.048057459294796
5.65 0.0478407293558121
5.675 0.0476545616984367
5.7 0.0474555492401123
5.725 0.0472662635147572
5.75 0.0471768081188202
5.775 0.0470796637237072
5.8 0.047109492123127
5.825 0.0471557639539242
5.85 0.0473250932991505
5.875 0.0475196242332458
5.9 0.0477788113057613
5.925 0.0479980371892452
5.95 0.0481975264847279
5.975 0.0483948029577732
6 0.0485725589096546
6.025 0.0487351417541504
6.05 0.0488741956651211
6.075 0.0490149296820164
6.1 0.0490998551249504
6.125 0.049280259758234
6.15 0.049505203962326
6.175 0.0496655255556107
6.2 0.0497769899666309
6.225 0.0499703735113144
6.25 0.0501581951975822
6.275 0.0502982474863529
6.3 0.0504869893193245
6.325 0.0505555421113968
6.35 0.0507003962993622
6.375 0.0508891269564629
6.4 0.0509419366717339
6.425 0.0510693863034248
6.45 0.0511247925460339
6.475 0.0512521788477898
6.5 0.0513264387845993
6.525 0.0514348223805428
6.55 0.051535289734602
6.575 0.0516682267189026
6.6 0.0518536940217018
6.625 0.0520101152360439
6.65 0.0522233843803406
6.675 0.0524019226431847
6.7 0.0526455007493496
6.725 0.0529379360377789
6.75 0.0531475245952606
6.775 0.0535141080617905
6.8 0.0538123771548271
6.825 0.0541013665497303
6.85 0.0544465519487858
6.875 0.0547425784170628
6.9 0.0550600625574589
6.925 0.0553903169929981
6.95 0.0556763559579849
6.975 0.0559542365372181
7 0.0563160963356495
7.025 0.0565835162997246
7.05 0.0569059252738953
7.075 0.0571883395314217
7.1 0.0573540143668652
7.125 0.0575999096035957
7.15 0.0578237809240818
7.175 0.0580247640609741
7.2 0.0582889169454575
7.225 0.058446753770113
7.25 0.0585784614086151
7.275 0.0586658827960491
7.3 0.0587167479097843
7.325 0.0586831644177437
7.35 0.0587266869843006
7.375 0.0586189813911915
7.4 0.0585337802767754
7.425 0.0582130402326584
7.45 0.0578917525708675
7.475 0.057419128715992
7.5 0.0568727105855942
7.525 0.0562002584338188
7.55 0.0554432757198811
7.575 0.0546262711286545
7.6 0.0537059307098389
7.625 0.0528667829930782
7.65 0.0518794618546963
7.675 0.0509127303957939
7.7 0.0500014871358871
7.725 0.0491094589233398
7.75 0.0483165830373764
7.775 0.0475391633808613
7.8 0.0469183214008808
7.825 0.0463466346263885
7.85 0.0457866080105305
7.875 0.0452989488840103
7.9 0.0449240133166313
7.925 0.0445943996310234
7.95 0.044320534914732
7.975 0.0440833829343319
8 0.0439229868352413
8.025 0.0437728688120842
8.05 0.0437143333256245
8.075 0.0436144545674324
8.1 0.0435698218643665
8.125 0.043541431427002
8.15 0.0435646623373032
8.175 0.0435647182166576
8.2 0.0435960665345192
8.225 0.0436159297823906
8.25 0.0436863601207733
8.275 0.0437189750373363
8.3 0.0438146404922009
8.325 0.0438806191086769
8.35 0.0439238734543324
8.375 0.0440133139491081
8.4 0.0440903455018997
8.425 0.0441934950649738
8.45 0.0442935489118099
8.475 0.0443747453391552
8.5 0.0444970987737179
8.525 0.0445798970758915
8.55 0.0446883477270603
8.575 0.0447917133569717
8.6 0.0448791570961475
8.625 0.0449804626405239
8.65 0.0450715720653534
8.675 0.0451857782900333
8.7 0.0453023985028267
8.725 0.0454058237373829
8.75 0.0454924367368221
8.775 0.0456169508397579
8.8 0.0457219965755939
8.825 0.0457948967814445
8.85 0.0458986833691597
8.875 0.0460361167788506
8.9 0.0461570918560028
8.925 0.0462784618139267
8.95 0.046352531760931
8.975 0.0465001091361046
9 0.046617578715086
9.025 0.0467367842793465
9.05 0.0468312650918961
9.075 0.0469899773597717
9.1 0.0470906533300877
9.125 0.0472484678030014
9.15 0.0473489463329315
9.175 0.0474615432322025
9.2 0.0476336292922497
9.225 0.047747440636158
9.25 0.0478903651237488
9.275 0.0480486899614334
9.3 0.0481571517884731
9.325 0.0482882298529148
9.35 0.0484494864940643
9.375 0.0485595613718033
9.4 0.0486933887004852
9.425 0.0488503761589527
9.45 0.048949871212244
9.475 0.0490605682134628
9.5 0.0491955690085888
9.525 0.0493315532803535
9.55 0.049444168806076
9.575 0.0495945923030376
9.6 0.0497346445918083
9.625 0.0498238392174244
9.65 0.0499237515032291
9.675 0.0500892736017704
9.7 0.0502149872481823
9.725 0.050341185182333
9.75 0.0504724308848381
9.775 0.0505867004394531
9.8 0.0507011115550995
9.825 0.050887193530798
9.85 0.0510107986629009
9.875 0.0511565878987312
9.9 0.0513185001909733
9.925 0.051443450152874
9.95 0.051624521613121
9.975 0.0517688356339931
10 0.0519410222768784
};
\addlegendentry{\tiny \lilans}
\addplot [semithick, darkorange25512714]
table {%
0.025 0.135219365358353
0.05 0.122156023979187
0.075 0.110890604555607
0.1 0.101515397429466
0.125 0.0937189683318138
0.15 0.0874467566609383
0.175 0.082554392516613
0.2 0.0788861140608788
0.225 0.076045922935009
0.25 0.0737511590123177
0.275 0.0720005184412003
0.3 0.0705590322613716
0.325 0.0692182779312134
0.35 0.0681807473301888
0.375 0.067278228700161
0.4 0.0664247199892998
0.425 0.0656800791621208
0.45 0.0649906843900681
0.475 0.0644383206963539
0.5 0.0638704523444176
0.525 0.0632670521736145
0.55 0.0627388954162598
0.575 0.0621509253978729
0.6 0.061605803668499
0.625 0.0610168352723122
0.65 0.0603335686028004
0.675 0.0598017498850822
0.7 0.0592182725667953
0.725 0.0586621575057507
0.75 0.0581322275102139
0.775 0.0576676353812218
0.8 0.0572984106838703
0.825 0.0569096319377422
0.85 0.0566711239516735
0.875 0.0564549714326859
0.9 0.0563547685742378
0.925 0.0563522055745125
0.95 0.0563125349581242
0.975 0.05632234364748
1 0.0563673973083496
1.025 0.0564016997814178
1.05 0.0564217083156109
1.075 0.0563621409237385
1.1 0.056270245462656
1.125 0.0561529323458672
1.15 0.0560196563601494
1.175 0.0557576343417168
1.2 0.0555400364100933
1.225 0.0553000643849373
1.25 0.0550837330520153
1.275 0.0549179464578629
1.3 0.0548869296908379
1.325 0.0549279190599918
1.35 0.0551087930798531
1.375 0.0554485507309437
1.4 0.0559713244438171
1.425 0.056560855358839
1.45 0.0572164431214333
1.475 0.0579220950603485
1.5 0.0586516708135605
1.525 0.0593659467995167
1.55 0.0600039176642895
1.575 0.0606607384979725
1.6 0.0612781383097172
1.625 0.061835952103138
1.65 0.062324732542038
1.675 0.0626983791589737
1.7 0.0630173087120056
1.725 0.0632045418024063
1.75 0.0633082464337349
1.775 0.0633869171142578
1.8 0.0633544027805328
1.825 0.0633265897631645
1.85 0.0631686300039291
1.875 0.0629491060972214
1.9 0.0627524107694626
1.925 0.0624938867986202
1.95 0.0621794871985912
1.975 0.0619199648499489
2 0.0617437623441219
2.025 0.0616820566356182
2.05 0.0616361983120441
2.075 0.0617479011416435
2.1 0.0617582127451897
2.125 0.0617751404643059
2.15 0.0616815127432346
2.175 0.0615414716303349
2.2 0.0613128840923309
2.225 0.0610236711800098
2.25 0.0606853105127811
2.275 0.0603206269443035
2.3 0.059936199337244
2.325 0.0594631806015968
2.35 0.0590522214770317
2.375 0.0586429499089718
2.4 0.0583314225077629
2.425 0.0580195114016533
2.45 0.0578499212861061
2.475 0.0579104237258434
2.5 0.0581924617290497
2.525 0.0584658794105053
2.55 0.0586482957005501
2.575 0.0588117055594921
2.6 0.0589349940419197
2.625 0.0589922703802586
2.65 0.0589869953691959
2.675 0.0589386075735092
2.7 0.0588947758078575
2.725 0.0588678196072578
2.75 0.058867584913969
2.775 0.058960746973753
2.8 0.0591433383524418
2.825 0.059316873550415
2.85 0.059567965567112
2.875 0.0598080158233643
2.9 0.0600773319602013
2.925 0.0603454858064651
2.95 0.0605954453349113
2.975 0.0608439296483994
3 0.0611344985663891
3.025 0.061393566429615
3.05 0.0616594962775707
3.075 0.0619484968483448
3.1 0.0622164569795132
3.125 0.0625672787427902
3.15 0.0628047436475754
3.175 0.0631060749292374
3.2 0.0633373036980629
3.225 0.0635213628411293
3.25 0.0637123957276344
3.275 0.0638751983642578
3.3 0.0641519799828529
3.325 0.0644637569785118
3.35 0.0648060142993927
3.375 0.0651812702417374
3.4 0.0655472576618195
3.425 0.0659078434109688
3.45 0.0662241652607918
3.475 0.0665818452835083
3.5 0.0668771415948868
3.525 0.0672271922230721
3.55 0.0675201639533043
3.575 0.0677883252501488
3.6 0.0680027008056641
3.625 0.0681975707411766
3.65 0.068231426179409
3.675 0.068222276866436
3.7 0.068162776529789
3.725 0.0679947659373283
3.75 0.0677951723337173
3.775 0.0675586014986038
3.8 0.0673236027359962
3.825 0.0670902654528618
3.85 0.0668729022145271
3.875 0.0666823834180832
3.9 0.0664570704102516
3.925 0.0662469863891602
3.95 0.0661002323031425
3.975 0.0659233331680298
4 0.065833568572998
4.025 0.0657760426402092
4.05 0.0657822340726852
4.075 0.0657669752836227
4.1 0.0657678842544556
4.125 0.0657400861382484
4.15 0.0656726583838463
4.175 0.065572664141655
4.2 0.0654710829257965
4.225 0.0653733313083649
4.25 0.0652077496051788
4.275 0.0650237649679184
4.3 0.0648896098136902
4.325 0.0646683573722839
4.35 0.0644237995147705
4.375 0.064141221344471
4.4 0.0638536289334297
4.425 0.0634814277291298
4.45 0.0631058663129807
4.475 0.0627013444900513
4.5 0.0622516460716724
4.525 0.0618441104888916
4.55 0.0613877810537815
4.575 0.0609768554568291
4.6 0.0605346001684666
4.625 0.0601309016346931
4.65 0.0598068311810493
4.675 0.0594235993921757
4.7 0.0591032952070236
4.725 0.058853380382061
4.75 0.0586542934179306
4.775 0.0585278309881687
4.8 0.0584772191941738
4.825 0.0584694370627403
4.85 0.0584639720618725
4.875 0.058429691940546
4.9 0.0583779811859131
4.925 0.058263260871172
4.95 0.0581240989267826
4.975 0.0579505898058414
5 0.0577158257365227
5.025 0.057465385645628
5.05 0.0571943558752537
5.075 0.0568765439093113
5.1 0.0566239878535271
5.125 0.0563496500253677
5.15 0.0561015009880066
5.175 0.0558496192097664
5.2 0.0556808263063431
5.225 0.0554784052073956
5.25 0.0553837008774281
5.275 0.0552973672747612
5.3 0.0552266873419285
5.325 0.0552160739898682
5.35 0.0552206486463547
5.375 0.0553200505673885
5.4 0.0554046891629696
5.425 0.0555358678102493
5.45 0.0557024292647839
5.475 0.0558439828455448
5.5 0.0559803619980812
5.525 0.0561223588883877
5.55 0.0562584474682808
5.575 0.0563424564898014
5.6 0.0564454011619091
5.625 0.0564941726624966
5.65 0.0565762929618359
5.675 0.0565698370337486
5.7 0.0566167905926704
5.725 0.0565874464809895
5.75 0.0565698742866516
5.775 0.0565197914838791
5.8 0.0564785189926624
5.825 0.0564155615866184
5.85 0.0563378483057022
5.875 0.0562331900000572
5.9 0.0561550594866276
5.925 0.0560317821800709
5.95 0.055891178548336
5.975 0.0557929873466492
6 0.0556601323187351
6.025 0.0555243752896786
6.05 0.0553973652422428
6.075 0.0552455373108387
6.1 0.0551269501447678
6.125 0.054977647960186
6.15 0.0548784285783768
6.175 0.0547721087932587
6.2 0.0546335652470589
6.225 0.0545311830937862
6.25 0.0543764010071754
6.275 0.0542825348675251
6.3 0.0541360303759575
6.325 0.0540347099304199
6.35 0.0538965202867985
6.375 0.0537729188799858
6.4 0.0536418706178665
6.425 0.0535244941711426
6.45 0.053395003080368
6.475 0.0532398484647274
6.5 0.0531126409769058
6.525 0.053016159683466
6.55 0.0528899766504765
6.575 0.052777674049139
6.6 0.0526561588048935
6.625 0.0525843501091003
6.65 0.0524995699524879
6.675 0.0523804277181625
6.7 0.0522819235920906
6.725 0.0522114597260952
6.75 0.0521121881902218
6.775 0.0520428121089935
6.8 0.0519770421087742
6.825 0.0518877319991589
6.85 0.0518320985138416
6.875 0.0517842657864094
6.9 0.0517185106873512
6.925 0.0516476780176163
6.95 0.0516012459993362
6.975 0.0515289530158043
7 0.0514690354466438
7.025 0.0514043495059013
7.05 0.0513231307268143
7.075 0.051257636398077
7.1 0.0511568263173103
7.125 0.051078986376524
7.15 0.050978098064661
7.175 0.0508348271250725
7.2 0.0507249981164932
7.225 0.0505948252975941
7.25 0.0504721067845821
7.275 0.0503389574587345
7.3 0.0501903109252453
7.325 0.050000537186861
7.35 0.0498062036931515
7.375 0.0496394485235214
7.4 0.0494425371289253
7.425 0.0491782948374748
7.45 0.0489561632275581
7.475 0.0487350560724735
7.5 0.0484627187252045
7.525 0.0482489913702011
7.55 0.0479866415262222
7.575 0.047768522053957
7.6 0.0475410893559456
7.625 0.0473350808024406
7.65 0.0471589788794518
7.675 0.0469977259635925
7.7 0.0468561388552189
7.725 0.0467386804521084
7.75 0.046637125313282
7.775 0.0465613752603531
7.8 0.0464875586330891
7.825 0.0464356429874897
7.85 0.0463771149516106
7.875 0.0463201217353344
7.9 0.0462703518569469
7.925 0.0462216474115849
7.95 0.0461767241358757
7.975 0.0461343564093113
8 0.0460836626589298
8.025 0.0460295863449574
8.05 0.0459783636033535
8.075 0.045923475176096
8.1 0.0458711050450802
8.125 0.0458122864365578
8.15 0.04575439915061
8.175 0.0456947572529316
8.2 0.045636385679245
8.225 0.045574989169836
8.25 0.0455251596868038
8.275 0.0454537533223629
8.3 0.0453925356268883
8.325 0.0453326962888241
8.35 0.0452714823186398
8.375 0.0452011674642563
8.4 0.0451438277959824
8.425 0.0450705327093601
8.45 0.0450115762650967
8.475 0.0449449308216572
8.5 0.0448930859565735
8.525 0.0448146723210812
8.55 0.0447539985179901
8.575 0.0446942411363125
8.6 0.0446284525096416
8.625 0.0445712096989155
8.65 0.044514711946249
8.675 0.0444552227854729
8.7 0.0443866513669491
8.725 0.0443349108099937
8.75 0.0442825704813004
8.775 0.0442212373018265
8.8 0.0441641882061958
8.825 0.0441104657948017
8.85 0.0440625138580799
8.875 0.0440022125840187
8.9 0.0439529307186604
8.925 0.0439085997641087
8.95 0.0438551567494869
8.975 0.0438058525323868
9 0.0437646396458149
9.025 0.043711818754673
9.05 0.0436719693243504
9.075 0.0436308942735195
9.1 0.0435915365815163
9.125 0.0435477904975414
9.15 0.0435097739100456
9.175 0.0434665940701962
9.2 0.0434280335903168
9.225 0.0433897115290165
9.25 0.0433605499565601
9.275 0.0433276258409023
9.3 0.0433042980730534
9.325 0.0432712696492672
9.35 0.0432456880807877
9.375 0.0432158447802067
9.4 0.0431982539594173
9.425 0.0431678779423237
9.45 0.0431602671742439
9.475 0.0431309416890144
9.5 0.0431195423007011
9.525 0.0431066676974297
9.55 0.0430960655212402
9.575 0.0430857837200165
9.6 0.0430720672011375
9.625 0.0430660061538219
9.65 0.0430648475885391
9.675 0.0430585667490959
9.7 0.0430566780269146
9.725 0.0430524535477161
9.75 0.0430537797510624
9.775 0.0430609136819839
9.8 0.0430652461946011
9.825 0.0430775433778763
9.85 0.0430800057947636
9.875 0.043102502822876
9.9 0.043113000690937
9.925 0.0431265719234943
9.95 0.0431500412523746
9.975 0.043174110352993
10 0.0431922227144241
};
\addlegendentry{\tiny Sulzer and Buck I}
\addplot [semithick, forestgreen4416044]
table {%
0.025 0.135061681270599
0.05 0.121567100286484
0.075 0.110037416219711
0.1 0.100315660238266
0.125 0.0921881496906281
0.15 0.0859382972121239
0.175 0.0810360983014107
0.2 0.0773584097623825
0.225 0.0745895951986313
0.25 0.0726761966943741
0.275 0.0712007582187653
0.3 0.0701353624463081
0.325 0.0692899152636528
0.35 0.068495474755764
0.375 0.0678076893091202
0.4 0.0670010223984718
0.425 0.0662663951516151
0.45 0.0655727535486221
0.475 0.0648422166705132
0.5 0.064121276140213
0.525 0.0633820295333862
0.55 0.0627116486430168
0.575 0.0620840676128864
0.6 0.0615630373358727
0.625 0.0608532316982746
0.65 0.0601968355476856
0.675 0.0595149397850037
0.7 0.0588676854968071
0.725 0.0582267567515373
0.75 0.0576274581253529
0.775 0.0570255629718304
0.8 0.0564933568239212
0.825 0.0560355335474014
0.85 0.0556079484522343
0.875 0.0552254766225815
0.9 0.054945707321167
0.925 0.0546534806489944
0.95 0.0544675625860691
0.975 0.0542414076626301
1 0.0540272556245327
1.025 0.0539051853120327
1.05 0.0537761673331261
1.075 0.0536314845085144
1.1 0.0535103343427181
1.125 0.053387425839901
1.15 0.0531646236777306
1.175 0.0529428757727146
1.2 0.0526832193136215
1.225 0.0524287484586239
1.25 0.0520878620445728
1.275 0.0517955757677555
1.3 0.0515171624720097
1.325 0.0512468218803406
1.35 0.051052950322628
1.375 0.0510052852332592
1.4 0.0510929748415947
1.425 0.05124381929636
1.45 0.0514715164899826
1.475 0.0518240742385387
1.5 0.0522732548415661
1.525 0.0527383238077164
1.55 0.0532554537057877
1.575 0.0538392066955566
1.6 0.0543949976563454
1.625 0.0549823641777039
1.65 0.0555706210434437
1.675 0.0560473948717117
1.7 0.0564538575708866
1.725 0.0567742697894573
1.75 0.0570358447730541
1.775 0.0571757815778255
1.8 0.0572582520544529
1.825 0.0573873221874237
1.85 0.0574476607143879
1.875 0.0573814362287521
1.9 0.0573512054979801
1.925 0.0572909973561764
1.95 0.0572420209646225
1.975 0.0573639012873173
2 0.0574560910463333
2.025 0.0577697418630123
2.05 0.0580590292811394
2.075 0.0584126450121403
2.1 0.0587038807570934
2.125 0.0588681250810623
2.15 0.059107456356287
2.175 0.0592451840639114
2.2 0.0593327134847641
2.225 0.0593749023973942
2.25 0.0593176819384098
2.275 0.0592943467199802
2.3 0.0592071563005447
2.325 0.0590769946575165
2.35 0.0589640699326992
2.375 0.0588692277669907
2.4 0.0588170886039734
2.425 0.0588254258036613
2.45 0.0590757168829441
2.475 0.0594243369996548
2.5 0.0597912780940533
2.525 0.060102429240942
2.55 0.0603448748588562
2.575 0.0605278015136719
2.6 0.0605516359210014
2.625 0.0605595372617245
2.65 0.0604581907391548
2.675 0.0603507310152054
2.7 0.060124110430479
2.725 0.0599207356572151
2.75 0.0597476847469807
2.775 0.0596084482967854
2.8 0.0595801137387753
2.825 0.0595583282411098
2.85 0.0595857426524162
2.875 0.0596868209540844
2.9 0.0598127357661724
2.925 0.0599500834941864
2.95 0.0601492188870907
2.975 0.060347069054842
3 0.0605663284659386
3.025 0.0609020888805389
3.05 0.0612677857279778
3.075 0.0616610459983349
3.1 0.0621639564633369
3.125 0.0626407936215401
3.15 0.0631089881062508
3.175 0.0635626316070557
3.2 0.0640544891357422
3.225 0.0645355433225632
3.25 0.0650735348463058
3.275 0.0656817927956581
3.3 0.0663018524646759
3.325 0.0668796300888062
3.35 0.0674236789345741
3.375 0.0678979977965355
3.4 0.0683961734175682
3.425 0.0688521936535835
3.45 0.0693566128611565
3.475 0.0699085220694542
3.5 0.0704358667135239
3.525 0.0709711015224457
3.55 0.0714004933834076
3.575 0.0717489868402481
3.6 0.0720746442675591
3.625 0.0722475573420525
3.65 0.0723727270960808
3.675 0.0724640339612961
3.7 0.0724358558654785
3.725 0.0723851248621941
3.75 0.0723102912306786
3.775 0.072192020714283
3.8 0.072017714381218
3.825 0.0717895328998566
3.85 0.0714906230568886
3.875 0.0712513625621796
3.9 0.0709224343299866
3.925 0.0706431195139885
3.95 0.0703958049416542
3.975 0.0702441856265068
4 0.0700861811637878
4.025 0.0699202716350555
4.05 0.0697622299194336
4.075 0.0696063712239265
4.1 0.0693992227315903
4.125 0.0692098289728165
4.15 0.0689145624637604
4.175 0.0686686858534813
4.2 0.0684061497449875
4.225 0.0681055933237076
4.25 0.0678061321377754
4.275 0.067542739212513
4.3 0.0672041997313499
4.325 0.0669057071208954
4.35 0.0665393993258476
4.375 0.0662035271525383
4.4 0.0658344849944115
4.425 0.0654009059071541
4.45 0.0649433583021164
4.475 0.0644932612776756
4.5 0.0640090554952621
4.525 0.0635843947529793
4.55 0.0631007924675941
4.575 0.0626349374651909
4.6 0.0622125715017319
4.625 0.061822947114706
4.65 0.0614821799099445
4.675 0.0611147433519363
4.7 0.0607610829174519
4.725 0.0604636743664742
4.75 0.0603136867284775
4.775 0.0601684637367725
4.8 0.0600932575762272
4.825 0.0600334294140339
4.85 0.0599839761853218
4.875 0.0599509328603745
4.9 0.0598859861493111
4.925 0.0597571544349194
4.95 0.0595505647361279
4.975 0.0593486689031124
5 0.0590509884059429
5.025 0.0586965307593346
5.05 0.0582775101065636
5.075 0.0578456372022629
5.1 0.0573899410665035
5.125 0.056909553706646
5.15 0.0564243160188198
5.175 0.0559731908142567
5.2 0.0555319301784039
5.225 0.0551330707967281
5.25 0.0547628626227379
5.275 0.0544825941324234
5.3 0.0542389489710331
5.325 0.0540568083524704
5.35 0.0539481341838837
5.375 0.0538124367594719
5.4 0.0537758804857731
5.425 0.0536844059824944
5.45 0.0536454953253269
5.475 0.0536368116736412
5.5 0.0536567308008671
5.525 0.0536937192082405
5.55 0.053726751357317
5.575 0.0537630841135979
5.6 0.053828988224268
5.625 0.0538669750094414
5.65 0.0539142154157162
5.675 0.0539242699742317
5.7 0.0539744943380356
5.725 0.0539572574198246
5.75 0.053939163684845
5.775 0.0539050623774529
5.8 0.0538684576749802
5.825 0.0538472086191177
5.85 0.0537622421979904
5.875 0.0536888763308525
5.9 0.0536063127219677
5.925 0.0534990355372429
5.95 0.0533837750554085
5.975 0.0532746687531471
6 0.0531719103455544
6.025 0.0530161559581757
6.05 0.0528921037912369
6.075 0.0527702569961548
6.1 0.0526445172727108
6.125 0.0525405444204807
6.15 0.0524133667349815
6.175 0.0522926226258278
6.2 0.0521559119224548
6.225 0.0520656853914261
6.25 0.051923431456089
6.275 0.0518163368105888
6.3 0.0516980476677418
6.325 0.0515716597437859
6.35 0.0514351651072502
6.375 0.0513062924146652
6.4 0.0511520057916641
6.425 0.0510276891291142
6.45 0.0508800148963928
6.475 0.0507489703595638
6.5 0.0505737476050854
6.525 0.0504706092178822
6.55 0.0503276996314526
6.575 0.0501966811716557
6.6 0.050071008503437
6.625 0.0499476715922356
6.65 0.0498350225389004
6.675 0.0497189983725548
6.7 0.0496155619621277
6.725 0.0495096668601036
6.75 0.049414224922657
6.775 0.0493088029325008
6.8 0.0492155626416206
6.825 0.0491154007613659
6.85 0.0490628816187382
6.875 0.0489626415073872
6.9 0.0488637499511242
6.925 0.0487986654043198
6.95 0.0486905984580517
6.975 0.0486143194139004
7 0.0485344417393208
7.025 0.0484763123095036
7.05 0.0483531504869461
7.075 0.0482820756733418
7.1 0.0481970459222794
7.125 0.0480870679020882
7.15 0.0479837469756603
7.175 0.0478571504354477
7.2 0.0477691777050495
7.225 0.047633346170187
7.25 0.0474871136248112
7.275 0.0473993085324764
7.3 0.0472259223461151
7.325 0.0470858477056026
7.35 0.0469506978988647
7.375 0.0467740148305893
7.4 0.0466448105871677
7.425 0.0464722700417042
7.45 0.0463092885911465
7.475 0.0461433194577694
7.5 0.045985534787178
7.525 0.0458067543804646
7.55 0.0456597208976746
7.575 0.0455261319875717
7.6 0.0453715249896049
7.625 0.0452554784715176
7.65 0.0451274178922176
7.675 0.0450316183269024
7.7 0.0449324920773506
7.725 0.0448246076703072
7.75 0.0447488389909267
7.775 0.0446596518158913
7.8 0.0445884503424168
7.825 0.0445112586021423
7.85 0.0444293580949306
7.875 0.0443617813289165
7.9 0.0442902334034443
7.925 0.0442207045853138
7.95 0.0441600605845451
7.975 0.0440881960093975
8 0.0440108142793179
8.025 0.0439447090029716
8.05 0.0438730046153069
8.075 0.0438040234148502
8.1 0.0437317453324795
8.125 0.0436684042215347
8.15 0.0435910485684872
8.175 0.0435261651873589
8.2 0.0434564091265202
8.225 0.0433852486312389
8.25 0.0433136150240898
8.275 0.0432473942637444
8.3 0.0431803204119205
8.325 0.0431134179234505
8.35 0.0430464632809162
8.375 0.0429790578782558
8.4 0.0429159440100193
8.425 0.0428531430661678
8.45 0.0427874587476254
8.475 0.042731735855341
8.5 0.0426710471510887
8.525 0.0426110401749611
8.55 0.0425571613013744
8.575 0.0425028689205647
8.6 0.0424453765153885
8.625 0.0423929020762444
8.65 0.042341485619545
8.675 0.0422943122684956
8.7 0.0422494225203991
8.725 0.0421976745128632
8.75 0.0421522222459316
8.775 0.0421102233231068
8.8 0.0420673601329327
8.825 0.0420244783163071
8.85 0.0419895835220814
8.875 0.0419478490948677
8.9 0.0419129617512226
8.925 0.0418782308697701
8.95 0.0418480969965458
8.975 0.0418120957911015
9 0.0417857505381107
9.025 0.0417586900293827
9.05 0.041735053062439
9.075 0.041711013764143
9.1 0.0416929125785828
9.125 0.0416748076677322
9.15 0.041651539504528
9.175 0.0416378863155842
9.2 0.0416200049221516
9.225 0.0416096188127995
9.25 0.0415974669158459
9.275 0.041589941829443
9.3 0.0415808483958244
9.325 0.0415735505521297
9.35 0.0415685847401619
9.375 0.0415663197636604
9.4 0.0415668822824955
9.425 0.0415656976401806
9.45 0.0415706299245358
9.475 0.0415744185447693
9.5 0.041581429541111
9.525 0.0415970794856548
9.55 0.041600625962019
9.575 0.0416146442294121
9.6 0.0416277125477791
9.625 0.0416399389505386
9.65 0.0416612513363361
9.675 0.041678674519062
9.7 0.041700217872858
9.725 0.0417289361357689
9.75 0.0417543165385723
9.775 0.0417720638215542
9.8 0.0418033823370934
9.825 0.041838388890028
9.85 0.041868831962347
9.875 0.0419060923159122
9.9 0.0419440679252148
9.925 0.0419807508587837
9.95 0.042029146105051
9.975 0.0420693047344685
10 0.0421076193451881
};
\addlegendentry{\tiny Sulzer and Buck II}
\addplot [semithick, crimson2143940]
table {%
0.025 0.370055377483368
0.05 0.334892153739929
0.075 0.302526414394379
0.1 0.272827088832855
0.125 0.245750114321709
0.15 0.22116507589817
0.175 0.198937997221947
0.2 0.179038107395172
0.225 0.161487385630608
0.25 0.146422415971756
0.275 0.134006842970848
0.3 0.124840565025806
0.325 0.119525544345379
0.35 0.117352977395058
0.375 0.117325142025948
0.4 0.119426056742668
0.425 0.123422846198082
0.45 0.12896665930748
0.475 0.135656535625458
0.5 0.142931595444679
0.525 0.15028940141201
0.55 0.157529547810555
0.575 0.164336487650871
0.6 0.170791164040565
0.625 0.17680686712265
0.65 0.182379379868507
0.675 0.187635228037834
0.7 0.192446529865265
0.725 0.196843490004539
0.75 0.200830534100533
0.775 0.204562336206436
0.8 0.207932502031326
0.825 0.21104171872139
0.85 0.21388341486454
0.875 0.216473028063774
0.9 0.218853294849396
0.925 0.221043467521667
0.95 0.223007470369339
0.975 0.22481632232666
1 0.226447984576225
1.025 0.227901414036751
1.05 0.229171931743622
1.075 0.230289608240128
1.1 0.231245741248131
1.125 0.232030361890793
1.15 0.232675999403
1.175 0.23316952586174
1.2 0.233520463109016
1.225 0.233745530247688
1.25 0.233848109841347
1.275 0.233827546238899
1.3 0.233741521835327
1.325 0.233547553420067
1.35 0.233293905854225
1.375 0.233015432953835
1.4 0.232744172215462
1.425 0.232499867677689
1.45 0.232338786125183
1.475 0.232226029038429
1.5 0.232146188616753
1.525 0.232037723064423
1.55 0.231906697154045
1.575 0.231763854622841
1.6 0.231602162122726
1.625 0.231444537639618
1.65 0.231270313262939
1.675 0.231135234236717
1.7 0.230986550450325
1.725 0.230878755450249
1.75 0.230812236666679
1.775 0.230794101953506
1.8 0.230834648013115
1.825 0.230910927057266
1.85 0.23108384013176
1.875 0.231317177414894
1.9 0.231630891561508
1.925 0.231992617249489
1.95 0.232432812452316
1.975 0.232940137386322
2 0.233512714505196
2.025 0.234121963381767
2.05 0.234793767333031
2.075 0.235488817095757
2.1 0.236241891980171
2.125 0.236985698342323
2.15 0.237745434045792
2.175 0.238510310649872
2.2 0.239249646663666
2.225 0.239986598491669
2.25 0.240699633955956
2.275 0.241381391882896
2.3 0.242036774754524
2.325 0.242657124996185
2.35 0.243232354521751
2.375 0.243774220347404
2.4 0.244272455573082
2.425 0.244740650057793
2.45 0.245177134871483
2.475 0.245598718523979
2.5 0.245994314551353
2.525 0.246379643678665
2.55 0.246772035956383
2.575 0.247168958187103
2.6 0.247566044330597
2.625 0.24801878631115
2.65 0.24849446117878
2.675 0.249011605978012
2.7 0.249580726027489
2.725 0.250205814838409
2.75 0.250875443220139
2.775 0.251626938581467
2.8 0.252445816993713
2.825 0.253313660621643
2.85 0.254243582487106
2.875 0.255232572555542
2.9 0.256277561187744
2.925 0.257354110479355
2.95 0.258455753326416
2.975 0.25958189368248
3 0.260718137025833
3.025 0.261855840682983
3.05 0.262973576784134
3.075 0.264066427946091
3.1 0.265129685401917
3.125 0.266161233186722
3.15 0.267135292291641
3.175 0.268074214458466
3.2 0.268937647342682
3.225 0.26974755525589
3.25 0.270498842000961
3.275 0.271180987358093
3.3 0.271794587373734
3.325 0.272352755069733
3.35 0.272838920354843
3.375 0.273263424634933
3.4 0.273641109466553
3.425 0.27397346496582
3.45 0.274236768484116
3.475 0.27447298169136
3.5 0.274648398160934
3.525 0.274823516607285
3.55 0.274950623512268
3.575 0.275030046701431
3.6 0.275131195783615
3.625 0.275162130594254
3.65 0.275199800729752
3.675 0.275192439556122
3.7 0.275182843208313
3.725 0.275109082460403
3.75 0.275052279233932
3.775 0.274937003850937
3.8 0.27482146024704
3.825 0.274649083614349
3.85 0.274477660655975
3.875 0.274256497621536
3.9 0.274030566215515
3.925 0.273786246776581
3.95 0.27351513504982
3.975 0.273236036300659
4 0.272932320833206
4.025 0.272651851177216
4.05 0.27234622836113
4.075 0.272057384252548
4.1 0.27176558971405
4.125 0.271438717842102
4.15 0.271140098571777
4.175 0.270835846662521
4.2 0.27054288983345
4.225 0.270265221595764
4.25 0.269993156194687
4.275 0.269741863012314
4.3 0.269509136676788
4.325 0.269261747598648
4.35 0.269017100334167
4.375 0.268820405006409
4.4 0.268625050783157
4.425 0.268433898687363
4.45 0.268274188041687
4.475 0.268120616674423
4.5 0.267964422702789
4.525 0.267808765172958
4.55 0.267691344022751
4.575 0.267578065395355
4.6 0.267463803291321
4.625 0.267357468605042
4.65 0.267244905233383
4.675 0.267127007246017
4.7 0.267010688781738
4.725 0.266927748918533
4.75 0.266823559999466
4.775 0.266731381416321
4.8 0.266640186309814
4.825 0.266557604074478
4.85 0.266491115093231
4.875 0.266415804624557
4.9 0.266358107328415
4.925 0.266308903694153
4.95 0.266283422708511
4.975 0.266259044408798
5 0.266243636608124
5.025 0.266227245330811
5.05 0.266245901584625
5.075 0.266249567270279
5.1 0.266266316175461
5.125 0.266271471977234
5.15 0.266288548707962
5.175 0.266301065683365
5.2 0.266315460205078
5.225 0.266313672065735
5.25 0.266304463148117
5.275 0.266286104917526
5.3 0.26625719666481
5.325 0.266218662261963
5.35 0.266172707080841
5.375 0.266104608774185
5.4 0.266021728515625
5.425 0.26594266295433
5.45 0.265837877988815
5.475 0.265732973814011
5.5 0.26559990644455
5.525 0.265474170446396
5.55 0.265343844890594
5.575 0.265192985534668
5.6 0.265046685934067
5.625 0.264904677867889
5.65 0.26474592089653
5.675 0.264585167169571
5.7 0.264429301023483
5.725 0.264279067516327
5.75 0.264115512371063
5.775 0.263968855142593
5.8 0.263807982206345
5.825 0.263663083314896
5.85 0.26350599527359
5.875 0.263369828462601
5.9 0.263223946094513
5.925 0.263089209794998
5.95 0.262944757938385
5.975 0.262805968523026
6 0.262668341398239
6.025 0.262535721063614
6.05 0.262410283088684
6.075 0.262282967567444
6.1 0.262154787778854
6.125 0.26203128695488
6.15 0.261902391910553
6.175 0.261773824691772
6.2 0.261650562286377
6.225 0.261523395776749
6.25 0.261397957801819
6.275 0.261272698640823
6.3 0.261144250631332
6.325 0.261022001504898
6.35 0.260885238647461
6.375 0.2607461810112
6.4 0.260630130767822
6.425 0.260490328073502
6.45 0.260349452495575
6.475 0.260219305753708
6.5 0.26007205247879
6.525 0.25993800163269
6.55 0.259780377149582
6.575 0.259639620780945
6.6 0.259508848190308
6.625 0.259343594312668
6.65 0.259186714887619
6.675 0.259044140577316
6.7 0.258881151676178
6.725 0.258727192878723
6.75 0.258561342954636
6.775 0.258408933877945
6.8 0.258241266012192
6.825 0.258081555366516
6.85 0.257915914058685
6.875 0.257737100124359
6.9 0.257581353187561
6.925 0.257415264844894
6.95 0.257258683443069
6.975 0.257096141576767
7 0.256921947002411
7.025 0.256758511066437
7.05 0.256601780653
7.075 0.256434708833694
7.1 0.256271570920944
7.125 0.256115406751633
7.15 0.255956143140793
7.175 0.255812764167786
7.2 0.255662709474564
7.225 0.2555011510849
7.25 0.255357086658478
7.275 0.255206406116486
7.3 0.255064338445663
7.325 0.254938900470734
7.35 0.254794090986252
7.375 0.254654109477997
7.4 0.25452333688736
7.425 0.254405438899994
7.45 0.254288703203201
7.475 0.254159271717072
7.5 0.254052549600601
7.525 0.253940343856812
7.55 0.253839671611786
7.575 0.253728359937668
7.6 0.253646105527878
7.625 0.253543019294739
7.65 0.253457546234131
7.675 0.253367841243744
7.7 0.253263592720032
7.725 0.253166735172272
7.75 0.25308832526207
7.775 0.253001123666763
7.8 0.252913266420364
7.825 0.252825915813446
7.85 0.252735197544098
7.875 0.252648681402206
7.9 0.252570599317551
7.925 0.252491533756256
7.95 0.252418011426926
7.975 0.252330213785172
8 0.252244561910629
8.025 0.25218254327774
8.05 0.252115935087204
8.075 0.252042382955551
8.1 0.251977622509003
8.125 0.251909375190735
8.15 0.251856714487076
8.175 0.251802325248718
8.2 0.251746743917465
8.225 0.251687228679657
8.25 0.251628756523132
8.275 0.25158417224884
8.3 0.251542270183563
8.325 0.251485824584961
8.35 0.251442492008209
8.375 0.251404613256454
8.4 0.251367896795273
8.425 0.251326322555542
8.45 0.251286655664444
8.475 0.251252144575119
8.5 0.2512087225914
8.525 0.251183301210403
8.55 0.251146018505096
8.575 0.251120150089264
8.6 0.251087576150894
8.625 0.251054912805557
8.65 0.251034468412399
8.675 0.251009434461594
8.7 0.250983327627182
8.725 0.250955194234848
8.75 0.250936269760132
8.775 0.250910669565201
8.8 0.250889748334885
8.825 0.250880390405655
8.85 0.250852853059769
8.875 0.250835001468658
8.9 0.250824391841888
8.925 0.250804245471954
8.95 0.250796675682068
8.975 0.250792354345322
9 0.250767350196838
9.025 0.250763207674026
9.05 0.250753253698349
9.075 0.250750124454498
9.1 0.250750601291656
9.125 0.250749379396439
9.15 0.250749886035919
9.175 0.250747561454773
9.2 0.250740051269531
9.225 0.250750690698624
9.25 0.250763475894928
9.275 0.250773310661316
9.3 0.250781059265137
9.325 0.25078222155571
9.35 0.250807106494904
9.375 0.250821173191071
9.4 0.250836282968521
9.425 0.250862181186676
9.45 0.250877797603607
9.475 0.250893771648407
9.5 0.250931888818741
9.525 0.250964760780334
9.55 0.251008868217468
9.575 0.251031458377838
9.6 0.251058608293533
9.625 0.251094192266464
9.65 0.251146912574768
9.675 0.251180738210678
9.7 0.251242846250534
9.725 0.251282334327698
9.75 0.251345813274384
9.775 0.251409709453583
9.8 0.251460522413254
9.825 0.251514196395874
9.85 0.251578122377396
9.875 0.251641422510147
9.9 0.251723885536194
9.925 0.251789301633835
9.95 0.251879543066025
9.975 0.251956731081009
10 0.25204598903656
};
\addlegendentry{\tiny DeepONet}
\addplot [semithick, mediumpurple148103189]
table {%
0.025 0.00981780141592026
0.05 0.0182719640433788
0.075 0.0255891438573599
0.1 0.0319436006247997
0.125 0.0374726615846157
0.15 0.0422839596867561
0.175 0.0464643761515617
0.2 0.0500762462615967
0.225 0.0531701408326626
0.25 0.0557875521481037
0.275 0.057970717549324
0.3 0.0597675740718842
0.325 0.0612246990203857
0.35 0.0623887293040752
0.375 0.0632999688386917
0.4 0.0639981031417847
0.425 0.064516969025135
0.45 0.0648834779858589
0.475 0.0651255324482918
0.5 0.0652638226747513
0.525 0.0653185322880745
0.55 0.0653059631586075
0.575 0.0652405694127083
0.6 0.0651388168334961
0.625 0.065015435218811
0.65 0.0648845210671425
0.675 0.0647621378302574
0.7 0.0646611452102661
0.725 0.0645945370197296
0.75 0.0645714104175568
0.775 0.064600758254528
0.8 0.0646844357252121
0.825 0.064825750887394
0.85 0.0650213286280632
0.875 0.0652681514620781
0.9 0.0655612200498581
0.925 0.0658943876624107
0.95 0.0662591978907585
0.975 0.066650964319706
1 0.0670629516243935
1.025 0.0674920901656151
1.05 0.0679388642311096
1.075 0.0684102475643158
1.1 0.0689126029610634
1.125 0.0694549381732941
1.15 0.0700417757034302
1.175 0.0706734955310822
1.2 0.071343868970871
1.225 0.0720493122935295
1.25 0.0727861747145653
1.275 0.0735532641410828
1.3 0.0743484795093536
1.325 0.0751690119504929
1.35 0.0759982466697693
1.375 0.076815590262413
1.4 0.07761350274086
1.425 0.0783983916044235
1.45 0.0791943892836571
1.475 0.080043837428093
1.5 0.0809961035847664
1.525 0.0820600837469101
1.55 0.0831959620118141
1.575 0.0843480378389359
1.6 0.0854664221405983
1.625 0.0865108594298363
1.65 0.0874553695321083
1.675 0.0882860273122787
1.7 0.0890001654624939
1.725 0.0896047353744507
1.75 0.0901119560003281
1.775 0.0905337482690811
1.8 0.0908765569329262
1.825 0.091148242354393
1.85 0.0913582816720009
1.875 0.0915219858288765
1.9 0.0916559770703316
1.925 0.0917786508798599
1.95 0.0919088125228882
1.975 0.0920629277825356
2 0.0922576040029526
2.025 0.092512920498848
2.05 0.0928568616509438
2.075 0.0933263748884201
2.1 0.0939375832676888
2.125 0.09464992582798
2.15 0.0953921899199486
2.175 0.0961079224944115
2.2 0.0967623144388199
2.225 0.0973367989063263
2.25 0.0978244766592979
2.275 0.098225362598896
2.3 0.0985449180006981
2.325 0.0987917855381966
2.35 0.0989756733179092
2.375 0.0991078540682793
2.4 0.0991991683840752
2.425 0.099260501563549
2.45 0.0993032976984978
2.475 0.0993398651480675
2.5 0.0993853136897087
2.525 0.0994564518332481
2.55 0.0995683521032333
2.575 0.0997258946299553
2.6 0.0999201536178589
2.625 0.100131928920746
2.65 0.100340031087399
2.675 0.100525893270969
2.7 0.100674629211426
2.725 0.100775010883808
2.75 0.100818939507008
2.775 0.100802578032017
2.8 0.100725047290325
2.825 0.100588627159595
2.85 0.100398488342762
2.875 0.100161693990231
2.9 0.0998861789703369
2.925 0.0995819047093391
2.95 0.0992599576711655
2.975 0.0989319980144501
3 0.0986106097698212
3.025 0.0983068272471428
3.05 0.0980287417769432
3.075 0.0977826490998268
3.1 0.0975716561079025
3.125 0.0973983183503151
3.15 0.0972636938095093
3.175 0.0971696153283119
3.2 0.0971166417002678
3.225 0.0971053317189217
3.25 0.0971342101693153
3.275 0.0972007140517235
3.3 0.0973017066717148
3.325 0.0974327996373177
3.35 0.0975900739431381
3.375 0.0977702662348747
3.4 0.0979704484343529
3.425 0.0981880202889442
3.45 0.0984209030866623
3.475 0.098667599260807
3.5 0.0989257618784904
3.525 0.0991936698555946
3.55 0.0994682088494301
3.575 0.0997472256422043
3.6 0.100027441978455
3.625 0.100305698812008
3.65 0.100578583776951
3.675 0.100842796266079
3.7 0.101095080375671
3.725 0.101332172751427
3.75 0.101551376283169
3.775 0.101750284433365
3.8 0.101926632225513
3.825 0.102078706026077
3.85 0.102205671370029
3.875 0.102307394146919
3.9 0.102384299039841
3.925 0.102437369525433
3.95 0.102468319237232
3.975 0.102479472756386
4 0.102473065257072
4.025 0.102451615035534
4.05 0.102418169379234
4.075 0.102376475930214
4.1 0.102329477667809
4.125 0.102281406521797
4.15 0.102236412465572
4.175 0.102197974920273
4.2 0.102169215679169
4.225 0.102151907980442
4.25 0.102146610617638
4.275 0.102151468396187
4.3 0.102162949740887
4.325 0.102175109088421
4.35 0.102180078625679
4.375 0.102168574929237
4.4 0.102130942046642
4.425 0.10205802321434
4.45 0.101941056549549
4.475 0.101771861314774
4.5 0.101544104516506
4.525 0.101251617074013
4.55 0.10089036822319
4.575 0.100457020103931
4.6 0.0999507382512093
4.625 0.0993729159235954
4.65 0.0987282544374466
4.675 0.098024383187294
4.7 0.0972718521952629
4.725 0.0964842140674591
4.75 0.0956762656569481
4.775 0.0948638319969177
4.8 0.094062551856041
4.825 0.093286819756031
4.85 0.0925496965646744
4.875 0.0918617323040962
4.9 0.0912311896681786
4.925 0.0906637758016586
4.95 0.0901627913117409
4.975 0.0897292271256447
5 0.0893628001213074
5.025 0.0890614092350006
5.05 0.0888224467635155
5.075 0.0886427089571953
5.1 0.0885183289647102
5.125 0.0884442403912544
5.15 0.088415190577507
5.175 0.0884259194135666
5.2 0.0884703919291496
5.225 0.0885427668690681
5.25 0.0886369124054909
5.275 0.0887470543384552
5.3 0.0888675600290298
5.325 0.0889932960271835
5.35 0.0891194120049477
5.375 0.0892419219017029
5.4 0.0893573388457298
5.425 0.0894632115960121
5.45 0.0895577818155289
5.475 0.0896396636962891
5.5 0.089708499610424
5.525 0.089764229953289
5.55 0.0898070111870766
5.575 0.0898374542593956
5.6 0.0898560434579849
5.625 0.0898636803030968
5.65 0.0898609980940819
5.675 0.0898488536477089
5.7 0.0898278579115868
5.725 0.089798741042614
5.75 0.0897619128227234
5.775 0.0897178426384926
5.8 0.0896672531962395
5.825 0.0896100625395775
5.85 0.0895465388894081
5.875 0.0894770398736
5.9 0.0894016921520233
5.925 0.0893212333321571
5.95 0.0892358496785164
5.975 0.0891463235020638
6 0.0890535116195679
6.025 0.088958777487278
6.05 0.0888632386922836
6.075 0.0887685790657997
6.1 0.0886766761541367
6.125 0.0885892063379288
6.15 0.0885078907012939
6.175 0.0884342864155769
6.2 0.0883693546056747
6.225 0.0883135795593262
6.25 0.0882671326398849
6.275 0.0882298424839973
6.3 0.0882013291120529
6.325 0.088181160390377
6.35 0.0881686583161354
6.375 0.0881636589765549
6.4 0.0881657227873802
6.425 0.0881746485829353
6.45 0.0881900191307068
6.475 0.0882115215063095
6.5 0.0882391929626465
6.525 0.0882726684212685
6.55 0.0883118212223053
6.575 0.0883563086390495
6.6 0.0884054750204086
6.625 0.0884591117501259
6.65 0.0885165557265282
6.675 0.0885772183537483
6.7 0.0886405482888222
6.725 0.0887058600783348
6.75 0.088772751390934
6.775 0.0888403803110123
6.8 0.0889082700014114
6.825 0.0889757499098778
6.85 0.0890423953533173
6.875 0.0891074314713478
6.9 0.0891704857349396
6.925 0.0892312303185463
6.95 0.0892891734838486
6.975 0.0893439725041389
7 0.0893951207399368
7.025 0.089442603290081
7.05 0.0894861370325089
7.075 0.0895252376794815
7.1 0.0895601660013199
7.125 0.0895903408527374
7.15 0.0896154493093491
7.175 0.0896351113915443
7.2 0.0896487906575203
7.225 0.0896560251712799
7.25 0.0896557942032814
7.275 0.0896474346518517
7.3 0.089629702270031
7.325 0.0896011739969254
7.35 0.0895605608820915
7.375 0.0895064696669579
7.4 0.0894374549388885
7.425 0.0893522650003433
7.45 0.0892500653862953
7.475 0.0891309157013893
7.5 0.0889955312013626
7.525 0.0888459160923958
7.55 0.0886848792433739
7.575 0.0885163322091103
7.6 0.0883448570966721
7.625 0.088175006210804
7.65 0.0880112946033478
7.675 0.0878573209047318
7.7 0.0877156034111977
7.725 0.087587982416153
7.75 0.0874751433730125
7.775 0.0873770490288734
7.8 0.0872931182384491
7.825 0.0872222483158112
7.85 0.0871630683541298
7.875 0.0871144458651543
7.9 0.0870751142501831
7.925 0.0870439484715462
7.95 0.0870199874043465
7.975 0.0870020911097527
8 0.0869897529482841
8.025 0.0869822800159454
8.05 0.0869792997837067
8.075 0.0869801491498947
8.1 0.086984746158123
8.125 0.0869927778840065
8.15 0.0870041027665138
8.175 0.0870184674859047
8.2 0.0870358124375343
8.225 0.0870558395981789
8.25 0.0870787426829338
8.275 0.0871042832732201
8.3 0.087132453918457
8.325 0.0871632546186447
8.35 0.0871966555714607
8.375 0.0872325301170349
8.4 0.0872708931565285
8.425 0.087311714887619
8.45 0.08735491335392
8.475 0.0874004065990448
8.5 0.0874482393264771
8.525 0.0874983668327332
8.55 0.0875507518649101
8.575 0.0876053422689438
8.6 0.087662048637867
8.625 0.0877208486199379
8.65 0.0877816453576088
8.675 0.0878444835543633
8.7 0.0879092589020729
8.725 0.0879759788513184
8.75 0.0880444496870041
8.775 0.0881147459149361
8.8 0.0881869420409203
8.825 0.0882609561085701
8.85 0.0883368626236916
8.875 0.0884148553013802
8.9 0.0884947255253792
8.925 0.0885764956474304
8.95 0.0886604264378548
8.975 0.0887463837862015
9 0.0888345539569855
9.025 0.0889248251914978
9.05 0.0890171825885773
9.075 0.0891117826104164
9.1 0.08920868486166
9.125 0.0893077477812767
9.15 0.0894090756773949
9.175 0.0895125493407249
9.2 0.0896181017160416
9.225 0.0897256731987
9.25 0.0898352637887001
9.275 0.0899467766284943
9.3 0.0900602117180824
9.325 0.0901754349470139
9.35 0.0902923718094826
9.375 0.0904109105467796
9.4 0.0905310213565826
9.425 0.0906528383493423
9.45 0.090776152908802
9.475 0.0909008979797363
9.5 0.0910270288586617
9.525 0.091154545545578
9.55 0.091283492743969
9.575 0.091413676738739
9.6 0.0915452018380165
9.625 0.0916780009865761
9.65 0.091811902821064
9.675 0.0919469743967056
9.7 0.0920832976698875
9.725 0.0922208204865456
9.75 0.092359222471714
9.775 0.092498742043972
9.8 0.0926393941044807
9.825 0.0927809700369835
9.85 0.092923603951931
9.875 0.0930672809481621
9.9 0.0932119265198708
9.925 0.0933574810624123
9.95 0.0935040712356567
9.975 0.0936516299843788
10 0.0938001647591591
};
\addlegendentry{\tiny NODE}
\end{axis}

\end{tikzpicture}

%% file: DataReduction.tex
\subsection{Improving Training Efficiency of \lilan}
\label{sec:DR}
Thus far, all previous sections have presented results for \lilans training in the data-rich regime. In each problem considered, data is relatively cheap, but this may not hold true for other engineering tasks which require surrogate modeling. Therefore, it provides valuable insight to investigate the accuracy of the \lilans model in the data-poor regime. On the other hand, a data-intensive training procedure can pose a heavy computational burden for training any model. This is especially the case when using large networks for high-dimensional problems (for example, the PDE problems in \cref{sec:AC,,sec:CH}). In situations where the available memory or training time budget have strict limitations, data reduction is an effective tool to meet project requirements. This being the case, the accuracy of \lilans surrogates under different data reduction schemes must be examined. In this section, we consider two data reduction procedures and their ramifications for \lilans:
\begin{enumerate}
    \item The first strategy involves reducing the training data and investigating the effect on the generalization performance of \lilan. In particular, 
    we study how reducing the number of sample trajectories (i.e. the parameter $N$ in \eqref{train_form}) affects the performance of \lilan. Note that smaller training datasets significantly improve the computational efficiency during training while also reducing the data generation time. Additionally, in some applications, data cannot be generated at will and is limited to data from real-world sensor observations. For these reasons, it is highly desirable that the \lilans architecture is robust and can achieve a high degree of generalization accuracy with varying amounts of data.
    \item   The second strategy involves the use of  coarsened time discretizations in training, meaning reduction of the parameter $M$ in \eqref{train_form}) and investigating the impact on the generalization performance of \lilan. Reducing the discretization of training data means potentially massive savings on memory during training. This, in turn, means that one can use a greater number of samples for a fixed training memory budget or can simply reap the rewards of lower memory cost and faster training speed.
\end{enumerate}




\subsubsection{Reducing Training Sample Size $N$ in \eqref{train_form}}
\label{sec:RTSZ}

The first strategy for improving the training efficiency of \lilan\  considers reducing the number of sample trajectories (i.e. the parameter $N$ in \eqref{train_form}). For the Robertson (\cref{sec:Robertson}) and full CR models (\cref{sec:full-CR}), the original network was trained with $4096$ samples from a three-dimensional parameter space. To understand how the prediction accuracy degrades with training data size, several smaller datasets were created by repeatedly reducing the number of training samples by a factor of two. For the CR charge state model (\cref{sec:chargestate}), the initial training utilized $15625$ samples from a three-dimensional parameter input space.  For this problem, smaller datasets were created by repeatedly reducing the number of training samples, this time by a factor of five. The smallest dataset, for each ODE problem considered, contained only the eight corners of the original 3-dimensional parameter space.

 
From \cref{datared} (left subfigure) we see that for the Robertson model (\cref{sec:Robertson}), using only $128$ samples, a \lilans can be trained to achieve error on the same order of magnitude as a network of the same architecture trained with $4096$ samples (a reduction of $32$x). For the CR charge state model (\cref{datared}, (middle subfigure)), $125$ samples were required to train a network which achieved error on the same order of magnitude of the original network trained with $15625$ samples. Finally, after $256$ training samples, the \lilan's final accuracy on the full CR test dataset does not significantly improve if more samples are provided, as seen in \cref{datared} (right subfigure). From these ODE problem results, we observe that the amount of data used for training \lilans surrogates can be reduced significantly while maintaining accuracy, indicating that data reduction via reducing the training samples is a viable data reduction strategy.

For completeness, the same sample reduction experiment was performed on each of the two PDE problems (\cref{sec:AC,,sec:CH}). In each case, the original dataset of $900$ samples was repeatedly reduced by $100$ samples and used to train a new model. The smallest dataset contained $100$ total sample trajectories. The change in accuracy with respect to sample size is shown in \cref{pde_datared}. \Cref{pde_datared} shows that for each of the PDE problems, adding more samples typically resulted in a significant decrease in the relative error, though this decrease does slow after the first few hundred samples are added in both cases. However, this does suggest that reducing the training data may not be a favorable option for improving the computational efficiency, as these problems might require a large number of samples to guarantee a low relative error.

\begin{figure}[h!t!b!] 
\begin{minipage}{0.32\textwidth}
    \centering
    \resizebox{1\linewidth}{!}{\input{figs/ROBER_dr_errors}}
\end{minipage}
\begin{minipage}{0.32\textwidth}
    \centering
    \resizebox{1\linewidth}{!}{\input{figs/Charge_dr_errors}}
\end{minipage}
\begin{minipage}{0.32\textwidth}
    \centering
    \resizebox{1\linewidth}{!}{\input{figs/CR_dr_errors}}
\end{minipage}
\caption{Left to Right: Average point-wise relative error over time \eqref{eq:R2} when evaluating Robertson chemical kinetics model (\cref{sec:Robertson}) on test data (each plotted point shows the accuracy in relation to the number of training samples); Average point-wise relative error over time \eqref{eq:R2} when evaluating CR charge state model (\cref{sec:CR-model}) on test data; Average point-wise relative error over time \eqref{eq:R2} when evaluating full CR model (\cref{sec:CR-model}) on test data.}
 \label{datared}
\end{figure}

\begin{figure}[h!t!b!] 
\begin{minipage}{0.48\textwidth}
    \centering
    \resizebox{1\linewidth}{!}{\input{figs/AC_dr_errors}}
\end{minipage}
\begin{minipage}{0.48\textwidth}
    \centering
    \resizebox{1\linewidth}{!}{\input{figs/CH_dr_errors}}
\end{minipage}
\caption{Left to Right: Average point-wise relative error over time \eqref{eq:R2} when evaluating Allen-Cahn PDE (\cref{sec:AC}) on test data (each plotted point shows the accuracy in relation to the number of training samples); Average point-wise relative error over time \eqref{eq:R2} when evaluating Cahn-Hilliard PDE (\cref{sec:CH}) on test data (each plotted point shows the accuracy in relation to the number of training samples) \eqref{eq:R2} on test data.}
 \label{pde_datared}
\end{figure}

\subsubsection{Using Coarse Time Discretizations in Training}
\label{coarse_time_s}
An additional strategy to reduce the data and memory requirements of training the network is to consider a coarse time grid for the sample trajectories. In this section, we investigate the effect of reducing the parameter $M$ in \eqref{train_form} on the performance of \lilan. We consider different time discretizations by removing large numbers of time steps from the training and validation data and using this new dataset to train the networks. Examples of different time grids considered for training neural networks are shown in \cref{Robertson_time,,Charge_time,,CR_time,,ac_time,,ch_time} (left subfigures). For assessing the performance, we retain the original test data, which still has a fine discretization in time. The main purpose of this study 
is to investigate whether coarsening the discretization in time is an effective means to reduce data, while maintaining the test accuracy of \lilan.

Our results indicate that, in all test problems, data reduction by coarsening the underlying time discretization of data trajectories proves to be an effective path to reducing computational cost while maintaining test accuracy. The results of our experiments are shown in \cref{Robertson_time,,Charge_time,,CR_time,,ac_time,,ch_time}. The left subfigure in each of the aforementioned figures depicts the new time discretizations used in training, while the right subfigure shows the change in relative error compared to how many steps were skipped. The x-axis of the right subfigures, labeled "Skipped Steps," indicates the number of time steps which were removed between steps in the original discretization. For example, if skipped steps $= 10$, then we retain the time steps $\{x_i\}_{i=0,\, 10,\, 20,\, \cdots,\, M-10,\, M}$ for each training and validation sample. It is important to note that the accuracy reported is still evaluated on the test dataset, which is not downsampled in time. In each of \cref{Robertson_time,,Charge_time,,CR_time,,ac_time,,ch_time}, the results indicate that a significant number of time steps can be removed from the training discretization before accuracy degrades significantly. Specifically, \cref{Robertson_time}, shows that on the Robertson ODE problem (\cref{sec:Robertson}), we may take skipped steps = 8 before seeing the first major drop in accuracy. For the CR charge states and full CR models (\cref{sec:CR-model}), \cref{Charge_time,,CR_time} show that generalization error does not decrease until 20 steps are skipped between each preserved step. In the case of the Allen-Cahn PDE (\cref{sec:AC}), \cref{ac_time} suggests that accuracy deteriorates very slowly in relation to the number of skipped steps, again with a falling off observed after gaps of 20 are removed from the data. For the final problem considered, in \cref{ch_time} we see that skipping a remarkable 40 steps results in almost no loss of test performance, and finally after 80 steps there is a larger decline in accuracy for the Cahn-Hilliard PDE (\cref{sec:CH}). These results lead us to believe that coarsening the time discretization of the data is a very effective data reduction technique for \lilans surrogates. However, the ability to train on coarse discretizations naturally begs the question of why \lilans is so effective after data is reduced in this manner. To attempt to answer this question we devised an additional experiment to determine whether the time transformation network, $\boldsymbol{\tau}$, plays a significant role in \lilans robustness to training on temporally downsampled discretizations of the original training data.

\subsection{Importance of Learned Time Transformation}

\label{time_import}

In this section, we investigate the importance of the nonlinear time transformation ($\boldsymbol{\tau}\LRp{t,\ \xb_0,\ \mathbf{p}; \ \boldsymbol{\nu}}$ in \eqref{solution_latent}) on all problems considered in this work. The following experiments were inspired by the results on temporally downsampling training data, presented in \cref{datared}. Additionally, since one of the key differences between \lilans and the methodology of Sulzer and Buck \cite{sulzer2023speedingastrochemicalreactionnetworks} is the learned, nonlinear transformation of the time variable, we thought it informative to test the results of the \lilans architecture if the time transformation was removed. In particular, instead of \eqref{solution_latent}, we consider the solution in the latent space to have the following form: \begin{equation}
    \yb\LRp{t,\ \xb_0,\ \mathbf{p}}= \boldsymbol{\sE}(\xb_0, \mathbf{p};\ \boldsymbol{\beta}) +  t\ \boldsymbol{\bs{c}}(\xb_0, \mathbf{p};\ \boldsymbol{\alpha}).
    \label{eq:no_time_transform}
\end{equation}
where time, $t$, is first log scaled with the base $10$ logarithm and then rescaled to the range $[0, 1]$. This new transform of the time variable is the same as used in \cite{sulzer2023speedingastrochemicalreactionnetworks}. The right subfigure of \cref{Robertson_time,,Charge_time,,CR_time,,ac_time,,ch_time}  explore the benefit of the time transformation in the context of data reduction described in \cref{coarse_time_s}. The black circles represent the error for \lilan s trained with a learned time transformation, while the red dots indicate the error for \lilan s trained with the simple, rescaled time variable. For the rest of this section, we will refer to the \lilan s trained without using a time transformation $\boldsymbol{\tau}$ network as $\lilan_{\cancel{\boldsymbol{\tau}}}$.  Observing the right subfigures of \cref{Charge_time,,ac_time,,ch_time},  we see a clear benefit in terms of final relative error achieved when using \lilans over $\lilan_{\cancel{\boldsymbol{\tau}}}$. Specifically, there are two important features to recognize in each figure. First, in each case, the error is lower for \lilan. Second, when coarsening the time discretization of training data, the accuracy of $\lilan_{\cancel{\boldsymbol{\tau}}}$ degrades significantly more quickly. On the CR charge state model (\cref{Charge_time} right subfigure),  \lilans with skipped steps $=20$ outperformed  $\lilan_{\cancel{\boldsymbol{\tau}}}$ with skipped steps $=12$. For the Allen-Cahn PDE (\cref{ac_time} right subfigure)  \lilans with skipped steps $=20$ achieved lower error than a $\lilan_{\cancel{\boldsymbol{\tau}}}$ with skipped steps $=10$. For the Cahn-Hilliard PDE (\cref{ch_time} right subfigure)   \lilans with skipped steps $=50$ outperformed a $\lilan_{\cancel{\boldsymbol{\tau}}}$ with skipped steps $=25$. These results suggest that the learned time transformation helps the \lilans architecture in modeling the time dependence of the underlying dynamical system with fewer temporal data points. This result also reinforces the claims from \cref{sec:full-CR,,sec:AC,,sec:CH} that \lilans is very good at modeling time dependent behaviors. In situations where memory budget is very limited or a large number of initial conditions must be trained over, \lilans will be the better choice of architecture, in comparison to $\lilan_{\cancel{\boldsymbol{\tau}}}$, since each sample trajectory will require fewer snapshots in time, which greatly reduces the memory impact in training for each sample trajectory.

For the Robertson ODE and full CR model, shown in the right subfigures of \cref{Robertson_time,,CR_time} respectively, we did not observe significant benefits of a learned time transformation. 
Therefore, we conclude that the importance of the learned time transformation is problem dependent, however in the majority of cases investigated in our work, \lilans allowed for greater data reduction than $\lilan_{\cancel{\boldsymbol{\tau}}}$. 
Considering the fact that the additional training cost associated with training the time transformation network $\boldsymbol{\tau}$ is negligible,  the use of \lilans (with time transformation $\boldsymbol{\tau}$) is generally advisable, instead of $\lilan_{\cancel{\boldsymbol{\tau}}}$.

\begin{figure}[h!t!b!] 
\begin{minipage}{0.48\textwidth}
    \centering
    \resizebox{1\linewidth}{!}{\input{figs/ROBER_skip_discretizations}}
\end{minipage}
\begin{minipage}{0.48\textwidth}
    \centering
    \resizebox{1\linewidth}{!}{\input{figs/ROBER_tr_errors}}
\end{minipage}
\caption{Effect of downsampling data in time (for training) on the prediction accuracy for the Robertson ODE (\cref{sec:Robertson}). Left to Right: Examples of different time grids considered for training neural networks; Average point-wise relative error over time \eqref{eq:R2}  on the test data. The relationship shown indicates that significant downsampling in time can be performed without losing accuracy. For this problem, \lilans and $\lilan_{\cancel{\boldsymbol{\tau}}}$ have a similar performance decay under temporal data coarsening.}
 \label{Robertson_time}
\end{figure}

\begin{figure}[h!t!b!] 
\begin{minipage}{0.48\textwidth}
    \centering
    \resizebox{1\linewidth}{!}{\input{figs/Charge_skip_discretizations}}
\end{minipage}
\begin{minipage}{0.48\textwidth}
    \centering
    \resizebox{1\linewidth}{!}{\input{figs/Charge_tr_errors}}
\end{minipage}
\caption{Effect of downsampling data in time (for training) on the prediction accuracy for CR charge state model (\cref{sec:CR-model}). Left to Right: Examples of different time grids considered for training neural networks; Average point-wise relative error over time \eqref{eq:R2} on test data. The relationship shown indicates that significant downsampling in time can be performed without losing accuracy. For this problem, \lilans has a slower accuracy decay than $\lilan_{\cancel{\boldsymbol{\tau}}}$ under temporal data coarsening.}
 \label{Charge_time}
\end{figure}

\begin{figure}[h!t!b!] 
\begin{minipage}{0.48\textwidth}
    \centering
    \resizebox{1\linewidth}{!}{\input{figs/CR_skip_discretizations}}
\end{minipage}
\begin{minipage}{0.48\textwidth}
    \centering
    \resizebox{1\linewidth}{!}{\input{figs/CR_tr_errors}}
\end{minipage}
\caption{Effect of downsampling data in time (for training) on the prediction accuracy for the full CR model (\cref{sec:CR-model}). Left to Right: Examples of different time grids considered for training neural networks; Average point-wise relative error over time \eqref{eq:R2} when evaluating full CR model on test data. The relationship shown indicates that significant downsampling in time can be performed without losing accuracy. For this problem, \lilans and $\lilan_{\cancel{\boldsymbol{\tau}}}$ have a similar performance decay under temporal data coarsening.}
 \label{CR_time}
\end{figure}

\begin{figure}[h!t!b!] 
\begin{minipage}{0.48\textwidth}
    \centering
    \resizebox{1\linewidth}{!}{\input{figs/AC_skip_discretizations}}
\end{minipage}
\begin{minipage}{0.48\textwidth}
    \centering
    \resizebox{1\linewidth}{!}{\input{figs/ac_skip}}
\end{minipage}
\caption{Effect of downsampling data in time (for training) on the prediction accuracy for  the Allen-Cahn PDE (\cref{sec:AC}). Left to Right: Examples of different time grids considered for training neural networks; Average point-wise relative error over time \eqref{eq:R2} on test data. The relationship shown indicates that significant downsampling in time can be performed without losing accuracy. For this problem, \lilans has a slower accuracy decay than $\lilan_{\cancel{\boldsymbol{\tau}}}$ under temporal data coarsening.}
 \label{ac_time}
\end{figure}

\begin{figure}[h!t!b!] 
\begin{minipage}{0.48\textwidth}
    \centering
    \resizebox{1\linewidth}{!}{\input{figs/CH_skip_discretizations}}
\end{minipage}
\begin{minipage}{0.48\textwidth}
    \centering
    \resizebox{1\linewidth}{!}{\input{figs/CH_skip}}
    \label{fig:ch_time}
\end{minipage}
\caption{Effect of downsampling data in time (for training) on the prediction accuracy for the Cahn-Hilliard PDE (\cref{sec:CH}). Left to Right: Examples of different time grids considered for training neural networks; Average point-wise relative error over time \eqref{eq:R2} on test data. The relationship shown indicates that significant downsampling in time can be performed without losing accuracy. For this problem, \lilans has a slower accuracy decay than $\lilan_{\cancel{\boldsymbol{\tau}}}$ under temporal data coarsening.}
 \label{ch_time}
\end{figure}

%% file: figs/ROBER_dr_errors.tex
\begin{tikzpicture}

\definecolor{darkgray176}{RGB}{176,176,176}

\begin{axis}[
log basis x={2},
log basis y={10},
tick align=outside,
tick pos=left,
title={Relative Error on Test Dataset},
x grid style={darkgray176},
xlabel={Training Samples},
xmin=6, xmax=6000,
xmode=log,
xtick style={color=black},
xtick={1,4,16,64,256,1024,4096,16384,65536},
xticklabels={
  \(\displaystyle {2^{0}}\),
  \(\displaystyle {2^{2}}\),
  \(\displaystyle {2^{4}}\),
  \(\displaystyle {2^{6}}\),
  \(\displaystyle {2^{8}}\),
  \(\displaystyle {2^{10}}\),
  \(\displaystyle {2^{12}}\),
  \(\displaystyle {2^{14}}\),
  \(\displaystyle {2^{16}}\)
},
y grid style={darkgray176},
ylabel={Relative Error},
ymin=0.000201404200900347, ymax=0.283163924536979,
ymode=log,
ytick style={color=black},
ytick={1e-05,0.0001,0.001,0.01,0.1,1,10},
yticklabels={
  \(\displaystyle {10^{-5}}\),
  \(\displaystyle {10^{-4}}\),
  \(\displaystyle {10^{-3}}\),
  \(\displaystyle {10^{-2}}\),
  \(\displaystyle {10^{-1}}\),
  \(\displaystyle {10^{0}}\),
  \(\displaystyle {10^{1}}\)
}
]
\addplot [draw=black, mark=o, only marks]
table{%
x  y
8 0.203680008649826
16 0.0626891255378723
32 0.00244017015211284
64 0.00117605680134147
128 0.000834547448903322
256 0.000527558499015868
512 0.000465635588625446
1024 0.000371808651834726
2048 0.00036562405875884
4096 0.00028000000747852
};
\end{axis}

\end{tikzpicture}

%% file: figs/Charge_dr_errors.tex
\begin{tikzpicture}

\definecolor{darkgray176}{RGB}{176,176,176}

\begin{axis}[
log basis x={5},
log basis y={10},
tick align=outside,
tick pos=left,
title={Relative Error on Test Dataset},
x grid style={darkgray176},
xlabel={Training Samples},
xmin=4, xmax=20000,
xmode=log,
xtick style={color=black},
xtick={0.2,1,5,25,125,625,3125,15625,78125,390625},
xticklabels={
  \(\displaystyle {5^{-1}}\),
  \(\displaystyle {5^{0}}\),
  \(\displaystyle {5^{1}}\),
  \(\displaystyle {5^{2}}\),
  \(\displaystyle {5^{3}}\),
  \(\displaystyle {5^{4}}\),
  \(\displaystyle {5^{5}}\),
  \(\displaystyle {5^{6}}\),
  \(\displaystyle {5^{7}}\),
  \(\displaystyle {5^{8}}\)
},
y grid style={darkgray176},
ylabel={Relative Error},
ymin=0.00142276144523841, ymax=0.539956967650392,
ymode=log,
ytick style={color=black},
ytick={0.0001,0.001,0.01,0.1,1,10},
yticklabels={
  \(\displaystyle {10^{-4}}\),
  \(\displaystyle {10^{-3}}\),
  \(\displaystyle {10^{-2}}\),
  \(\displaystyle {10^{-1}}\),
  \(\displaystyle {10^{0}}\),
  \(\displaystyle {10^{1}}\)
}
]
\addplot [draw=black, mark=o, only marks]
table{%
x  y
8 0.412212878465652
25 0.0695414543151855
125 0.00598676130175591
625 0.00254386640153825
3125 0.00186367286369205
15625 0.00198099995031953
};
\end{axis}

\end{tikzpicture}

%% file: figs/CR_dr_errors.tex
\begin{tikzpicture}

\definecolor{darkgray176}{RGB}{176,176,176}

\begin{axis}[
log basis x={2},
log basis y={10},
tick align=outside,
tick pos=left,
title={Relative Error on Test Dataset},
x grid style={darkgray176},
xlabel={Training Samples},
xmin=6, xmax=6000,
xmode=log,
xtick style={color=black},
xtick={1,4,16,64,256,1024,4096,16384,65536},
xticklabels={
  \(\displaystyle {2^{0}}\),
  \(\displaystyle {2^{2}}\),
  \(\displaystyle {2^{4}}\),
  \(\displaystyle {2^{6}}\),
  \(\displaystyle {2^{8}}\),
  \(\displaystyle {2^{10}}\),
  \(\displaystyle {2^{12}}\),
  \(\displaystyle {2^{14}}\),
  \(\displaystyle {2^{16}}\)
},
y grid style={darkgray176},
ylabel={Relative Error},
ymin=0.00579443425613876, ymax=0.663176535482284,
ymode=log,
ytick style={color=black},
ytick={0.0001,0.001,0.01,0.1,1,10},
yticklabels={
  \(\displaystyle {10^{-4}}\),
  \(\displaystyle {10^{-3}}\),
  \(\displaystyle {10^{-2}}\),
  \(\displaystyle {10^{-1}}\),
  \(\displaystyle {10^{0}}\),
  \(\displaystyle {10^{1}}\)
}
]
\addplot [draw=black, mark=o, only marks]
table{%
x  y
8 0.534632802009583
16 0.30271577835083
32 0.0763823986053467
64 0.039219107478857
128 0.0189175773411989
256 0.0105170635506511
512 0.00992537382990122
1024 0.00882995035499334
2048 0.00718761142343283
4096 0.00748600019142032
};
\end{axis}

\end{tikzpicture}

%% file: figs/AC_dr_errors.tex
\begin{tikzpicture}

\definecolor{darkgray176}{RGB}{176,176,176}

\begin{axis}[
log basis y={10},
minor ytick={0.002,0.003,0.004,0.005,0.006,0.007,0.008,0.009,0.02,0.03,0.04,0.05,0.06,0.07,0.08,0.09,0.2,0.3,0.4,0.5,0.6,0.7,0.8,0.9,2,3,4,5,6,7,8,9,20,30,40,50,60,70,80,90},
tick align=outside,
tick pos=left,
title={Relative Error on Holdout Dataset},
x grid style={darkgray176},
xlabel={Training Samples},
xmin=50, xmax=950,
xtick style={color=black},
y grid style={darkgray176},
ylabel={Relative Error},
ymin=0.00944699228428446, ymax=0.152029022173976,
ymode=log,
ytick style={color=black},
ytick={0.0001,0.001,0.01,0.1,1,10},
yticklabels={
  \(\displaystyle {10^{-4}}\),
  \(\displaystyle {10^{-3}}\),
  \(\displaystyle {10^{-2}}\),
  \(\displaystyle {10^{-1}}\),
  \(\displaystyle {10^{0}}\),
  \(\displaystyle {10^{1}}\)
}
]
\addplot [draw=black, mark=o, only marks]
table{%
x  y
100 0.133992210030556
200 0.0698755532503128
300 0.0460527464747429
400 0.032859742641449
500 0.0243665538728237
600 0.0192273873835802
700 0.0160225853323936
800 0.0114724021404982
900 0.0107186604291201
};
\end{axis}

\end{tikzpicture}

%% file: figs/CH_dr_errors.tex
\begin{tikzpicture}

\definecolor{darkgray176}{RGB}{176,176,176}

\begin{axis}[
log basis y={10},
minor ytick={0.002,0.003,0.004,0.005,0.006,0.007,0.008,0.009,0.02,0.03,0.04,0.05,0.06,0.07,0.08,0.09,0.2,0.3,0.4,0.5,0.6,0.7,0.8,0.9,2,3,4,5,6,7,8,9,20,30,40,50,60,70,80,90},
tick align=outside,
tick pos=left,
title={Relative Error on Test Dataset},
x grid style={darkgray176},
xlabel={Training Samples},
xmin=50, xmax=950,
xtick style={color=black},
y grid style={darkgray176},
ylabel={Relative Error},
ymin=0.0464419001300416, ymax=0.348884623817227,
ymode=log,
ytick style={color=black},
ytick={0.001,0.01,0.1,1,10},
yticklabels={
  \(\displaystyle {10^{-3}}\),
  \(\displaystyle {10^{-2}}\),
  \(\displaystyle {10^{-1}}\),
  \(\displaystyle {10^{0}}\),
  \(\displaystyle {10^{1}}\)
}
]
\addplot [draw=black, mark=o, only marks]
table{%
x  y
100 0.318327397108078
200 0.177455902099609
300 0.146176785230637
400 0.120832175016403
500 0.0873842015862465
600 0.0716001912951469
700 0.0653960183262825
800 0.0557465627789497
900 0.0509000010788441
};
\end{axis}

\end{tikzpicture}

%% file: figs/ROBER_skip_discretizations.tex
\begin{tikzpicture}

\definecolor{darkgray176}{RGB}{176,176,176}

\begin{axis}[
log basis x={10},
log basis y={10},
tick align=outside,
tick pos=left,
title={Various Training Time Discretizations},
x grid style={darkgray176},
xlabel={Times (s)},
xmin=1.25e-05, xmax=125000,
xmode=log,
xtick style={color=black},
xtick={1e-07,1e-05,0.001,0.1,10,1000,100000,10000000},
xticklabels={
  \(\displaystyle {10^{-7}}\),
  \(\displaystyle {10^{-5}}\),
  \(\displaystyle {10^{-3}}\),
  \(\displaystyle {10^{-1}}\),
  \(\displaystyle {10^{1}}\),
  \(\displaystyle {10^{3}}\),
  \(\displaystyle {10^{5}}\),
  \(\displaystyle {10^{7}}\)
},
y grid style={darkgray176},
ylabel={Steps Retained},
ymin=3.52901242060987, ymax=55.5396175018642,
ymode=log,
ytick style={color=black},
ytick={1,2,3,4,5,6,7,8,9,10,20,30,40,50,60,70,80,90,100,200,300,400},
yticklabels={
  \(\displaystyle {10^{0}}\),
  \(\displaystyle {}\),
  \(\displaystyle {}\),
  \(\displaystyle {}\),
  \(\displaystyle {}\),
  \(\displaystyle {}\),
  \(\displaystyle {}\),
  \(\displaystyle {}\),
  \(\displaystyle {}\),
  \(\displaystyle {10^{1}}\),
  \(\displaystyle {}\),
  \(\displaystyle {}\),
  \(\displaystyle {}\),
  \(\displaystyle {}\),
  \(\displaystyle {}\),
  \(\displaystyle {}\),
  \(\displaystyle {}\),
  \(\displaystyle {}\),
  \(\displaystyle {10^{2}}\),
  \(\displaystyle {}\),
  \(\displaystyle {}\),
  \(\displaystyle {}\),
}
]
\addplot [draw=black, fill=black, mark=*, only marks]
table{%
x  y
1.59985829668585e-05 49
2.55954673775705e-05 49
4.09491221944336e-05 49
6.55128678772599e-05 49
0.000104811369965319 49
0.000167683348990977 49
0.000268269592197612 49
0.000429193605668843 49
0.000686648942064494 49
0.0010985410772264 49
0.00175751012284309 49
0.00281176739372313 49
0.0044984295964241 49
0.00719685014337301 49
0.0115139465779066 49
0.0184206832200289 49
0.0294704865664244 49
0.0471486672759056 49
0.075431190431118 49
0.120679251849651 49
0.193069815635681 49
0.308884352445602 49
0.494171231985092 49
0.790603935718536 49
1.26485502719879 49
2.02358889579773 49
3.23745584487915 49
5.17947053909302 49
8.2864236831665 49
13.2571115493774 49
21.2095012664795 49
33.9321975708008 49
54.2867126464844 49
86.8510513305664 49
138.949447631836 49
222.299560546875 49
355.647827148438 49
568.986145019531 49
910.297241210938 49
1456.34826660156 49
2329.95092773438 49
3727.59350585938 49
5963.62158203125 49
9540.94921875 49
15264.16015625 49
24420.494140625 49
39069.33203125 49
62505.46484375 49
100000 49
};
\addplot [draw=black, fill=black, mark=*, only marks]
table{%
x  y
0 26
1.59985829668585e-05 26
4.09491221944336e-05 26
0.000104811369965319 26
0.000268269592197612 26
0.000686648942064494 26
0.00175751012284309 26
0.0044984295964241 26
0.0115139465779066 26
0.0294704865664244 26
0.075431190431118 26
0.193069815635681 26
0.494171231985092 26
1.26485502719879 26
3.23745584487915 26
8.2864236831665 26
21.2095012664795 26
54.2867126464844 26
138.949447631836 26
355.647827148438 26
910.297241210938 26
2329.95092773438 26
5963.62158203125 26
15264.16015625 26
39069.33203125 26
100000 26
};
\addplot [draw=black, fill=black, mark=*, only marks]
table{%
x  y
0 14
1.59985829668585e-05 14
0.000104811369965319 14
0.000686648942064494 14
0.0044984295964241 14
0.0294704865664244 14
0.193069815635681 14
1.26485502719879 14
8.2864236831665 14
54.2867126464844 14
355.647827148438 14
2329.95092773438 14
15264.16015625 14
100000 14
};
\addplot [draw=black, fill=black, mark=*, only marks]
table{%
x  y
0 10
1.59985829668585e-05 10
0.000268269592197612 10
0.0044984295964241 10
0.075431190431118 10
1.26485502719879 10
21.2095012664795 10
355.647827148438 10
5963.62158203125 10
100000 10
};
\addplot [draw=black, fill=black, mark=*, only marks]
table{%
x  y
0 8
1.59985829668585e-05 8
0.000686648942064494 8
0.0294704865664244 8
1.26485502719879 8
54.2867126464844 8
2329.95092773438 8
100000 8
};
\addplot [draw=black, fill=black, mark=*, only marks]
table{%
x  y
0 6
1.59985829668585e-05 6
0.0044984295964241 6
1.26485502719879 6
355.647827148438 6
100000 6
};
\addplot [draw=black, fill=black, mark=*, only marks]
table{%
x  y
0 5
1.59985829668585e-05 5
0.0294704865664244 5
54.2867126464844 5
100000 5
};
\addplot [draw=black, fill=black, mark=*, only marks]
table{%
x  y
0 4
1.59985829668585e-05 4
1.26485502719879 4
100000 4
};
\end{axis}

\end{tikzpicture}

%% file: figs/ROBER_tr_errors.tex

\begin{tikzpicture}

\definecolor{darkgray176}{RGB}{176,176,176}
\definecolor{lightgray204}{RGB}{204,204,204}

\begin{axis}[
legend cell align={left},
legend style={
  fill opacity=0.8,
  draw opacity=1,
  text opacity=1,
  at={(0.03,0.97)},
  anchor=north west,
  draw=lightgray204
},
log basis x={10},
log basis y={10},
tick align=outside,
tick pos=left,
title={Relative Error on Test Dataset},
x grid style={darkgray176},
xlabel={Skipped Steps},
xmin=0.853079366789105, xmax=28.1333729713019,
xmode=log,
xtick style={color=black},
xtick={1,2,3,4,5,6,7,8,9,10,20,30,40,50,60, 70,80,90,100,200,300},
xticklabels={
  \(\displaystyle {10^{0}}\),
  \(\displaystyle {}\),
  \(\displaystyle {}\),
  \(\displaystyle {}\),
  \(\displaystyle {}\),
  \(\displaystyle {}\),
  \(\displaystyle {}\),
  \(\displaystyle {}\),
  \(\displaystyle {}\),
  \(\displaystyle {10^{1}}\),
  \(\displaystyle {}\),
  \(\displaystyle {}\),
  \(\displaystyle {}\),
  \(\displaystyle {}\),
  \(\displaystyle {}\),
  \(\displaystyle {}\),
  \(\displaystyle {}\),
  \(\displaystyle {}\),
  \(\displaystyle {10^{2}}\),
  \(\displaystyle {}\),
  \(\displaystyle {}\),
},
y grid style={darkgray176},
ylabel={Relative Error},
ymin=0.00056525672997053, ymax=0.340571094231357,
ymode=log,
ytick style={color=black},
ytick={0.001,0.002,0.003,0.004,0.005,0.006,0.007,0.008,0.009,0.01,0.02,0.03,0.04,0.05,0.06,0.07,0.08,0.09,0.1,0.2,0.3,0.4,0.5},
yticklabels={
  \(\displaystyle {10^{-3}}\),
  \(\displaystyle {}\),
  \(\displaystyle {}\),
  \(\displaystyle {}\),
  \(\displaystyle {}\),
  \(\displaystyle {}\),
  \(\displaystyle {}\),
  \(\displaystyle {}\),
  \(\displaystyle {}\),
  \(\displaystyle {10^{-2}}\),
  \(\displaystyle {}\),
  \(\displaystyle {}\),
  \(\displaystyle {}\),
  \(\displaystyle {}\),
  \(\displaystyle {}\),
  \(\displaystyle {}\),
  \(\displaystyle {}\),
  \(\displaystyle {}\),
  \(\displaystyle {10^{-1}}\),
  \(\displaystyle {}\),
  \(\displaystyle {}\),
  \(\displaystyle {}\),
  \(\displaystyle {}\),
}
]
\addplot [draw=black, mark=o, only marks]
table{%
x  y
1 0.000834547448903322
2 0.000987709267064929
4 0.000827469048090279
6 0.00115811603609473
8 0.00142520433291793
12 0.0241827107965946
16 0.0573260225355625
24 0.245794907212257
};
\addlegendentry{\lilan}
\addplot [draw=red, fill=red, mark=*, only marks]
table{%
x  y
1 0.000904617481864989
2 0.000756149471271783
4 0.000926435168366879
6 0.00108375179115683
8 0.0022453349083662
12 0.020584961399436
16 0.055563673377037
24 0.254592657089233
};
\addlegendentry{$\lilan_{\cancel{\boldsymbol{\tau}}}$}
\end{axis}

\end{tikzpicture}

%% file: figs/Charge_skip_discretizations.tex
\begin{tikzpicture}

\definecolor{darkgray176}{RGB}{176,176,176}

\begin{axis}[
log basis x={10},
log basis y={10},
tick align=outside,
tick pos=left,
title={Various Training Time Discretizations},
x grid style={darkgray176},
xlabel={Times (s)},
xmin=0.75e-16, xmax=1.40,
xmode=log,
xtick style={color=black},
xtick={1e-18,1e-16,1e-14,1e-12,1e-10,1e-08,1e-06,0.0001,0.01,1,100},
xticklabels={
  \(\displaystyle {10^{-18}}\),
  \(\displaystyle {10^{-16}}\),
  \(\displaystyle {10^{-14}}\),
  \(\displaystyle {10^{-12}}\),
  \(\displaystyle {10^{-10}}\),
  \(\displaystyle {10^{-8}}\),
  \(\displaystyle {10^{-6}}\),
  \(\displaystyle {10^{-4}}\),
  \(\displaystyle {10^{-2}}\),
  \(\displaystyle {10^{0}}\),
  \(\displaystyle {10^{2}}\)
},
y grid style={darkgray176},
ylabel={Steps Retained},
ymin=3.17771062518461, ymax=502.248375717749,
ymode=log,
ytick style={color=black},
ytick={1,2,3,4,5,6,7,8,9,10,20,30,40,50,60,70,80,90,100,200,300,400},
yticklabels={
  \(\displaystyle {10^{0}}\),
  \(\displaystyle {}\),
  \(\displaystyle {}\),
  \(\displaystyle {}\),
  \(\displaystyle {}\),
  \(\displaystyle {}\),
  \(\displaystyle {}\),
  \(\displaystyle {}\),
  \(\displaystyle {}\),
  \(\displaystyle {10^{1}}\),
  \(\displaystyle {}\),
  \(\displaystyle {}\),
  \(\displaystyle {}\),
  \(\displaystyle {}\),
  \(\displaystyle {}\),
  \(\displaystyle {}\),
  \(\displaystyle {}\),
  \(\displaystyle {}\),
  \(\displaystyle {10^{2}}\),
  \(\displaystyle {}\),
  \(\displaystyle {}\),
  \(\displaystyle {}\),
}
]
\addplot [draw=black, fill=black, mark=*, only marks]
table{%
x  y
1.09673091474103e-16 399
1.20281881353442e-16 399
1.31916858265826e-16 399
1.44677621374448e-16 399
1.58672418842992e-16 399
1.74020947014816e-16 399
1.90854165124498e-16 399
2.0931567172638e-16 399
2.29562975243979e-16 399
2.51769373324349e-16 399
2.76123264623805e-16 399
3.02832913368252e-16 399
3.32126224360062e-16 399
3.64253125258233e-16 399
3.99487657690983e-16 399
4.38131444796872e-16 399
4.80512314796267e-16 399
5.26992718381255e-16 399
5.77969225789828e-16 399
6.33876735500841e-16 399
6.95192232942718e-16 399
7.62440507965842e-16 399
8.36192090199748e-16 399
9.17077701552784e-16 399
1.00578756799785e-15 399
1.10307830171507e-15 399
1.20978010087187e-15 399
1.32680331412833e-15 399
1.45514945206545e-15 399
1.59590748141604e-15 399
1.75028103599815e-15 399
1.91958735792387e-15 399
2.10527095600565e-15 399
2.30892075817339e-15 399
2.53226492339775e-15 399
2.77721329921654e-15 399
3.04585572766887e-15 399
3.34048427991002e-15 399
3.66361231445272e-15 399
4.01799734705676e-15 399
4.40667154391508e-15 399
4.83293325132428e-15 399
5.3004272520564e-15 399
5.81314264777653e-15 399
6.3754535166243e-15 399
6.99215702969647e-15 399
7.66853147280387e-15 399
8.41031634119709e-15 399
9.22385421758517e-15 399
1.01160856899379e-14 399
1.10946247374182e-14 399
1.21678180226882e-14 399
1.33448228189408e-14 399
1.46357128890176e-14 399
1.60514386748608e-14 399
1.76041093594842e-14 399
1.93069708441171e-14 399
2.11745531319366e-14 399
2.32228380405916e-14 399
2.54692049920938e-14 399
2.79328663877528e-14 399
3.063483711499e-14 399
3.35981751046956e-14 399
3.68481592081474e-14 399
4.0412514507775e-14 399
4.43216562626483e-14 399
4.86090388860522e-14 399
5.33110373608734e-14 399
5.84678654233159e-14 399
6.412351796466e-14 399
7.03262453697132e-14 399
7.71291396636092e-14 399
8.45899142832424e-14 399
9.27723745264633e-14 399
1.01746336236917e-13 399
1.11588350864243e-13 399
1.22382396642966e-13 399
1.34220569771373e-13 399
1.4720416183743e-13 399
1.61443368439443e-13 399
1.77059945481381e-13 399
1.94187114305189e-13 399
2.1297101181119e-13 399
2.33572411629038e-13 399
2.56166100701463e-13 399
2.80945297408131e-13 399
3.0812139406762e-13 399
3.37926247314518e-13 399
3.70614183873436e-13 399
4.06464060543135e-13 399
4.45781703651421e-13 399
4.88903676657687e-13 399
5.36195762269875e-13 399
5.8806247631854e-13 399
6.44946335920382e-13 399
7.07334147850847e-13 399
7.7575527382065e-13 399
8.50794803532173e-13 399
9.33093012578379e-13 399
1.02335197607606e-12 399
1.12234180282833e-12 399
1.23090697790734e-12 399
1.34997373522905e-12 399
1.48055797222907e-12 399
1.6237772845068e-12 399
1.78084684585716e-12 399
1.95310976593144e-12 399
2.14204066810442e-12 399
2.34924232844769e-12 399
2.57648676699196e-12 399
2.82571275406207e-12 399
3.0990466269587e-12 399
3.39882028771443e-12 399
3.72759158609459e-12 399
4.0881647025981e-12 399
4.48361660299046e-12 399
4.91732132443246e-12 399
5.39299030780094e-12 399
5.91465938726277e-12 399
6.48679027159815e-12 399
7.11427878413717e-12 399
7.80244965858934e-12 399
8.5571879443469e-12 399
9.38493293495268e-12 399
1.02927466896774e-11 399
1.12883738814795e-11 399
1.23803085838592e-11 399
1.35778679966059e-11 399
1.48912670389079e-11 399
1.63317501883364e-11 399
1.79115351023329e-11 399
1.96441352767751e-11 399
2.15443791279357e-11 399
2.36283857063535e-11 399
2.59139827651911e-11 399
2.84206668615949e-11 399
3.11698236665769e-11 399
3.41849118457027e-11 399
3.74916486745036e-11 399
4.11182546344868e-11 399
4.5095659845229e-11 399
4.94578059062611e-11 399
5.42420240667862e-11 399
5.94889068561422e-11 399
6.52433246317585e-11 399
7.15545331320122e-11 399
7.84760728622658e-11 399
8.60671325875195e-11 399
9.43924868823665e-11 399
1.03523162264807e-10 399
1.13537054757806e-10 399
1.24519602962003e-10 399
1.36564495822888e-10 399
1.49774512236966e-10 399
1.64262714541508e-10 399
1.8015199054755e-10 399
1.97578273186672e-10 399
2.16690679288867e-10 399
2.3765137080467e-10 399
2.60639620996983e-10 399
2.85851536885318e-10 399
3.13502224180695e-10 399
3.43827577520273e-10 399
3.77086362135515e-10 399
4.13562267764789e-10 399
4.53566545433048e-10 399
4.97440477698063e-10 399
5.45559541986762e-10 399
5.98331995060875e-10 399
6.56209253602213e-10 399
7.19686532590913e-10 399
7.89302512238521e-10 399
8.65652494130842e-10 399
9.49387901627574e-10 399
1.04122310773391e-09 399
1.14194154132718e-09 399
1.25240273707306e-09 399
1.37354871920792e-09 399
1.50641343843461e-09 399
1.65213387415264e-09 399
1.81194637072224e-09 399
1.98721772370902e-09 399
2.17944773339696e-09 399
2.39026798354303e-09 399
2.62148081020541e-09 399
2.87505907969887e-09 399
3.15316617260919e-09 399
3.4581750796292e-09 399
3.79268749739481e-09 399
4.15955803134693e-09 399
4.56191573405818e-09 399
5.00320540552934e-09 399
5.48716982962105e-09 399
6.01794925003674e-09 399
6.60007115627081e-09 399
7.2385177851686e-09 399
7.93870658100104e-09 399
8.70662564267377e-09 399
9.54882573012128e-09 399
1.04724922067589e-08 399
1.14855067678832e-08 399
1.25965105013393e-08 399
1.38150131334669e-08 399
1.5151352172893e-08 399
1.66169566995222e-08 399
1.82243304891472e-08 399
1.9987187016568e-08 399
2.19205666951439e-08 399
2.40409647744855e-08 399
2.63665267397073e-08 399
2.89169861389382e-08 399
3.17141548578093e-08 399
3.47818946977441e-08 399
3.81463785004144e-08 399
4.18363157450585e-08 399
4.58832829508538e-08 399
5.0321613542792e-08 399
5.51892718192448e-08 399
6.05277818976901e-08 399
6.63826895674902e-08 399
7.28039566411098e-08 399
7.98465222828781e-08 399
8.75701573477272e-08 399
9.60408996775186e-08 399
1.05331025679334e-07 399
1.15519796395347e-07 399
1.26694132518423e-07 399
1.38949687311651e-07 399
1.52390413177272e-07 399
1.67131290140787e-07 399
1.83298055844716e-07 399
2.01028640844925e-07 399
2.20474333900711e-07 399
2.41801018319165e-07 399
2.65191260950814e-07 399
2.90843445327482e-07 399
3.18977015467681e-07 399
3.49831964285841e-07 399
3.83671533654706e-07 399
4.2078445972038e-07 399
4.61488326664039e-07 399
5.06128515098681e-07 399
5.55086842268793e-07 399
6.08780908351036e-07 399
6.6766887130143e-07 399
7.32253113255865e-07 399
8.03086379619344e-07 399
8.80769675859483e-07 399
9.65967387855926e-07 399
1.05940637240565e-06 399
1.16188368792791e-06 399
1.27427381357847e-06 399
1.39753854000446e-06 399
1.53272378611291e-06 399
1.68098563335661e-06 399
1.84358896149206e-06 399
2.02192109099997e-06 399
2.21750337914273e-06 399
2.43200452132442e-06 399
2.66726055997424e-06 399
2.92526715384156e-06 399
3.20823096444656e-06 399
3.51856624547509e-06 399
3.85892053600401e-06 399
4.23219762524241e-06 399
4.64159211333026e-06 399
5.09057781528099e-06 399
5.58299416297814e-06 399
6.12304256719653e-06 399
6.71533007334801e-06 399
7.36491028874298e-06 399
8.07734340924071e-06 399
8.8586721176398e-06 399
9.71557983575622e-06 399
1.06553770820028e-05 399
1.16860819616704e-05 399
1.28164874695358e-05 399
1.40562688102364e-05 399
1.54159461089876e-05 399
1.69071445270674e-05 399
1.85425888048485e-05 399
2.03362305910559e-05 399
2.23033730435418e-05 399
2.44607999775326e-05 399
2.68269741354743e-05 399
2.94219735224033e-05 399
3.22679879900534e-05 399
3.53893010469619e-05 399
3.88125408790074e-05 399
4.2566916818032e-05 399
4.66845558548812e-05 399
5.12003971380182e-05 399
5.61530614504591e-05 399
6.15847966400906e-05 399
6.75419578328729e-05 399
7.40753530408256e-05 399
8.1240912550129e-05 399
8.9099419710692e-05 399
9.77180898189545e-05 399
0.000107170453702565 399
0.000117537150799762 399
0.000128906627651304 399
0.000141376207466237 399
0.0001550516608404 399
0.000170049956068397 399
0.000186499048140831 399
0.000204539275728166 399
0.000224324554437771 399
0.000246023671934381 399
0.00026982236886397 399
0.000295922538498417 399
0.000324547407217324 399
0.000355941185262054 399
0.000390371715184301 399
0.000428132741944864 399
0.000469547463580966 399
0.00051496725063771 399
0.000564780493732542 399
0.000619412225205451 399
0.000679328572005033 399
0.00074504065560177 399
0.000817110994830728 399
0.00089615088654682 399
0.000982836354523897 399
0.00107790704350919 399
0.00118217407725751 399
0.00129652686882764 399
0.00142194423824549 399
0.00155949033796787 399
0.00171034119557589 399
0.00187578413169831 399
0.00205723056569695 399
0.00225622835569084 399
0.0024744754191488 399
0.00271383975632489 399
0.00297635211609304 399
0.00326425745151937 399
0.00358001212589443 399
0.00392631022259593 399
0.00430610589683056 399
0.00472265016287565 399
0.0051794764585793 399
0.00568049214780331 399
0.00622997153550386 399
0.0068326024338603 399
0.00749352620914578 399
0.00821840018033981 399
0.00901337433606386 399
0.00988524593412876 399
0.0108414553105831 399
0.011890159919858 399
0.0130403060466051 399
0.0143017387017608 399
0.0156851597130299 399
0.0172023996710777 399
0.0188664030283689 399
0.0206913687288761 399
0.0226928647607565 399
0.0248879659920931 399
0.0272954627871513 399
0.0299357790499926 399
0.0328314937651157 399
0.0360073186457157 399
0.0394903384149075 399
0.0433102771639824 399
0.0474998243153095 399
0.0520945303142071 399
0.0571336820721626 399
0.062660276889801 399
0.0687214657664299 399
0.0753689557313919 399
0.0826596468687057 399
0.0906553938984871 399
0.0994245782494545 399
0.109042011201382 399
0.119589745998383 399
0.131157770752907 399
0.143845111131668 399
0.157759383320808 399
0.173019587993622 399
0.189755946397781 399
0.208111211657524 399
0.228242009878159 399
0.250320613384247 399
0.274534374475479 399
0.301090329885483 399
0.330215096473694 399
0.362157106399536 399
0.397188901901245 399
0.435609370470047 399
0.477747321128845 399
0.523960292339325 399
0.574643433094025 399
0.630229234695435 399
0.691191911697388 399
0.758051574230194 399
0.831380426883698 399
0.911800682544708 399
1 399
};
\addplot [draw=black, fill=black, mark=*, only marks]
table{%
x  y
0 201
1.09673091474103e-16 201
1.31916858265826e-16 201
1.58672418842992e-16 201
1.90854165124498e-16 201
2.29562975243979e-16 201
2.76123264623805e-16 201
3.32126224360062e-16 201
3.99487657690983e-16 201
4.80512314796267e-16 201
5.77969225789828e-16 201
6.95192232942718e-16 201
8.36192090199748e-16 201
1.00578756799785e-15 201
1.20978010087187e-15 201
1.45514945206545e-15 201
1.75028103599815e-15 201
2.10527095600565e-15 201
2.53226492339775e-15 201
3.04585572766887e-15 201
3.66361231445272e-15 201
4.40667154391508e-15 201
5.3004272520564e-15 201
6.3754535166243e-15 201
7.66853147280387e-15 201
9.22385421758517e-15 201
1.10946247374182e-14 201
1.33448228189408e-14 201
1.60514386748608e-14 201
1.93069708441171e-14 201
2.32228380405916e-14 201
2.79328663877528e-14 201
3.35981751046956e-14 201
4.0412514507775e-14 201
4.86090388860522e-14 201
5.84678654233159e-14 201
7.03262453697132e-14 201
8.45899142832424e-14 201
1.01746336236917e-13 201
1.22382396642966e-13 201
1.4720416183743e-13 201
1.77059945481381e-13 201
2.1297101181119e-13 201
2.56166100701463e-13 201
3.0812139406762e-13 201
3.70614183873436e-13 201
4.45781703651421e-13 201
5.36195762269875e-13 201
6.44946335920382e-13 201
7.7575527382065e-13 201
9.33093012578379e-13 201
1.12234180282833e-12 201
1.34997373522905e-12 201
1.6237772845068e-12 201
1.95310976593144e-12 201
2.34924232844769e-12 201
2.82571275406207e-12 201
3.39882028771443e-12 201
4.0881647025981e-12 201
4.91732132443246e-12 201
5.91465938726277e-12 201
7.11427878413717e-12 201
8.5571879443469e-12 201
1.02927466896774e-11 201
1.23803085838592e-11 201
1.48912670389079e-11 201
1.79115351023329e-11 201
2.15443791279357e-11 201
2.59139827651911e-11 201
3.11698236665769e-11 201
3.74916486745036e-11 201
4.5095659845229e-11 201
5.42420240667862e-11 201
6.52433246317585e-11 201
7.84760728622658e-11 201
9.43924868823665e-11 201
1.13537054757806e-10 201
1.36564495822888e-10 201
1.64262714541508e-10 201
1.97578273186672e-10 201
2.3765137080467e-10 201
2.85851536885318e-10 201
3.43827577520273e-10 201
4.13562267764789e-10 201
4.97440477698063e-10 201
5.98331995060875e-10 201
7.19686532590913e-10 201
8.65652494130842e-10 201
1.04122310773391e-09 201
1.25240273707306e-09 201
1.50641343843461e-09 201
1.81194637072224e-09 201
2.17944773339696e-09 201
2.62148081020541e-09 201
3.15316617260919e-09 201
3.79268749739481e-09 201
4.56191573405818e-09 201
5.48716982962105e-09 201
6.60007115627081e-09 201
7.93870658100104e-09 201
9.54882573012128e-09 201
1.14855067678832e-08 201
1.38150131334669e-08 201
1.66169566995222e-08 201
1.9987187016568e-08 201
2.40409647744855e-08 201
2.89169861389382e-08 201
3.47818946977441e-08 201
4.18363157450585e-08 201
5.0321613542792e-08 201
6.05277818976901e-08 201
7.28039566411098e-08 201
8.75701573477272e-08 201
1.05331025679334e-07 201
1.26694132518423e-07 201
1.52390413177272e-07 201
1.83298055844716e-07 201
2.20474333900711e-07 201
2.65191260950814e-07 201
3.18977015467681e-07 201
3.83671533654706e-07 201
4.61488326664039e-07 201
5.55086842268793e-07 201
6.6766887130143e-07 201
8.03086379619344e-07 201
9.65967387855926e-07 201
1.16188368792791e-06 201
1.39753854000446e-06 201
1.68098563335661e-06 201
2.02192109099997e-06 201
2.43200452132442e-06 201
2.92526715384156e-06 201
3.51856624547509e-06 201
4.23219762524241e-06 201
5.09057781528099e-06 201
6.12304256719653e-06 201
7.36491028874298e-06 201
8.8586721176398e-06 201
1.06553770820028e-05 201
1.28164874695358e-05 201
1.54159461089876e-05 201
1.85425888048485e-05 201
2.23033730435418e-05 201
2.68269741354743e-05 201
3.22679879900534e-05 201
3.88125408790074e-05 201
4.66845558548812e-05 201
5.61530614504591e-05 201
6.75419578328729e-05 201
8.1240912550129e-05 201
9.77180898189545e-05 201
0.000117537150799762 201
0.000141376207466237 201
0.000170049956068397 201
0.000204539275728166 201
0.000246023671934381 201
0.000295922538498417 201
0.000355941185262054 201
0.000428132741944864 201
0.00051496725063771 201
0.000619412225205451 201
0.00074504065560177 201
0.00089615088654682 201
0.00107790704350919 201
0.00129652686882764 201
0.00155949033796787 201
0.00187578413169831 201
0.00225622835569084 201
0.00271383975632489 201
0.00326425745151937 201
0.00392631022259593 201
0.00472265016287565 201
0.00568049214780331 201
0.0068326024338603 201
0.00821840018033981 201
0.00988524593412876 201
0.011890159919858 201
0.0143017387017608 201
0.0172023996710777 201
0.0206913687288761 201
0.0248879659920931 201
0.0299357790499926 201
0.0360073186457157 201
0.0433102771639824 201
0.0520945303142071 201
0.062660276889801 201
0.0753689557313919 201
0.0906553938984871 201
0.109042011201382 201
0.131157770752907 201
0.157759383320808 201
0.189755946397781 201
0.228242009878159 201
0.274534374475479 201
0.330215096473694 201
0.397188901901245 201
0.477747321128845 201
0.574643433094025 201
0.691191911697388 201
0.831380426883698 201
1 201
};
\addplot [draw=black, fill=black, mark=*, only marks]
table{%
x  y
1.09673091474103e-16 134
1.31916858265826e-16 134
1.74020947014816e-16 134
2.29562975243979e-16 134
3.02832913368252e-16 134
3.99487657690983e-16 134
5.26992718381255e-16 134
6.95192232942718e-16 134
9.17077701552784e-16 134
1.20978010087187e-15 134
1.59590748141604e-15 134
2.10527095600565e-15 134
2.77721329921654e-15 134
3.66361231445272e-15 134
4.83293325132428e-15 134
6.3754535166243e-15 134
8.41031634119709e-15 134
1.10946247374182e-14 134
1.46357128890176e-14 134
1.93069708441171e-14 134
2.54692049920938e-14 134
3.35981751046956e-14 134
4.43216562626483e-14 134
5.84678654233159e-14 134
7.71291396636092e-14 134
1.01746336236917e-13 134
1.34220569771373e-13 134
1.77059945481381e-13 134
2.33572411629038e-13 134
3.0812139406762e-13 134
4.06464060543135e-13 134
5.36195762269875e-13 134
7.07334147850847e-13 134
9.33093012578379e-13 134
1.23090697790734e-12 134
1.6237772845068e-12 134
2.14204066810442e-12 134
2.82571275406207e-12 134
3.72759158609459e-12 134
4.91732132443246e-12 134
6.48679027159815e-12 134
8.5571879443469e-12 134
1.12883738814795e-11 134
1.48912670389079e-11 134
1.96441352767751e-11 134
2.59139827651911e-11 134
3.41849118457027e-11 134
4.5095659845229e-11 134
5.94889068561422e-11 134
7.84760728622658e-11 134
1.03523162264807e-10 134
1.36564495822888e-10 134
1.8015199054755e-10 134
2.3765137080467e-10 134
3.13502224180695e-10 134
4.13562267764789e-10 134
5.45559541986762e-10 134
7.19686532590913e-10 134
9.49387901627574e-10 134
1.25240273707306e-09 134
1.65213387415264e-09 134
2.17944773339696e-09 134
2.87505907969887e-09 134
3.79268749739481e-09 134
5.00320540552934e-09 134
6.60007115627081e-09 134
8.70662564267377e-09 134
1.14855067678832e-08 134
1.5151352172893e-08 134
1.9987187016568e-08 134
2.63665267397073e-08 134
3.47818946977441e-08 134
4.58832829508538e-08 134
6.05277818976901e-08 134
7.98465222828781e-08 134
1.05331025679334e-07 134
1.38949687311651e-07 134
1.83298055844716e-07 134
2.41801018319165e-07 134
3.18977015467681e-07 134
4.2078445972038e-07 134
5.55086842268793e-07 134
7.32253113255865e-07 134
9.65967387855926e-07 134
1.27427381357847e-06 134
1.68098563335661e-06 134
2.21750337914273e-06 134
2.92526715384156e-06 134
3.85892053600401e-06 134
5.09057781528099e-06 134
6.71533007334801e-06 134
8.8586721176398e-06 134
1.16860819616704e-05 134
1.54159461089876e-05 134
2.03362305910559e-05 134
2.68269741354743e-05 134
3.53893010469619e-05 134
4.66845558548812e-05 134
6.15847966400906e-05 134
8.1240912550129e-05 134
0.000107170453702565 134
0.000141376207466237 134
0.000186499048140831 134
0.000246023671934381 134
0.000324547407217324 134
0.000428132741944864 134
0.000564780493732542 134
0.00074504065560177 134
0.000982836354523897 134
0.00129652686882764 134
0.00171034119557589 134
0.00225622835569084 134
0.00297635211609304 134
0.00392631022259593 134
0.0051794764585793 134
0.0068326024338603 134
0.00901337433606386 134
0.011890159919858 134
0.0156851597130299 134
0.0206913687288761 134
0.0272954627871513 134
0.0360073186457157 134
0.0474998243153095 134
0.062660276889801 134
0.0826596468687057 134
0.109042011201382 134
0.143845111131668 134
0.189755946397781 134
0.250320613384247 134
0.330215096473694 134
0.435609370470047 134
0.574643433094025 134
0.758051574230194 134
1 134
};
\addplot [draw=black, fill=black, mark=*, only marks]
table{%
x  y
0 101
1.31916858265826e-16 101
1.90854165124498e-16 101
2.76123264623805e-16 101
3.99487657690983e-16 101
5.77969225789828e-16 101
8.36192090199748e-16 101
1.20978010087187e-15 101
1.75028103599815e-15 101
2.53226492339775e-15 101
3.66361231445272e-15 101
5.3004272520564e-15 101
7.66853147280387e-15 101
1.10946247374182e-14 101
1.60514386748608e-14 101
2.32228380405916e-14 101
3.35981751046956e-14 101
4.86090388860522e-14 101
7.03262453697132e-14 101
1.01746336236917e-13 101
1.4720416183743e-13 101
2.1297101181119e-13 101
3.0812139406762e-13 101
4.45781703651421e-13 101
6.44946335920382e-13 101
9.33093012578379e-13 101
1.34997373522905e-12 101
1.95310976593144e-12 101
2.82571275406207e-12 101
4.0881647025981e-12 101
5.91465938726277e-12 101
8.5571879443469e-12 101
1.23803085838592e-11 101
1.79115351023329e-11 101
2.59139827651911e-11 101
3.74916486745036e-11 101
5.42420240667862e-11 101
7.84760728622658e-11 101
1.13537054757806e-10 101
1.64262714541508e-10 101
2.3765137080467e-10 101
3.43827577520273e-10 101
4.97440477698063e-10 101
7.19686532590913e-10 101
1.04122310773391e-09 101
1.50641343843461e-09 101
2.17944773339696e-09 101
3.15316617260919e-09 101
4.56191573405818e-09 101
6.60007115627081e-09 101
9.54882573012128e-09 101
1.38150131334669e-08 101
1.9987187016568e-08 101
2.89169861389382e-08 101
4.18363157450585e-08 101
6.05277818976901e-08 101
8.75701573477272e-08 101
1.26694132518423e-07 101
1.83298055844716e-07 101
2.65191260950814e-07 101
3.83671533654706e-07 101
5.55086842268793e-07 101
8.03086379619344e-07 101
1.16188368792791e-06 101
1.68098563335661e-06 101
2.43200452132442e-06 101
3.51856624547509e-06 101
5.09057781528099e-06 101
7.36491028874298e-06 101
1.06553770820028e-05 101
1.54159461089876e-05 101
2.23033730435418e-05 101
3.22679879900534e-05 101
4.66845558548812e-05 101
6.75419578328729e-05 101
9.77180898189545e-05 101
0.000141376207466237 101
0.000204539275728166 101
0.000295922538498417 101
0.000428132741944864 101
0.000619412225205451 101
0.00089615088654682 101
0.00129652686882764 101
0.00187578413169831 101
0.00271383975632489 101
0.00392631022259593 101
0.00568049214780331 101
0.00821840018033981 101
0.011890159919858 101
0.0172023996710777 101
0.0248879659920931 101
0.0360073186457157 101
0.0520945303142071 101
0.0753689557313919 101
0.109042011201382 101
0.157759383320808 101
0.228242009878159 101
0.330215096473694 101
0.477747321128845 101
0.691191911697388 101
1 101
};
\addplot [draw=black, fill=black, mark=*, only marks]
table{%
x  y
0 68
1.31916858265826e-16 68
2.29562975243979e-16 68
3.99487657690983e-16 68
6.95192232942718e-16 68
1.20978010087187e-15 68
2.10527095600565e-15 68
3.66361231445272e-15 68
6.3754535166243e-15 68
1.10946247374182e-14 68
1.93069708441171e-14 68
3.35981751046956e-14 68
5.84678654233159e-14 68
1.01746336236917e-13 68
1.77059945481381e-13 68
3.0812139406762e-13 68
5.36195762269875e-13 68
9.33093012578379e-13 68
1.6237772845068e-12 68
2.82571275406207e-12 68
4.91732132443246e-12 68
8.5571879443469e-12 68
1.48912670389079e-11 68
2.59139827651911e-11 68
4.5095659845229e-11 68
7.84760728622658e-11 68
1.36564495822888e-10 68
2.3765137080467e-10 68
4.13562267764789e-10 68
7.19686532590913e-10 68
1.25240273707306e-09 68
2.17944773339696e-09 68
3.79268749739481e-09 68
6.60007115627081e-09 68
1.14855067678832e-08 68
1.9987187016568e-08 68
3.47818946977441e-08 68
6.05277818976901e-08 68
1.05331025679334e-07 68
1.83298055844716e-07 68
3.18977015467681e-07 68
5.55086842268793e-07 68
9.65967387855926e-07 68
1.68098563335661e-06 68
2.92526715384156e-06 68
5.09057781528099e-06 68
8.8586721176398e-06 68
1.54159461089876e-05 68
2.68269741354743e-05 68
4.66845558548812e-05 68
8.1240912550129e-05 68
0.000141376207466237 68
0.000246023671934381 68
0.000428132741944864 68
0.00074504065560177 68
0.00129652686882764 68
0.00225622835569084 68
0.00392631022259593 68
0.0068326024338603 68
0.011890159919858 68
0.0206913687288761 68
0.0360073186457157 68
0.062660276889801 68
0.109042011201382 68
0.189755946397781 68
0.330215096473694 68
0.574643433094025 68
1 68
};
\addplot [draw=black, fill=black, mark=*, only marks]
table{%
x  y
0 35
1.31916858265826e-16 35
3.99487657690983e-16 35
1.20978010087187e-15 35
3.66361231445272e-15 35
1.10946247374182e-14 35
3.35981751046956e-14 35
1.01746336236917e-13 35
3.0812139406762e-13 35
9.33093012578379e-13 35
2.82571275406207e-12 35
8.5571879443469e-12 35
2.59139827651911e-11 35
7.84760728622658e-11 35
2.3765137080467e-10 35
7.19686532590913e-10 35
2.17944773339696e-09 35
6.60007115627081e-09 35
1.9987187016568e-08 35
6.05277818976901e-08 35
1.83298055844716e-07 35
5.55086842268793e-07 35
1.68098563335661e-06 35
5.09057781528099e-06 35
1.54159461089876e-05 35
4.66845558548812e-05 35
0.000141376207466237 35
0.000428132741944864 35
0.00129652686882764 35
0.00392631022259593 35
0.011890159919858 35
0.0360073186457157 35
0.109042011201382 35
0.330215096473694 35
1 35
};
\addplot [draw=black, fill=black, mark=*, only marks]
table{%
x  y
0 24
1.31916858265826e-16 24
6.95192232942718e-16 24
3.66361231445272e-15 24
1.93069708441171e-14 24
1.01746336236917e-13 24
5.36195762269875e-13 24
2.82571275406207e-12 24
1.48912670389079e-11 24
7.84760728622658e-11 24
4.13562267764789e-10 24
2.17944773339696e-09 24
1.14855067678832e-08 24
6.05277818976901e-08 24
3.18977015467681e-07 24
1.68098563335661e-06 24
8.8586721176398e-06 24
4.66845558548812e-05 24
0.000246023671934381 24
0.00129652686882764 24
0.0068326024338603 24
0.0360073186457157 24
0.189755946397781 24
1 24
};
\addplot [draw=black, fill=black, mark=*, only marks]
table{%
x  y
0 20
1.31916858265826e-16 20
1.00578756799785e-15 20
7.66853147280387e-15 20
5.84678654233159e-14 20
4.45781703651421e-13 20
3.39882028771443e-12 20
2.59139827651911e-11 20
1.97578273186672e-10 20
1.50641343843461e-09 20
1.14855067678832e-08 20
8.75701573477272e-08 20
6.6766887130143e-07 20
5.09057781528099e-06 20
3.88125408790074e-05 20
0.000295922538498417 20
0.00225622835569084 20
0.0172023996710777 20
0.131157770752907 20
1 20
};
\addplot [draw=black, fill=black, mark=*, only marks]
table{%
x  y
0 14
1.31916858265826e-16 14
2.77721329921654e-15 14
5.84678654233159e-14 14
1.23090697790734e-12 14
2.59139827651911e-11 14
5.45559541986762e-10 14
1.14855067678832e-08 14
2.41801018319165e-07 14
5.09057781528099e-06 14
0.000107170453702565 14
0.00225622835569084 14
0.0474998243153095 14
1 14
};
\addplot [draw=black, fill=black, mark=*, only marks]
table{%
x  y
0 11
1.31916858265826e-16 11
7.66853147280387e-15 11
4.45781703651421e-13 11
2.59139827651911e-11 11
1.50641343843461e-09 11
8.75701573477272e-08 11
5.09057781528099e-06 11
0.000295922538498417 11
0.0172023996710777 11
1 11
};
\addplot [draw=black, fill=black, mark=*, only marks]
table{%
x  y
0 8
1.31916858265826e-16 8
5.84678654233159e-14 8
2.59139827651911e-11 8
1.14855067678832e-08 8
5.09057781528099e-06 8
0.00225622835569084 8
1 8
};
\addplot [draw=black, fill=black, mark=*, only marks]
table{%
x  y
0 6
1.31916858265826e-16 6
1.23090697790734e-12 6
1.14855067678832e-08 6
0.000107170453702565 6
1 6
};
\addplot [draw=black, fill=black, mark=*, only marks]
table{%
x  y
0 5
1.31916858265826e-16 5
2.59139827651911e-11 5
5.09057781528099e-06 5
1 5
};
\addplot [draw=black, fill=black, mark=*, only marks]
table{%
x  y
0 4
1.09673091474103e-16 4
1.04724922067589e-08 4
1 4
};
\end{axis}

\end{tikzpicture}

%% file: figs/Charge_tr_errors.tex
\begin{tikzpicture}

\definecolor{darkgray176}{RGB}{176,176,176}
\definecolor{lightgray204}{RGB}{204,204,204}

\begin{axis}[
legend cell align={left},
legend style={
  fill opacity=0.8,
  draw opacity=1,
  text opacity=1,
  at={(0.03,0.97)},
  anchor=north west,
  draw=lightgray204
},
log basis x={10},
log basis y={10},
tick align=outside,
tick pos=left,
title={Relative Error on Test Dataset},
x grid style={darkgray176},
xlabel={Skipped Steps},
xmin=0.767462821883893, xmax=259.295948058454,
xmode=log,
xtick style={color=black},
xtick={1,2,3,4,5,6,7,8,9,10,20,30,40,50,60, 70,80,90,100,200,300},
xticklabels={
  \(\displaystyle {10^{0}}\),
  \(\displaystyle {}\),
  \(\displaystyle {}\),
  \(\displaystyle {}\),
  \(\displaystyle {}\),
  \(\displaystyle {}\),
  \(\displaystyle {}\),
  \(\displaystyle {}\),
  \(\displaystyle {}\),
  \(\displaystyle {10^{1}}\),
  \(\displaystyle {}\),
  \(\displaystyle {}\),
  \(\displaystyle {}\),
  \(\displaystyle {}\),
  \(\displaystyle {}\),
  \(\displaystyle {}\),
  \(\displaystyle {}\),
  \(\displaystyle {}\),
  \(\displaystyle {10^{2}}\),
  \(\displaystyle {}\),
  \(\displaystyle {}\),
},
y grid style={darkgray176},
ylabel={Relative Error},
ymin=0.00462842559952551, ymax=0.358618156847402,
ymode=log,
ytick style={color=black},
ytick={0.001,0.002,0.003,0.004,0.005,0.006,0.007,0.008,0.009,0.01,0.02,0.03,0.04,0.05,0.06,0.07,0.08,0.09,0.1,0.2,0.3,0.4,0.5},
yticklabels={
  \(\displaystyle {10^{-3}}\),
  \(\displaystyle {}\),
  \(\displaystyle {}\),
  \(\displaystyle {}\),
  \(\displaystyle {}\),
  \(\displaystyle {}\),
  \(\displaystyle {}\),
  \(\displaystyle {}\),
  \(\displaystyle {}\),
  \(\displaystyle {10^{-2}}\),
  \(\displaystyle {}\),
  \(\displaystyle {}\),
  \(\displaystyle {}\),
  \(\displaystyle {}\),
  \(\displaystyle {}\),
  \(\displaystyle {}\),
  \(\displaystyle {}\),
  \(\displaystyle {}\),
  \(\displaystyle {10^{-1}}\),
  \(\displaystyle {}\),
  \(\displaystyle {}\),
  \(\displaystyle {}\),
  \(\displaystyle {}\),
}
]
\addplot [draw=black, mark=o, only marks]
table{%
x  y
1 0.00598676130175591
2 0.00658358307555318
3 0.00739271147176623
4 0.00564034888520837
6 0.00606388878077269
12 0.00673612952232361
18 0.0149689428508282
22 0.0206915475428104
33 0.0452025160193443
44 0.0971512645483017
66 0.146276101469994
99 0.198783710598946
132 0.273149400949478
199 0.240097910165787
};
\addlegendentry{\lilan}
\addplot [draw=red, fill=red, mark=*, only marks]
table{%
x  y
1 0.0122664300724864
2 0.0131125878542662
3 0.0108915455639362
4 0.0126963630318642
6 0.010609608143568
12 0.0163874980062246
18 0.0217119920998812
22 0.0275648236274719
33 0.042033389210701
44 0.0839584767818451
66 0.0890510082244873
99 0.190380483865738
132 0.294279217720032
199 0.239834621548653
};
\addlegendentry{$\lilan_{\cancel{\boldsymbol{\tau}}}$}
\end{axis}

\end{tikzpicture}

%% file: figs/CR_skip_discretizations.tex
\begin{tikzpicture}

\definecolor{darkgray176}{RGB}{176,176,176}

\begin{axis}[
log basis x={10},
log basis y={10},
tick align=outside,
tick pos=left,
title={Various Training Time Discretizations},
x grid style={darkgray176},
xlabel={Times (s)},
xmin=0.75e-16, xmax=1.40,
xmode=log,
xtick style={color=black},
xtick={1e-18,1e-16,1e-14,1e-12,1e-10,1e-08,1e-06,0.0001,0.01,1,100},
xticklabels={
  \(\displaystyle {10^{-18}}\),
  \(\displaystyle {10^{-16}}\),
  \(\displaystyle {10^{-14}}\),
  \(\displaystyle {10^{-12}}\),
  \(\displaystyle {10^{-10}}\),
  \(\displaystyle {10^{-8}}\),
  \(\displaystyle {10^{-6}}\),
  \(\displaystyle {10^{-4}}\),
  \(\displaystyle {10^{-2}}\),
  \(\displaystyle {10^{0}}\),
  \(\displaystyle {10^{2}}\)
},
y grid style={darkgray176},
ylabel={Steps Retained},
ymin=3.17771062518461, ymax=502.248375717749,
ymode=log,
ytick style={color=black},
ytick={1,2,3,4,5,6,7,8,9,10,20,30,40,50,60,70,80,90,100,200,300,400},
yticklabels={
  \(\displaystyle {10^{0}}\),
  \(\displaystyle {}\),
  \(\displaystyle {}\),
  \(\displaystyle {}\),
  \(\displaystyle {}\),
  \(\displaystyle {}\),
  \(\displaystyle {}\),
  \(\displaystyle {}\),
  \(\displaystyle {}\),
  \(\displaystyle {10^{1}}\),
  \(\displaystyle {}\),
  \(\displaystyle {}\),
  \(\displaystyle {}\),
  \(\displaystyle {}\),
  \(\displaystyle {}\),
  \(\displaystyle {}\),
  \(\displaystyle {}\),
  \(\displaystyle {}\),
  \(\displaystyle {10^{2}}\),
  \(\displaystyle {}\),
  \(\displaystyle {}\),
  \(\displaystyle {}\),
}
]
\addplot [draw=black, fill=black, mark=*, only marks]
table{%
x  y
1.09673091474103e-16 399
1.20281881353442e-16 399
1.31916858265826e-16 399
1.44677621374448e-16 399
1.58672418842992e-16 399
1.74020947014816e-16 399
1.90854165124498e-16 399
2.0931567172638e-16 399
2.29562975243979e-16 399
2.51769373324349e-16 399
2.76123264623805e-16 399
3.02832913368252e-16 399
3.32126224360062e-16 399
3.64253125258233e-16 399
3.99487657690983e-16 399
4.38131444796872e-16 399
4.80512314796267e-16 399
5.26992718381255e-16 399
5.77969225789828e-16 399
6.33876735500841e-16 399
6.95192232942718e-16 399
7.62440507965842e-16 399
8.36192090199748e-16 399
9.17077701552784e-16 399
1.00578756799785e-15 399
1.10307830171507e-15 399
1.20978010087187e-15 399
1.32680331412833e-15 399
1.45514945206545e-15 399
1.59590748141604e-15 399
1.75028103599815e-15 399
1.91958735792387e-15 399
2.10527095600565e-15 399
2.30892075817339e-15 399
2.53226492339775e-15 399
2.77721329921654e-15 399
3.04585572766887e-15 399
3.34048427991002e-15 399
3.66361231445272e-15 399
4.01799734705676e-15 399
4.40667154391508e-15 399
4.83293325132428e-15 399
5.3004272520564e-15 399
5.81314264777653e-15 399
6.3754535166243e-15 399
6.99215702969647e-15 399
7.66853147280387e-15 399
8.41031634119709e-15 399
9.22385421758517e-15 399
1.01160856899379e-14 399
1.10946247374182e-14 399
1.21678180226882e-14 399
1.33448228189408e-14 399
1.46357128890176e-14 399
1.60514386748608e-14 399
1.76041093594842e-14 399
1.93069708441171e-14 399
2.11745531319366e-14 399
2.32228380405916e-14 399
2.54692049920938e-14 399
2.79328663877528e-14 399
3.063483711499e-14 399
3.35981751046956e-14 399
3.68481592081474e-14 399
4.0412514507775e-14 399
4.43216562626483e-14 399
4.86090388860522e-14 399
5.33110373608734e-14 399
5.84678654233159e-14 399
6.412351796466e-14 399
7.03262453697132e-14 399
7.71291396636092e-14 399
8.45899142832424e-14 399
9.27723745264633e-14 399
1.01746336236917e-13 399
1.11588350864243e-13 399
1.22382396642966e-13 399
1.34220569771373e-13 399
1.4720416183743e-13 399
1.61443368439443e-13 399
1.77059945481381e-13 399
1.94187114305189e-13 399
2.1297101181119e-13 399
2.33572411629038e-13 399
2.56166100701463e-13 399
2.80945297408131e-13 399
3.0812139406762e-13 399
3.37926247314518e-13 399
3.70614183873436e-13 399
4.06464060543135e-13 399
4.45781703651421e-13 399
4.88903676657687e-13 399
5.36195762269875e-13 399
5.8806247631854e-13 399
6.44946335920382e-13 399
7.07334147850847e-13 399
7.7575527382065e-13 399
8.50794803532173e-13 399
9.33093012578379e-13 399
1.02335197607606e-12 399
1.12234180282833e-12 399
1.23090697790734e-12 399
1.34997373522905e-12 399
1.48055797222907e-12 399
1.6237772845068e-12 399
1.78084684585716e-12 399
1.95310976593144e-12 399
2.14204066810442e-12 399
2.34924232844769e-12 399
2.57648676699196e-12 399
2.82571275406207e-12 399
3.0990466269587e-12 399
3.39882028771443e-12 399
3.72759158609459e-12 399
4.0881647025981e-12 399
4.48361660299046e-12 399
4.91732132443246e-12 399
5.39299030780094e-12 399
5.91465938726277e-12 399
6.48679027159815e-12 399
7.11427878413717e-12 399
7.80244965858934e-12 399
8.5571879443469e-12 399
9.38493293495268e-12 399
1.02927466896774e-11 399
1.12883738814795e-11 399
1.23803085838592e-11 399
1.35778679966059e-11 399
1.48912670389079e-11 399
1.63317501883364e-11 399
1.79115351023329e-11 399
1.96441352767751e-11 399
2.15443791279357e-11 399
2.36283857063535e-11 399
2.59139827651911e-11 399
2.84206668615949e-11 399
3.11698236665769e-11 399
3.41849118457027e-11 399
3.74916486745036e-11 399
4.11182546344868e-11 399
4.5095659845229e-11 399
4.94578059062611e-11 399
5.42420240667862e-11 399
5.94889068561422e-11 399
6.52433246317585e-11 399
7.15545331320122e-11 399
7.84760728622658e-11 399
8.60671325875195e-11 399
9.43924868823665e-11 399
1.03523162264807e-10 399
1.13537054757806e-10 399
1.24519602962003e-10 399
1.36564495822888e-10 399
1.49774512236966e-10 399
1.64262714541508e-10 399
1.8015199054755e-10 399
1.97578273186672e-10 399
2.16690679288867e-10 399
2.3765137080467e-10 399
2.60639620996983e-10 399
2.85851536885318e-10 399
3.13502224180695e-10 399
3.43827577520273e-10 399
3.77086362135515e-10 399
4.13562267764789e-10 399
4.53566545433048e-10 399
4.97440477698063e-10 399
5.45559541986762e-10 399
5.98331995060875e-10 399
6.56209253602213e-10 399
7.19686532590913e-10 399
7.89302512238521e-10 399
8.65652494130842e-10 399
9.49387901627574e-10 399
1.04122310773391e-09 399
1.14194154132718e-09 399
1.25240273707306e-09 399
1.37354871920792e-09 399
1.50641343843461e-09 399
1.65213387415264e-09 399
1.81194637072224e-09 399
1.98721772370902e-09 399
2.17944773339696e-09 399
2.39026798354303e-09 399
2.62148081020541e-09 399
2.87505907969887e-09 399
3.15316617260919e-09 399
3.4581750796292e-09 399
3.79268749739481e-09 399
4.15955803134693e-09 399
4.56191573405818e-09 399
5.00320540552934e-09 399
5.48716982962105e-09 399
6.01794925003674e-09 399
6.60007115627081e-09 399
7.2385177851686e-09 399
7.93870658100104e-09 399
8.70662564267377e-09 399
9.54882573012128e-09 399
1.04724922067589e-08 399
1.14855067678832e-08 399
1.25965105013393e-08 399
1.38150131334669e-08 399
1.5151352172893e-08 399
1.66169566995222e-08 399
1.82243304891472e-08 399
1.9987187016568e-08 399
2.19205666951439e-08 399
2.40409647744855e-08 399
2.63665267397073e-08 399
2.89169861389382e-08 399
3.17141548578093e-08 399
3.47818946977441e-08 399
3.81463785004144e-08 399
4.18363157450585e-08 399
4.58832829508538e-08 399
5.0321613542792e-08 399
5.51892718192448e-08 399
6.05277818976901e-08 399
6.63826895674902e-08 399
7.28039566411098e-08 399
7.98465222828781e-08 399
8.75701573477272e-08 399
9.60408996775186e-08 399
1.05331025679334e-07 399
1.15519796395347e-07 399
1.26694132518423e-07 399
1.38949687311651e-07 399
1.52390413177272e-07 399
1.67131290140787e-07 399
1.83298055844716e-07 399
2.01028640844925e-07 399
2.20474333900711e-07 399
2.41801018319165e-07 399
2.65191260950814e-07 399
2.90843445327482e-07 399
3.18977015467681e-07 399
3.49831964285841e-07 399
3.83671533654706e-07 399
4.2078445972038e-07 399
4.61488326664039e-07 399
5.06128515098681e-07 399
5.55086842268793e-07 399
6.08780908351036e-07 399
6.6766887130143e-07 399
7.32253113255865e-07 399
8.03086379619344e-07 399
8.80769675859483e-07 399
9.65967387855926e-07 399
1.05940637240565e-06 399
1.16188368792791e-06 399
1.27427381357847e-06 399
1.39753854000446e-06 399
1.53272378611291e-06 399
1.68098563335661e-06 399
1.84358896149206e-06 399
2.02192109099997e-06 399
2.21750337914273e-06 399
2.43200452132442e-06 399
2.66726055997424e-06 399
2.92526715384156e-06 399
3.20823096444656e-06 399
3.51856624547509e-06 399
3.85892053600401e-06 399
4.23219762524241e-06 399
4.64159211333026e-06 399
5.09057781528099e-06 399
5.58299416297814e-06 399
6.12304256719653e-06 399
6.71533007334801e-06 399
7.36491028874298e-06 399
8.07734340924071e-06 399
8.8586721176398e-06 399
9.71557983575622e-06 399
1.06553770820028e-05 399
1.16860819616704e-05 399
1.28164874695358e-05 399
1.40562688102364e-05 399
1.54159461089876e-05 399
1.69071445270674e-05 399
1.85425888048485e-05 399
2.03362305910559e-05 399
2.23033730435418e-05 399
2.44607999775326e-05 399
2.68269741354743e-05 399
2.94219735224033e-05 399
3.22679879900534e-05 399
3.53893010469619e-05 399
3.88125408790074e-05 399
4.2566916818032e-05 399
4.66845558548812e-05 399
5.12003971380182e-05 399
5.61530614504591e-05 399
6.15847966400906e-05 399
6.75419578328729e-05 399
7.40753530408256e-05 399
8.1240912550129e-05 399
8.9099419710692e-05 399
9.77180898189545e-05 399
0.000107170453702565 399
0.000117537150799762 399
0.000128906627651304 399
0.000141376207466237 399
0.0001550516608404 399
0.000170049956068397 399
0.000186499048140831 399
0.000204539275728166 399
0.000224324554437771 399
0.000246023671934381 399
0.00026982236886397 399
0.000295922538498417 399
0.000324547407217324 399
0.000355941185262054 399
0.000390371715184301 399
0.000428132741944864 399
0.000469547463580966 399
0.00051496725063771 399
0.000564780493732542 399
0.000619412225205451 399
0.000679328572005033 399
0.00074504065560177 399
0.000817110994830728 399
0.00089615088654682 399
0.000982836354523897 399
0.00107790704350919 399
0.00118217407725751 399
0.00129652686882764 399
0.00142194423824549 399
0.00155949033796787 399
0.00171034119557589 399
0.00187578413169831 399
0.00205723056569695 399
0.00225622835569084 399
0.0024744754191488 399
0.00271383975632489 399
0.00297635211609304 399
0.00326425745151937 399
0.00358001212589443 399
0.00392631022259593 399
0.00430610589683056 399
0.00472265016287565 399
0.0051794764585793 399
0.00568049214780331 399
0.00622997153550386 399
0.0068326024338603 399
0.00749352620914578 399
0.00821840018033981 399
0.00901337433606386 399
0.00988524593412876 399
0.0108414553105831 399
0.011890159919858 399
0.0130403060466051 399
0.0143017387017608 399
0.0156851597130299 399
0.0172023996710777 399
0.0188664030283689 399
0.0206913687288761 399
0.0226928647607565 399
0.0248879659920931 399
0.0272954627871513 399
0.0299357790499926 399
0.0328314937651157 399
0.0360073186457157 399
0.0394903384149075 399
0.0433102771639824 399
0.0474998243153095 399
0.0520945303142071 399
0.0571336820721626 399
0.062660276889801 399
0.0687214657664299 399
0.0753689557313919 399
0.0826596468687057 399
0.0906553938984871 399
0.0994245782494545 399
0.109042011201382 399
0.119589745998383 399
0.131157770752907 399
0.143845111131668 399
0.157759383320808 399
0.173019587993622 399
0.189755946397781 399
0.208111211657524 399
0.228242009878159 399
0.250320613384247 399
0.274534374475479 399
0.301090329885483 399
0.330215096473694 399
0.362157106399536 399
0.397188901901245 399
0.435609370470047 399
0.477747321128845 399
0.523960292339325 399
0.574643433094025 399
0.630229234695435 399
0.691191911697388 399
0.758051574230194 399
0.831380426883698 399
0.911800682544708 399
1 399
};
\addplot [draw=black, fill=black, mark=*, only marks]
table{%
x  y
0 201
1.09673091474103e-16 201
1.31916858265826e-16 201
1.58672418842992e-16 201
1.90854165124498e-16 201
2.29562975243979e-16 201
2.76123264623805e-16 201
3.32126224360062e-16 201
3.99487657690983e-16 201
4.80512314796267e-16 201
5.77969225789828e-16 201
6.95192232942718e-16 201
8.36192090199748e-16 201
1.00578756799785e-15 201
1.20978010087187e-15 201
1.45514945206545e-15 201
1.75028103599815e-15 201
2.10527095600565e-15 201
2.53226492339775e-15 201
3.04585572766887e-15 201
3.66361231445272e-15 201
4.40667154391508e-15 201
5.3004272520564e-15 201
6.3754535166243e-15 201
7.66853147280387e-15 201
9.22385421758517e-15 201
1.10946247374182e-14 201
1.33448228189408e-14 201
1.60514386748608e-14 201
1.93069708441171e-14 201
2.32228380405916e-14 201
2.79328663877528e-14 201
3.35981751046956e-14 201
4.0412514507775e-14 201
4.86090388860522e-14 201
5.84678654233159e-14 201
7.03262453697132e-14 201
8.45899142832424e-14 201
1.01746336236917e-13 201
1.22382396642966e-13 201
1.4720416183743e-13 201
1.77059945481381e-13 201
2.1297101181119e-13 201
2.56166100701463e-13 201
3.0812139406762e-13 201
3.70614183873436e-13 201
4.45781703651421e-13 201
5.36195762269875e-13 201
6.44946335920382e-13 201
7.7575527382065e-13 201
9.33093012578379e-13 201
1.12234180282833e-12 201
1.34997373522905e-12 201
1.6237772845068e-12 201
1.95310976593144e-12 201
2.34924232844769e-12 201
2.82571275406207e-12 201
3.39882028771443e-12 201
4.0881647025981e-12 201
4.91732132443246e-12 201
5.91465938726277e-12 201
7.11427878413717e-12 201
8.5571879443469e-12 201
1.02927466896774e-11 201
1.23803085838592e-11 201
1.48912670389079e-11 201
1.79115351023329e-11 201
2.15443791279357e-11 201
2.59139827651911e-11 201
3.11698236665769e-11 201
3.74916486745036e-11 201
4.5095659845229e-11 201
5.42420240667862e-11 201
6.52433246317585e-11 201
7.84760728622658e-11 201
9.43924868823665e-11 201
1.13537054757806e-10 201
1.36564495822888e-10 201
1.64262714541508e-10 201
1.97578273186672e-10 201
2.3765137080467e-10 201
2.85851536885318e-10 201
3.43827577520273e-10 201
4.13562267764789e-10 201
4.97440477698063e-10 201
5.98331995060875e-10 201
7.19686532590913e-10 201
8.65652494130842e-10 201
1.04122310773391e-09 201
1.25240273707306e-09 201
1.50641343843461e-09 201
1.81194637072224e-09 201
2.17944773339696e-09 201
2.62148081020541e-09 201
3.15316617260919e-09 201
3.79268749739481e-09 201
4.56191573405818e-09 201
5.48716982962105e-09 201
6.60007115627081e-09 201
7.93870658100104e-09 201
9.54882573012128e-09 201
1.14855067678832e-08 201
1.38150131334669e-08 201
1.66169566995222e-08 201
1.9987187016568e-08 201
2.40409647744855e-08 201
2.89169861389382e-08 201
3.47818946977441e-08 201
4.18363157450585e-08 201
5.0321613542792e-08 201
6.05277818976901e-08 201
7.28039566411098e-08 201
8.75701573477272e-08 201
1.05331025679334e-07 201
1.26694132518423e-07 201
1.52390413177272e-07 201
1.83298055844716e-07 201
2.20474333900711e-07 201
2.65191260950814e-07 201
3.18977015467681e-07 201
3.83671533654706e-07 201
4.61488326664039e-07 201
5.55086842268793e-07 201
6.6766887130143e-07 201
8.03086379619344e-07 201
9.65967387855926e-07 201
1.16188368792791e-06 201
1.39753854000446e-06 201
1.68098563335661e-06 201
2.02192109099997e-06 201
2.43200452132442e-06 201
2.92526715384156e-06 201
3.51856624547509e-06 201
4.23219762524241e-06 201
5.09057781528099e-06 201
6.12304256719653e-06 201
7.36491028874298e-06 201
8.8586721176398e-06 201
1.06553770820028e-05 201
1.28164874695358e-05 201
1.54159461089876e-05 201
1.85425888048485e-05 201
2.23033730435418e-05 201
2.68269741354743e-05 201
3.22679879900534e-05 201
3.88125408790074e-05 201
4.66845558548812e-05 201
5.61530614504591e-05 201
6.75419578328729e-05 201
8.1240912550129e-05 201
9.77180898189545e-05 201
0.000117537150799762 201
0.000141376207466237 201
0.000170049956068397 201
0.000204539275728166 201
0.000246023671934381 201
0.000295922538498417 201
0.000355941185262054 201
0.000428132741944864 201
0.00051496725063771 201
0.000619412225205451 201
0.00074504065560177 201
0.00089615088654682 201
0.00107790704350919 201
0.00129652686882764 201
0.00155949033796787 201
0.00187578413169831 201
0.00225622835569084 201
0.00271383975632489 201
0.00326425745151937 201
0.00392631022259593 201
0.00472265016287565 201
0.00568049214780331 201
0.0068326024338603 201
0.00821840018033981 201
0.00988524593412876 201
0.011890159919858 201
0.0143017387017608 201
0.0172023996710777 201
0.0206913687288761 201
0.0248879659920931 201
0.0299357790499926 201
0.0360073186457157 201
0.0433102771639824 201
0.0520945303142071 201
0.062660276889801 201
0.0753689557313919 201
0.0906553938984871 201
0.109042011201382 201
0.131157770752907 201
0.157759383320808 201
0.189755946397781 201
0.228242009878159 201
0.274534374475479 201
0.330215096473694 201
0.397188901901245 201
0.477747321128845 201
0.574643433094025 201
0.691191911697388 201
0.831380426883698 201
1 201
};
\addplot [draw=black, fill=black, mark=*, only marks]
table{%
x  y
1.09673091474103e-16 134
1.31916858265826e-16 134
1.74020947014816e-16 134
2.29562975243979e-16 134
3.02832913368252e-16 134
3.99487657690983e-16 134
5.26992718381255e-16 134
6.95192232942718e-16 134
9.17077701552784e-16 134
1.20978010087187e-15 134
1.59590748141604e-15 134
2.10527095600565e-15 134
2.77721329921654e-15 134
3.66361231445272e-15 134
4.83293325132428e-15 134
6.3754535166243e-15 134
8.41031634119709e-15 134
1.10946247374182e-14 134
1.46357128890176e-14 134
1.93069708441171e-14 134
2.54692049920938e-14 134
3.35981751046956e-14 134
4.43216562626483e-14 134
5.84678654233159e-14 134
7.71291396636092e-14 134
1.01746336236917e-13 134
1.34220569771373e-13 134
1.77059945481381e-13 134
2.33572411629038e-13 134
3.0812139406762e-13 134
4.06464060543135e-13 134
5.36195762269875e-13 134
7.07334147850847e-13 134
9.33093012578379e-13 134
1.23090697790734e-12 134
1.6237772845068e-12 134
2.14204066810442e-12 134
2.82571275406207e-12 134
3.72759158609459e-12 134
4.91732132443246e-12 134
6.48679027159815e-12 134
8.5571879443469e-12 134
1.12883738814795e-11 134
1.48912670389079e-11 134
1.96441352767751e-11 134
2.59139827651911e-11 134
3.41849118457027e-11 134
4.5095659845229e-11 134
5.94889068561422e-11 134
7.84760728622658e-11 134
1.03523162264807e-10 134
1.36564495822888e-10 134
1.8015199054755e-10 134
2.3765137080467e-10 134
3.13502224180695e-10 134
4.13562267764789e-10 134
5.45559541986762e-10 134
7.19686532590913e-10 134
9.49387901627574e-10 134
1.25240273707306e-09 134
1.65213387415264e-09 134
2.17944773339696e-09 134
2.87505907969887e-09 134
3.79268749739481e-09 134
5.00320540552934e-09 134
6.60007115627081e-09 134
8.70662564267377e-09 134
1.14855067678832e-08 134
1.5151352172893e-08 134
1.9987187016568e-08 134
2.63665267397073e-08 134
3.47818946977441e-08 134
4.58832829508538e-08 134
6.05277818976901e-08 134
7.98465222828781e-08 134
1.05331025679334e-07 134
1.38949687311651e-07 134
1.83298055844716e-07 134
2.41801018319165e-07 134
3.18977015467681e-07 134
4.2078445972038e-07 134
5.55086842268793e-07 134
7.32253113255865e-07 134
9.65967387855926e-07 134
1.27427381357847e-06 134
1.68098563335661e-06 134
2.21750337914273e-06 134
2.92526715384156e-06 134
3.85892053600401e-06 134
5.09057781528099e-06 134
6.71533007334801e-06 134
8.8586721176398e-06 134
1.16860819616704e-05 134
1.54159461089876e-05 134
2.03362305910559e-05 134
2.68269741354743e-05 134
3.53893010469619e-05 134
4.66845558548812e-05 134
6.15847966400906e-05 134
8.1240912550129e-05 134
0.000107170453702565 134
0.000141376207466237 134
0.000186499048140831 134
0.000246023671934381 134
0.000324547407217324 134
0.000428132741944864 134
0.000564780493732542 134
0.00074504065560177 134
0.000982836354523897 134
0.00129652686882764 134
0.00171034119557589 134
0.00225622835569084 134
0.00297635211609304 134
0.00392631022259593 134
0.0051794764585793 134
0.0068326024338603 134
0.00901337433606386 134
0.011890159919858 134
0.0156851597130299 134
0.0206913687288761 134
0.0272954627871513 134
0.0360073186457157 134
0.0474998243153095 134
0.062660276889801 134
0.0826596468687057 134
0.109042011201382 134
0.143845111131668 134
0.189755946397781 134
0.250320613384247 134
0.330215096473694 134
0.435609370470047 134
0.574643433094025 134
0.758051574230194 134
1 134
};
\addplot [draw=black, fill=black, mark=*, only marks]
table{%
x  y
0 68
1.31916858265826e-16 68
2.29562975243979e-16 68
3.99487657690983e-16 68
6.95192232942718e-16 68
1.20978010087187e-15 68
2.10527095600565e-15 68
3.66361231445272e-15 68
6.3754535166243e-15 68
1.10946247374182e-14 68
1.93069708441171e-14 68
3.35981751046956e-14 68
5.84678654233159e-14 68
1.01746336236917e-13 68
1.77059945481381e-13 68
3.0812139406762e-13 68
5.36195762269875e-13 68
9.33093012578379e-13 68
1.6237772845068e-12 68
2.82571275406207e-12 68
4.91732132443246e-12 68
8.5571879443469e-12 68
1.48912670389079e-11 68
2.59139827651911e-11 68
4.5095659845229e-11 68
7.84760728622658e-11 68
1.36564495822888e-10 68
2.3765137080467e-10 68
4.13562267764789e-10 68
7.19686532590913e-10 68
1.25240273707306e-09 68
2.17944773339696e-09 68
3.79268749739481e-09 68
6.60007115627081e-09 68
1.14855067678832e-08 68
1.9987187016568e-08 68
3.47818946977441e-08 68
6.05277818976901e-08 68
1.05331025679334e-07 68
1.83298055844716e-07 68
3.18977015467681e-07 68
5.55086842268793e-07 68
9.65967387855926e-07 68
1.68098563335661e-06 68
2.92526715384156e-06 68
5.09057781528099e-06 68
8.8586721176398e-06 68
1.54159461089876e-05 68
2.68269741354743e-05 68
4.66845558548812e-05 68
8.1240912550129e-05 68
0.000141376207466237 68
0.000246023671934381 68
0.000428132741944864 68
0.00074504065560177 68
0.00129652686882764 68
0.00225622835569084 68
0.00392631022259593 68
0.0068326024338603 68
0.011890159919858 68
0.0206913687288761 68
0.0360073186457157 68
0.062660276889801 68
0.109042011201382 68
0.189755946397781 68
0.330215096473694 68
0.574643433094025 68
1 68
};
\addplot [draw=black, fill=black, mark=*, only marks]
table{%
x  y
0 35
1.31916858265826e-16 35
3.99487657690983e-16 35
1.20978010087187e-15 35
3.66361231445272e-15 35
1.10946247374182e-14 35
3.35981751046956e-14 35
1.01746336236917e-13 35
3.0812139406762e-13 35
9.33093012578379e-13 35
2.82571275406207e-12 35
8.5571879443469e-12 35
2.59139827651911e-11 35
7.84760728622658e-11 35
2.3765137080467e-10 35
7.19686532590913e-10 35
2.17944773339696e-09 35
6.60007115627081e-09 35
1.9987187016568e-08 35
6.05277818976901e-08 35
1.83298055844716e-07 35
5.55086842268793e-07 35
1.68098563335661e-06 35
5.09057781528099e-06 35
1.54159461089876e-05 35
4.66845558548812e-05 35
0.000141376207466237 35
0.000428132741944864 35
0.00129652686882764 35
0.00392631022259593 35
0.011890159919858 35
0.0360073186457157 35
0.109042011201382 35
0.330215096473694 35
1 35
};
\addplot [draw=black, fill=black, mark=*, only marks]
table{%
x  y
0 24
1.31916858265826e-16 24
6.95192232942718e-16 24
3.66361231445272e-15 24
1.93069708441171e-14 24
1.01746336236917e-13 24
5.36195762269875e-13 24
2.82571275406207e-12 24
1.48912670389079e-11 24
7.84760728622658e-11 24
4.13562267764789e-10 24
2.17944773339696e-09 24
1.14855067678832e-08 24
6.05277818976901e-08 24
3.18977015467681e-07 24
1.68098563335661e-06 24
8.8586721176398e-06 24
4.66845558548812e-05 24
0.000246023671934381 24
0.00129652686882764 24
0.0068326024338603 24
0.0360073186457157 24
0.189755946397781 24
1 24
};
\addplot [draw=black, fill=black, mark=*, only marks]
table{%
x  y
0 20
1.31916858265826e-16 20
1.00578756799785e-15 20
7.66853147280387e-15 20
5.84678654233159e-14 20
4.45781703651421e-13 20
3.39882028771443e-12 20
2.59139827651911e-11 20
1.97578273186672e-10 20
1.50641343843461e-09 20
1.14855067678832e-08 20
8.75701573477272e-08 20
6.6766887130143e-07 20
5.09057781528099e-06 20
3.88125408790074e-05 20
0.000295922538498417 20
0.00225622835569084 20
0.0172023996710777 20
0.131157770752907 20
1 20
};
\addplot [draw=black, fill=black, mark=*, only marks]
table{%
x  y
0 14
1.31916858265826e-16 14
2.77721329921654e-15 14
5.84678654233159e-14 14
1.23090697790734e-12 14
2.59139827651911e-11 14
5.45559541986762e-10 14
1.14855067678832e-08 14
2.41801018319165e-07 14
5.09057781528099e-06 14
0.000107170453702565 14
0.00225622835569084 14
0.0474998243153095 14
1 14
};
\addplot [draw=black, fill=black, mark=*, only marks]
table{%
x  y
0 11
1.31916858265826e-16 11
7.66853147280387e-15 11
4.45781703651421e-13 11
2.59139827651911e-11 11
1.50641343843461e-09 11
8.75701573477272e-08 11
5.09057781528099e-06 11
0.000295922538498417 11
0.0172023996710777 11
1 11
};
\addplot [draw=black, fill=black, mark=*, only marks]
table{%
x  y
0 8
1.31916858265826e-16 8
5.84678654233159e-14 8
2.59139827651911e-11 8
1.14855067678832e-08 8
5.09057781528099e-06 8
0.00225622835569084 8
1 8
};
\addplot [draw=black, fill=black, mark=*, only marks]
table{%
x  y
0 6
1.31916858265826e-16 6
1.23090697790734e-12 6
1.14855067678832e-08 6
0.000107170453702565 6
1 6
};
\addplot [draw=black, fill=black, mark=*, only marks]
table{%
x  y
0 5
1.31916858265826e-16 5
2.59139827651911e-11 5
5.09057781528099e-06 5
1 5
};
\addplot [draw=black, fill=black, mark=*, only marks]
table{%
x  y
0 4
1.09673091474103e-16 4
1.04724922067589e-08 4
1 4
};
\end{axis}

\end{tikzpicture}

%% file: figs/CR_tr_errors.tex
\begin{tikzpicture}

\definecolor{darkgray176}{RGB}{176,176,176}
\definecolor{lightgray204}{RGB}{204,204,204}

\begin{axis}[
legend cell align={left},
legend style={
  fill opacity=0.8,
  draw opacity=1,
  text opacity=1,
  at={(0.03,0.97)},
  anchor=north west,
  draw=lightgray204
},
log basis x={10},
log basis y={10},
tick align=outside,
tick pos=left,
title={Relative Error on Test Dataset},
x grid style={darkgray176},
xlabel={Skipped Steps},
xmin=0.767462821883893, xmax=259.295948058454,
xmode=log,
xtick style={color=black},
xtick={1,2,3,4,5,6,7,8,9,10,20,30,40,50,60, 70,80,90,100,200,300},
xticklabels={
  \(\displaystyle {10^{0}}\),
  \(\displaystyle {}\),
  \(\displaystyle {}\),
  \(\displaystyle {}\),
  \(\displaystyle {}\),
  \(\displaystyle {}\),
  \(\displaystyle {}\),
  \(\displaystyle {}\),
  \(\displaystyle {}\),
  \(\displaystyle {10^{1}}\),
  \(\displaystyle {}\),
  \(\displaystyle {}\),
  \(\displaystyle {}\),
  \(\displaystyle {}\),
  \(\displaystyle {}\),
  \(\displaystyle {}\),
  \(\displaystyle {}\),
  \(\displaystyle {}\),
  \(\displaystyle {10^{2}}\),
  \(\displaystyle {}\),
  \(\displaystyle {}\),
},
y grid style={darkgray176},
ylabel={Relative Error},
ymin=0.00778720349364555, ymax=0.378434080539559,
ymode=log,
ytick style={color=black},
ytick={0.001,0.002,0.003,0.004,0.005,0.006,0.007,0.008,0.009,0.01,0.02,0.03,0.04,0.05,0.06,0.07,0.08,0.09,0.1,0.2,0.3,0.4,0.5},
yticklabels={
  \(\displaystyle {10^{-3}}\),
  \(\displaystyle {}\),
  \(\displaystyle {}\),
  \(\displaystyle {}\),
  \(\displaystyle {}\),
  \(\displaystyle {}\),
  \(\displaystyle {}\),
  \(\displaystyle {}\),
  \(\displaystyle {}\),
  \(\displaystyle {10^{-2}}\),
  \(\displaystyle {}\),
  \(\displaystyle {}\),
  \(\displaystyle {}\),
  \(\displaystyle {}\),
  \(\displaystyle {}\),
  \(\displaystyle {}\),
  \(\displaystyle {}\),
  \(\displaystyle {}\),
  \(\displaystyle {10^{-1}}\),
  \(\displaystyle {}\),
  \(\displaystyle {}\),
  \(\displaystyle {}\),
  \(\displaystyle {}\),
}
]
\addplot [draw=black, mark=o, only marks]
table{%
x  y
1 0.0105170635506511
2 0.0103083522990346
3 0.0118501670658588
6 0.00993235222995281
12 0.0184565428644419
18 0.0373448096215725
22 0.0543615706264973
33 0.0755683481693268
44 0.123591206967831
66 0.13884973526001
99 0.216244474053383
132 0.31511116027832
199 0.31056821346283
};
\addlegendentry{\lilan}
\addplot [draw=red, fill=red, mark=*, only marks]
table{%
x  y
1 0.0151419593021274
2 0.0104826996102929
3 0.00929063837975264
6 0.0110042514279485
12 0.0185055788606405
18 0.0270573291927576
22 0.0415490716695786
33 0.0688614696264267
44 0.103406637907028
66 0.150497421622276
99 0.211341351270676
132 0.317194908857346
199 0.236711084842682
};
\addlegendentry{$\lilan_{\cancel{\boldsymbol{\tau}}}$}
\end{axis}

\end{tikzpicture}

%% file: figs/AC_skip_discretizations.tex
\begin{tikzpicture}

\definecolor{darkgray176}{RGB}{176,176,176}

\begin{axis}[
log basis y={10},
tick align=outside,
tick pos=left,
title={Various Training Time Discretizations},
x grid style={darkgray176},
xlabel={Times (s)},
xmin=0, xmax=10.1,
xtick style={color=black},
y grid style={darkgray176},
ylabel={Steps Retained},
ymin=2.5175454367322, ymax=119.163688417637,
ymode=log,
ytick style={color=black},
ytick={1,2,3,4,5,6,7,8,9,10,20,30,40,50,60,70,80,90,100},
yticklabels={
  \(\displaystyle {10^{0}}\),
  \(\displaystyle {}\),
  \(\displaystyle {}\),
  \(\displaystyle {}\),
  \(\displaystyle {}\),
  \(\displaystyle {}\),
  \(\displaystyle {}\),
  \(\displaystyle {}\),
  \(\displaystyle {}\),
  \(\displaystyle {10^{1}}\),
  \(\displaystyle {}\),
  \(\displaystyle {}\),
  \(\displaystyle {}\),
  \(\displaystyle {}\),
  \(\displaystyle {}\),
  \(\displaystyle {}\),
  \(\displaystyle {}\),
  \(\displaystyle {}\),
  \(\displaystyle {10^{2}}\),
}
]
\addplot [draw=black, fill=black, mark=*, only marks]
table{%
x  y
0.1 100
0.2 100
0.3 100
0.4 100
0.5 100
0.6 100
0.7 100
0.8 100
0.9 100
1 100
1.1 100
1.2 100
1.3 100
1.4 100
1.5 100
1.6 100
1.7 100
1.8 100
1.9 100
2 100
2.1 100
2.2 100
2.3 100
2.4 100
2.5 100
2.6 100
2.7 100
2.8 100
2.9 100
3 100
3.1 100
3.2 100
3.3 100
3.4 100
3.5 100
3.6 100
3.7 100
3.8 100
3.9 100
4 100
4.1 100
4.2 100
4.3 100
4.4 100
4.5 100
4.6 100
4.7 100
4.8 100
4.9 100
5 100
5.1 100
5.2 100
5.3 100
5.4 100
5.5 100
5.6 100
5.7 100
5.8 100
5.9 100
6 100
6.1 100
6.2 100
6.3 100
6.4 100
6.5 100
6.6 100
6.7 100
6.8 100
6.9 100
7 100
7.1 100
7.2 100
7.3 100
7.4 100
7.5 100
7.6 100
7.7 100
7.8 100
7.9 100
8 100
8.1 100
8.2 100
8.3 100
8.4 100
8.5 100
8.6 100
8.7 100
8.8 100
8.9 100
9 100
9.1 100
9.2 100
9.3 100
9.4 100
9.5 100
9.6 100
9.7 100
9.8 100
9.9 100
10 100
};
\addplot [draw=black, fill=black, mark=*, only marks]
table{%
x  y
0.100000001490116 51
0.200000002980232 51
0.400000005960464 51
0.600000023841858 51
0.800000011920929 51
1 51
1.20000004768372 51
1.39999997615814 51
1.60000002384186 51
1.79999995231628 51
2 51
2.20000004768372 51
2.40000009536743 51
2.59999990463257 51
2.79999995231628 51
3 51
3.20000004768372 51
3.40000009536743 51
3.59999990463257 51
3.79999995231628 51
4 51
4.19999980926514 51
4.40000009536743 51
4.59999990463257 51
4.80000019073486 51
5 51
5.19999980926514 51
5.40000009536743 51
5.59999990463257 51
5.80000019073486 51
6 51
6.19999980926514 51
6.40000009536743 51
6.59999990463257 51
6.80000019073486 51
7 51
7.19999980926514 51
7.40000009536743 51
7.59999990463257 51
7.80000019073486 51
8 51
8.19999980926514 51
8.39999961853027 51
8.60000038146973 51
8.80000019073486 51
9 51
9.19999980926514 51
9.39999961853027 51
9.60000038146973 51
9.80000019073486 51
10 51
};
\addplot [draw=black, fill=black, mark=*, only marks]
table{%
x  y
0.100000001490116 26
0.400000005960464 26
0.800000011920929 26
1.20000004768372 26
1.60000002384186 26
2 26
2.40000009536743 26
2.79999995231628 26
3.20000004768372 26
3.59999990463257 26
4 26
4.40000009536743 26
4.80000019073486 26
5.19999980926514 26
5.59999990463257 26
6 26
6.40000009536743 26
6.80000019073486 26
7.19999980926514 26
7.59999990463257 26
8 26
8.39999961853027 26
8.80000019073486 26
9.19999980926514 26
9.60000038146973 26
10 26
};
\addplot [draw=black, fill=black, mark=*, only marks]
table{%
x  y
0.100000001490116 21
0.5 21
1 21
1.5 21
2 21
2.5 21
3 21
3.5 21
4 21
4.5 21
5 21
5.5 21
6 21
6.5 21
7 21
7.5 21
8 21
8.5 21
9 21
9.5 21
10 21
};
\addplot [draw=black, fill=black, mark=*, only marks]
table{%
x  y
0.100000001490116 11
1 11
2 11
3 11
4 11
5 11
6 11
7 11
8 11
9 11
10 11
};
\addplot [draw=black, fill=black, mark=*, only marks]
table{%
x  y
0.100000001490116 6
2 6
4 6
6 6
8 6
10 6
};
\addplot [draw=black, fill=black, mark=*, only marks]
table{%
x  y
0.100000001490116 5
2.5 5
5 5
7.5 5
10 5
};
\addplot [draw=black, fill=black, mark=*, only marks]
table{%
x  y
0.100000001490116 4
3.40000009536743 4
6.69999980926514 4
10 4
};
\addplot [draw=black, fill=black, mark=*, only marks]
table{%
x  y
0.100000001490116 3
5 3
10 3
};
\end{axis}

\end{tikzpicture}

%% file: figs/ac_skip.tex
\begin{tikzpicture}

\definecolor{darkgray176}{RGB}{176,176,176}
\definecolor{lightgray204}{RGB}{204,204,204}

\begin{axis}[
legend cell align={left},
legend style={
  fill opacity=0.8,
  draw opacity=1,
  text opacity=1,
  at={(0.03,0.97)},
  anchor=north west,
  draw=lightgray204
},
log basis x={10},
log basis y={10},
tick align=outside,
tick pos=left,
title={Relative Error on Test Data},
x grid style={darkgray176},
xlabel={Skipped Steps},
xmin=0.899999976158142, xmax=55,
xmode=log,
xtick style={color=black},
xtick={1,2,3,4,5,6,7,8,9,10,20,30,40,50},
xticklabels={
  \(\displaystyle {10^{0}}\),
  \(\displaystyle {}\),
  \(\displaystyle {}\),
  \(\displaystyle {}\),
  \(\displaystyle {}\),
  \(\displaystyle {}\),
  \(\displaystyle {}\),
  \(\displaystyle {}\),
  \(\displaystyle {}\),
  \(\displaystyle {10^{1}}\),
  \(\displaystyle {}\),
  \(\displaystyle {}\),
  \(\displaystyle {}\),
  \(\displaystyle {}\),
},
y grid style={darkgray176},
ylabel={Relative L2 Error},
ymin=0.0154382191943308, ymax=0.55134185842599295,
ymode=log,
ytick style={color=black},
ytick={0.01,0.02,0.03,0.04,0.05,0.06,0.07,0.08,0.09,0.1,0.2,0.3,0.4,0.5},
yticklabels={
  \(\displaystyle {10^{-2}}\),
  \(\displaystyle {}\),
  \(\displaystyle {}\),
  \(\displaystyle {}\),
  \(\displaystyle {}\),
  \(\displaystyle {}\),
  \(\displaystyle {}\),
  \(\displaystyle {}\),
  \(\displaystyle {}\),
  \(\displaystyle {10^{-1}}\),
  \(\displaystyle {}\),
  \(\displaystyle {}\),
  \(\displaystyle {}\),
  \(\displaystyle {}\),
}
]
\addplot [draw=black, mark=o, only marks]
table{%
x  y
1 0.0290755089372396
2 0.0285974740982056
4 0.0311386156827211
5 0.0305384490638971
10 0.0306355189532042
20 0.0398451648652554
25 0.0533566884696484
33 0.049382533878088
50 0.0850562676787376
};
\addlegendentry{\lilan}
\addplot [draw=red, fill=red, mark=*, only marks]
table{%
x  y
1 0.0355829782783985
2 0.0359165668487549
4 0.0364157110452652
5 0.0363361723721027
10 0.0422995947301388
20 0.0830635279417038
25 0.0679647251963615
33 0.134664416313171
50 0.297267258167267
};
\addlegendentry{$\lilan_{\cancel{\boldsymbol{\tau}}}$}
\end{axis}

\end{tikzpicture}

%% file: figs/CH_skip_discretizations.tex
\begin{tikzpicture}

\definecolor{darkgray176}{RGB}{176,176,176}

\begin{axis}[
log basis y={10},
tick align=outside,
tick pos=left,
title={Various Training Time Discretizations},
x grid style={darkgray176},
xlabel={Times (s)},
xmin=-0.1, xmax=10.1,
xtick style={color=black},
y grid style={darkgray176},
ylabel={Steps Retained},
ymin=2.34895295016408, ymax=510.865915775867,
ymode=log,
ytick style={color=black},
ytick={1,2,3,4,5,6,7,8,9,10,20,30,40,50,60,70,80,90,100,200,300,400},
yticklabels={
  \(\displaystyle {10^{0}}\),
  \(\displaystyle {}\),
  \(\displaystyle {}\),
  \(\displaystyle {}\),
  \(\displaystyle {}\),
  \(\displaystyle {}\),
  \(\displaystyle {}\),
  \(\displaystyle {}\),
  \(\displaystyle {}\),
  \(\displaystyle {10^{1}}\),
  \(\displaystyle {}\),
  \(\displaystyle {}\),
  \(\displaystyle {}\),
  \(\displaystyle {}\),
  \(\displaystyle {}\),
  \(\displaystyle {}\),
  \(\displaystyle {}\),
  \(\displaystyle {}\),
  \(\displaystyle {10^{2}}\),
  \(\displaystyle {}\),
  \(\displaystyle {}\),
  \(\displaystyle {}\),
}
]
\addplot [draw=black, fill=black, mark=*, only marks]
table{%
x  y
0.025 400
0.05 400
0.075 400
0.1 400
0.125 400
0.15 400
0.175 400
0.2 400
0.225 400
0.25 400
0.275 400
0.3 400
0.325 400
0.35 400
0.375 400
0.4 400
0.425 400
0.45 400
0.475 400
0.5 400
0.525 400
0.55 400
0.575 400
0.6 400
0.625 400
0.65 400
0.675 400
0.7 400
0.725 400
0.75 400
0.775 400
0.8 400
0.825 400
0.85 400
0.875 400
0.9 400
0.925 400
0.95 400
0.975 400
1 400
1.025 400
1.05 400
1.075 400
1.1 400
1.125 400
1.15 400
1.175 400
1.2 400
1.225 400
1.25 400
1.275 400
1.3 400
1.325 400
1.35 400
1.375 400
1.4 400
1.425 400
1.45 400
1.475 400
1.5 400
1.525 400
1.55 400
1.575 400
1.6 400
1.625 400
1.65 400
1.675 400
1.7 400
1.725 400
1.75 400
1.775 400
1.8 400
1.825 400
1.85 400
1.875 400
1.9 400
1.925 400
1.95 400
1.975 400
2 400
2.025 400
2.05 400
2.075 400
2.1 400
2.125 400
2.15 400
2.175 400
2.2 400
2.225 400
2.25 400
2.275 400
2.3 400
2.325 400
2.35 400
2.375 400
2.4 400
2.425 400
2.45 400
2.475 400
2.5 400
2.525 400
2.55 400
2.575 400
2.6 400
2.625 400
2.65 400
2.675 400
2.7 400
2.725 400
2.75 400
2.775 400
2.8 400
2.825 400
2.85 400
2.875 400
2.9 400
2.925 400
2.95 400
2.975 400
3 400
3.025 400
3.05 400
3.075 400
3.1 400
3.125 400
3.15 400
3.175 400
3.2 400
3.225 400
3.25 400
3.275 400
3.3 400
3.325 400
3.35 400
3.375 400
3.4 400
3.425 400
3.45 400
3.475 400
3.5 400
3.525 400
3.55 400
3.575 400
3.6 400
3.625 400
3.65 400
3.675 400
3.7 400
3.725 400
3.75 400
3.775 400
3.8 400
3.825 400
3.85 400
3.875 400
3.9 400
3.925 400
3.95 400
3.975 400
4 400
4.025 400
4.05 400
4.075 400
4.1 400
4.125 400
4.15 400
4.175 400
4.2 400
4.225 400
4.25 400
4.275 400
4.3 400
4.325 400
4.35 400
4.375 400
4.4 400
4.425 400
4.45 400
4.475 400
4.5 400
4.525 400
4.55 400
4.575 400
4.6 400
4.625 400
4.65 400
4.675 400
4.7 400
4.725 400
4.75 400
4.775 400
4.8 400
4.825 400
4.85 400
4.875 400
4.9 400
4.925 400
4.95 400
4.975 400
5 400
5.025 400
5.05 400
5.075 400
5.1 400
5.125 400
5.15 400
5.175 400
5.2 400
5.225 400
5.25 400
5.275 400
5.3 400
5.325 400
5.35 400
5.375 400
5.4 400
5.425 400
5.45 400
5.475 400
5.5 400
5.525 400
5.55 400
5.575 400
5.6 400
5.625 400
5.65 400
5.675 400
5.7 400
5.725 400
5.75 400
5.775 400
5.8 400
5.825 400
5.85 400
5.875 400
5.9 400
5.925 400
5.95 400
5.975 400
6 400
6.025 400
6.05 400
6.075 400
6.1 400
6.125 400
6.15 400
6.175 400
6.2 400
6.225 400
6.25 400
6.275 400
6.3 400
6.325 400
6.35 400
6.375 400
6.4 400
6.425 400
6.45 400
6.475 400
6.5 400
6.525 400
6.55 400
6.575 400
6.6 400
6.625 400
6.65 400
6.675 400
6.7 400
6.725 400
6.75 400
6.775 400
6.8 400
6.825 400
6.85 400
6.875 400
6.9 400
6.925 400
6.95 400
6.975 400
7 400
7.025 400
7.05 400
7.075 400
7.1 400
7.125 400
7.15 400
7.175 400
7.2 400
7.225 400
7.25 400
7.275 400
7.3 400
7.325 400
7.35 400
7.375 400
7.4 400
7.425 400
7.45 400
7.475 400
7.5 400
7.525 400
7.55 400
7.575 400
7.6 400
7.625 400
7.65 400
7.675 400
7.7 400
7.725 400
7.75 400
7.775 400
7.8 400
7.825 400
7.85 400
7.875 400
7.9 400
7.925 400
7.95 400
7.975 400
8 400
8.025 400
8.05 400
8.075 400
8.1 400
8.125 400
8.15 400
8.175 400
8.2 400
8.225 400
8.25 400
8.275 400
8.3 400
8.325 400
8.35 400
8.375 400
8.4 400
8.425 400
8.45 400
8.475 400
8.5 400
8.525 400
8.55 400
8.575 400
8.6 400
8.625 400
8.65 400
8.675 400
8.7 400
8.725 400
8.75 400
8.775 400
8.8 400
8.825 400
8.85 400
8.875 400
8.9 400
8.925 400
8.95 400
8.975 400
9 400
9.025 400
9.05 400
9.075 400
9.1 400
9.125 400
9.15 400
9.175 400
9.2 400
9.225 400
9.25 400
9.275 400
9.3 400
9.325 400
9.35 400
9.375 400
9.4 400
9.425 400
9.45 400
9.475 400
9.5 400
9.525 400
9.55 400
9.575 400
9.6 400
9.625 400
9.65 400
9.675 400
9.7 400
9.725 400
9.75 400
9.775 400
9.8 400
9.825 400
9.85 400
9.875 400
9.9 400
9.925 400
9.95 400
9.975 400
10 400
};
\addplot [draw=black, fill=black, mark=*, only marks]
table{%
x  y
0.025000000372529 201
0.0500000007450581 201
0.100000001490116 201
0.150000005960464 201
0.200000002980232 201
0.25 201
0.300000011920929 201
0.349999994039536 201
0.400000005960464 201
0.449999988079071 201
0.5 201
0.550000011920929 201
0.600000023841858 201
0.649999976158142 201
0.699999988079071 201
0.75 201
0.800000011920929 201
0.850000023841858 201
0.899999976158142 201
0.949999988079071 201
1 201
1.04999995231628 201
1.10000002384186 201
1.14999997615814 201
1.20000004768372 201
1.25 201
1.29999995231628 201
1.35000002384186 201
1.39999997615814 201
1.45000004768372 201
1.5 201
1.54999995231628 201
1.60000002384186 201
1.64999997615814 201
1.70000004768372 201
1.75 201
1.79999995231628 201
1.85000002384186 201
1.89999997615814 201
1.95000004768372 201
2 201
2.04999995231628 201
2.09999990463257 201
2.15000009536743 201
2.20000004768372 201
2.25 201
2.29999995231628 201
2.34999990463257 201
2.40000009536743 201
2.45000004768372 201
2.5 201
2.54999995231628 201
2.59999990463257 201
2.65000009536743 201
2.70000004768372 201
2.75 201
2.79999995231628 201
2.84999990463257 201
2.90000009536743 201
2.95000004768372 201
3 201
3.04999995231628 201
3.09999990463257 201
3.15000009536743 201
3.20000004768372 201
3.25 201
3.29999995231628 201
3.34999990463257 201
3.40000009536743 201
3.45000004768372 201
3.5 201
3.54999995231628 201
3.59999990463257 201
3.65000009536743 201
3.70000004768372 201
3.75 201
3.79999995231628 201
3.84999990463257 201
3.90000009536743 201
3.95000004768372 201
4 201
4.05000019073486 201
4.09999990463257 201
4.15000009536743 201
4.19999980926514 201
4.25 201
4.30000019073486 201
4.34999990463257 201
4.40000009536743 201
4.44999980926514 201
4.5 201
4.55000019073486 201
4.59999990463257 201
4.65000009536743 201
4.69999980926514 201
4.75 201
4.80000019073486 201
4.84999990463257 201
4.90000009536743 201
4.94999980926514 201
5 201
5.05000019073486 201
5.09999990463257 201
5.15000009536743 201
5.19999980926514 201
5.25 201
5.30000019073486 201
5.34999990463257 201
5.40000009536743 201
5.44999980926514 201
5.5 201
5.55000019073486 201
5.59999990463257 201
5.65000009536743 201
5.69999980926514 201
5.75 201
5.80000019073486 201
5.84999990463257 201
5.90000009536743 201
5.94999980926514 201
6 201
6.05000019073486 201
6.09999990463257 201
6.15000009536743 201
6.19999980926514 201
6.25 201
6.30000019073486 201
6.34999990463257 201
6.40000009536743 201
6.44999980926514 201
6.5 201
6.55000019073486 201
6.59999990463257 201
6.65000009536743 201
6.69999980926514 201
6.75 201
6.80000019073486 201
6.84999990463257 201
6.90000009536743 201
6.94999980926514 201
7 201
7.05000019073486 201
7.09999990463257 201
7.15000009536743 201
7.19999980926514 201
7.25 201
7.30000019073486 201
7.34999990463257 201
7.40000009536743 201
7.44999980926514 201
7.5 201
7.55000019073486 201
7.59999990463257 201
7.65000009536743 201
7.69999980926514 201
7.75 201
7.80000019073486 201
7.84999990463257 201
7.90000009536743 201
7.94999980926514 201
8 201
8.05000019073486 201
8.10000038146973 201
8.14999961853027 201
8.19999980926514 201
8.25 201
8.30000019073486 201
8.35000038146973 201
8.39999961853027 201
8.44999980926514 201
8.5 201
8.55000019073486 201
8.60000038146973 201
8.64999961853027 201
8.69999980926514 201
8.75 201
8.80000019073486 201
8.85000038146973 201
8.89999961853027 201
8.94999980926514 201
9 201
9.05000019073486 201
9.10000038146973 201
9.14999961853027 201
9.19999980926514 201
9.25 201
9.30000019073486 201
9.35000038146973 201
9.39999961853027 201
9.44999980926514 201
9.5 201
9.55000019073486 201
9.60000038146973 201
9.64999961853027 201
9.69999980926514 201
9.75 201
9.80000019073486 201
9.85000038146973 201
9.89999961853027 201
9.94999980926514 201
10 201
};
\addplot [draw=black, fill=black, mark=*, only marks]
table{%
x  y
0.025000000372529 101
0.100000001490116 101
0.200000002980232 101
0.300000011920929 101
0.400000005960464 101
0.5 101
0.600000023841858 101
0.699999988079071 101
0.800000011920929 101
0.899999976158142 101
1 101
1.10000002384186 101
1.20000004768372 101
1.29999995231628 101
1.39999997615814 101
1.5 101
1.60000002384186 101
1.70000004768372 101
1.79999995231628 101
1.89999997615814 101
2 101
2.09999990463257 101
2.20000004768372 101
2.29999995231628 101
2.40000009536743 101
2.5 101
2.59999990463257 101
2.70000004768372 101
2.79999995231628 101
2.90000009536743 101
3 101
3.09999990463257 101
3.20000004768372 101
3.29999995231628 101
3.40000009536743 101
3.5 101
3.59999990463257 101
3.70000004768372 101
3.79999995231628 101
3.90000009536743 101
4 101
4.09999990463257 101
4.19999980926514 101
4.30000019073486 101
4.40000009536743 101
4.5 101
4.59999990463257 101
4.69999980926514 101
4.80000019073486 101
4.90000009536743 101
5 101
5.09999990463257 101
5.19999980926514 101
5.30000019073486 101
5.40000009536743 101
5.5 101
5.59999990463257 101
5.69999980926514 101
5.80000019073486 101
5.90000009536743 101
6 101
6.09999990463257 101
6.19999980926514 101
6.30000019073486 101
6.40000009536743 101
6.5 101
6.59999990463257 101
6.69999980926514 101
6.80000019073486 101
6.90000009536743 101
7 101
7.09999990463257 101
7.19999980926514 101
7.30000019073486 101
7.40000009536743 101
7.5 101
7.59999990463257 101
7.69999980926514 101
7.80000019073486 101
7.90000009536743 101
8 101
8.10000038146973 101
8.19999980926514 101
8.30000019073486 101
8.39999961853027 101
8.5 101
8.60000038146973 101
8.69999980926514 101
8.80000019073486 101
8.89999961853027 101
9 101
9.10000038146973 101
9.19999980926514 101
9.30000019073486 101
9.39999961853027 101
9.5 101
9.60000038146973 101
9.69999980926514 101
9.80000019073486 101
9.89999961853027 101
10 101
};
\addplot [draw=black, fill=black, mark=*, only marks]
table{%
x  y
0.025000000372529 81
0.125 81
0.25 81
0.375 81
0.5 81
0.625 81
0.75 81
0.875 81
1 81
1.125 81
1.25 81
1.375 81
1.5 81
1.625 81
1.75 81
1.875 81
2 81
2.125 81
2.25 81
2.375 81
2.5 81
2.625 81
2.75 81
2.875 81
3 81
3.125 81
3.25 81
3.375 81
3.5 81
3.625 81
3.75 81
3.875 81
4 81
4.125 81
4.25 81
4.375 81
4.5 81
4.625 81
4.75 81
4.875 81
5 81
5.125 81
5.25 81
5.375 81
5.5 81
5.625 81
5.75 81
5.875 81
6 81
6.125 81
6.25 81
6.375 81
6.5 81
6.625 81
6.75 81
6.875 81
7 81
7.125 81
7.25 81
7.375 81
7.5 81
7.625 81
7.75 81
7.875 81
8 81
8.125 81
8.25 81
8.375 81
8.5 81
8.625 81
8.75 81
8.875 81
9 81
9.125 81
9.25 81
9.375 81
9.5 81
9.625 81
9.75 81
9.875 81
10 81
};
\addplot [draw=black, fill=black, mark=*, only marks]
table{%
x  y
0.025000000372529 41
0.25 41
0.5 41
0.75 41
1 41
1.25 41
1.5 41
1.75 41
2 41
2.25 41
2.5 41
2.75 41
3 41
3.25 41
3.5 41
3.75 41
4 41
4.25 41
4.5 41
4.75 41
5 41
5.25 41
5.5 41
5.75 41
6 41
6.25 41
6.5 41
6.75 41
7 41
7.25 41
7.5 41
7.75 41
8 41
8.25 41
8.5 41
8.75 41
9 41
9.25 41
9.5 41
9.75 41
10 41
};
\addplot [draw=black, fill=black, mark=*, only marks]
table{%
x  y
0.025000000372529 21
0.5 21
1 21
1.5 21
2 21
2.5 21
3 21
3.5 21
4 21
4.5 21
5 21
5.5 21
6 21
6.5 21
7 21
7.5 21
8 21
8.5 21
9 21
9.5 21
10 21
};
\addplot [draw=black, fill=black, mark=*, only marks]
table{%
x  y
0.025000000372529 17
0.625 17
1.25 17
1.875 17
2.5 17
3.125 17
3.75 17
4.375 17
5 17
5.625 17
6.25 17
6.875 17
7.5 17
8.125 17
8.75 17
9.375 17
10 17
};
\addplot [draw=black, fill=black, mark=*, only marks]
table{%
x  y
0.025000000372529 11
1 11
2 11
3 11
4 11
5 11
6 11
7 11
8 11
9 11
10 11
};
\addplot [draw=black, fill=black, mark=*, only marks]
table{%
x  y
0.025000000372529 9
1.25 9
2.5 9
3.75 9
5 9
6.25 9
7.5 9
8.75 9
10 9
};
\addplot [draw=black, fill=black, mark=*, only marks]
table{%
x  y
0.025000000372529 6
2 6
4 6
6 6
8 6
10 6
};
\addplot [draw=black, fill=black, mark=*, only marks]
table{%
x  y
0.025000000372529 5
2.5 5
5 5
7.5 5
10 5
};
\addplot [draw=black, fill=black, mark=*, only marks]
table{%
x  y
0.025000000372529 4
3.34999990463257 4
6.67500019073486 4
10 4
};
\addplot [draw=black, fill=black, mark=*, only marks]
table{%
x  y
0.025000000372529 3
5 3
10 3
};
\end{axis}

\end{tikzpicture}

%% file: figs/CH_skip.tex
\begin{tikzpicture}

\definecolor{darkgray176}{RGB}{176,176,176}
\definecolor{lightgray204}{RGB}{204,204,204}

\begin{axis}[
legend cell align={left},
legend style={
  fill opacity=0.8,
  draw opacity=1,
  text opacity=1,
  at={(0.03,0.97)},
  anchor=north west,
  draw=lightgray204
},
log basis x={10},
log basis y={10},
minor ytick={0.002,0.003,0.004,0.005,0.006,0.007,0.008,0.009,0.02,0.03,0.04,0.05,0.06,0.07,0.08,0.09,0.2,0.3,0.4,0.5,0.6,0.7,0.8,0.9,2,3,4,5,6,7,8,9,20,30,40,50,60,70,80,90},
tick align=outside,
tick pos=left,
title={Relative Error on Test Data},
x grid style={darkgray176},
xlabel={Skipped Steps},
xmin=0.899999976158142, xmax=240,
xmode=log,
xtick style={color=black},
xtick={1,2,3,4,5,6,7,8,9,10,20,30,40,50,60,70,80,90,100,200,300},
xticklabels={
  \(\displaystyle {10^{0}}\),
  \(\displaystyle {}\),
  \(\displaystyle {}\),
  \(\displaystyle {}\),
  \(\displaystyle {}\),
  \(\displaystyle {}\),
  \(\displaystyle {}\),
  \(\displaystyle {}\),
  \(\displaystyle {}\),
  \(\displaystyle {10^{1}}\),
  \(\displaystyle {}\),
  \(\displaystyle {}\),
  \(\displaystyle {}\),
  \(\displaystyle {}\),
  \(\displaystyle {}\),
  \(\displaystyle {}\),
  \(\displaystyle {}\),
  \(\displaystyle {}\),
  \(\displaystyle {10^{2}}\),
  \(\displaystyle {}\),
  \(\displaystyle {}\),
},
y grid style={darkgray176},
ylabel={Relative L2 Error},
ymin=0.0410280248307637, ymax=0.354511174998557,
ymode=log,
ytick style={color=black},
ytick={0.01,0.02,0.03,0.04,0.05,0.06,0.07,0.08,0.09,0.1,0.2,0.3,0.4,0.5},
yticklabels={
  \(\displaystyle {10^{-2}}\),
  \(\displaystyle {}\),
  \(\displaystyle {}\),
  \(\displaystyle {}\),
  \(\displaystyle {}\),
  \(\displaystyle {}\),
  \(\displaystyle {}\),
  \(\displaystyle {}\),
  \(\displaystyle {}\),
  \(\displaystyle {10^{-1}}\),
  \(\displaystyle {}\),
  \(\displaystyle {}\),
  \(\displaystyle {}\),
  \(\displaystyle {}\),
}
]
\addplot [draw=black, mark=o, only marks]
table{%
x  y
1 0.0452533811330795
2 0.0483093485236168
4 0.0462674051523209
5 0.0480397641658783
10 0.0454860627651215
20 0.0485944449901581
25 0.0470181740820408
40 0.0513629727065563
50 0.0556519106030464
80 0.0730401501059532
100 0.085146002471447
133 0.120074115693569
200 0.162173762917519
};
\addlegendentry{\lilan}
\addplot [draw=red, fill=red, mark=*, only marks]
table{%
x  y
1 0.0528070628643036
2 0.0522412471473217
4 0.051187090575695
5 0.052790954709053
10 0.0521610490977764
20 0.0551119297742844
25 0.057461604475975
40 0.0675373375415802
50 0.0788189768791199
80 0.12363813072443
100 0.174845963716507
133 0.280819594860077
200 0.321410089731216
};
\addlegendentry{$\lilan_{\cancel{\boldsymbol{\tau}}}$}
\end{axis}

\end{tikzpicture}

%% file: Appendix.tex
\subsection{Data Transformations}
\label{sec:tranforms}

\vspace{0.5cm}

{\bf{ODE Problems:}} \cref{sec:Robertson,,sec:CR-model}
\newline To address the multiscale nature of the data in the three ODE problems, as a preprocessing step, all time series data is log scaled using the base $10$ logarithm. This procedure reduces the scale separation of different components of the solution. As an additional training aid, all network inputs, apart from time, are scaled independently to the range $[-1,1]$. Finally, the time discretization points are log scaled, using the base 10 logarithm, and then rescaled to fit the range $[0,1]$ seconds. Additionally, since each of the ODE problems is conservative (Robertson ODE has conservation of total concentration and CR ODEs have conservation of total number of electrons), we implement conservation constraints implicitly, by applying a softmax to the output of the decoder and multiplying by the appropriate constant (known from $\xb_0$) to obtain a solution satisfying the conservation constraint.

\vspace{0.5cm}

{\bf{PDE Problems:}} \cref{sec:AC,,sec:CH}
\newline For both PDE models, the data all lies on the range $[-1,1]$, so no rescaling is required. However, the time discretization points are rescaled to fit the range $[0,1]$ seconds.

\subsection{Training for Multiscale Stiff Systems}
\label{sec:loss}
In all of the ODE (\cref{sec:Robertson,,sec:CR-model}) test problems presented in this paper, the unique degrees of freedom in the solution vary by several orders of magnitude. This led to challenges in training with traditional loss functions such as mean squared error (MSE). In training, such loss functions were heavily biased toward learning the large magnitude components of the solution. To remedy this issue, we considered an absolute relative error loss function. The use of relative error balances out the magnitudes of errors for each individual degree of freedom.
\begin{equation}\label{eq:J_old_rel}
    \mathcal{L}(\mathbf{X}; \boldsymbol{\alpha}, \boldsymbol{\beta}, \boldsymbol{\nu}, \boldsymbol{\theta}) = \frac{1}{(N+1) \cdot (M+1)}\sum_{i=0}^{N}\sum_{j=0}^{M}\left|\frac{\mathbf{x}_{i}(t_{j}) - \mathbf{\hat{x}}_{i}(t_{j})}{\mathbf{x}_{i}(t_{j})}\right|.
\end{equation}
where,  the predicted state $\hat{\xb}_{i}(t_j)$  for time $t_j$ and for the $i^{th}$ training input $\{\LRp{\xb_0}_i, \mathbf{p}_i\}$ is given by \eqref{for_prop}. ${\xb}_{i}(t_j)$ denotes the actual state at time $t_j$ for the $i^{th}$ input.
However, this loss function still has the problem that overshooting errors are punished much more harshly than undershooting errors. Finally, we settled on the following loss function which resolves the unbalanced penalties between overshooting and undershooting.
\begin{equation}\label{eq:J}
    \mathcal{L}(\mathbf{X}; \boldsymbol{\alpha}, \boldsymbol{\beta}, \boldsymbol{\nu}, \boldsymbol{\theta}) = \frac{1}{N+1}\sum_{i=0}^{N}\prod_{j=0}^{M}10^{\frac{1}{M+1}\left|\mathbf{x}_{i}(t_{j}) - \mathbf{\hat{x}}_{i}(t_{j})\right|},
\end{equation} 
\begin{equation}
\mathbf{\hat{x}}_{i}(t_{j}) = \bs{d}\left( \bs{e}\LRp{\LRp{\mathbf{x}_0}_{i},\mathbf{p}_{i}; \boldsymbol{\alpha}} + \boldsymbol{\tau}(t_j, \LRp{\mathbf{x}_0}_{i},\mathbf{p}_{i}; \boldsymbol{\nu}) \ \circ \ \bs{c}(\mathbf{x}_{0_i},\mathbf{p}_{i}; \boldsymbol{\beta});\boldsymbol{\theta}\right).
\label{for_prop}
\end{equation}
In \eqref{eq:J_old_rel} and \eqref{eq:J}, $\mathbf{X}$ is a third order tensor containing $\LRs{\mathbf{x}(t_{j})}_{i}$ for all samples, $s = 0, \ldots, N_s$, and all times, $t = 0, \ldots, N_t$, in the dataset.

The above loss function is used for the ODE problems which are trained using a log scaled version of the original data. For the PDE problems (\cref{sec:AC,,sec:CH}), the data is not log scaled so a different loss function was chosen. We do not use an absolute error style loss function, since some of the true data is 0, meaning the relative error will be undefined at those points. Instead, the evaluation metric \eqref{eq:R2} was used as the loss function in training.

\subsection{Error Evaluation Metrics}

The plots, in this paper, depicting point-wise relative error over time, report the relative error in the Euclidean norm, averaged over all $N$ samples, given as
\begin{equation}\label{eq:R1}
    \mathcal{R}_1 = \frac{1}{N+1}\sum_{i=0}^{N}\frac{\left|\left|\mathbf{x}_{i}(t_{j}) - \mathbf{\hat{x}}_{i}(t_{j})\right|\right|_{2}}{\left|\left|\mathbf{x}_{i}(t_{j})\right|\right|_{2}}.
\end{equation}
\Cref{fig:err_table} reports the point-wise relative error in the Euclidean norm, averaged over all $N$ samples and $M$ time steps, given as:
\begin{equation}\label{eq:R2}
    \mathcal{R}_2 = \frac{1}{(N+1) \cdot (M+1)}\sum_{i=0}^{N}\sum_{j=0}^{M}\frac{\left|\left|\mathbf{x}_{i}(t_{j}) - \mathbf{\hat{x}}_{i}(t_{j})\right|\right|_{2}}{\left|\left|\mathbf{x}_{i}(t_{j})\right|\right|_{2}}.
\end{equation}

\subsection{Description of methods adopted for comparison}
 \lilans is compared with the four approaches described below:  
\begin{equation*}
 \begin{aligned}
\text{DeepONet}&: \text{An operator learning approach with branch networks, taking the state as its input, and trunk}\\ 
&\text{ \ \ networks, taking time as its input, with the inner product of their respective outputs being the}\\
&\text{ \ \ predicted solution, as described in \cite{goswami2023learningstiffchemicalkinetics}.}\\
\text{Neural ODE}&: \text{A scheme for approximating the right hand side of a parameterized ODE with a neural network}\\ 
&\text{\ \ \  and solving using traditional numerical integration schemes, as described in \cite{Lee_2021}. The solver used}\\ 
&\text{ \ \ with the Neural ODEs implemented in our work is an implicit Euler scheme. Data and time steps}\\ 
&\text{ \ \ are rescaled in ways resembling the methodology in \cite{Kim_2021}.}\\
\text{Sulzer and Buck I}&: \text{A Neural ODE approach in a reduced dimension ($n_x \ge m$) latent space, described in \cite{sulzer2023speedingastrochemicalreactionnetworks}, in which}\\
&\text{\ \ \ the right hand side is considered a constant function of the encoded initial condition.}\\
\text{Sulzer and Buck II}&: \text{A Neural ODE approach in an expanded dimension ($n_x \le m$) latent space, in which the right hand}\\
&\text{\ \ \ side is considered a constant function of the encoded initial condition. This is a modified version}\\
&\text{\ \ \ of the approach described in \cite{sulzer2023speedingastrochemicalreactionnetworks}.}\\
\end{aligned}  
\end{equation*}

\subsection{Architectural Details}
\Cref{arc_det_1,arc_det_2,arc_det_3,arc_det_4,arc_det_5}  show the neural network architectures used for each machine learning method and problem setting. Each network uses the $tanh$ activation function at each hidden layer and has a linear output layer. The notation, [*,...,*], indicates the number of neurons in each layer for a multilayer perceptron network, including the initial input layer. Recall that, since we use the independent architectural variants for testing each ODE problem, there is one network of each type per degree of freedom in the system of ODEs for \lilans and DeepONet. Networks following the "independent" architectural variant from \cref{sec:variations} are labeled with a $*$ in the following table. All other networks are implemented as the "full" variant from \cref{sec:variations}. The labels (DR1) and (DR2) denote the architectures which were used in the data reduction experiments, described in \cref{sec:DR}. The label (DR1) indicates our approach, while (DR2) indicates the omission of the time transform. For problems, in the table without a (DR1) entry, the same architecture used in \cref{num_re} was used for the data reduction experiments.

\begin{table}[h!]
\caption{Architecture for \lilans for each problem.}
\label{arc_det_1}
\begin{tabular}{ |p{2.5cm}||p{2.0cm}|p{2.0cm}|p{3.0cm}|p{2.0cm}||p{2.0cm}|  }
 \hline
 \multicolumn{6}{|c|}{\lilans Approach} \\
 \hline
 \tiny Problem& \tiny Encoder $\bs{e}$ & \tiny Encoder $\bs{c}$ & \tiny Encoder $\boldsymbol{\tau}$ & \tiny Decoder $\bs{d}$ & \tiny Total Parameters \\
 \hline
 \tiny Robertson$^*$ & \tiny [$3$,$20$,$5$] & \tiny [$3$,$20$,$5$] & \tiny [$4$,$20$,$5$], $5$ & \tiny [$5$,$20$,$20$,$1$] & \tiny $3,423$ \\
 \hline
 \tiny Robertson$^*$ (DR2) & \tiny [$3$,$23$,$5$] & \tiny [$3$,$23$,$5$] & \tiny N/A & \tiny [$5$,$23$,$23$,$1$] & \tiny $3,414$ \\
 \hline
 \tiny CR Charge States$^*$ & \tiny [$3$,$20$,$5$] & \tiny [$3$,$20$,$5$] & \tiny [$4$,$20$,$5$], $5$ &  \tiny [$5$,$20$,$20$,$1$] & \tiny $4,564$\\
 \hline
 \tiny CR Charge States$^*$ (DR2) & \tiny [$3$,$23$,$5$] & \tiny [$3$,$23$,$5$] & \tiny N/A &  \tiny [$5$,$23$,$23$,$1$] & \tiny $4,552$\\
 \hline
 \tiny Full CR$^*$ & \tiny [$3$,$40$,$20$] & \tiny [$3$,$40$,$20$] & \tiny [$4$,$40$,$20$], $20$ & \tiny [$20$,$40$,$40$,$1$] & \tiny $518,974$ \\
 \hline
 \tiny Full CR$^*$ (DR1) & \tiny [$3$,$40$,$5$] & \tiny [$3$,$40$,$5$] & \tiny [$4$,$40$,$5$], $5$ & \tiny [$5$,$40$,$40$,$1$] & \tiny $287,734$ \\
 \hline
 \tiny Full CR$^*$ (DR2) & \tiny [$3$,$44$,$5$] & \tiny [$3$,$44$,$5$] & \tiny N/A & \tiny [$5$,$44$,$44$,$1$] & \tiny $290,554$ \\
 \hline
 \tiny Allen-Cahn & \tiny [$201$,$201$,$201$] & \tiny [$201$,$201$,$201$] & \tiny [$202$,$201$,$201$], 201 & \tiny [$201$,$201$,$201$,$201$] & \tiny $365,820$ \\
 \hline
 \tiny Allen-Cahn (DR1) & \tiny [$201$,$100$,$20$] & \tiny [$201$,$100$,$20$] & \tiny [$202$,$100$,$20$], 20 & \tiny [$20$,$100$,$100$,$201$] & \tiny $99,281$ \\
 \hline
 \tiny Allen-Cahn (DR2) & \tiny [$201$,$155$,$20$] & \tiny [$201$,$155$,$20$] & \tiny N/A & \tiny [$202$,$155$,$155$,$20$] & \tiny $99,596$ \\
 \hline
 \tiny Cahn-Hilliard & \tiny [$201$,$201$,$201$] & \tiny [$201$,$201$,$201$] & \tiny [$202$,$201$,$201$], 201 & \tiny [$201$,$201$,$201$,$201$] & \tiny $365,820$ \\
 \hline
 \tiny Cahn-Hilliard (DR1) & \tiny [$201$,$100$,$20$] & \tiny [$201$,$100$,$20$] & \tiny [$20$,$100$,$201$], 20 & \tiny [$20$,$100$,$100$,$201$] & \tiny $99,281$ \\
 \hline
 \tiny Cahn-Hilliard (DR2) & \tiny [$201$,$155$,$20$] & \tiny [$201$,$155$,$20$] & \tiny N/A & \tiny [$20$,$155$,$155$,$201$] & \tiny $99,596$ \\
 \hline
\end{tabular}
\end{table}

\begin{table}[h!]
\caption{Architecture for Sulzer and Buck I for each problem.}
\label{arc_det_2}
\begin{tabular}{ |p{2.5cm}||p{2.5cm}|p{2.5cm}|p{2.5cm}||p{2.5cm}|  }
 \hline
 \multicolumn{5}{|c|}{Sulzer and Buck I} \\
 \hline
 \tiny Problem& \tiny Encoder & \tiny Neural ODE & \tiny Decoder & \tiny Total Parameters \\
 \hline
 \tiny Robertson & \tiny [3,49,3] & \tiny [3,49,3] & \tiny [3,49,49,3] & \tiny $3,488$ \\
 \hline
 \tiny CR Charge States & \tiny [3,55,4] & \tiny [4,55,4] & \tiny [4,55,55,4] & \tiny $4,522$ \\
 \hline
 \tiny Full CR & \tiny [3,645,16] & \tiny [16,645,16] & \tiny [16,645,645,94] & \tiny $522,576$ \\
 \hline
 \tiny Allen-Cahn & \tiny [201,390,34] & \tiny [34,390,34] & \tiny [34,390,390,201] & \tiny $363,749$ \\
 \hline
 \tiny Cahn-Hilliard & \tiny [201,390,34] & \tiny [34,390,34] & \tiny [34,390,390,201] & \tiny $363,749$ \\
 \hline
\end{tabular}
\end{table}

\begin{table}[h!]
\caption{Architecture for Sulzer and Buck II for each problem.}
\label{arc_det_3}
\begin{tabular}{ |p{2.5cm}||p{2.5cm}|p{2.5cm}|p{2.5cm}||p{2.5cm}|  }
 \hline
 \multicolumn{5}{|c|}{Sulzer and Buck II} \\
 \hline
 \tiny Problem& \tiny Encoder & \tiny Neural ODE & \tiny Decoder & \tiny Total Parameters \\
 \hline
 \tiny Robertson & \tiny [3,33,15] & \tiny [15,33,15] & \tiny [15,33,33,3] & \tiny $3,432$ \\
 \hline
 \tiny CR Charge States & \tiny [3,36,20] & \tiny [20,36,20] & \tiny [20,36,36,4] & \tiny $4,616$  \\
 \hline
 \tiny Full CR & \tiny [3,67,1880] & \tiny [1880,67,1880] & \tiny [1880,67,67,94] & \tiny $518,950$ \\
 \hline
 \tiny Allen-Cahn & \tiny [201,250,201] & \tiny [201,250,201] & \tiny [201,250,250,201] & \tiny $365,603$ \\
 \hline
 \tiny Cahn-Hilliard & \tiny [201,250,201] & \tiny [201,250,201] & \tiny [201,250,250,201] & \tiny $365,603$ \\
 \hline
\end{tabular}
\end{table}

\begin{table}[h!]
\caption{Architecture for DeepONet for each problem.}
\label{arc_det_4}
\begin{tabular}{ |p{2.5cm}||p{2.5cm}|p{2.5cm}||p{2.5cm}|  }
 \hline
 \multicolumn{4}{|c|}{DeepONet} \\
 \hline
 \tiny Problem& \tiny Branch Network & \tiny Trunk Network & \tiny Total Parameters \\
 \hline
 \tiny Robertson & \tiny [$3$,$20$,$20$,$5$] & \tiny [$1$,$20$,$20$,$5$] & \tiny $3,510$ \\
 \hline
 \tiny CR Charge States & \tiny [$3$,$20$,$20$,$5$] & \tiny [$1$,$20$,$20$,$5$] & \tiny $4,680$ \\
 \hline
 \tiny Full CR & \tiny [$3$,$42$,$42$,$20$] & \tiny [$1$,$42$,$42$,$20$] & \tiny $524,896$ \\
 \hline
 \tiny Allen-Cahn & \tiny [$201$,$8$,$8$,$8$] & \tiny [$1$,$8$,$8$,$8$] & \tiny $356,976$ \\
 \hline
 \tiny Cahn-Hilliard & \tiny [$201$,$8$,$8$,$8$] & \tiny [$1$,$8$,$8$,$8$] & \tiny $356,976$ \\
 \hline
\end{tabular}
\end{table}

\begin{table}[h!]
\caption{Architecture for Neural ODE for each problem.}
\label{arc_det_5}
\begin{tabular}{ |p{2.5cm}||p{3.5cm}||p{2.5cm}|  }
 \hline
 \multicolumn{3}{|c|}{Neural ODE} \\
 \hline
 \tiny Problem& \tiny Network & \tiny Total Parameters\\
 \hline
 \tiny Robertson & \tiny [$6$,$39$,$39$,$39$,$3$] & \tiny $3,513$ \\
 \hline
 \tiny CR Charge States & \tiny [$7$,$45$,$45$,$45$,$4$] & \tiny $4,684$\\
 \hline
 \tiny Allen-Cahn & \tiny [$201$,$338$,$338$,$338$,$201$] & \tiny $365,579$ \\
 \hline
 \tiny Cahn-Hilliard & \tiny [$201$,$338$,$338$,$338$,$201$] & \tiny $365,579$ \\
 \hline
\end{tabular}
\end{table}

\subsection{Choice of Latent Dimension $m$ for each problem}
\label{appen_late}
The  choice of latent dimension $m$ for each problem and for different approaches is shown in \cref{late}.
\begin{table}[h!]
\caption{ Latent dimension $m$ for each problem.}
\label{late}
\begin{tabular}{ |p{2.5cm}||p{3.5cm}||p{2.5cm}||p{2.5cm}||p{2.5cm}|  }
 \hline
 & \tiny \lilans & \tiny Sulzer and Buck I& \tiny Sulzer and Buck II& \tiny DeepONet\\
 \hline
\tiny Robertson & \tiny 15 & \tiny 3 & \tiny 15 & \tiny 15 \\
 \hline
 \tiny CR Charge States & \tiny 20 & \tiny 4 & \tiny 20 &\tiny 20 \\
 \hline 
\tiny  Full CR & \tiny 1880 & \tiny 16 & \tiny 1880 &\tiny 1880 \\
 \hline
\tiny Allen-Cahn & \tiny 201 & \tiny 34 & \tiny 201 &\tiny 201 \\
 \hline
\tiny Cahn-Hilliard & \tiny 201 & \tiny 34 & \tiny 201 &\tiny 201 \\
 \hline
\end{tabular}
\end{table}

\subsection{Hyperparameter Settings}

\Cref{hyper_p} shows the hyperparameter settings used in training for each method and problem setting. In each ODE problem, the batch size is approximately $5\%$ of the total samples. In the PDE problems (\cref{sec:AC,,sec:CH}), the entire training set is given to the network as a single batch. The label (DR) in the table indicates the hyperparameter setting used in \cref{time_import}, which contains experiments on coarsening the time discretization of training data.

\begin{table}[h!]
\caption{ Hyperparameter settings for each problem.}
\label{hyper_p}
\begin{tabular}{ |p{3.0cm}||p{2.5cm}|p{2.5cm}|p{2.5cm}|p{2.5cm}|  }
 \hline
 \multicolumn{5}{|c|}{Training Hyperparameter Settings} \\
 \hline
 \tiny Problem& \tiny Learning Rate & \tiny Training Samples & \tiny Batch Size & \tiny Training epochs \\
 \hline
 \tiny Robertson & \tiny $10^{-5}$ & \tiny $4,096$ & \tiny $204$ & \tiny $10,000$ \\
 \hline
 \tiny Robertson (DR) & \tiny $10^{-5}$ & \tiny $128$ & \tiny $6$ & \tiny $10,000$ \\
 \hline
 \tiny CR Charge States & \tiny $10^{-4}$ & \tiny $15,625$ & \tiny $781$ & \tiny $10,000$ \\
 \hline
 \tiny CR Charge States (DR) & \tiny $10^{-4}$ & \tiny $125$ & \tiny $6$ & \tiny $10,000$ \\
 \hline
 \tiny Full CR & \tiny $10^{-4}$ & \tiny $4,096$ & \tiny $204$ & \tiny $10,000$ \\
 \hline
 \tiny Full CR (DR) & \tiny $10^{-4}$ & \tiny $256$ & \tiny $12$ & \tiny $10,000$ \\
 \hline
 \tiny Allen-Cahn & \tiny $10^{-4}$ & \tiny $900$ & \tiny $900$ & \tiny $20,000$ \\
\hline
 \tiny Allen-Cahn (DR) & \tiny $10^{-4}$ & \tiny $900$ & \tiny $900$ & \tiny $50,000$ \\
 \hline
 \tiny Cahn-Hilliard & \tiny $10^{-4}$ & \tiny $900$ & \tiny $900$ & \tiny $20,000$ \\
 \hline
 \tiny Cahn-Hilliard (DR) & \tiny $10^{-4}$ & \tiny $900$ & \tiny $900$ & \tiny $100,000$ \\
 \hline
\end{tabular}
\end{table}




\subsection{Initialization Sensitivity Studies}
\vspace{0.5cm}
\subsubsection{Robertson Chemical Reaction ODE}

For the Robertson ODE (\cref{sec:Robertson}), we trained networks considering 100 random parameter initializations. In \cref{fig:rob_init}, the left plot shows the mean relative error over time \eqref{eq:R1} on a test dataset of 512 samples, while the right plot shows the width of the error band over time. The width of the error band generally stays on the same scale as the error.

\begin{figure}[h!t!b!] 
\begin{minipage}{0.48\textwidth}
    \centering
    \resizebox{1\linewidth}{!}{\input{figs/rob_rand_err}}
    \centering
\end{minipage}
\begin{minipage}{0.48\textwidth}
    \centering
    \resizebox{1\linewidth}{!}{\input{figs/rob_err_band_width}}
    \centering
\end{minipage}
\caption{Left to Right: Mean relative error for Robertson ODE (\cref{sec:Robertson}) test dataset, calculated by \eqref{eq:R1}, across 100 random network parameter initializations, with min-max range; Maximum - Minimum error band width over time}
\label{fig:rob_init}
\end{figure}

\subsubsection{Collisional-Radiative Model}
For the CR charge state model (\cref{sec:CR-model}), we, again, trained networks from 100 random parameter initializations. In \cref{fig:charge_init}, the left plot shows the mean relative error over time \eqref{eq:R1} on a test dataset of 4096 samples, while the right plot shows the width of the error band over time. The width of the error band generally stays on the same scale as the error.

\begin{figure}[h!t!b!] 
\begin{minipage}{0.48\textwidth}
    \centering
    \resizebox{1\linewidth}{!}{\input{figs/charge_rand_err}}
    \centering
\end{minipage}
\begin{minipage}{0.48\textwidth}
    \centering
    \resizebox{1\linewidth}{!}{\input{figs/charge_err_band_width}}
    \centering
\end{minipage}
\caption{Left to Right: Mean relative error for CR charge state model (\cref{sec:CR-model}) test dataset, calculated by \eqref{eq:R1}, across 100 random network parameter initializations, with min-max range; Maximum - Minimum error band width over time}
\label{fig:charge_init}
\end{figure}

For the full CR model (\cref{sec:CR-model}), we trained networks from 15 random parameter initializations. Fewer networks were trained because the networks are much more costly to train on this problem. In \cref{fig:CR_init}, the left plot shows the mean relative error over time \eqref{eq:R1} on a test dataset of 2500 samples, while the right plot shows the width of the error band over time. The width of the error band generally stays on the same scale as the error.

\begin{figure}[h!t!b!] 
\begin{minipage}{0.48\textwidth}
    \centering
    \resizebox{1\linewidth}{!}{\input{figs/CR_rand_err}}
    \centering
\end{minipage}
\begin{minipage}{0.48\textwidth}
    \centering
    \resizebox{1\linewidth}{!}{\input{figs/CR_err_band_width}}
    \centering
\end{minipage}
\caption{Left to Right: Mean relative error for full CR model (\cref{sec:CR-model}) test dataset, calculated by \eqref{eq:R1}, across 100 random network parameter initializations, with min-max range; Maximum - Minimum error band width over time}
\label{fig:CR_init}
\end{figure}

\subsubsection{Allen-Cahn PDE}

For the Allen-Cahn PDE (\cref{sec:AC}), we trained networks from 100 random parameter initializations. In \cref{fig:AC_init}, the left plot shows the mean relative error over time \eqref{eq:R1} on a test dataset of 200 samples, while the right plot shows the width of the error band over time. The width of the error band generally stays on the same scale as the error.

\begin{figure}[h!t!b!] 
\begin{minipage}{0.48\textwidth}
    \centering
    \resizebox{1\linewidth}{!}{\input{figs/ac_rand_err}}
    \centering
\end{minipage}
\begin{minipage}{0.48\textwidth}
    \centering
    \resizebox{1\linewidth}{!}{\input{figs/ac_err_band_width}}
    \centering
\end{minipage}
\caption{Left to Right: Mean relative error for Allen-Cahn 
(\cref{sec:AC}) test dataset, calculated by \eqref{eq:R1}, across 100 random network parameter initializations, with min-max range; Maximum - Minimum error band width over time}
\label{fig:AC_init}
\end{figure}

\subsubsection{Cahn-Hilliard PDE}

For the Cahn-Hilliard PDE (\cref{sec:CH}), we trained networks from 100 random parameter initializations. In \cref{fig:CH_init}, the left plot shows the mean relative error over time \eqref{eq:R1} on a test dataset of 200 samples, while the right plot shows the width of the error band over time. The width of the error band generally stays on the same scale as the error.

\begin{figure}[h!t!b!] 
\begin{minipage}{0.48\textwidth}
    \centering
    \resizebox{1\linewidth}{!}{\input{figs/ch_rand_err}}
    \centering
\end{minipage}
\begin{minipage}{0.48\textwidth}
    \centering
    \resizebox{1\linewidth}{!}{\input{figs/ch_err_band_width}}
    \centering
\end{minipage}
\caption{Left to Right: Mean relative error for Cahn-Hilliard (\cref{sec:CH}) test dataset, calculated by \eqref{eq:R1}, across 100 random network parameter initializations, with min-max range; Maximum - Minimum error band width over time}
\label{fig:CH_init}
\end{figure}

%% file: figs/rob_rand_err.tex
\begin{tikzpicture}

\definecolor{darkgray176}{RGB}{176,176,176}
\definecolor{lightgray204}{RGB}{204,204,204}
\definecolor{steelblue31119180}{RGB}{31,119,180}

\begin{axis}[
legend cell align={left},
legend style={fill opacity=0.8, draw opacity=1,at={(0.98,0.2)}, text opacity=1, draw=lightgray204},
log basis x={10},
log basis y={10},
tick align=outside,
tick pos=left,
title={Error Range of Networks Trained from 100 Random Initializations},
x grid style={darkgray176},
xlabel={Time (s)},
xmin=1.59985847858479e-05, xmax=100000,
xmode=log,
xtick style={color=black},
xtick={1e-07,1e-05,0.001,0.1,10,1000,100000,10000000},
xticklabels={
  \(\displaystyle {10^{-7}}\),
  \(\displaystyle {10^{-5}}\),
  \(\displaystyle {10^{-3}}\),
  \(\displaystyle {10^{-1}}\),
  \(\displaystyle {10^{1}}\),
  \(\displaystyle {10^{3}}\),
  \(\displaystyle {10^{5}}\),
  \(\displaystyle {10^{7}}\)
},
y grid style={darkgray176},
ylabel={Relative L2-Norm Error},
ymin=8.55954025036687e-09, ymax=0.00322221338671679,
ymode=log,
ytick style={color=black},
ytick={1e-10,1e-09,1e-08,1e-07,1e-06,1e-05,0.0001,0.001,0.01,0.1},
yticklabels={
  \(\displaystyle {10^{-10}}\),
  \(\displaystyle {10^{-9}}\),
  \(\displaystyle {10^{-8}}\),
  \(\displaystyle {10^{-7}}\),
  \(\displaystyle {10^{-6}}\),
  \(\displaystyle {10^{-5}}\),
  \(\displaystyle {10^{-4}}\),
  \(\displaystyle {10^{-3}}\),
  \(\displaystyle {10^{-2}}\),
  \(\displaystyle {10^{-1}}\)
}
]
\path [fill=steelblue31119180, fill opacity=0.2]
(axis cs:1.59985847858479e-05,3.92078440825117e-08)
--(axis cs:1.59985847858479e-05,3.01260811852444e-08)
--(axis cs:2.55954673775705e-05,1.82279293881038e-08)
--(axis cs:4.09491185564548e-05,1.53423371784811e-08)
--(axis cs:6.55128751532175e-05,1.71902989620776e-08)
--(axis cs:0.000104811362689361,4.58178881501681e-08)
--(axis cs:0.000167683334439062,3.70253552262056e-08)
--(axis cs:0.000268269592197612,4.50507258165089e-08)
--(axis cs:0.000429193605668843,5.51184378139169e-08)
--(axis cs:0.000686648942064494,6.70187674245426e-08)
--(axis cs:0.0010985410772264,9.57441557147831e-08)
--(axis cs:0.00175751012284309,1.46232267184132e-07)
--(axis cs:0.00281176739372313,2.25589317892627e-07)
--(axis cs:0.00449842913076282,2.67790198904549e-07)
--(axis cs:0.00719684967771173,4.71345515506982e-07)
--(axis cs:0.0115139465779066,6.15704493611702e-07)
--(axis cs:0.018420685082674,1.26298573377426e-06)
--(axis cs:0.0294704847037792,1.67395899097755e-06)
--(axis cs:0.0471486635506153,2.99199359687918e-06)
--(axis cs:0.075431190431118,4.57270107290242e-06)
--(axis cs:0.120679251849651,7.91490674600936e-06)
--(axis cs:0.193069815635681,1.3484898772731e-05)
--(axis cs:0.308884382247925,1.58087586896727e-05)
--(axis cs:0.494171231985092,2.33413520618342e-05)
--(axis cs:0.790603995323181,4.51856867584866e-05)
--(axis cs:1.26485514640808,7.15483547537588e-05)
--(axis cs:2.02358889579773,9.83645368251018e-05)
--(axis cs:3.23745584487915,0.000143614117405377)
--(axis cs:5.17947053909302,0.000198646856006235)
--(axis cs:8.28642463684082,0.00023226166376844)
--(axis cs:13.2571115493774,0.00028452547849156)
--(axis cs:21.2095012664795,0.0003360616392456)
--(axis cs:33.9321975708008,0.000391320878406987)
--(axis cs:54.2867088317871,0.000402244128054008)
--(axis cs:86.8510513305664,0.000412489345762879)
--(axis cs:138.949447631836,0.000415148591855541)
--(axis cs:222.299545288086,0.000532686768565327)
--(axis cs:355.647796630859,0.000560918648261577)
--(axis cs:568.986145019531,0.000514143786858767)
--(axis cs:910.297180175781,0.000469661026727408)
--(axis cs:1456.34826660156,0.000454304739832878)
--(axis cs:2329.95092773438,0.00042579259024933)
--(axis cs:3727.59326171875,0.000367503525922075)
--(axis cs:5963.62158203125,0.000280333828413859)
--(axis cs:9540.9501953125,0.000228891905862838)
--(axis cs:15264.16015625,0.000192414445336908)
--(axis cs:24420.494140625,0.000156412541400641)
--(axis cs:39069.33203125,0.000125992213725112)
--(axis cs:62505.4609375,8.67112321429886e-05)
--(axis cs:100000,8.50956566864625e-05)
--(axis cs:100000,0.000332257885020226)
--(axis cs:100000,0.000332257885020226)
--(axis cs:62505.4609375,0.000282255641650409)
--(axis cs:39069.33203125,0.000389905355405062)
--(axis cs:24420.494140625,0.000523737573530525)
--(axis cs:15264.16015625,0.000685930775944144)
--(axis cs:9540.9501953125,0.000775350141339004)
--(axis cs:5963.62158203125,0.000816537125501782)
--(axis cs:3727.59326171875,0.0009762400877662)
--(axis cs:2329.95092773438,0.00120824493933469)
--(axis cs:1456.34826660156,0.00146149459760636)
--(axis cs:910.297180175781,0.00172557192854583)
--(axis cs:568.986145019531,0.00179768342059106)
--(axis cs:355.647796630859,0.00162584136705846)
--(axis cs:222.299545288086,0.00158142112195492)
--(axis cs:138.949447631836,0.00157361326273531)
--(axis cs:86.8510513305664,0.00141706434078515)
--(axis cs:54.2867088317871,0.00115257070865482)
--(axis cs:33.9321975708008,0.00112004624679685)
--(axis cs:21.2095012664795,0.000998957199044526)
--(axis cs:13.2571115493774,0.000897002930287272)
--(axis cs:8.28642463684082,0.000657471595332026)
--(axis cs:5.17947053909302,0.000561432156246156)
--(axis cs:3.23745584487915,0.000482009869301692)
--(axis cs:2.02358889579773,0.000367656321031973)
--(axis cs:1.26485514640808,0.00025034326245077)
--(axis cs:0.790603995323181,0.000151451982674189)
--(axis cs:0.494171231985092,0.000105046594399028)
--(axis cs:0.308884382247925,7.44117714930326e-05)
--(axis cs:0.193069815635681,4.23958554165438e-05)
--(axis cs:0.120679251849651,2.49530112341745e-05)
--(axis cs:0.075431190431118,1.62408632604638e-05)
--(axis cs:0.0471486635506153,9.85263068287168e-06)
--(axis cs:0.0294704847037792,5.68373025089386e-06)
--(axis cs:0.018420685082674,4.31858552474296e-06)
--(axis cs:0.0115139465779066,2.61805348600319e-06)
--(axis cs:0.00719684967771173,1.48676599565079e-06)
--(axis cs:0.00449842913076282,1.25573262721446e-06)
--(axis cs:0.00281176739372313,8.00613463525224e-07)
--(axis cs:0.00175751012284309,1.10241444417625e-06)
--(axis cs:0.0010985410772264,6.30725423889089e-07)
--(axis cs:0.000686648942064494,5.31654052338126e-07)
--(axis cs:0.000429193605668843,3.44260342899361e-07)
--(axis cs:0.000268269592197612,2.11078017287036e-07)
--(axis cs:0.000167683334439062,2.34269620591476e-07)
--(axis cs:0.000104811362689361,1.32492104398807e-07)
--(axis cs:6.55128751532175e-05,4.53431532321247e-08)
--(axis cs:4.09491185564548e-05,8.99874876836293e-08)
--(axis cs:2.55954673775705e-05,7.34891685283401e-08)
--(axis cs:1.59985847858479e-05,3.92078440825117e-08)
--cycle;
\addlegendimage{area legend, fill=steelblue31119180, fill opacity=0.2}
\addlegendentry{Min/Max Error Range}

\addplot [semithick, steelblue31119180]
table {%
1.59985847858479e-05 3.06305167896426e-08
2.55954673775705e-05 2.16664979291181e-08
4.09491185564548e-05 2.08750030594729e-08
6.55128751532175e-05 2.39412916158699e-08
0.000104811362689361 5.17129308263975e-08
0.000167683334439062 5.59583490655768e-08
0.000268269592197612 7.00530691233325e-08
0.000429193605668843 1.04180166715651e-07
0.000686648942064494 1.80604459387723e-07
0.0010985410772264 2.27159461019255e-07
0.00175751012284309 3.44062129897793e-07
0.00281176739372313 4.04692457323108e-07
0.00449842913076282 5.77341324969893e-07
0.00719684967771173 8.00659734068176e-07
0.0115139465779066 1.26900044961076e-06
0.018420685082674 2.03558738576248e-06
0.0294704847037792 3.27530688082334e-06
0.0471486635506153 5.49523338122526e-06
0.075431190431118 8.6434720287798e-06
0.120679251849651 1.38872956085834e-05
0.193069815635681 2.50741813943023e-05
0.308884382247925 4.30582695116755e-05
0.494171231985092 6.49832654744387e-05
0.790603995323181 9.33832488954067e-05
1.26485514640808 0.000136628543259576
2.02358889579773 0.000197185290744528
3.23745584487915 0.000267336727119982
5.17947053909302 0.000342410581652075
8.28642463684082 0.000422411772888154
13.2571115493774 0.000503674906212837
21.2095012664795 0.00057962368009612
33.9321975708008 0.000648639688733965
54.2867088317871 0.00070312200114131
86.8510513305664 0.000754981767386198
138.949447631836 0.000825213326606899
222.299545288086 0.000913278898224235
355.647796630859 0.00099486974067986
568.986145019531 0.00103351287543774
910.297180175781 0.00100567704066634
1456.34826660156 0.000914052710868418
2329.95092773438 0.000782687624450773
3727.59326171875 0.000647563370876014
5963.62158203125 0.000529235054273158
9540.9501953125 0.000430336687713861
15264.16015625 0.00034500626497902
24420.494140625 0.00026783932116814
39069.33203125 0.000204076743102632
62505.4609375 0.000166661062394269
100000 0.000166331985383295
};
\addlegendentry{Mean Error}
\end{axis}

\end{tikzpicture}

%% file: figs/rob_err_band_width.tex
\begin{tikzpicture}

\definecolor{darkgray176}{RGB}{176,176,176}
\definecolor{lightgray204}{RGB}{204,204,204}
\definecolor{steelblue31119180}{RGB}{31,119,180}

\begin{axis}[
legend cell align={left},
legend style={fill opacity=0.8, draw opacity=1,at={(0.98,0.2)}, text opacity=1, draw=lightgray204},
log basis x={10},
log basis y={10},
tick align=outside,
tick pos=left,
title={Error Range of Networks Trained from 100 Random Initializations},
x grid style={darkgray176},
xlabel={Time (s)},
xmin=1.59985847858479e-05, xmax=100000,
xmode=log,
xtick style={color=black},
xtick={1e-07,1e-05,0.001,0.1,10,1000,100000,10000000},
xticklabels={
  \(\displaystyle {10^{-7}}\),
  \(\displaystyle {10^{-5}}\),
  \(\displaystyle {10^{-3}}\),
  \(\displaystyle {10^{-1}}\),
  \(\displaystyle {10^{1}}\),
  \(\displaystyle {10^{3}}\),
  \(\displaystyle {10^{5}}\),
  \(\displaystyle {10^{7}}\)
},
y grid style={darkgray176},
ylabel={Width of Error Band},
ymin=5.01947392072087e-09, ymax=0.00232231571187365,
ymode=log,
ytick style={color=black},
ytick={1e-10,1e-09,1e-08,1e-07,1e-06,1e-05,0.0001,0.001,0.01,0.1},
yticklabels={
  \(\displaystyle {10^{-10}}\),
  \(\displaystyle {10^{-9}}\),
  \(\displaystyle {10^{-8}}\),
  \(\displaystyle {10^{-7}}\),
  \(\displaystyle {10^{-6}}\),
  \(\displaystyle {10^{-5}}\),
  \(\displaystyle {10^{-4}}\),
  \(\displaystyle {10^{-3}}\),
  \(\displaystyle {10^{-2}}\),
  \(\displaystyle {10^{-1}}\)
}
]
\addplot [semithick, steelblue31119180]
table {%
1.59985847858479e-05 9.08176289726725e-09
2.55954673775705e-05 5.52612391402363e-08
4.09491185564548e-05 7.46451505051482e-08
6.55128751532175e-05 2.81528542700471e-08
0.000104811362689361 8.66742198013526e-08
0.000167683334439062 1.97244261812557e-07
0.000268269592197612 1.66027291470527e-07
0.000429193605668843 2.89141894427303e-07
0.000686648942064494 4.64635292019011e-07
0.0010985410772264 5.34981268174306e-07
0.00175751012284309 9.56182134359551e-07
0.00281176739372313 5.75024159843451e-07
0.00449842913076282 9.87942485153326e-07
0.00719684967771173 1.0154204801438e-06
0.0115139465779066 2.00234899239149e-06
0.018420685082674 3.0555997909687e-06
0.0294704847037792 4.00977114622947e-06
0.0471486635506153 6.86063685861882e-06
0.075431190431118 1.16681621875614e-05
0.120679251849651 1.70381044881651e-05
0.193069815635681 2.89109557343181e-05
0.308884382247925 5.86030146223493e-05
0.494171231985092 8.17052423371933e-05
0.790603995323181 0.000106266292277724
1.26485514640808 0.000178794900421053
2.02358889579773 0.000269291776930913
3.23745584487915 0.00033839576644823
5.17947053909302 0.000362785300239921
8.28642463684082 0.000425209931563586
13.2571115493774 0.000612477422691882
21.2095012664795 0.000662895559798926
33.9321975708008 0.000728725339286029
54.2867088317871 0.000750326551496983
86.8510513305664 0.00100457505322993
138.949447631836 0.0011584646999836
222.299545288086 0.00104873441159725
355.647796630859 0.00106492266058922
568.986145019531 0.00128353969193995
910.297180175781 0.00125591084361076
1456.34826660156 0.00100718985777348
2329.95092773438 0.000782452349085361
3727.59326171875 0.000608736532740295
5963.62158203125 0.000536203267984092
9540.9501953125 0.000546458235476166
15264.16015625 0.000493516330607235
24420.494140625 0.000367325032129884
39069.33203125 0.000263913127128035
62505.4609375 0.000195544416783378
100000 0.000247162242885679
};
\addlegendentry{Width of Error Band}
\end{axis}

\end{tikzpicture}

%% file: figs/charge_err_band_width.tex
\begin{tikzpicture}

\definecolor{darkgray176}{RGB}{176,176,176}
\definecolor{lightgray204}{RGB}{204,204,204}
\definecolor{steelblue31119180}{RGB}{31,119,180}

\begin{axis}[
legend cell align={left},
legend style={fill opacity=0.8, draw opacity=1,at={(0.98,0.2)}, text opacity=1, draw=lightgray204},
log basis x={10},
log basis y={10},
tick align=outside,
tick pos=left,
title={Error Range of Networks Trained from 100 Random Initializations},
x grid style={darkgray176},
xlabel={Time (s)},
xmin=1.09673091474103e-16, xmax=1,
xmode=log,
xtick style={color=black},
xtick={1e-18,1e-16,1e-14,1e-12,1e-10,1e-08,1e-06,0.0001,0.01,1,100},
xticklabels={
  \(\displaystyle {10^{-18}}\),
  \(\displaystyle {10^{-16}}\),
  \(\displaystyle {10^{-14}}\),
  \(\displaystyle {10^{-12}}\),
  \(\displaystyle {10^{-10}}\),
  \(\displaystyle {10^{-8}}\),
  \(\displaystyle {10^{-6}}\),
  \(\displaystyle {10^{-4}}\),
  \(\displaystyle {10^{-2}}\),
  \(\displaystyle {10^{0}}\),
  \(\displaystyle {10^{2}}\)
},
y grid style={darkgray176},
ylabel={Width of Error Band},
ymin=7.77942906272676e-06, ymax=0.0218269023674996,
ymode=log,
ytick style={color=black},
ytick={1e-07,1e-06,1e-05,0.0001,0.001,0.01,0.1,1},
yticklabels={
  \(\displaystyle {10^{-7}}\),
  \(\displaystyle {10^{-6}}\),
  \(\displaystyle {10^{-5}}\),
  \(\displaystyle {10^{-4}}\),
  \(\displaystyle {10^{-3}}\),
  \(\displaystyle {10^{-2}}\),
  \(\displaystyle {10^{-1}}\),
  \(\displaystyle {10^{0}}\)
}
]
\addplot [semithick, steelblue31119180]
table {%
1.09673091474103e-16 1.73699208971811e-05
1.20281868118553e-16 1.75432724063285e-05
1.31916845030936e-16 1.77043439180125e-05
1.44677608139559e-16 1.79002217919333e-05
1.58672418842992e-16 1.81040813913569e-05
1.74020947014816e-16 1.82858948392095e-05
1.90854165124498e-16 1.85241624421906e-05
2.0931567172638e-16 1.87293571798364e-05
2.29562948774199e-16 1.89623478945578e-05
2.51769346854569e-16 1.92078150575981e-05
2.76123238154025e-16 1.94836156879319e-05
3.02832913368252e-16 1.97315494006034e-05
3.32126224360062e-16 2.00101803784491e-05
3.64253125258233e-16 2.02957508008694e-05
3.99487657690983e-16 2.05602045753039e-05
4.38131418327092e-16 2.08253022719873e-05
4.80512314796267e-16 2.11312653846107e-05
5.26992771320814e-16 2.14422470889986e-05
5.77969172850268e-16 2.17323631659383e-05
6.33876735500841e-16 2.20301208173623e-05
6.95192232942718e-16 2.23478073166916e-05
7.62440507965842e-16 2.26570282393368e-05
8.36192090199748e-16 2.29626875807298e-05
9.17077807431903e-16 2.33041300816694e-05
1.00578756799785e-15 2.36146261158865e-05
1.10307830171507e-15 2.39165347011294e-05
1.20978010087187e-15 2.42243277170928e-05
1.32680331412833e-15 2.45038572757039e-05
1.45514945206545e-15 2.47551542997826e-05
1.59590737553692e-15 2.50130651693325e-05
1.75028093011903e-15 2.53122943831841e-05
1.91958735792387e-15 2.55503837252036e-05
2.10527074424741e-15 2.58197760558687e-05
2.30892096993163e-15 2.60713968600612e-05
2.53226492339775e-15 2.6310875909985e-05
2.77721329921654e-15 2.65307862719055e-05
3.04585551591063e-15 2.67263403657125e-05
3.34048427991002e-15 2.69143019977491e-05
3.66361231445272e-15 2.71154840447707e-05
4.01799650002381e-15 2.72694414888974e-05
4.40667196743155e-15 2.74636295216624e-05
4.83293282780781e-15 2.76344744634116e-05
5.30042682853993e-15 2.78024708677549e-05
5.81314264777653e-15 2.79746800515568e-05
6.37545309310782e-15 2.81468310276978e-05
6.99215702969647e-15 2.83631416095886e-05
7.66853231983682e-15 2.86030044662766e-05
8.41031634119709e-15 2.88078445009887e-05
9.22385337055222e-15 2.90812422463205e-05
1.01160856899379e-14 2.93815446639201e-05
1.10946238903852e-14 2.96736507152673e-05
1.21678180226882e-14 2.99744770018151e-05
1.33448228189408e-14 3.03218657791149e-05
1.46357128890176e-14 3.06732072203886e-05
1.60514386748608e-14 3.1030362151796e-05
1.76041076654183e-14 3.13471537083387e-05
1.93069691500512e-14 3.16699297400191e-05
2.11745514378707e-14 3.19902719638776e-05
2.32228380405916e-14 3.23398053296842e-05
2.54692049920938e-14 3.26629960909486e-05
2.79328663877528e-14 3.29612194036599e-05
3.06348405031218e-14 3.32536910718773e-05
3.35981751046956e-14 3.35722506861202e-05
3.68481558200157e-14 3.39153593813535e-05
4.04125178959068e-14 3.42621715390123e-05
4.43216562626483e-14 3.46023953170516e-05
4.86090388860522e-14 3.49936963175423e-05
5.33110373608734e-14 3.53763280145358e-05
5.84678654233159e-14 3.57314856955782e-05
6.412351796466e-14 3.60954836651217e-05
7.03262453697132e-14 3.64560837624595e-05
7.71291396636092e-14 3.67877037206199e-05
8.45899142832424e-14 3.70926209143363e-05
9.27723677501997e-14 3.73895818484016e-05
1.01746329460653e-13 3.7683013943024e-05
1.1158834408798e-13 3.79864686692599e-05
1.22382396642966e-13 3.82573271053843e-05
1.34220556218846e-13 3.85179300792515e-05
1.47204175389958e-13 3.87967738788575e-05
1.6144338199197e-13 3.90631248592399e-05
1.77059931928854e-13 3.93493137380574e-05
1.94187100752662e-13 3.96395880670752e-05
2.12970998258663e-13 3.9905629819259e-05
2.33572411629038e-13 4.01617980969604e-05
2.56166100701463e-13 4.04048478230834e-05
2.80945297408131e-13 4.06310864491388e-05
3.08121366962566e-13 4.08334963140078e-05
3.37926247314518e-13 4.10096581617836e-05
3.70614156768381e-13 4.11708133469801e-05
4.06464033438081e-13 4.13101843150798e-05
4.45781703651421e-13 4.14437599829398e-05
4.88903676657687e-13 4.1553117625881e-05
5.36195762269875e-13 4.16593175032176e-05
5.8806247631854e-13 4.17152405134402e-05
6.44946281710274e-13 4.17339288105723e-05
7.07334147850847e-13 4.1741212044144e-05
7.75755219610541e-13 4.17442606703844e-05
8.50794803532173e-13 4.16998045693617e-05
9.33093012578379e-13 4.16443981521297e-05
1.02335208449628e-12 4.15749018429779e-05
1.12234180282833e-12 4.15173853980377e-05
1.23090686948713e-12 4.14064597862307e-05
1.34997373522905e-12 4.12764857173897e-05
1.48055786380885e-12 4.11079199693631e-05
1.62377739292702e-12 4.09196800319478e-05
1.78084684585716e-12 4.07387924497016e-05
1.95310998277187e-12 4.05339451390319e-05
2.14204066810442e-12 4.03468693548348e-05
2.34924232844769e-12 4.01713259634562e-05
2.57648676699196e-12 3.99676719098352e-05
2.82571253722164e-12 3.97475487261545e-05
3.09904641011827e-12 3.94941562262829e-05
3.39882028771443e-12 3.92320253013168e-05
3.72759158609459e-12 3.89656925108284e-05
4.0881647025981e-12 3.86602405342273e-05
4.48361703667133e-12 3.83846381737385e-05
4.91732132443246e-12 3.80802848667372e-05
5.39299030780094e-12 3.77888682123739e-05
5.91465938726277e-12 3.74425035261083e-05
6.48679027159815e-12 3.70540510630235e-05
7.11427878413717e-12 3.66254389518872e-05
7.80244965858934e-12 3.62152059096843e-05
8.5571879443469e-12 3.5758985177381e-05
9.38493380231442e-12 3.52789320459124e-05
1.02927458223157e-11 3.48013090842869e-05
1.12883738814795e-11 3.43059582519345e-05
1.23803077164975e-11 3.37534693244379e-05
1.35778671292441e-11 3.32172821799759e-05
1.48912670389079e-11 3.26735753333196e-05
1.63317501883364e-11 3.21124753099866e-05
1.79115351023329e-11 3.14754943246953e-05
1.96441352767751e-11 3.08527742163278e-05
2.15443791279357e-11 3.02233729598811e-05
2.3628387441077e-11 2.95616537187016e-05
2.59139827651911e-11 2.88435039692558e-05
2.84206651268715e-11 2.80919084616471e-05
3.11698236665769e-11 2.73601199296536e-05
3.41849118457027e-11 2.66616516455542e-05
3.74916521439506e-11 2.59552580246236e-05
4.11182476955929e-11 2.51056462730048e-05
4.50956563757821e-11 2.43069534917595e-05
4.94578059062611e-11 2.34979434026172e-05
5.42420240667862e-11 2.26947104238207e-05
5.94889137950361e-11 2.19093490159139e-05
6.52433246317585e-11 2.09766258194577e-05
7.15545331320122e-11 2.01686398213496e-05
7.84760728622658e-11 1.92699262697715e-05
8.60671325875195e-11 1.83992142410716e-05
9.43924868823665e-11 1.74584783962928e-05
1.03523162264807e-10 1.65862438734621e-05
1.13537054757806e-10 1.57232825586107e-05
1.24519602962003e-10 1.48610561154783e-05
1.36564509700676e-10 1.39004068842041e-05
1.49774512236966e-10 1.29725058286567e-05
1.6426270066372e-10 1.20888880701386e-05
1.80151976669762e-10 1.11603294499218e-05
1.97578273186672e-10 1.12573825390427e-05
2.16690665411079e-10 1.14304348244332e-05
2.3765137080467e-10 1.23289364637458e-05
2.60639620996983e-10 1.33037965497351e-05
2.85851509129742e-10 1.4264361198002e-05
3.13502224180695e-10 1.53167256939923e-05
3.43827577520273e-10 1.6486595995957e-05
3.77086334379939e-10 1.75978857441805e-05
4.13562267764789e-10 1.87391924555413e-05
4.53566517677473e-10 1.9948525732616e-05
4.97440422186912e-10 2.12268096220214e-05
5.45559486475611e-10 2.24509021791164e-05
5.98332050572026e-10 2.37941949308151e-05
6.56209253602213e-10 2.51292749453569e-05
7.19686532590913e-10 2.64890732069034e-05
7.89302512238521e-10 2.79329760815017e-05
8.65652494130842e-10 2.9397302569123e-05
9.49387790605272e-10 3.09109600493684e-05
1.04122299671161e-09 3.23205749737099e-05
1.14194165234949e-09 3.38297941198107e-05
1.25240273707306e-09 3.53239847754594e-05
1.37354871920792e-09 3.69217414117884e-05
1.50641343843461e-09 3.83013175451197e-05
1.65213387415264e-09 3.97429394070059e-05
1.81194625969994e-09 4.09511267207563e-05
1.98721772370902e-09 4.21003351220861e-05
2.17944773339696e-09 4.36963091487996e-05
2.39026776149842e-09 4.51516825705767e-05
2.62148081020541e-09 4.66000601591077e-05
2.87505907969887e-09 4.80893286294304e-05
3.15316617260919e-09 4.97695291414857e-05
3.45817485758459e-09 5.17076150572393e-05
3.79268749739481e-09 5.39389839104842e-05
4.15955758725772e-09 5.68829455005471e-05
4.56191573405818e-09 6.13264201092534e-05
5.00320540552934e-09 6.66871928842738e-05
5.48716982962105e-09 7.85652227932587e-05
6.01794880594753e-09 0.000102709986094851
6.6000702680924e-09 0.000134460016852245
7.23851734107939e-09 0.000175369612406939
7.93870658100104e-09 0.000226271076826379
8.70662475449535e-09 0.000288051378447562
9.54882484194286e-09 0.000362125661922619
1.04724913185805e-08 0.000458430149592459
1.14855049915263e-08 0.0005804393440485
1.25965096131608e-08 0.000726782018318772
1.38150122452885e-08 0.000894299650099128
1.5151352172893e-08 0.00107103865593672
1.66169566995222e-08 0.00125381373800337
1.82243304891472e-08 0.00146720267366618
1.9987187016568e-08 0.00169658684171736
2.19205666951439e-08 0.00198221579194069
2.40409629981286e-08 0.00243740295991302
2.63665267397073e-08 0.00327062443830073
2.89169861389382e-08 0.00550302024930716
3.17141548578093e-08 0.00892180111259222
3.47818946977441e-08 0.0123222470283508
3.81463785004144e-08 0.0126566179096699
4.18363121923448e-08 0.0102479178458452
4.58832829508538e-08 0.00734824128448963
5.0321613542792e-08 0.00765089225023985
5.51892718192448e-08 0.00901636015623808
6.05277818976901e-08 0.00909562781453133
6.63826895674902e-08 0.0078900707885623
7.28039495356825e-08 0.00609751511365175
7.98465151774508e-08 0.00529119465500116
8.75701502422999e-08 0.00471476186066866
9.60408925720913e-08 0.00439048325642943
1.05331025679334e-07 0.00437162164598703
1.1551978928992e-07 0.0047098696231842
1.26694132518423e-07 0.00433208793401718
1.38949673100797e-07 0.00398158421739936
1.52390427388127e-07 0.00476308912038803
1.67131275929933e-07 0.00511795142665505
1.83298055844716e-07 0.00445945281535387
2.01028640844925e-07 0.00470628077164292
2.20474348111566e-07 0.00703052617609501
2.41801018319165e-07 0.00744301732629538
2.65191260950814e-07 0.00574670592322946
2.90843473749192e-07 0.00661271065473557
3.18977015467681e-07 0.0116394646465778
3.49831935864131e-07 0.0152146797627211
3.83671533654706e-07 0.0127433799207211
4.20784431298671e-07 0.00881195813417435
4.61488326664039e-07 0.00630461983382702
5.06128515098681e-07 0.00584143260493875
5.55086785425374e-07 0.00623701233416796
6.08780965194455e-07 0.00701754353940487
6.67668814458011e-07 0.00767989084124565
7.32253113255865e-07 0.00807971972972155
8.03086379619344e-07 0.0080535039305687
8.80769675859483e-07 0.00775555288419127
9.65967274169088e-07 0.00776941236108541
1.05940625871881e-06 0.00788660906255245
1.16188368792791e-06 0.00806067325174809
1.27427381357847e-06 0.00828584283590317
1.39753854000446e-06 0.00869813747704029
1.53272378611291e-06 0.00904517620801926
1.68098563335661e-06 0.00942842476069927
1.84358884780522e-06 0.00994935631752014
2.02192109099997e-06 0.0105871390551329
2.21750315176905e-06 0.0112145179882646
2.43200452132442e-06 0.011894297786057
2.66726055997424e-06 0.0124830659478903
2.92526715384156e-06 0.0128609044477344
3.20823119182023e-06 0.0128875700756907
3.51856601810141e-06 0.012684078887105
3.85892053600401e-06 0.012437466531992
4.23219717049506e-06 0.0122367031872272
4.64159211333026e-06 0.0122871976345778
5.09057781528099e-06 0.0122782867401838
5.58299416297814e-06 0.0121370544657111
6.12304302194389e-06 0.0119100073352456
6.71533007334801e-06 0.0116755403578281
7.36491028874298e-06 0.0115494728088379
8.07734340924071e-06 0.0113291274756193
8.8586712081451e-06 0.0110060097649693
9.71557983575622e-06 0.010675523430109
1.06553761725081e-05 0.010278282687068
1.16860819616704e-05 0.00973415561020374
1.28164874695358e-05 0.00912221148610115
1.40562688102364e-05 0.00850425846874714
1.54159461089876e-05 0.00811492465436459
1.69071445270674e-05 0.00776604795828462
1.85425888048485e-05 0.00741439452394843
2.03362305910559e-05 0.0069755595177412
2.23033730435418e-05 0.00652071414515376
2.44607999775326e-05 0.00607561459764838
2.68269723164849e-05 0.00585700431838632
2.94219735224033e-05 0.0057008033618331
3.22679879900534e-05 0.00550549756735563
3.53893046849407e-05 0.00524992821738124
3.88125408790074e-05 0.00492947455495596
4.25669131800532e-05 0.00484606856480241
4.66845558548812e-05 0.00467914761975408
5.12003971380182e-05 0.00449552899226546
5.61530578124803e-05 0.00426610466092825
6.15848111920059e-05 0.00396517291665077
6.75419578328729e-05 0.00362414517439902
7.40753530408256e-05 0.00349701382219791
8.12409052741714e-05 0.00343466829508543
8.9099419710692e-05 0.00340771139599383
9.77180898189545e-05 0.00341128371655941
0.000107170446426608 0.00351431081071496
0.000117537150799762 0.00369541579857469
0.000128906642203219 0.00399197172373533
0.000141376207466237 0.00436158757656813
0.0001550516608404 0.00474663730710745
0.000170049941516481 0.00512075331062078
0.000186499033588916 0.00550002884119749
0.000204539261176251 0.00595319736748934
0.000224324539885856 0.00634841807186604
0.000246023701038212 0.00666230171918869
0.00026982236886397 0.0068760016001761
0.000295922538498417 0.00702366977930069
0.000324547407217324 0.00709732808172703
0.000355941185262054 0.00702908355742693
0.000390371715184301 0.00694481749087572
0.000428132712841034 0.00683483062312007
0.000469547463580966 0.00670943781733513
0.00051496725063771 0.00651964591816068
0.000564780493732542 0.00629085721448064
0.000619412225205451 0.00603849440813065
0.000679328513797373 0.00579702574759722
0.00074504065560177 0.00555308535695076
0.000817110936623067 0.00533322896808386
0.000896150828339159 0.00513986218720675
0.000982836470939219 0.00493483804166317
0.00107790704350919 0.00469073187559843
0.00118217407725751 0.00443237088620663
0.00129652675241232 0.00420287251472473
0.00142194435466081 0.00395722314715385
0.00155949033796787 0.00369771616533399
0.00171034119557589 0.00349131366237998
0.00187578424811363 0.0033529857173562
0.00205723056569695 0.00329784885980189
0.00225622835569084 0.00320978928357363
0.0024744754191488 0.00312263518571854
0.00271383975632489 0.00298594613559544
0.00297635211609304 0.00301071302965283
0.00326425745151937 0.00300576351583004
0.00358001212589443 0.00297086453065276
0.00392631022259593 0.00294348085299134
0.00430610589683056 0.00294111203402281
0.00472265016287565 0.00301172025501728
0.00517947599291801 0.00316039239987731
0.00568049214780331 0.00330589944496751
0.00622997106984258 0.0034443293698132
0.00683260289952159 0.00353885325603187
0.00749352620914578 0.00359923322685063
0.00821840018033981 0.00367025844752789
0.00901337433606386 0.00375269236974418
0.00988524686545134 0.00384349096566439
0.0108414553105831 0.00394602492451668
0.0118901589885354 0.00405628373846412
0.0130403051152825 0.00416943477466702
0.0143017387017608 0.00428209733217955
0.0156851597130299 0.00439251726493239
0.0172023996710777 0.00449840677902102
0.0188664030283689 0.00459858402609825
0.0206913687288761 0.00469188066199422
0.0226928628981113 0.00477807596325874
0.0248879659920931 0.00485674478113651
0.0272954646497965 0.0049284971319139
0.0299357790499926 0.00499624339863658
0.0328314937651157 0.00506384624168277
0.0360073149204254 0.00513836834579706
0.0394903384149075 0.00522773014381528
0.0433102734386921 0.00533651746809483
0.0474998243153095 0.00546417757868767
0.0520945265889168 0.00560606457293034
0.0571336783468723 0.0057538440451026
0.0626602694392204 0.00590258464217186
0.0687214583158493 0.00605320278555155
0.0753689557313919 0.00620608776807785
0.0826596468687057 0.00634133210405707
0.0906553938984871 0.00644782464951277
0.0994245782494545 0.00651253666728735
0.109042011201382 0.00652942014858127
0.119589745998383 0.00650060689076781
0.131157770752907 0.00643951212987304
0.143845111131668 0.00632099993526936
0.157759368419647 0.00614602351561189
0.173019587993622 0.00591466668993235
0.18975593149662 0.0056993761099875
0.208111211657524 0.00564809981733561
0.228241994976997 0.00552268046885729
0.250320613384247 0.00546316616237164
0.274534374475479 0.00551537331193686
0.301090329885483 0.00553902145475149
0.330215096473694 0.00558790657669306
0.362157106399536 0.00589169142767787
0.397188931703568 0.00617851410061121
0.435609370470047 0.0064473575912416
0.477747321128845 0.00670118862763047
0.52396023273468 0.00694469409063458
0.574643433094025 0.00721360091120005
0.630229234695435 0.00752086471766233
0.691191911697388 0.00786723848432302
0.758051514625549 0.00819237343966961
0.831380367279053 0.00845716893672943
0.911800682544708 0.00877487193793058
1 0.00909443385899067
};
\addlegendentry{\tiny Width of Error Band}
\end{axis}

\end{tikzpicture}

%% file: figs/CR_err_band_width.tex
\begin{tikzpicture}

\definecolor{darkgray176}{RGB}{176,176,176}
\definecolor{lightgray204}{RGB}{204,204,204}
\definecolor{steelblue31119180}{RGB}{31,119,180}

\begin{axis}[
legend cell align={left},
legend style={fill opacity=0.8, draw opacity=1,at={(0.98,0.2)}, text opacity=1, draw=lightgray204},
log basis x={10},
log basis y={10},
tick align=outside,
tick pos=left,
title={Error Range of Networks Trained from 100 Random Initializations},
x grid style={darkgray176},
xlabel={Time (s)},
xmin=1.09673091474103e-16, xmax=1,
xmode=log,
xtick style={color=black},
xtick={1e-18,1e-16,1e-14,1e-12,1e-10,1e-08,1e-06,0.0001,0.01,1,100},
xticklabels={
  \(\displaystyle {10^{-18}}\),
  \(\displaystyle {10^{-16}}\),
  \(\displaystyle {10^{-14}}\),
  \(\displaystyle {10^{-12}}\),
  \(\displaystyle {10^{-10}}\),
  \(\displaystyle {10^{-8}}\),
  \(\displaystyle {10^{-6}}\),
  \(\displaystyle {10^{-4}}\),
  \(\displaystyle {10^{-2}}\),
  \(\displaystyle {10^{0}}\),
  \(\displaystyle {10^{2}}\)
},
y grid style={darkgray176},
ylabel={Width of Error Band},
ymin=1.91630082576165e-05, ymax=0.0445867878674018,
ymode=log,
ytick style={color=black},
ytick={0.001,0.002,0.003,0.004,0.005,0.006,0.007,0.008,0.009,0.01,0.02,0.03,0.04,0.05,0.06,0.07,0.08,0.09,0.1,0.2,0.3,0.4,0.5},
yticklabels={
  \(\displaystyle {10^{-3}}\),
  \(\displaystyle {}\),
  \(\displaystyle {}\),
  \(\displaystyle {}\),
  \(\displaystyle {}\),
  \(\displaystyle {}\),
  \(\displaystyle {}\),
  \(\displaystyle {}\),
  \(\displaystyle {}\),
  \(\displaystyle {10^{-2}}\),
  \(\displaystyle {}\),
  \(\displaystyle {}\),
  \(\displaystyle {}\),
  \(\displaystyle {}\),
  \(\displaystyle {}\),
  \(\displaystyle {}\),
  \(\displaystyle {}\),
  \(\displaystyle {}\),
  \(\displaystyle {10^{-1}}\),
  \(\displaystyle {}\),
  \(\displaystyle {}\),
  \(\displaystyle {}\),
  \(\displaystyle {}\),
}
]
\addplot [semithick, steelblue31119180]
table {%
1.09673091474103e-16 4.41276861238293e-05
1.20281868118553e-16 4.41452393715736e-05
1.31916845030936e-16 4.40783333033323e-05
1.44677608139559e-16 4.396611257107e-05
1.58672418842992e-16 4.37596536357887e-05
1.74020947014816e-16 4.34749708801974e-05
1.90854165124498e-16 4.31505068263505e-05
2.0931567172638e-16 4.27650920755696e-05
2.29562948774199e-16 4.22940574935637e-05
2.51769346854569e-16 4.17621704400517e-05
2.76123238154025e-16 4.12079170928337e-05
3.02832913368252e-16 4.1341787436977e-05
3.32126224360062e-16 4.1495863115415e-05
3.64253125258233e-16 4.16503971791826e-05
3.99487657690983e-16 4.17742194258608e-05
4.38131418327092e-16 4.18804920627736e-05
4.80512314796267e-16 4.19916032114998e-05
5.26992771320814e-16 4.21064760303125e-05
5.77969172850268e-16 4.21815457229968e-05
6.33876735500841e-16 4.22869597969111e-05
6.95192232942718e-16 4.23327473981772e-05
7.62440507965842e-16 4.24185782321729e-05
8.36192090199748e-16 4.24597965320572e-05
9.17077807431903e-16 4.25126199843362e-05
1.00578756799785e-15 4.25283215008676e-05
1.10307830171507e-15 4.25353755417746e-05
1.20978010087187e-15 4.25796242780052e-05
1.32680331412833e-15 4.26730257458985e-05
1.45514945206545e-15 4.27305640187114e-05
1.59590737553692e-15 4.27936574851628e-05
1.75028093011903e-15 4.28140119765885e-05
1.91958735792387e-15 4.28668099630158e-05
2.10527074424741e-15 4.28916573582683e-05
2.30892096993163e-15 4.29257815994788e-05
2.53226492339775e-15 4.29468855145387e-05
2.77721329921654e-15 4.29541687481105e-05
3.04585551591063e-15 4.29677056672517e-05
3.34048427991002e-15 4.29653155151755e-05
3.66361231445272e-15 4.29463580076117e-05
4.01799650002381e-15 4.29221581725869e-05
4.40667196743155e-15 4.29004358011298e-05
4.83293282780781e-15 4.28617750003468e-05
5.30042682853993e-15 4.28529201599304e-05
5.81314264777653e-15 4.28180319431704e-05
6.37545309310782e-15 4.27751983806957e-05
6.99215702969647e-15 4.27271334046964e-05
7.66853231983682e-15 4.26750630140305e-05
8.41031634119709e-15 4.2637711885618e-05
9.22385337055222e-15 4.25746184191667e-05
1.01160856899379e-14 4.25324469688348e-05
1.10946238903852e-14 4.24790669057984e-05
1.21678180226882e-14 4.24354948336259e-05
1.33448228189408e-14 4.23772216890939e-05
1.46357128890176e-14 4.23398450948298e-05
1.60514386748608e-14 4.22633675043471e-05
1.76041076654183e-14 4.22281154897064e-05
1.93069691500512e-14 4.21961667598225e-05
2.11745514378707e-14 4.21802942582872e-05
2.32228380405916e-14 4.21704935433809e-05
2.54692049920938e-14 4.21325603383593e-05
2.79328663877528e-14 4.21038421336561e-05
3.06348405031218e-14 4.20904980273917e-05
3.35981751046956e-14 4.20549840782769e-05
3.68481558200157e-14 4.20275500800926e-05
4.04125178959068e-14 4.19809839513618e-05
4.43216562626483e-14 4.19405987486243e-05
4.86090388860522e-14 4.19045209127944e-05
5.33110373608734e-14 4.18559648096561e-05
5.84678654233159e-14 4.18433664890472e-05
6.412351796466e-14 4.1831150156213e-05
7.03262453697132e-14 4.18265117332339e-05
7.71291396636092e-14 4.18051095039118e-05
8.45899142832424e-14 4.17766495957039e-05
9.27723677501997e-14 4.17718001699541e-05
1.01746329460653e-13 4.17474875575863e-05
1.1158834408798e-13 4.17543269577436e-05
1.22382396642966e-13 4.17258197558112e-05
1.34220556218846e-13 4.17049704992678e-05
1.47204175389958e-13 4.17004630435258e-05
1.6144338199197e-13 4.16924813180231e-05
1.77059931928854e-13 4.1645147575764e-05
1.94187100752662e-13 4.16574366681743e-05
2.12970998258663e-13 4.16408256569412e-05
2.33572411629038e-13 4.15974900533911e-05
2.56166100701463e-13 4.15522445109673e-05
2.80945297408131e-13 4.15086433349643e-05
3.08121366962566e-13 4.14661808463279e-05
3.37926247314518e-13 4.1608334868215e-05
3.70614156768381e-13 4.17500232288148e-05
4.06464033438081e-13 4.1937229980249e-05
4.45781703651421e-13 4.21093536715489e-05
4.88903676657687e-13 4.22368975705467e-05
5.36195762269875e-13 4.23609271820169e-05
5.8806247631854e-13 4.2451069020899e-05
6.44946281710274e-13 4.25358884967864e-05
7.07334147850847e-13 4.26459846494254e-05
7.75755219610541e-13 4.27099184889812e-05
8.50794803532173e-13 4.27500235673506e-05
9.33093012578379e-13 4.27515406045131e-05
1.02335208449628e-12 4.27802187914494e-05
1.12234180282833e-12 4.27749000664335e-05
1.23090686948713e-12 4.27081104135141e-05
1.34997373522905e-12 4.27417835453525e-05
1.48055786380885e-12 4.26640835939907e-05
1.62377739292702e-12 4.25776634074282e-05
1.78084684585716e-12 4.24650716013275e-05
1.95310998277187e-12 4.23601231887005e-05
2.14204066810442e-12 4.22464363509789e-05
2.34924232844769e-12 4.21365912188776e-05
2.57648676699196e-12 4.2014020436909e-05
2.82571253722164e-12 4.19316493207589e-05
3.09904641011827e-12 4.1798128222581e-05
3.39882028771443e-12 4.16285474784672e-05
3.72759158609459e-12 4.14606220147107e-05
4.0881647025981e-12 4.12992885685526e-05
4.48361703667133e-12 4.11194414482452e-05
4.91732132443246e-12 4.08465311920736e-05
5.39299030780094e-12 4.05786122428253e-05
5.91465938726277e-12 4.03204940084834e-05
6.48679027159815e-12 4.00793760491069e-05
7.11427878413717e-12 3.97446419810876e-05
7.80244965858934e-12 3.95758215745445e-05
8.5571879443469e-12 3.91646426578518e-05
9.38493380231442e-12 3.88291882700287e-05
1.02927458223157e-11 3.8516656786669e-05
1.12883738814795e-11 3.814370938926e-05
1.23803077164975e-11 3.79383382096421e-05
1.35778671292441e-11 3.75486815755721e-05
1.48912670389079e-11 3.73674265574664e-05
1.63317501883364e-11 3.67720749636646e-05
1.79115351023329e-11 3.63236758857965e-05
1.96441352767751e-11 3.60823250957765e-05
2.15443791279357e-11 3.57453136530239e-05
2.3628387441077e-11 3.53881405317225e-05
2.59139827651911e-11 3.479489532765e-05
2.84206651268715e-11 3.45240368915256e-05
3.11698236665769e-11 3.41088816639967e-05
3.41849118457027e-11 3.3686839742586e-05
3.74916521439506e-11 3.32059171341825e-05
4.11182476955929e-11 3.27806701534428e-05
4.50956563757821e-11 3.24182692565955e-05
4.94578059062611e-11 3.22696723742411e-05
5.42420240667862e-11 3.19471182592679e-05
5.94889137950361e-11 3.15965662593953e-05
6.52433246317585e-11 3.133489008178e-05
7.15545331320122e-11 3.10938412440009e-05
7.84760728622658e-11 3.08830167341512e-05
8.60671325875195e-11 3.06673646264244e-05
9.43924868823665e-11 3.04567456623772e-05
1.03523162264807e-10 3.02420030493522e-05
1.13537054757806e-10 2.98771283269161e-05
1.24519602962003e-10 2.97792194032809e-05
1.36564509700676e-10 2.94288256554864e-05
1.49774512236966e-10 2.91852338705212e-05
1.6426270066372e-10 2.88380433630664e-05
1.80151976669762e-10 2.86507074633846e-05
1.97578273186672e-10 2.8324022423476e-05
2.16690665411079e-10 2.79202249657828e-05
2.3765137080467e-10 2.76390728686238e-05
2.60639620996983e-10 2.72582165052881e-05
2.85851509129742e-10 2.77057224593591e-05
3.13502224180695e-10 2.85226851701736e-05
3.43827577520273e-10 2.95534737233538e-05
3.77086334379939e-10 3.06782858388033e-05
4.13562267764789e-10 3.19015307468362e-05
4.53566517677473e-10 3.32539966620971e-05
4.97440422186912e-10 3.46199376508594e-05
5.45559486475611e-10 3.60577250830829e-05
5.98332050572026e-10 3.7515663279919e-05
6.56209253602213e-10 3.893043322023e-05
7.19686532590913e-10 4.04008533223532e-05
7.89302512238521e-10 4.18321287725121e-05
8.65652494130842e-10 4.36985355918296e-05
9.49387790605272e-10 4.65737903141417e-05
1.04122299671161e-09 4.94410414830782e-05
1.14194165234949e-09 5.23092239745893e-05
1.25240273707306e-09 5.50516524526756e-05
1.37354871920792e-09 5.76782222196925e-05
1.50641343843461e-09 5.99195809627417e-05
1.65213387415264e-09 6.89594162395224e-05
1.81194625969994e-09 8.3291255577933e-05
1.98721772370902e-09 0.000103813814348541
2.17944773339696e-09 0.000133394423755817
2.39026776149842e-09 0.000172617161297239
2.62148081020541e-09 0.000222223374294117
2.87505907969887e-09 0.00028144646785222
3.15316617260919e-09 0.000348602770827711
3.45817485758459e-09 0.000418971932958812
3.79268749739481e-09 0.00048773197340779
4.15955758725772e-09 0.000550377415493131
4.56191573405818e-09 0.000604088476393372
5.00320540552934e-09 0.000647258420940489
5.48716982962105e-09 0.000679623859468848
6.01794880594753e-09 0.000701290497090667
6.6000702680924e-09 0.000712146167643368
7.23851734107939e-09 0.000711203087121248
7.93870658100104e-09 0.000699194148182869
8.70662475449535e-09 0.000678223615977913
9.54882484194286e-09 0.000650844303891063
1.04724913185805e-08 0.000621134182438254
1.14855049915263e-08 0.000595593359321356
1.25965096131608e-08 0.000581128464546055
1.38150122452885e-08 0.000574425735976547
1.5151352172893e-08 0.000578681298065931
1.66169566995222e-08 0.000590682378970087
1.82243304891472e-08 0.000608098634984344
1.9987187016568e-08 0.000629417540039867
2.19205666951439e-08 0.000653828610666096
2.40409629981286e-08 0.000679965771269053
2.63665267397073e-08 0.000704652746208012
2.89169861389382e-08 0.000724912970326841
3.17141548578093e-08 0.00074098922777921
3.47818946977441e-08 0.000753897649701685
3.81463785004144e-08 0.000791576225310564
4.18363121923448e-08 0.000833487662021071
4.58832829508538e-08 0.000878634746186435
5.0321613542792e-08 0.000930923735722899
5.51892718192448e-08 0.000979732489213347
6.05277818976901e-08 0.000993235036730766
6.63826895674902e-08 0.00110370805487037
7.28039495356825e-08 0.00136741122696549
7.98465151774508e-08 0.00174633774440736
8.75701502422999e-08 0.00230900291353464
9.60408925720913e-08 0.00284122861921787
1.05331025679334e-07 0.00374443782493472
1.1551978928992e-07 0.0047496035695076
1.26694132518423e-07 0.00533786928281188
1.38949673100797e-07 0.00746920332312584
1.52390427388127e-07 0.0119906738400459
1.67131275929933e-07 0.0178838726133108
1.83298055844716e-07 0.0228163357824087
2.01028640844925e-07 0.0265520885586739
2.20474348111566e-07 0.0300430785864592
2.41801018319165e-07 0.0313453003764153
2.65191260950814e-07 0.0305236708372831
2.90843473749192e-07 0.0266958102583885
3.18977015467681e-07 0.0268485154956579
3.49831935864131e-07 0.025467811152339
3.83671533654706e-07 0.0198396183550358
4.20784431298671e-07 0.0212751142680645
4.61488326664039e-07 0.0156296081840992
5.06128515098681e-07 0.0136351138353348
5.55086785425374e-07 0.0127361426129937
6.08780965194455e-07 0.0102360192686319
6.67668814458011e-07 0.00714317057281733
7.32253113255865e-07 0.00760105764493346
8.03086379619344e-07 0.00703639211133122
8.80769675859483e-07 0.00567347044125199
9.65967274169088e-07 0.00377506157383323
1.05940625871881e-06 0.00365865090861917
1.16188368792791e-06 0.00406295899301767
1.27427381357847e-06 0.00437928596511483
1.39753854000446e-06 0.00467769522219896
1.53272378611291e-06 0.00492402352392673
1.68098563335661e-06 0.00510090729221702
1.84358884780522e-06 0.00518445624038577
2.02192109099997e-06 0.00516766216605902
2.21750315176905e-06 0.0051003722473979
2.43200452132442e-06 0.00540416594594717
2.66726055997424e-06 0.00576669396832585
2.92526715384156e-06 0.00599369173869491
3.20823119182023e-06 0.00605053640902042
3.51856601810141e-06 0.0059581222012639
3.85892053600401e-06 0.00598953431472182
4.23219717049506e-06 0.00664378562942147
4.64159211333026e-06 0.00708753662183881
5.09057781528099e-06 0.00733523583039641
5.58299416297814e-06 0.00743259489536285
6.12304302194389e-06 0.00731839379295707
6.71533007334801e-06 0.00761328730732203
7.36491028874298e-06 0.00846522301435471
8.07734340924071e-06 0.00926876533776522
8.8586712081451e-06 0.0100697176530957
9.71557983575622e-06 0.0105415359139442
1.06553761725081e-05 0.0106110479682684
1.16860819616704e-05 0.0103330686688423
1.28164874695358e-05 0.0102095957845449
1.40562688102364e-05 0.0113654639571905
1.54159461089876e-05 0.0122800869867206
1.69071445270674e-05 0.0124790705740452
1.85425888048485e-05 0.0120357237756252
2.03362305910559e-05 0.011477586813271
2.23033730435418e-05 0.0114791449159384
2.44607999775326e-05 0.0112474802881479
2.68269723164849e-05 0.0106570459902287
2.94219735224033e-05 0.00968101434409618
3.22679879900534e-05 0.0090955588966608
3.53893046849407e-05 0.00832868833094835
3.88125408790074e-05 0.00756073230877519
4.25669131800532e-05 0.00677685579285026
4.66845558548812e-05 0.00592984957620502
5.12003971380182e-05 0.00520552368834615
5.61530578124803e-05 0.00469727208837867
6.15848111920059e-05 0.0049249310977757
6.75419578328729e-05 0.00501072593033314
7.40753530408256e-05 0.00499437749385834
8.12409052741714e-05 0.00487409206107259
8.9099419710692e-05 0.00483178626745939
9.77180898189545e-05 0.00494827004149556
0.000107170446426608 0.00519323348999023
0.000117537150799762 0.00532206054776907
0.000128906642203219 0.00534327980130911
0.000141376207466237 0.00541031546890736
0.0001550516608404 0.00547915045171976
0.000170049941516481 0.00551944505423307
0.000186499033588916 0.00538397766649723
0.000204539261176251 0.00525210425257683
0.000224324539885856 0.00535980984568596
0.000246023701038212 0.00570973195135593
0.00026982236886397 0.00621145498007536
0.000295922538498417 0.00667407177388668
0.000324547407217324 0.00733683910220861
0.000355941185262054 0.00804840587079525
0.000390371715184301 0.00872577074915171
0.000428132712841034 0.00930394046008587
0.000469547463580966 0.00975603610277176
0.00051496725063771 0.0101920571178198
0.000564780493732542 0.0106155732646585
0.000619412225205451 0.0112624196335673
0.000679328513797373 0.0118843279778957
0.00074504065560177 0.0123286107555032
0.000817110936623067 0.0126170637086034
0.000896150828339159 0.0129941375926137
0.000982836470939219 0.0132592488080263
0.00107790704350919 0.0133663453161716
0.00118217407725751 0.0133899189531803
0.00129652675241232 0.0136453900486231
0.00142194435466081 0.0137937013059855
0.00155949033796787 0.0136694815009832
0.00171034119557589 0.013238400220871
0.00187578424811363 0.0131935868412256
0.00205723056569695 0.0138222835958004
0.00225622835569084 0.0144809447228909
0.0024744754191488 0.0150398388504982
0.00271383975632489 0.0158579275012016
0.00297635211609304 0.0166022628545761
0.00326425745151937 0.0173454266041517
0.00358001212589443 0.0181223005056381
0.00392631022259593 0.0189011041074991
0.00430610589683056 0.01965606585145
0.00472265016287565 0.0203189700841904
0.00517947599291801 0.0208531748503447
0.00568049214780331 0.0212452746927738
0.00622997106984258 0.0214671902358532
0.00683260289952159 0.0214977785944939
0.00749352620914578 0.0218547508120537
0.00821840018033981 0.0223338454961777
0.00901337433606386 0.0227806400507689
0.00988524686545134 0.0231899973005056
0.0108414553105831 0.0235571209341288
0.0118901589885354 0.0238246005028486
0.0130403051152825 0.0239434987306595
0.0143017387017608 0.023970253765583
0.0156851597130299 0.0239377804100513
0.0172023996710777 0.0238319039344788
0.0188664030283689 0.0236664470285177
0.0206913687288761 0.0235132891684771
0.0226928628981113 0.0234055928885937
0.0248879659920931 0.0233229584991932
0.0272954646497965 0.0232515186071396
0.0299357790499926 0.0231739822775126
0.0328314937651157 0.0230660159140825
0.0360073149204254 0.0228860229253769
0.0394903384149075 0.0232084207236767
0.0433102734386921 0.0236317720264196
0.0474998243153095 0.0240379981696606
0.0520945265889168 0.0244257561862469
0.0571336783468723 0.0247926786541939
0.0626602694392204 0.0251359287649393
0.0687214583158493 0.0254546254873276
0.0753689557313919 0.0257475730031729
0.0826596468687057 0.0260134562849998
0.0906553938984871 0.0262504406273365
0.0994245782494545 0.0264563821256161
0.109042011201382 0.0266314186155796
0.119589745998383 0.026994938030839
0.131157770752907 0.0274085737764835
0.143845111131668 0.0277774445712566
0.157759368419647 0.0280926395207644
0.173019587993622 0.0283441245555878
0.18975593149662 0.0285339243710041
0.208111211657524 0.0286579020321369
0.228241994976997 0.0287195201963186
0.250320613384247 0.0287306383252144
0.274534374475479 0.0286968983709812
0.301090329885483 0.0286207348108292
0.330215096473694 0.0285033416002989
0.362157106399536 0.0283442307263613
0.397188931703568 0.0281414836645126
0.435609370470047 0.0278936121612787
0.477747321128845 0.0276071969419718
0.52396023273468 0.0273006688803434
0.574643433094025 0.0269743371754885
0.630229234695435 0.0266538374125957
0.691191911697388 0.0263401083648205
0.758051514625549 0.0260181240737438
0.831380367279053 0.0256842989474535
0.911800682544708 0.0255830585956573
1 0.0257035605609417
};
\addlegendentry{\tiny Width of Error Band}
\end{axis}

\end{tikzpicture}

%% file: figs/ac_rand_err.tex
\begin{tikzpicture}

\definecolor{darkgray176}{RGB}{176,176,176}
\definecolor{lightgray204}{RGB}{204,204,204}
\definecolor{steelblue31119180}{RGB}{31,119,180}

\begin{axis}[
legend cell align={left},
legend style={
  fill opacity=0.8,
  draw opacity=1,
  text opacity=1,
  at={(0.03,0.97)},
  anchor=north west,
  draw=lightgray204
},
log basis y={10},
tick align=outside,
tick pos=left,
title={Error Range of Networks Trained from 100 Random Initializations},
x grid style={darkgray176},
xlabel={Time (s)},
xmin=0.1, xmax=10,
xtick style={color=black},
y grid style={darkgray176},
ylabel={Relative L2-Norm Error},
ymin=0.00577651067981189, ymax=0.0860214825668982,
ymode=log,
ytick style={color=black},
ytick={0.001,0.002,0.003,0.004,0.005,0.006,0.007,0.008,0.009,0.01,0.02,0.03,0.04,0.05,0.06,0.07,0.08,0.09,0.1,0.2,0.3,0.4,0.5},
yticklabels={
  \(\displaystyle {10^{-3}}\),
  \(\displaystyle {}\),
  \(\displaystyle {}\),
  \(\displaystyle {}\),
  \(\displaystyle {}\),
  \(\displaystyle {}\),
  \(\displaystyle {}\),
  \(\displaystyle {}\),
  \(\displaystyle {}\),
  \(\displaystyle {10^{-2}}\),
  \(\displaystyle {}\),
  \(\displaystyle {}\),
  \(\displaystyle {}\),
  \(\displaystyle {}\),
  \(\displaystyle {}\),
  \(\displaystyle {}\),
  \(\displaystyle {}\),
  \(\displaystyle {}\),
  \(\displaystyle {10^{-1}}\),
  \(\displaystyle {}\),
  \(\displaystyle {}\),
  \(\displaystyle {}\),
  \(\displaystyle {}\),
}
]
\path [fill=steelblue31119180, fill opacity=0.2]
(axis cs:0.1,0.0140104128944052)
--(axis cs:0.1,0.0103976894923605)
--(axis cs:0.2,0.00966981243424112)
--(axis cs:0.3,0.00903532827582287)
--(axis cs:0.4,0.00850835486744291)
--(axis cs:0.5,0.00807037237667045)
--(axis cs:0.6,0.00771186810721671)
--(axis cs:0.7,0.00742862334719387)
--(axis cs:0.8,0.00713890257213179)
--(axis cs:0.9,0.00691051235607854)
--(axis cs:1,0.00674092537753696)
--(axis cs:1.1,0.00662487498583887)
--(axis cs:1.2,0.00655677558594954)
--(axis cs:1.3,0.00653102124045442)
--(axis cs:1.4,0.00654226334582126)
--(axis cs:1.5,0.00658575649363163)
--(axis cs:1.6,0.00665753398445862)
--(axis cs:1.7,0.00675423972067853)
--(axis cs:1.8,0.00687268797951016)
--(axis cs:1.9,0.00700934302066712)
--(axis cs:2,0.00715989759250413)
--(axis cs:2.1,0.00731911285624171)
--(axis cs:2.2,0.00748104444757685)
--(axis cs:2.3,0.007639724773706)
--(axis cs:2.4,0.00779024774477227)
--(axis cs:2.5,0.00792573580446513)
--(axis cs:2.6,0.00802841276787282)
--(axis cs:2.7,0.00812858462995266)
--(axis cs:2.8,0.00823094036370227)
--(axis cs:2.9,0.00834135190408687)
--(axis cs:3,0.00846628609050648)
--(axis cs:3.1,0.00861224363942403)
--(axis cs:3.2,0.00878518770871639)
--(axis cs:3.3,0.00898815078532287)
--(axis cs:3.4,0.00918324254634704)
--(axis cs:3.5,0.00940570157985184)
--(axis cs:3.6,0.00965644702989708)
--(axis cs:3.7,0.00993660630419915)
--(axis cs:3.8,0.0102478762386854)
--(axis cs:3.9,0.010592619367875)
--(axis cs:4,0.0109736980487248)
--(axis cs:4.1,0.0113942053050737)
--(axis cs:4.2,0.0118571704292838)
--(axis cs:4.3,0.0123651083174785)
--(axis cs:4.4,0.0129194068270511)
--(axis cs:4.5,0.0135200101594766)
--(axis cs:4.6,0.0141657745936361)
--(axis cs:4.7,0.0148552276974062)
--(axis cs:4.8,0.015587213146162)
--(axis cs:4.9,0.0163610908833873)
--(axis cs:5,0.0171764816123942)
--(axis cs:5.1,0.0180328376404136)
--(axis cs:5.2,0.0189292134691917)
--(axis cs:5.3,0.019864406655382)
--(axis cs:5.4,0.0208355009082092)
--(axis cs:5.5,0.0218232550745526)
--(axis cs:5.6,0.022857451400195)
--(axis cs:5.7,0.0239329076542319)
--(axis cs:5.8,0.0250439085158797)
--(axis cs:5.9,0.0261845848718235)
--(axis cs:6,0.0273491743831749)
--(axis cs:6.1,0.0285322053115827)
--(axis cs:6.2,0.0297286217675122)
--(axis cs:6.3,0.0309337789795146)
--(axis cs:6.4,0.0321432316775206)
--(axis cs:6.5,0.0333524350551858)
--(axis cs:6.6,0.0345566122888927)
--(axis cs:6.7,0.0357508556762256)
--(axis cs:6.8,0.0369303173521379)
--(axis cs:6.9,0.0380595148861323)
--(axis cs:7,0.0391452019220625)
--(axis cs:7.1,0.0401964545437811)
--(axis cs:7.2,0.0412135720564887)
--(axis cs:7.3,0.0421980293720201)
--(axis cs:7.4,0.0431496302418489)
--(axis cs:7.5,0.0440662624310932)
--(axis cs:7.6,0.044946020691479)
--(axis cs:7.7,0.0457880314849333)
--(axis cs:7.8,0.0465924891579165)
--(axis cs:7.9,0.0473605665996011)
--(axis cs:8,0.0480941201394169)
--(axis cs:8.1,0.0487949605962069)
--(axis cs:8.2,0.0494640583511384)
--(axis cs:8.3,0.0501016087988086)
--(axis cs:8.4,0.0507077459005586)
--(axis cs:8.5,0.0512830814661521)
--(axis cs:8.6,0.0518288973584502)
--(axis cs:8.7,0.0523471589732086)
--(axis cs:8.8,0.0528405442033781)
--(axis cs:8.9,0.0533126098974734)
--(axis cs:9,0.0537680785117619)
--(axis cs:9.1,0.0542130660921199)
--(axis cs:9.2,0.0546549171652945)
--(axis cs:9.3,0.0551012917407721)
--(axis cs:9.4,0.0555582291035099)
--(axis cs:9.5,0.0560255610349919)
--(axis cs:9.6,0.0564889459323434)
--(axis cs:9.7,0.0569247756229741)
--(axis cs:9.8,0.0572911253733689)
--(axis cs:9.9,0.0574514106374623)
--(axis cs:10,0.0576120162033538)
--(axis cs:10,0.0760836620256292)
--(axis cs:10,0.0760836620256292)
--(axis cs:9.9,0.0758740105467022)
--(axis cs:9.8,0.0756329185293105)
--(axis cs:9.7,0.075358324430154)
--(axis cs:9.6,0.0750514912235895)
--(axis cs:9.5,0.0747177737578964)
--(axis cs:9.4,0.0743633735252419)
--(axis cs:9.3,0.0739909603935362)
--(axis cs:9.2,0.0735984554273215)
--(axis cs:9.1,0.0731807615790559)
--(axis cs:9,0.072731942637794)
--(axis cs:8.9,0.0722467584826819)
--(axis cs:8.8,0.0717215263305123)
--(axis cs:8.7,0.0711542485846677)
--(axis cs:8.6,0.0705439920057703)
--(axis cs:8.5,0.0698899298502944)
--(axis cs:8.4,0.0691906422620868)
--(axis cs:8.3,0.0684439389793774)
--(axis cs:8.2,0.0676471233459742)
--(axis cs:8.1,0.0667974919000013)
--(axis cs:8,0.0658927133300216)
--(axis cs:7.9,0.0649307542604418)
--(axis cs:7.8,0.0639099904284846)
--(axis cs:7.7,0.0628297605611156)
--(axis cs:7.6,0.0616889359933218)
--(axis cs:7.5,0.0604851506054501)
--(axis cs:7.4,0.0593147179909702)
--(axis cs:7.3,0.0581382163513596)
--(axis cs:7.2,0.0568887938602752)
--(axis cs:7.1,0.0555676068095054)
--(axis cs:7,0.0541774392267933)
--(axis cs:6.9,0.0527219791215837)
--(axis cs:6.8,0.0512050365304044)
--(axis cs:6.7,0.0496304920593295)
--(axis cs:6.6,0.0480029837357819)
--(axis cs:6.5,0.046328266863312)
--(axis cs:6.4,0.0446131863254434)
--(axis cs:6.3,0.0428656559078992)
--(axis cs:6.2,0.0410945439506013)
--(axis cs:6.1,0.0393092206663351)
--(axis cs:6,0.0375192318465336)
--(axis cs:5.9,0.0357347277741415)
--(axis cs:5.8,0.0339671177217614)
--(axis cs:5.7,0.0322292501189124)
--(axis cs:5.6,0.0305348510528113)
--(axis cs:5.5,0.0288969024686361)
--(axis cs:5.4,0.0273256319196038)
--(axis cs:5.3,0.0258278931262735)
--(axis cs:5.2,0.0244076232903529)
--(axis cs:5.1,0.0230819741278601)
--(axis cs:5,0.0218620336374263)
--(axis cs:4.9,0.0207008369612815)
--(axis cs:4.8,0.0195998610474784)
--(axis cs:4.7,0.0185606791259982)
--(axis cs:4.6,0.017584056444613)
--(axis cs:4.5,0.0166725223537618)
--(axis cs:4.4,0.0158557086919113)
--(axis cs:4.3,0.0151509227379041)
--(axis cs:4.2,0.0145547554374799)
--(axis cs:4.1,0.0140056291150251)
--(axis cs:4,0.0135018981394355)
--(axis cs:3.9,0.0130414717148238)
--(axis cs:3.8,0.0126216217309416)
--(axis cs:3.7,0.0122391557938277)
--(axis cs:3.6,0.0118908432143227)
--(axis cs:3.5,0.0115737427279219)
--(axis cs:3.4,0.0112852602395824)
--(axis cs:3.3,0.0110230319670229)
--(axis cs:3.2,0.0107847726015068)
--(axis cs:3.1,0.0105680851464664)
--(axis cs:3,0.0103701596162866)
--(axis cs:2.9,0.0101874422506394)
--(axis cs:2.8,0.0100155096748696)
--(axis cs:2.7,0.00984930829387542)
--(axis cs:2.6,0.00968372709366653)
--(axis cs:2.5,0.00951438306036931)
--(axis cs:2.4,0.00934272667283837)
--(axis cs:2.3,0.00917470039313442)
--(axis cs:2.2,0.0090051004215801)
--(axis cs:2.1,0.00884001600693535)
--(axis cs:2,0.00875401090680373)
--(axis cs:1.9,0.00869468822859321)
--(axis cs:1.8,0.00864940844293748)
--(axis cs:1.7,0.00862026660864543)
--(axis cs:1.6,0.00860896631389771)
--(axis cs:1.5,0.0086171307266751)
--(axis cs:1.4,0.00864680046929174)
--(axis cs:1.3,0.00870085592613868)
--(axis cs:1.2,0.00881110110158518)
--(axis cs:1.1,0.00901782941493705)
--(axis cs:1,0.00925943267836075)
--(axis cs:0.9,0.00953832312499017)
--(axis cs:0.8,0.00985759554361529)
--(axis cs:0.7,0.0102200043020323)
--(axis cs:0.6,0.0106266926432706)
--(axis cs:0.5,0.0110776401481456)
--(axis cs:0.4,0.0115750983218457)
--(axis cs:0.3,0.0121450601144103)
--(axis cs:0.2,0.0129979988874684)
--(axis cs:0.1,0.0140104128944052)
--cycle;
\addlegendimage{area legend, fill=steelblue31119180, fill opacity=0.2}
\addlegendentry{Min/Max Error Range}

\addplot [semithick, steelblue31119180]
table {%
0.1 0.0118267913973069
0.2 0.0110265436319726
0.3 0.0103586271994095
0.4 0.00978944832646171
0.5 0.00929863862512417
0.6 0.0088765048466688
0.7 0.00851980711740193
0.8 0.00822699906399783
0.9 0.00799479425004289
1 0.00781728257292028
1.1 0.00768703019535583
1.2 0.00759671356869136
1.3 0.00754031972795206
1.4 0.00751369160712504
1.5 0.00751453040458787
1.6 0.00754197711872403
1.7 0.00759586521084868
1.8 0.00767578656873029
1.9 0.00778020314411296
2 0.0079058835925946
2.1 0.00804788616077675
2.2 0.00820011304155466
2.3 0.00835624265470866
2.4 0.00851075417231791
2.5 0.00865979172533325
2.6 0.0088017183463224
2.7 0.00893730099525652
2.8 0.00906951989070431
2.9 0.00920305091757662
3 0.00934354221435807
3.1 0.00949686104208931
3.2 0.0096684864382461
3.3 0.00986315338599927
3.4 0.0100847536786434
3.5 0.0103364263713358
3.6 0.0106207430325788
3.7 0.0109399003809174
3.8 0.0112958745368494
3.9 0.0116905461422311
4 0.0121258002875585
4.1 0.0126036118628428
4.2 0.0131261615485906
4.3 0.0136959511908505
4.4 0.0143158369858573
4.5 0.0149889372391955
4.6 0.0157184251363077
4.7 0.016507281493603
4.8 0.0173580782133051
4.9 0.0182727742254255
5 0.0192525633194506
5.1 0.0202977834046447
5.2 0.0214077355166184
5.3 0.0225805134628712
5.4 0.0238129609476108
5.5 0.0251007887225019
5.6 0.0264387503854642
5.7 0.0278207303480213
5.8 0.0292397697321839
5.9 0.0306881732545045
6 0.0321577868675153
6.1 0.0336403326154675
6.2 0.0351275726529018
6.3 0.0366114354289843
6.4 0.0380842498028805
6.5 0.0395389735622229
6.6 0.0409693763228641
6.7 0.0423701407412322
6.8 0.0437368239327468
6.9 0.0450658257056666
7 0.0463544026628426
7.1 0.0476005963387656
7.2 0.0488030655939136
7.3 0.0499608645477184
7.4 0.0510732608323421
7.5 0.0521396766743202
7.6 0.0531597526720489
7.7 0.0541334471253709
7.8 0.0550610097409578
7.9 0.055942858742225
8 0.0567795687033693
8.1 0.0575718726676034
8.2 0.0583206672444007
8.3 0.0590272113326164
8.4 0.059693222452331
8.5 0.0603207787197968
8.6 0.0609121510716283
8.7 0.0614696343286807
8.8 0.0619954758664164
8.9 0.0624918470462127
9 0.062960829400078
9.1 0.0634044842371809
9.2 0.0638247831090488
9.3 0.0642234933312096
9.4 0.0646019922621487
9.5 0.064961375069161
9.6 0.0653026109637144
9.7 0.0656262289352252
9.8 0.065932379847439
9.9 0.0662208490356613
10 0.066491350976315
};
\addlegendentry{Mean Error}
\end{axis}

\end{tikzpicture}

%% file: figs/ac_err_band_width.tex
\begin{tikzpicture}

\definecolor{darkgray176}{RGB}{176,176,176}
\definecolor{lightgray204}{RGB}{204,204,204}
\definecolor{steelblue31119180}{RGB}{31,119,180}

\begin{axis}[
legend cell align={left},
legend style={
  fill opacity=0.8,
  draw opacity=1,
  text opacity=1,
  at={(0.03,0.97)},
  anchor=north west,
  draw=lightgray204
},
log basis y={10},
tick align=outside,
tick pos=left,
title={Error Range of Networks Trained from 100 Random Initializations},
x grid style={darkgray176},
xlabel={Time (s)},
xmin=0.1, xmax=10,
xtick style={color=black},
y grid style={darkgray176},
ylabel={Width of Error Band},
ymin=0.00134062066369237, ymax=0.021518412037619,
ymode=log,
ytick style={color=black},
ytick={0.001,0.002,0.003,0.004,0.005,0.006,0.007,0.008,0.009,0.01,0.02,0.03,0.04,0.05,0.06,0.07,0.08,0.09,0.1,0.2,0.3,0.4,0.5},
yticklabels={
  \(\displaystyle {10^{-3}}\),
  \(\displaystyle {}\),
  \(\displaystyle {}\),
  \(\displaystyle {}\),
  \(\displaystyle {}\),
  \(\displaystyle {}\),
  \(\displaystyle {}\),
  \(\displaystyle {}\),
  \(\displaystyle {}\),
  \(\displaystyle {10^{-2}}\),
  \(\displaystyle {}\),
  \(\displaystyle {}\),
  \(\displaystyle {}\),
  \(\displaystyle {}\),
  \(\displaystyle {}\),
  \(\displaystyle {}\),
  \(\displaystyle {}\),
  \(\displaystyle {}\),
  \(\displaystyle {10^{-1}}\),
  \(\displaystyle {}\),
  \(\displaystyle {}\),
  \(\displaystyle {}\),
  \(\displaystyle {}\),
}
]
\addplot [semithick, steelblue31119180]
table {%
0.1 0.00361272340204471
0.2 0.0033281864532273
0.3 0.00310973183858739
0.4 0.00306674345440277
0.5 0.00300726777147514
0.6 0.00291482453605393
0.7 0.00279138095483842
0.8 0.0027186929714835
0.9 0.00262781076891163
1 0.00251850730082379
1.1 0.00239295442909818
1.2 0.00225432551563564
1.3 0.00216983468568426
1.4 0.00210453712347048
1.5 0.00203137423304347
1.6 0.00195143232943909
1.7 0.00186602688796689
1.8 0.00177672046342732
1.9 0.00168534520792609
2 0.0015941133142996
2.1 0.00152090315069364
2.2 0.00152405597400325
2.3 0.00153497561942841
2.4 0.00155247892806609
2.5 0.00158864725590417
2.6 0.00165531432579371
2.7 0.00172072366392276
2.8 0.00178456931116733
2.9 0.00184609034655255
3 0.00190387352578008
3.1 0.00195584150704235
3.2 0.00199958489279044
3.3 0.00203488118169998
3.4 0.00210201769323539
3.5 0.00216804114807006
3.6 0.00223439618442567
3.7 0.00230254948962853
3.8 0.00237374549225615
3.9 0.00244885234694888
4 0.00252820009071069
4.1 0.00261142380995142
4.2 0.00269758500819615
4.3 0.00278581442042561
4.4 0.00293630186486016
4.5 0.00315251219428521
4.6 0.00341828185097691
4.7 0.00370545142859202
4.8 0.00401264790131635
4.9 0.00433974607789422
5 0.0046855520250321
5.1 0.00504913648744643
5.2 0.00547840982116114
5.3 0.0059634864708915
5.4 0.00649013101139463
5.5 0.00707364739408353
5.6 0.00767739965261632
5.7 0.00829634246468042
5.8 0.00892320920588164
5.9 0.00955014290231803
6 0.0101700574633587
6.1 0.0107770153547524
6.2 0.0113659221830891
6.3 0.0119318769283846
6.4 0.0124699546479228
6.5 0.0129758318081262
6.6 0.0134463714468891
6.7 0.0138796363831039
6.8 0.0142747191782665
6.9 0.0146624642354514
7 0.0150322373047308
7.1 0.0153711522657243
7.2 0.0156752218037865
7.3 0.0159401869793396
7.4 0.0161650877491213
7.5 0.0164188881743569
7.6 0.0167429153018428
7.7 0.0170417290761824
7.8 0.0173175012705681
7.9 0.0175701876608407
8 0.0177985931906047
8.1 0.0180025313037944
8.2 0.0181830649948358
8.3 0.0183423301805689
8.4 0.0184828963615281
8.5 0.0186068483841423
8.6 0.0187150946473202
8.7 0.0188070896114591
8.8 0.0188809821271342
8.9 0.0189341485852084
9 0.0189638641260322
9.1 0.018967695486936
9.2 0.018943538262027
9.3 0.0188896686527641
9.4 0.0188051444217321
9.5 0.0186922127229046
9.6 0.0185625452912461
9.7 0.01843354880718
9.8 0.0183417931559416
9.9 0.0184225999092399
10 0.0184716458222754
};
\addlegendentry{Width of Error Band}
\end{axis}

\end{tikzpicture}

%% file: figs/ch_rand_err.tex
\begin{tikzpicture}

\definecolor{darkgray176}{RGB}{176,176,176}
\definecolor{lightgray204}{RGB}{204,204,204}
\definecolor{steelblue31119180}{RGB}{31,119,180}

\begin{axis}[
legend cell align={left},
legend style={fill opacity=0.8, draw opacity=1,at={(0.98,0.2)}, text opacity=1, draw=lightgray204},
log basis y={10},
minor ytick={0.002,0.003,0.004,0.005,0.006,0.007,0.008,0.009,0.02,0.03,0.04,0.05,0.06,0.07,0.08,0.09,0.2,0.3,0.4,0.5,0.6,0.7,0.8,0.9,2,3,4,5,6,7,8,9,20,30,40,50,60,70,80,90},
tick align=outside,
tick pos=left,
title={Error Range of Networks Trained from 100 Random Initializations},
x grid style={darkgray176},
xlabel={Time (s)},
xmin=0.025, xmax=10,
xtick style={color=black},
y grid style={darkgray176},
ylabel={Relative L2-Norm Error},
ymin=0.0248957075117089, ymax=0.114278167912041,
ymode=log,
ytick style={color=black},
ytick={0.001,0.002,0.003,0.004,0.005,0.006,0.007,0.008,0.009,0.01,0.02,0.03,0.04,0.05,0.06,0.07,0.08,0.09,0.1,0.2,0.3,0.4,0.5},
yticklabels={
  \(\displaystyle {10^{-3}}\),
  \(\displaystyle {}\),
  \(\displaystyle {}\),
  \(\displaystyle {}\),
  \(\displaystyle {}\),
  \(\displaystyle {}\),
  \(\displaystyle {}\),
  \(\displaystyle {}\),
  \(\displaystyle {}\),
  \(\displaystyle {10^{-2}}\),
  \(\displaystyle {}\),
  \(\displaystyle {}\),
  \(\displaystyle {}\),
  \(\displaystyle {}\),
  \(\displaystyle {}\),
  \(\displaystyle {}\),
  \(\displaystyle {}\),
  \(\displaystyle {}\),
  \(\displaystyle {10^{-1}}\),
  \(\displaystyle {}\),
  \(\displaystyle {}\),
  \(\displaystyle {}\),
  \(\displaystyle {}\),
}
]
\path [fill=steelblue31119180, fill opacity=0.2]
(axis cs:0.025,0.0796453014798281)
--(axis cs:0.025,0.0600197731885495)
--(axis cs:0.05,0.0543437046285641)
--(axis cs:0.075,0.0494202236549991)
--(axis cs:0.1,0.045295479162474)
--(axis cs:0.125,0.0420499547237716)
--(axis cs:0.15,0.039720149912624)
--(axis cs:0.175,0.0381547429913314)
--(axis cs:0.2,0.037084394204043)
--(axis cs:0.225,0.0362920218168538)
--(axis cs:0.25,0.0356658785337224)
--(axis cs:0.275,0.0351532444020441)
--(axis cs:0.3,0.0347222755550585)
--(axis cs:0.325,0.0343472545517294)
--(axis cs:0.35,0.0340052213709825)
--(axis cs:0.375,0.0336770616779589)
--(axis cs:0.4,0.0333491756898017)
--(axis cs:0.425,0.0330137948727101)
--(axis cs:0.45,0.0326680303383618)
--(axis cs:0.475,0.0323124022944157)
--(axis cs:0.5,0.031949358908008)
--(axis cs:0.525,0.0315820409068101)
--(axis cs:0.55,0.0312134533516111)
--(axis cs:0.575,0.030846110545758)
--(axis cs:0.6,0.0304820815249418)
--(axis cs:0.625,0.0301232881401276)
--(axis cs:0.65,0.0297719186404465)
--(axis cs:0.675,0.0294308096347934)
--(axis cs:0.7,0.0291035867825357)
--(axis cs:0.725,0.02879435885149)
--(axis cs:0.75,0.0285069874651441)
--(axis cs:0.775,0.0282443272966252)
--(axis cs:0.8,0.0280079361382818)
--(axis cs:0.825,0.0277983916420086)
--(axis cs:0.85,0.0276158629171318)
--(axis cs:0.875,0.0274604207516233)
--(axis cs:0.9,0.0273318674629838)
--(axis cs:0.925,0.0272293143676787)
--(axis cs:0.95,0.0271508990753986)
--(axis cs:0.975,0.0270938237329066)
--(axis cs:1,0.027054592375085)
--(axis cs:1.025,0.0270292262645369)
--(axis cs:1.05,0.0270133686888484)
--(axis cs:1.075,0.0270024207607589)
--(axis cs:1.1,0.0269919992921703)
--(axis cs:1.125,0.0269788363034382)
--(axis cs:1.15,0.0269618113874358)
--(axis cs:1.175,0.0269426526250081)
--(axis cs:1.2,0.0269260902680635)
--(axis cs:1.225,0.026919500855466)
--(axis cs:1.25,0.0269318710926538)
--(axis cs:1.275,0.0268401224336639)
--(axis cs:1.3,0.0267423261856819)
--(axis cs:1.325,0.0266854562659452)
--(axis cs:1.35,0.0266813656538799)
--(axis cs:1.375,0.0267467734044229)
--(axis cs:1.4,0.0269003241036542)
--(axis cs:1.425,0.0271481242576925)
--(axis cs:1.45,0.0274762750250364)
--(axis cs:1.475,0.0278634988977416)
--(axis cs:1.5,0.0282866280013135)
--(axis cs:1.525,0.0287217933065687)
--(axis cs:1.55,0.029195127848465)
--(axis cs:1.575,0.0296957666600791)
--(axis cs:1.6,0.030214764397009)
--(axis cs:1.625,0.0307430232658537)
--(axis cs:1.65,0.0312696494175026)
--(axis cs:1.675,0.031781895104187)
--(axis cs:1.7,0.0322664512836819)
--(axis cs:1.725,0.0326377731397103)
--(axis cs:1.75,0.0329141433634551)
--(axis cs:1.775,0.0331272913072606)
--(axis cs:1.8,0.0332788248128199)
--(axis cs:1.825,0.0333770603241672)
--(axis cs:1.85,0.0334386024158844)
--(axis cs:1.875,0.0334844185229311)
--(axis cs:1.9,0.0335257285127411)
--(axis cs:1.925,0.0335611785367822)
--(axis cs:1.95,0.0335918157867094)
--(axis cs:1.975,0.0336326980220016)
--(axis cs:2,0.0337194556795937)
--(axis cs:2.025,0.0338831861074466)
--(axis cs:2.05,0.0341196366241943)
--(axis cs:2.075,0.0344384657156215)
--(axis cs:2.1,0.0348459025736853)
--(axis cs:2.125,0.0352981296181312)
--(axis cs:2.15,0.0357570813602613)
--(axis cs:2.175,0.0362057072549823)
--(axis cs:2.2,0.036637056174422)
--(axis cs:2.225,0.037049030471728)
--(axis cs:2.25,0.0374421772710474)
--(axis cs:2.275,0.0378182313199789)
--(axis cs:2.3,0.0381789042571209)
--(axis cs:2.325,0.0385250470089433)
--(axis cs:2.35,0.0386834271879283)
--(axis cs:2.375,0.0388253835587011)
--(axis cs:2.4,0.0389879956817255)
--(axis cs:2.425,0.0391868845054026)
--(axis cs:2.45,0.0394565555764194)
--(axis cs:2.475,0.0398364273198624)
--(axis cs:2.5,0.0402904739796368)
--(axis cs:2.525,0.040748607770478)
--(axis cs:2.55,0.0411717912874264)
--(axis cs:2.575,0.0415424752108565)
--(axis cs:2.6,0.0418531206179229)
--(axis cs:2.625,0.0421013540844692)
--(axis cs:2.65,0.0422881197980866)
--(axis cs:2.675,0.0424185051301978)
--(axis cs:2.7,0.042504695460237)
--(axis cs:2.725,0.0425667349841733)
--(axis cs:2.75,0.0426231499678173)
--(axis cs:2.775,0.0426785291185035)
--(axis cs:2.8,0.042505735484062)
--(axis cs:2.825,0.042272179207582)
--(axis cs:2.85,0.0420791998290818)
--(axis cs:2.875,0.0419199535472447)
--(axis cs:2.9,0.0417919849074934)
--(axis cs:2.925,0.0417146106568455)
--(axis cs:2.95,0.0416899347198955)
--(axis cs:2.975,0.0417091481037663)
--(axis cs:3,0.04177657838005)
--(axis cs:3.025,0.0419063206027756)
--(axis cs:3.05,0.0421111027657389)
--(axis cs:3.075,0.0423935029608203)
--(axis cs:3.1,0.0427438748178482)
--(axis cs:3.125,0.0431455880371212)
--(axis cs:3.15,0.0435823007243483)
--(axis cs:3.175,0.0440425581758958)
--(axis cs:3.2,0.0445210056733663)
--(axis cs:3.225,0.0450169975354685)
--(axis cs:3.25,0.0455314372503434)
--(axis cs:3.275,0.0460633751749396)
--(axis cs:3.3,0.0466085115197253)
--(axis cs:3.325,0.0471604914547342)
--(axis cs:3.35,0.0477134937775459)
--(axis cs:3.375,0.0482638179804789)
--(axis cs:3.4,0.0488095710817945)
--(axis cs:3.425,0.0493494415267597)
--(axis cs:3.45,0.0498816506057547)
--(axis cs:3.475,0.050402211252363)
--(axis cs:3.5,0.0509023961407925)
--(axis cs:3.525,0.0513690970681143)
--(axis cs:3.55,0.0517886705715443)
--(axis cs:3.575,0.05215010154539)
--(axis cs:3.6,0.0524458982721552)
--(axis cs:3.625,0.0526718534067172)
--(axis cs:3.65,0.0528266777321219)
--(axis cs:3.675,0.0529119274473438)
--(axis cs:3.7,0.0529324053140371)
--(axis cs:3.725,0.0528970812132462)
--(axis cs:3.75,0.05282010854586)
--(axis cs:3.775,0.0527201210117921)
--(axis cs:3.8,0.0526150811026111)
--(axis cs:3.825,0.0525155282801688)
--(axis cs:3.85,0.0524252257221701)
--(axis cs:3.875,0.052346950199355)
--(axis cs:3.9,0.0522832085132839)
--(axis cs:3.925,0.0522325636202661)
--(axis cs:3.95,0.0521890050127438)
--(axis cs:3.975,0.0521451316705853)
--(axis cs:4,0.0520948343290147)
--(axis cs:4.025,0.0520340478814093)
--(axis cs:4.05,0.0519605482440166)
--(axis cs:4.075,0.0518736004962048)
--(axis cs:4.1,0.0517737094694513)
--(axis cs:4.125,0.0516624719428035)
--(axis cs:4.15,0.051542444106289)
--(axis cs:4.175,0.0514168905569424)
--(axis cs:4.2,0.0512892705251016)
--(axis cs:4.225,0.0511624512617277)
--(axis cs:4.25,0.0510379646709988)
--(axis cs:4.275,0.0509158165303283)
--(axis cs:4.3,0.0507950293374514)
--(axis cs:4.325,0.050674604409235)
--(axis cs:4.35,0.0505544622260508)
--(axis cs:4.375,0.0504359836953366)
--(axis cs:4.4,0.0503218289194487)
--(axis cs:4.425,0.0502150225445476)
--(axis cs:4.45,0.0501179721537805)
--(axis cs:4.475,0.050032332601855)
--(axis cs:4.5,0.0499596967513669)
--(axis cs:4.525,0.0499020676202217)
--(axis cs:4.55,0.0498612218870623)
--(axis cs:4.575,0.0498371417736789)
--(axis cs:4.6,0.0497661838382623)
--(axis cs:4.625,0.049515164445767)
--(axis cs:4.65,0.0492763290291882)
--(axis cs:4.675,0.0490478861653239)
--(axis cs:4.7,0.0488304936463378)
--(axis cs:4.725,0.0486281524119261)
--(axis cs:4.75,0.0484484800705999)
--(axis cs:4.775,0.0483022965624031)
--(axis cs:4.8,0.0482018350234602)
--(axis cs:4.825,0.0481573472619409)
--(axis cs:4.85,0.0481756589177889)
--(axis cs:4.875,0.0482641886357122)
--(axis cs:4.9,0.0484329181489478)
--(axis cs:4.925,0.048681353946679)
--(axis cs:4.95,0.0489866770512665)
--(axis cs:4.975,0.0493183697540148)
--(axis cs:5,0.0496534683047534)
--(axis cs:5.025,0.049977786854876)
--(axis cs:5.05,0.0502830173610204)
--(axis cs:5.075,0.0505646307302488)
--(axis cs:5.1,0.050820766640911)
--(axis cs:5.125,0.0510515880515944)
--(axis cs:5.15,0.051258689660588)
--(axis cs:5.175,0.0514443342421562)
--(axis cs:5.2,0.0515121718820972)
--(axis cs:5.225,0.0513564834656583)
--(axis cs:5.25,0.0511623167677065)
--(axis cs:5.275,0.0509317494832744)
--(axis cs:5.3,0.0506680019810328)
--(axis cs:5.325,0.0503736695108896)
--(axis cs:5.35,0.0500501704033333)
--(axis cs:5.375,0.0496994756979366)
--(axis cs:5.4,0.0493263496028234)
--(axis cs:5.425,0.048939351339292)
--(axis cs:5.45,0.0485504119916895)
--(axis cs:5.475,0.0481732227347531)
--(axis cs:5.5,0.0478204767543438)
--(axis cs:5.525,0.0475005522643702)
--(axis cs:5.55,0.0472154461765275)
--(axis cs:5.575,0.0469616103892147)
--(axis cs:5.6,0.0467329650726287)
--(axis cs:5.625,0.046523792119307)
--(axis cs:5.65,0.0463302320793653)
--(axis cs:5.675,0.046150639058204)
--(axis cs:5.7,0.045985544330461)
--(axis cs:5.725,0.0458378758727606)
--(axis cs:5.75,0.0457141024086149)
--(axis cs:5.775,0.0456274443305201)
--(axis cs:5.8,0.0456041887688746)
--(axis cs:5.825,0.0456837291560311)
--(axis cs:5.85,0.0458804488480043)
--(axis cs:5.875,0.0461527868889893)
--(axis cs:5.9,0.0464517833274814)
--(axis cs:5.925,0.0467505130632099)
--(axis cs:5.95,0.0470373713432116)
--(axis cs:5.975,0.0473082278163237)
--(axis cs:6,0.0475628324314735)
--(axis cs:6.025,0.0478034005563018)
--(axis cs:6.05,0.0480340950365633)
--(axis cs:6.075,0.0482607995093909)
--(axis cs:6.1,0.0484906927013498)
--(axis cs:6.125,0.0487309854516064)
--(axis cs:6.15,0.0489864647836544)
--(axis cs:6.175,0.0492570636803045)
--(axis cs:6.2,0.0495379060478444)
--(axis cs:6.225,0.0498220069073021)
--(axis cs:6.25,0.0501030527310218)
--(axis cs:6.275,0.0503766535500667)
--(axis cs:6.3,0.0506404079184762)
--(axis cs:6.325,0.0508935374760543)
--(axis cs:6.35,0.0511365015484975)
--(axis cs:6.375,0.0513707212697309)
--(axis cs:6.4,0.0515984269834203)
--(axis cs:6.425,0.051822603547418)
--(axis cs:6.45,0.0520469878252555)
--(axis cs:6.475,0.0522760490461259)
--(axis cs:6.5,0.0525148537435272)
--(axis cs:6.525,0.05276870285356)
--(axis cs:6.55,0.0530424813514534)
--(axis cs:6.575,0.0533398358949584)
--(axis cs:6.6,0.0536625359514777)
--(axis cs:6.625,0.0540104274973825)
--(axis cs:6.65,0.0543461727125861)
--(axis cs:6.675,0.0546417501152301)
--(axis cs:6.7,0.0549543744840525)
--(axis cs:6.725,0.0552830448189851)
--(axis cs:6.75,0.0556274769866815)
--(axis cs:6.775,0.0559872402879606)
--(axis cs:6.8,0.0563609177236077)
--(axis cs:6.825,0.0567457441193329)
--(axis cs:6.85,0.0571378546253817)
--(axis cs:6.875,0.0575329123717855)
--(axis cs:6.9,0.0579267595748671)
--(axis cs:6.925,0.0583158607266681)
--(axis cs:6.95,0.058697486103279)
--(axis cs:6.975,0.0590696888483973)
--(axis cs:7,0.0594311554790583)
--(axis cs:7.025,0.0597809976189624)
--(axis cs:7.05,0.0601185305259572)
--(axis cs:7.075,0.0604430632966795)
--(axis cs:7.1,0.0607537099657633)
--(axis cs:7.125,0.0610492205377763)
--(axis cs:7.15,0.0613278256464527)
--(axis cs:7.175,0.061587086980135)
--(axis cs:7.2,0.0618237468809667)
--(axis cs:7.225,0.0620335740639917)
--(axis cs:7.25,0.0622112081938896)
--(axis cs:7.275,0.0623500145533483)
--(axis cs:7.3,0.0624419718653661)
--(axis cs:7.325,0.0624776316711829)
--(axis cs:7.35,0.0624462051674744)
--(axis cs:7.375,0.062335848767447)
--(axis cs:7.4,0.0621342245254305)
--(axis cs:7.425,0.0618293941719471)
--(axis cs:7.45,0.0614110543176872)
--(axis cs:7.475,0.0608720312630917)
--(axis cs:7.5,0.0602098390534432)
--(axis cs:7.525,0.0594279988849239)
--(axis cs:7.55,0.0585367746773155)
--(axis cs:7.575,0.0575530468426206)
--(axis cs:7.6,0.0564992322415471)
--(axis cs:7.625,0.0554014083936718)
--(axis cs:7.65,0.0542870138055268)
--(axis cs:7.675,0.0531825831724178)
--(axis cs:7.7,0.0521119095997558)
--(axis cs:7.725,0.0510948535306877)
--(axis cs:7.75,0.0501468239109787)
--(axis cs:7.775,0.0492788113338195)
--(axis cs:7.8,0.048497784158848)
--(axis cs:7.825,0.0478072581972665)
--(axis cs:7.85,0.047207892361629)
--(axis cs:7.875,0.0466980209003855)
--(axis cs:7.9,0.0462740897426742)
--(axis cs:7.925,0.0459310097242473)
--(axis cs:7.95,0.0456624666006189)
--(axis cs:7.975,0.0454612328459075)
--(axis cs:8,0.04528407596542)
--(axis cs:8.025,0.0450362432304204)
--(axis cs:8.05,0.0448346124067057)
--(axis cs:8.075,0.0446735048237092)
--(axis cs:8.1,0.0445476460572163)
--(axis cs:8.125,0.0444522255652644)
--(axis cs:8.15,0.0443829273888998)
--(axis cs:8.175,0.0443359360376238)
--(axis cs:8.2,0.0443079228536536)
--(axis cs:8.225,0.04429601853935)
--(axis cs:8.25,0.0442977770507438)
--(axis cs:8.275,0.0443111349668393)
--(axis cs:8.3,0.0443343691492323)
--(axis cs:8.325,0.0443660543831587)
--(axis cs:8.35,0.0444050219688945)
--(axis cs:8.375,0.0444338605835631)
--(axis cs:8.4,0.0444681263002497)
--(axis cs:8.425,0.0445092948803558)
--(axis cs:8.45,0.044557180830938)
--(axis cs:8.475,0.0446115403886774)
--(axis cs:8.5,0.0446719809731266)
--(axis cs:8.525,0.0447379082217607)
--(axis cs:8.55,0.0448085247810901)
--(axis cs:8.575,0.0448828798498032)
--(axis cs:8.6,0.0449599534756147)
--(axis cs:8.625,0.0450387520514675)
--(axis cs:8.65,0.0451183931955742)
--(axis cs:8.675,0.0451981659440683)
--(axis cs:8.7,0.0452775611465448)
--(axis cs:8.725,0.0453562738331462)
--(axis cs:8.75,0.0454341830656074)
--(axis cs:8.775,0.0455113159609652)
--(axis cs:8.8,0.0455878024430062)
--(axis cs:8.825,0.0456638270015257)
--(axis cs:8.85,0.0457395838373904)
--(axis cs:8.875,0.0458152418460291)
--(axis cs:8.9,0.0458909249017273)
--(axis cs:8.925,0.045966710002481)
--(axis cs:8.95,0.0460426413221324)
--(axis cs:8.975,0.0461187537607278)
--(axis cs:9,0.0461950972681127)
--(axis cs:9.025,0.0462717540880622)
--(axis cs:9.05,0.0463488444529746)
--(axis cs:9.075,0.0464265203577226)
--(axis cs:9.1,0.0465049501859421)
--(axis cs:9.125,0.0465842984334114)
--(axis cs:9.15,0.046664704792367)
--(axis cs:9.175,0.0467462660724578)
--(axis cs:9.2,0.0468290234076486)
--(axis cs:9.225,0.0469129562145266)
--(axis cs:9.25,0.0469979834663724)
--(axis cs:9.275,0.0470839719842991)
--(axis cs:9.3,0.0471707506216878)
--(axis cs:9.325,0.047258128519081)
--(axis cs:9.35,0.0473459151714216)
--(axis cs:9.375,0.0474339399891058)
--(axis cs:9.4,0.0475220693703869)
--(axis cs:9.425,0.0476102199496638)
--(axis cs:9.45,0.047698367481216)
--(axis cs:9.475,0.0477865515759885)
--(axis cs:9.5,0.0478748770749325)
--(axis cs:9.525,0.0479635131200026)
--(axis cs:9.55,0.0480526909392341)
--(axis cs:9.575,0.0481427010112675)
--(axis cs:9.6,0.0482338896732597)
--(axis cs:9.625,0.0483266544899213)
--(axis cs:9.65,0.0484214369964636)
--(axis cs:9.675,0.0485187110605565)
--(axis cs:9.7,0.0486189654518066)
--(axis cs:9.725,0.0487226805456238)
--(axis cs:9.75,0.0488303013107686)
--(axis cs:9.775,0.0489422110737201)
--(axis cs:9.8,0.0490587117461513)
--(axis cs:9.825,0.0491800151875691)
--(axis cs:9.85,0.0493062472150674)
--(axis cs:9.875,0.0494374618726356)
--(axis cs:9.9,0.0495736608326082)
--(axis cs:9.925,0.0497148123453493)
--(axis cs:9.95,0.0498608657386205)
--(axis cs:9.975,0.0500117599305755)
--(axis cs:10,0.0501674265578518)
--(axis cs:10,0.0658283292192838)
--(axis cs:10,0.0658283292192838)
--(axis cs:9.975,0.0656787479586791)
--(axis cs:9.95,0.0655317630908718)
--(axis cs:9.925,0.0653872574402035)
--(axis cs:9.9,0.0652451136880661)
--(axis cs:9.875,0.065105221758361)
--(axis cs:9.85,0.0649674836625804)
--(axis cs:9.825,0.064831815916127)
--(axis cs:9.8,0.0646981499888763)
--(axis cs:9.775,0.0645664314470858)
--(axis cs:9.75,0.0644366184850282)
--(axis cs:9.725,0.0643086804681257)
--(axis cs:9.7,0.064182596961602)
--(axis cs:9.675,0.0640583575420647)
--(axis cs:9.65,0.0639359625118332)
--(axis cs:9.625,0.0638154244671337)
--(axis cs:9.6,0.0636967705082483)
--(axis cs:9.575,0.0635800447186071)
--(axis cs:9.55,0.0634653103964927)
--(axis cs:9.525,0.0633526514517133)
--(axis cs:9.5,0.0632421724732336)
--(axis cs:9.475,0.0631339973216984)
--(axis cs:9.45,0.063028266689306)
--(axis cs:9.425,0.062925135649412)
--(axis cs:9.4,0.0628247722183966)
--(axis cs:9.375,0.0627273565615285)
--(axis cs:9.35,0.0626330770491929)
--(axis cs:9.325,0.06254211441747)
--(axis cs:9.3,0.0624546022092519)
--(axis cs:9.275,0.0623705579694812)
--(axis cs:9.25,0.06228980239303)
--(axis cs:9.225,0.0622119143873608)
--(axis cs:9.2,0.0621362752185394)
--(axis cs:9.175,0.0620622068309176)
--(axis cs:9.15,0.0619891396505523)
--(axis cs:9.125,0.061916724704255)
--(axis cs:9.1,0.0618448521160808)
--(axis cs:9.075,0.0617735986249859)
--(axis cs:9.05,0.0617031507710056)
--(axis cs:9.025,0.0616337386741603)
--(axis cs:9,0.0615655935735323)
--(axis cs:8.975,0.061498927488177)
--(axis cs:8.95,0.0614339279563521)
--(axis cs:8.925,0.0613707609769307)
--(axis cs:8.9,0.06130957730527)
--(axis cs:8.875,0.0612505192496837)
--(axis cs:8.85,0.0611937265175463)
--(axis cs:8.825,0.0611393405158945)
--(axis cs:8.8,0.0610875070218039)
--(axis cs:8.775,0.0610383774793851)
--(axis cs:8.75,0.0609921094504174)
--(axis cs:8.725,0.0609488669631148)
--(axis cs:8.7,0.0609088216333054)
--(axis cs:8.675,0.0608721554273009)
--(axis cs:8.65,0.0608390657774918)
--(axis cs:8.625,0.0608097734821896)
--(axis cs:8.6,0.0607845334939733)
--(axis cs:8.575,0.0607636484114244)
--(axis cs:8.55,0.0607474843029726)
--(axis cs:8.525,0.0607364884348894)
--(axis cs:8.5,0.0607312085349012)
--(axis cs:8.475,0.0607323133600407)
--(axis cs:8.45,0.0607406145076582)
--(axis cs:8.425,0.0607570895740973)
--(axis cs:8.4,0.0607829068965739)
--(axis cs:8.375,0.060819452183705)
--(axis cs:8.35,0.0608683573216092)
--(axis cs:8.325,0.0609315315116508)
--(axis cs:8.3,0.0610111946425025)
--(axis cs:8.275,0.0611099124382262)
--(axis cs:8.25,0.0612306325001751)
--(axis cs:8.225,0.0613767199413864)
--(axis cs:8.2,0.0615519909707867)
--(axis cs:8.175,0.06176074257531)
--(axis cs:8.15,0.0620077763827354)
--(axis cs:8.125,0.0622984148173656)
--(axis cs:8.1,0.0626385076657148)
--(axis cs:8.075,0.063034426966875)
--(axis cs:8.05,0.0634930475062809)
--(axis cs:8.025,0.0640217088931371)
--(axis cs:8,0.0646281530624662)
--(axis cs:7.975,0.0653204279969166)
--(axis cs:7.95,0.0661067446180593)
--(axis cs:7.925,0.0669952694809362)
--(axis cs:7.9,0.0679938317048665)
--(axis cs:7.875,0.0691095193675555)
--(axis cs:7.85,0.0703481396444436)
--(axis cs:7.825,0.0717135201745219)
--(axis cs:7.8,0.0732066393690722)
--(axis cs:7.775,0.0753180886722669)
--(axis cs:7.75,0.0777495634777484)
--(axis cs:7.725,0.0803220308569687)
--(axis cs:7.7,0.083007729162381)
--(axis cs:7.675,0.0857685251761798)
--(axis cs:7.65,0.0885560622520057)
--(axis cs:7.625,0.0913132859357206)
--(axis cs:7.6,0.0939775746400957)
--(axis cs:7.575,0.0964853607968683)
--(axis cs:7.55,0.0987776718839335)
--(axis cs:7.525,0.100805613489448)
--(axis cs:7.5,0.102534670766903)
--(axis cs:7.475,0.103946948098199)
--(axis cs:7.45,0.105041027624083)
--(axis cs:7.425,0.105829759609423)
--(axis cs:7.4,0.106336738848305)
--(axis cs:7.375,0.106592344441497)
--(axis cs:7.35,0.106630068348784)
--(axis cs:7.325,0.106483570378811)
--(axis cs:7.3,0.106184610943615)
--(axis cs:7.275,0.105761807114156)
--(axis cs:7.25,0.105331750059866)
--(axis cs:7.225,0.104859008826665)
--(axis cs:7.2,0.104315351336826)
--(axis cs:7.175,0.103715084854393)
--(axis cs:7.15,0.103069666683351)
--(axis cs:7.125,0.102388110418238)
--(axis cs:7.1,0.101677373536057)
--(axis cs:7.075,0.100942712162925)
--(axis cs:7.05,0.10018800249212)
--(axis cs:7.025,0.0994160371745036)
--(axis cs:7,0.0986288097608526)
--(axis cs:6.975,0.0978278009720691)
--(axis cs:6.95,0.0970142768302019)
--(axis cs:6.925,0.0961896002461162)
--(axis cs:6.9,0.0953555448024812)
--(axis cs:6.875,0.0945145827435081)
--(axis cs:6.85,0.0936700983884988)
--(axis cs:6.825,0.0928264526948203)
--(axis cs:6.8,0.0919888015203615)
--(axis cs:6.775,0.0911625851157132)
--(axis cs:6.75,0.0903527325097647)
--(axis cs:6.725,0.0895628823622667)
--(axis cs:6.7,0.0887951000155958)
--(axis cs:6.675,0.0880503272385518)
--(axis cs:6.65,0.0873292179374302)
--(axis cs:6.625,0.0866326776287171)
--(axis cs:6.6,0.0859616676644356)
--(axis cs:6.575,0.0853164080697226)
--(axis cs:6.55,0.0846955710314953)
--(axis cs:6.525,0.0840960445430017)
--(axis cs:6.5,0.0835133720418379)
--(axis cs:6.475,0.0829425170969325)
--(axis cs:6.45,0.0823785468858869)
--(axis cs:6.425,0.0818170502780557)
--(axis cs:6.4,0.0812543108461159)
--(axis cs:6.375,0.0806873346749237)
--(axis cs:6.35,0.0801138306920861)
--(axis cs:6.325,0.0795322146955132)
--(axis cs:6.3,0.0789416829880915)
--(axis cs:6.275,0.0783423800405144)
--(axis cs:6.25,0.0777356561706696)
--(axis cs:6.225,0.077124351234963)
--(axis cs:6.2,0.076512911648389)
--(axis cs:6.175,0.0759069574865542)
--(axis cs:6.15,0.0753118987318093)
--(axis cs:6.125,0.0747308486548404)
--(axis cs:6.1,0.0741631782551597)
--(axis cs:6.075,0.073604923900878)
--(axis cs:6.05,0.0730505689326736)
--(axis cs:6.025,0.0724948594131218)
--(axis cs:6,0.0719340637673491)
--(axis cs:5.975,0.0713669585761481)
--(axis cs:5.95,0.070796177337361)
--(axis cs:5.925,0.0702307250348542)
--(axis cs:5.9,0.0696901541514896)
--(axis cs:5.875,0.069207003764941)
--(axis cs:5.85,0.0688143472587204)
--(axis cs:5.825,0.0685192077743313)
--(axis cs:5.8,0.0683022864841962)
--(axis cs:5.775,0.068140987040611)
--(axis cs:5.75,0.0680197557154508)
--(axis cs:5.725,0.0679290596215745)
--(axis cs:5.7,0.0678627898919492)
--(axis cs:5.675,0.0678165948856716)
--(axis cs:5.65,0.0677870182088537)
--(axis cs:5.625,0.0677711034196646)
--(axis cs:5.6,0.0677662668974339)
--(axis cs:5.575,0.0677703386444024)
--(axis cs:5.55,0.067781702240777)
--(axis cs:5.525,0.0677994355151989)
--(axis cs:5.5,0.0678232707420927)
--(axis cs:5.475,0.0678531191896198)
--(axis cs:5.45,0.0678880211977106)
--(axis cs:5.425,0.0679248717218637)
--(axis cs:5.4,0.0679578464980596)
--(axis cs:5.375,0.0679792803133176)
--(axis cs:5.35,0.0679816303180599)
--(axis cs:5.325,0.0679591770056439)
--(axis cs:5.3,0.0679082560841319)
--(axis cs:5.275,0.0678261499166)
--(axis cs:5.25,0.0677102376899894)
--(axis cs:5.225,0.0675584302995665)
--(axis cs:5.2,0.0674129607101378)
--(axis cs:5.175,0.0673201637508073)
--(axis cs:5.15,0.0671927746220223)
--(axis cs:5.125,0.0670288410427885)
--(axis cs:5.1,0.0668250958156205)
--(axis cs:5.075,0.066579539552825)
--(axis cs:5.05,0.0662931203074777)
--(axis cs:5.025,0.0659706479047521)
--(axis cs:5,0.0656216281891186)
--(axis cs:4.975,0.0652611269333048)
--(axis cs:4.95,0.0649095844150813)
--(axis cs:4.925,0.0645891471663718)
--(axis cs:4.9,0.0643163371569708)
--(axis cs:4.875,0.0640980830060889)
--(axis cs:4.85,0.0639357564513592)
--(axis cs:4.825,0.0638295114208539)
--(axis cs:4.8,0.0637772638379249)
--(axis cs:4.775,0.0637725215695614)
--(axis cs:4.75,0.0638051258005569)
--(axis cs:4.725,0.0638641237298303)
--(axis cs:4.7,0.0639404202987044)
--(axis cs:4.675,0.0640282643666382)
--(axis cs:4.65,0.0641257771614611)
--(axis cs:4.625,0.064234587835795)
--(axis cs:4.6,0.0643585229833102)
--(axis cs:4.575,0.0645018704784951)
--(axis cs:4.55,0.064667913710928)
--(axis cs:4.525,0.0648575840855586)
--(axis cs:4.5,0.0650680835728444)
--(axis cs:4.475,0.0652927841327975)
--(axis cs:4.45,0.065523186226916)
--(axis cs:4.425,0.0657513554142121)
--(axis cs:4.4,0.0659713786986475)
--(axis cs:4.375,0.0661799137102622)
--(axis cs:4.35,0.0663763158376995)
--(axis cs:4.325,0.0665624763019796)
--(axis cs:4.3,0.0667421551271977)
--(axis cs:4.275,0.0669195678093879)
--(axis cs:4.25,0.0670975064752737)
--(axis cs:4.225,0.0672760047714119)
--(axis cs:4.2,0.067452380482062)
--(axis cs:4.175,0.0676223336174919)
--(axis cs:4.15,0.0677811544390514)
--(axis cs:4.125,0.0679245058780958)
--(axis cs:4.1,0.068048788905383)
--(axis cs:4.075,0.0681512838476304)
--(axis cs:4.05,0.0683809216176786)
--(axis cs:4.025,0.068585415088716)
--(axis cs:4,0.0687643732662716)
--(axis cs:3.975,0.068919098811682)
--(axis cs:3.95,0.0690523241855679)
--(axis cs:3.925,0.0691682924236881)
--(axis cs:3.9,0.0692724964537667)
--(axis cs:3.875,0.0693705724913442)
--(axis cs:3.85,0.0694661191187033)
--(axis cs:3.825,0.069558636412066)
--(axis cs:3.8,0.0696438729203287)
--(axis cs:3.775,0.0697168978558022)
--(axis cs:3.75,0.0697753751377867)
--(axis cs:3.725,0.0698205514648445)
--(axis cs:3.7,0.0698547136359764)
--(axis cs:3.675,0.0698757253453715)
--(axis cs:3.65,0.0698736682886763)
--(axis cs:3.625,0.0698342178266223)
--(axis cs:3.6,0.0697446380470557)
--(axis cs:3.575,0.0695967262522433)
--(axis cs:3.55,0.0693867436986487)
--(axis cs:3.525,0.0691145740005762)
--(axis cs:3.5,0.0687834159220116)
--(axis cs:3.475,0.0684001146823138)
--(axis cs:3.45,0.0679757847890645)
--(axis cs:3.425,0.0675263119086704)
--(axis cs:3.4,0.0670713279440678)
--(axis cs:3.375,0.0666283983556174)
--(axis cs:3.35,0.0662023980598409)
--(axis cs:3.325,0.0657816269744538)
--(axis cs:3.3,0.0653482478438825)
--(axis cs:3.275,0.0648903991605282)
--(axis cs:3.25,0.0644062977556662)
--(axis cs:3.225,0.0639032940337471)
--(axis cs:3.2,0.0633956127482654)
--(axis cs:3.175,0.062900954662909)
--(axis cs:3.15,0.062434553813621)
--(axis cs:3.125,0.062002386204712)
--(axis cs:3.1,0.0615998229036307)
--(axis cs:3.075,0.0612167017704492)
--(axis cs:3.05,0.0608428044086122)
--(axis cs:3.025,0.0604708682473875)
--(axis cs:3,0.0600976453020776)
--(axis cs:2.975,0.0597236327033349)
--(axis cs:2.95,0.059351910496291)
--(axis cs:2.925,0.0589863955320419)
--(axis cs:2.9,0.0586297842164848)
--(axis cs:2.875,0.0582818564049936)
--(axis cs:2.85,0.0579392048187253)
--(axis cs:2.825,0.0575967170629883)
--(axis cs:2.8,0.0572496498579495)
--(axis cs:2.775,0.0568948843720977)
--(axis cs:2.75,0.0565312301691156)
--(axis cs:2.725,0.0561598658787861)
--(axis cs:2.7,0.0557858061898404)
--(axis cs:2.675,0.0554185850629378)
--(axis cs:2.65,0.0550670445909214)
--(axis cs:2.625,0.0547304269374879)
--(axis cs:2.6,0.0543942933974729)
--(axis cs:2.575,0.0540349353340701)
--(axis cs:2.55,0.0536321224517515)
--(axis cs:2.525,0.0531748823161231)
--(axis cs:2.5,0.0526591600795928)
--(axis cs:2.475,0.0520852927402709)
--(axis cs:2.45,0.0514576894306875)
--(axis cs:2.425,0.0507863607054224)
--(axis cs:2.4,0.0500921727791469)
--(axis cs:2.375,0.0494262283489889)
--(axis cs:2.35,0.0488759309647058)
--(axis cs:2.325,0.0484260184540194)
--(axis cs:2.3,0.0481027951246353)
--(axis cs:2.275,0.0477548511498777)
--(axis cs:2.25,0.0473790682130989)
--(axis cs:2.225,0.0469733347817586)
--(axis cs:2.2,0.0465361773758313)
--(axis cs:2.175,0.0460669783442717)
--(axis cs:2.15,0.0455665531852671)
--(axis cs:2.125,0.0450380279912871)
--(axis cs:2.1,0.0444883361609797)
--(axis cs:2.075,0.0439317978123472)
--(axis cs:2.05,0.0433995877009973)
--(axis cs:2.025,0.0429527425305429)
--(axis cs:2,0.0426582705126706)
--(axis cs:1.975,0.0425055607335123)
--(axis cs:1.95,0.0424038593972358)
--(axis cs:1.925,0.0422946393915405)
--(axis cs:1.9,0.0421533766073829)
--(axis cs:1.875,0.0419717513532345)
--(axis cs:1.85,0.041758619004729)
--(axis cs:1.825,0.0415440104357801)
--(axis cs:1.8,0.0413300980738848)
--(axis cs:1.775,0.041074858618775)
--(axis cs:1.75,0.0408087571728748)
--(axis cs:1.725,0.0404668322043298)
--(axis cs:1.7,0.0400518279208152)
--(axis cs:1.675,0.0395735309293557)
--(axis cs:1.65,0.0390454820079597)
--(axis cs:1.625,0.038483677395811)
--(axis cs:1.6,0.0379044908780393)
--(axis cs:1.575,0.0373227469591358)
--(axis cs:1.55,0.0367510721683455)
--(axis cs:1.525,0.0361994197213436)
--(axis cs:1.5,0.0356742226152894)
--(axis cs:1.475,0.0351793813945526)
--(axis cs:1.45,0.0347189181873722)
--(axis cs:1.425,0.0342990539159017)
--(axis cs:1.4,0.0339296013331946)
--(axis cs:1.375,0.0336242422388622)
--(axis cs:1.35,0.0334611765276027)
--(axis cs:1.325,0.0333963739684606)
--(axis cs:1.3,0.0333827599170257)
--(axis cs:1.275,0.033406845757565)
--(axis cs:1.25,0.0334601121942764)
--(axis cs:1.225,0.0335347945121109)
--(axis cs:1.2,0.0336208164003538)
--(axis cs:1.175,0.0337065806829504)
--(axis cs:1.15,0.0337817551171109)
--(axis cs:1.125,0.0338432733927562)
--(axis cs:1.1,0.0339399004986849)
--(axis cs:1.075,0.0340253137521198)
--(axis cs:1.05,0.0341025582835284)
--(axis cs:1.025,0.0341765827603884)
--(axis cs:1,0.0342536972477342)
--(axis cs:0.975,0.0343408863109946)
--(axis cs:0.95,0.0344450605479756)
--(axis cs:0.925,0.0345725320749307)
--(axis cs:0.9,0.0347288420090227)
--(axis cs:0.875,0.0349185858694509)
--(axis cs:0.85,0.035144954872677)
--(axis cs:0.825,0.0354094336991937)
--(axis cs:0.8,0.0357121907426558)
--(axis cs:0.775,0.0360528400401257)
--(axis cs:0.75,0.0364308512725991)
--(axis cs:0.725,0.0368453881612461)
--(axis cs:0.7,0.0372948858328639)
--(axis cs:0.675,0.0377767560431845)
--(axis cs:0.65,0.0383975806726226)
--(axis cs:0.625,0.0390446641927025)
--(axis cs:0.6,0.0397048363967582)
--(axis cs:0.575,0.040375297467386)
--(axis cs:0.55,0.0410531860430056)
--(axis cs:0.525,0.0417356442605143)
--(axis cs:0.5,0.0424201023271955)
--(axis cs:0.475,0.0431048109053588)
--(axis cs:0.45,0.0437895861265555)
--(axis cs:0.425,0.0444766014721715)
--(axis cs:0.4,0.0451709569057717)
--(axis cs:0.375,0.0458809785993132)
--(axis cs:0.35,0.0466187087040794)
--(axis cs:0.325,0.0474009562240264)
--(axis cs:0.3,0.0482504472364465)
--(axis cs:0.275,0.049196641159732)
--(axis cs:0.25,0.0502770630937949)
--(axis cs:0.225,0.0515395335380641)
--(axis cs:0.2,0.0530444366020668)
--(axis cs:0.175,0.0548685839683773)
--(axis cs:0.15,0.0571124298524476)
--(axis cs:0.125,0.0599026572678667)
--(axis cs:0.1,0.0633705969262048)
--(axis cs:0.075,0.0676255529930947)
--(axis cs:0.05,0.0727524079582884)
--(axis cs:0.025,0.0796453014798281)
--cycle;
\addlegendimage{area legend, fill=steelblue31119180, fill opacity=0.2}
\addlegendentry{Min/Max Error Range}

\addplot [semithick, steelblue31119180]
table {%
0.025 0.0687026360152026
0.05 0.0625134094591808
0.075 0.0571776347391141
0.1 0.0527334806438078
0.125 0.0492102974754918
0.15 0.0465772478318199
0.175 0.0446853269139217
0.2 0.0433152936702975
0.225 0.0422809077881552
0.25 0.0414634674409086
0.275 0.0407918078237105
0.3 0.0402196483808201
0.325 0.0397133867159433
0.35 0.039247143904351
0.375 0.0388010907872609
0.4 0.0383609058046487
0.425 0.0379172999010879
0.45 0.0374652539726531
0.475 0.0370030557208826
0.5 0.0365312958827422
0.525 0.0360519745614625
0.55 0.0355678278611206
0.575 0.0350819060776984
0.6 0.0345973835077352
0.625 0.0341175593479193
0.65 0.0336459775745601
0.675 0.0331865447301073
0.7 0.0327435141214308
0.725 0.0323212985612863
0.75 0.0319242228631402
0.775 0.0315563450765826
0.8 0.0312213491681798
0.825 0.030922422382439
0.85 0.0306620276431584
0.875 0.0304415511113284
0.9 0.030260922111282
0.925 0.0301183381155299
0.95 0.0300101511727058
0.975 0.0299309356127124
1 0.0298737979434012
1.025 0.0298309682108694
1.05 0.0297945986620302
1.075 0.0297576152696324
1.1 0.0297144876721464
1.125 0.0296618675108023
1.15 0.0295991143421059
1.175 0.0295286569943975
1.2 0.0294559550695591
1.225 0.0293888327236173
1.25 0.0293362427599297
1.275 0.0293069522106994
1.3 0.0293089803360355
1.325 0.0293502012951818
1.35 0.0294393659550795
1.375 0.0295859475555977
1.4 0.0297975643279542
1.425 0.0300760752825127
1.45 0.030415835234578
1.475 0.0308057115379631
1.5 0.0312334648839411
1.525 0.0316895408209854
1.55 0.0321683303987962
1.575 0.0326667554976195
1.6 0.033181361807887
1.625 0.0337056510861077
1.65 0.0342282899255242
1.675 0.0347333291827492
1.7 0.0352035777830731
1.725 0.0356244336560998
1.75 0.0359856055580948
1.775 0.036281682754501
1.8 0.0365131282980799
1.825 0.0366887628080608
1.85 0.0368287663981559
1.875 0.0369581521309295
1.9 0.037084737441845
1.925 0.0371979438663535
1.95 0.0372939994626493
1.975 0.0373864573796778
2 0.0375108224121353
2.025 0.0377134184474548
2.05 0.0380188221771955
2.075 0.0384244461407635
2.1 0.038890528491591
2.125 0.0393682461561613
2.15 0.0398317232980009
2.175 0.0402705246632842
2.2 0.0406813251569307
2.225 0.04106465867445
2.25 0.0414233915918891
2.275 0.0417616616718407
2.3 0.0420840584959721
2.325 0.0423952070044064
2.35 0.0427002893758414
2.375 0.0430071068925299
2.4 0.0433305604518071
2.425 0.0437010351795474
2.45 0.0441522560601124
2.475 0.0446757734252896
2.5 0.0452174028174076
2.525 0.0457298960666361
2.55 0.046191147653629
2.575 0.0465912658103042
2.6 0.0469256999370824
2.625 0.0471932272945508
2.65 0.0473960283427442
2.675 0.047540472354075
2.7 0.0476384610950667
2.725 0.0477071594068535
2.75 0.0477649558042128
2.775 0.0478252677705083
2.8 0.0478930445637711
2.825 0.0479697097929333
2.85 0.0480570781666234
2.875 0.0481592711958522
2.9 0.0482838641504751
2.925 0.0484411838363026
2.95 0.0486422340293649
2.975 0.0488958159973994
3 0.0492056153364339
3.025 0.0495681470891331
3.05 0.0499733744303535
3.075 0.0504080873973376
3.1 0.0508603748921727
3.125 0.0513230263037581
3.15 0.0517949621764635
3.175 0.052280639315706
3.2 0.0527873946150598
3.225 0.0533209630879886
3.25 0.053880933741405
3.275 0.0544592465485803
3.3 0.0550436106409523
3.325 0.0556231517966824
3.35 0.0561918052428836
3.375 0.0567481855585396
3.4 0.057293363183238
3.425 0.057828561319346
3.45 0.0583533077876166
3.475 0.0588626401345757
3.5 0.0593447258296722
3.525 0.0597843971911861
3.55 0.0601686000695104
3.575 0.0604886106739592
3.6 0.0607400709992589
3.625 0.0609226459575966
3.65 0.0610398705832965
3.675 0.0610991091874752
3.7 0.0611113079223902
3.725 0.0610900845658881
3.75 0.0610494904475178
3.775 0.0609998387631865
3.8 0.0609445419630176
3.825 0.0608833052343279
3.85 0.0608178549313531
3.875 0.0607533284130147
3.9 0.0606952472958563
3.925 0.0606449166811696
3.95 0.0605975525509938
3.975 0.0605449073716046
4 0.0604791029459089
4.025 0.0603948805610968
4.05 0.0602896409946051
4.075 0.06016278587534
4.1 0.0600152475865754
4.125 0.0598492441601571
4.15 0.0596681364670488
4.175 0.0594762485794371
4.2 0.0592784781040617
4.225 0.0590795166859298
4.25 0.0588826779762484
4.275 0.0586887822013502
4.3 0.058495887147313
4.325 0.0583002484402551
4.35 0.0580979978039368
4.375 0.0578866503473402
4.4 0.0576659542559647
4.425 0.0574381505605422
4.45 0.0572079063475024
4.475 0.056982025000359
4.5 0.0567687289675256
4.525 0.0565761750990533
4.55 0.0564103041635303
4.575 0.0562729713687587
4.6 0.0561615125080824
4.625 0.0560700854196332
4.65 0.0559921977263504
4.675 0.0559233142831957
4.7 0.0558625931324936
4.725 0.0558135315000694
4.75 0.0557837122601831
4.775 0.055783513593579
4.8 0.0558233676583043
4.825 0.0559103870297773
4.85 0.0560470953137389
4.875 0.0562336694509678
4.9 0.0564704698537848
4.925 0.0567557728682849
4.95 0.0570801062869197
4.975 0.0574254852506529
5 0.0577705125845137
5.025 0.0580967285919199
5.05 0.0583913893093329
5.075 0.0586469122053823
5.1 0.0588596282542172
5.125 0.0590286780109072
5.15 0.0591548470299531
5.175 0.0592391925782174
5.2 0.0592819344092175
5.225 0.0592824899205841
5.25 0.0592407698044369
5.275 0.0591586102543874
5.3 0.0590400582274077
5.325 0.0588902429303976
5.35 0.0587140186580932
5.375 0.0585160147898509
5.4 0.0583020938670972
5.425 0.0580804599376741
5.45 0.0578610396552922
5.475 0.0576532536159418
5.5 0.0574634622757893
5.525 0.0572937024575642
5.55 0.057142435634759
5.575 0.0570065769312265
5.6 0.0568834108374162
5.625 0.0567715779100762
5.65 0.0566712624670154
5.675 0.0565840823637064
5.7 0.0565130662110169
5.725 0.0564629381440389
5.75 0.056440823251701
5.775 0.056457188464948
5.8 0.0565260007921827
5.825 0.0566622392865325
5.85 0.0568745029096213
5.875 0.0571568686583139
5.9 0.0574900881087435
5.925 0.0578508655957439
5.95 0.0582202817908492
5.975 0.0585867132409402
6 0.058944907167894
6.025 0.0592941018244423
6.05 0.0596366952452591
6.075 0.059977431461049
6.1 0.0603225419254807
6.125 0.0606781704078878
6.15 0.0610479326490839
6.175 0.0614309116541427
6.2 0.061821987318481
6.225 0.062214308712187
6.25 0.0626017292281087
6.275 0.0629799062566813
6.3 0.0633463536449574
6.325 0.0637001016244734
6.35 0.064041333163698
6.375 0.064371119412649
6.4 0.0646912716156418
6.425 0.0650042907128607
6.45 0.065313378556279
6.475 0.0656224557409874
6.5 0.0659361050485019
6.525 0.0662593355899116
6.55 0.0665970766954838
6.575 0.0669534238098425
6.6 0.0673308872570047
6.625 0.0677300783338033
6.65 0.0681501228085941
6.675 0.0685896165625221
6.7 0.0690475655048337
6.725 0.069523800235046
6.75 0.0700186812187096
6.775 0.0705322732834049
6.8 0.0710634596269011
6.825 0.0716095271716792
6.85 0.0721664444358431
6.875 0.072729592509952
6.9 0.0732945212962219
6.925 0.0738574519353023
6.95 0.0744154718610583
6.975 0.0749664935573604
7 0.0755090728578181
7.025 0.0760421630467904
7.05 0.0765648528943937
7.075 0.0770761128935706
7.1 0.0775745569127395
7.125 0.0780582157391162
7.15 0.0785243134171921
7.175 0.078969035929817
7.2 0.0793872842424883
7.225 0.079772410030635
7.25 0.0801159422462276
7.275 0.0804073254550676
7.3 0.0806337070267578
7.325 0.080779831379095
7.35 0.0808281249652608
7.375 0.0807590795166527
7.4 0.0805520515305422
7.425 0.0801865766194392
7.45 0.0796442294279961
7.475 0.0789109315631264
7.5 0.0779794326806973
7.525 0.0768515126934503
7.55 0.0755393598988074
7.575 0.0740656545989983
7.6 0.0724621562715311
7.625 0.0707669801551269
7.65 0.0690211019847496
7.675 0.0672647980939338
7.7 0.0655346569841592
7.725 0.0638615505848345
7.75 0.0622696559395903
7.775 0.0607763831022287
7.8 0.0593929432402257
7.825 0.0581252737566516
7.85 0.0569750860560389
7.875 0.0559408758773047
7.9 0.0550188074486664
7.925 0.054203436974173
7.95 0.0534882757911409
7.975 0.0528662129763502
8 0.0523298257087898
8.025 0.0518716063684487
8.05 0.0514841307669298
8.075 0.051160184913222
8.1 0.050892861194776
8.125 0.0506756306130104
8.15 0.0505023957247673
8.175 0.0503675278881828
8.2 0.0502658911205094
8.225 0.0501928536108203
8.25 0.0501442876297813
8.275 0.0501165587848684
8.3 0.0501065053050815
8.325 0.0501114079648458
8.35 0.050128952224023
8.375 0.0501571854644835
8.4 0.0501944728211986
8.425 0.0502394547888036
8.45 0.050291008610228
8.475 0.0503482135889097
8.5 0.0504103186719774
8.525 0.0504767102708175
8.55 0.0505468801522656
8.575 0.0506203965116532
8.6 0.0506968832923287
8.625 0.0507760106180001
8.65 0.0508574936103935
8.675 0.0509410930611652
8.7 0.0510266134392564
8.725 0.0511138989864748
8.75 0.051202830962956
8.775 0.0512933258084911
8.8 0.0513853290157988
8.825 0.0514788002214583
8.85 0.0515736942706004
8.875 0.051669951083661
8.9 0.0517675010764289
8.925 0.0518662794900862
8.95 0.0519662394018436
8.975 0.0520673588322239
9 0.0521696419034626
9.025 0.0522731131487995
9.05 0.0523777990320771
9.075 0.0524836953299982
9.1 0.0525907426512504
9.125 0.0526988349904
9.15 0.0528078501544935
9.175 0.0529176786989667
9.2 0.0530282442482353
9.225 0.0531395162561733
9.25 0.0532515191647479
9.275 0.0533643373217714
9.3 0.0534780947752282
9.325 0.0535928916955463
9.35 0.0537087691878102
9.375 0.0538257460333953
9.4 0.0539438457522571
9.425 0.0540630993646678
9.45 0.0541835531199783
9.475 0.05430527353902
9.5 0.0544283296829246
9.525 0.0545527625092182
9.55 0.0546785775822404
9.575 0.0548057681838013
9.6 0.0549343277401761
9.625 0.055064235818834
9.65 0.0551954501333107
9.675 0.0553279126052596
9.7 0.0554615542566988
9.725 0.0555962998448919
9.75 0.0557320747237038
9.775 0.0558688117851739
9.8 0.0560064572126616
9.825 0.0561449686596627
9.85 0.0562842957562913
9.875 0.0564243533306401
9.9 0.0565650257539935
9.925 0.0567062174277309
9.95 0.056847909055863
9.975 0.0569901769388386
10 0.0571331764305584
};
\addlegendentry{Mean Error}
\end{axis}

\end{tikzpicture}

%% file: figs/ch_err_band_width.tex
\begin{tikzpicture}

\definecolor{darkgray176}{RGB}{176,176,176}
\definecolor{lightgray204}{RGB}{204,204,204}
\definecolor{steelblue31119180}{RGB}{31,119,180}

\begin{axis}[
legend cell align={left},
legend style={
  fill opacity=0.8,
  draw opacity=1,
  text opacity=1,
  at={(0.03,0.97)},
  anchor=north west,
  draw=lightgray204
},
log basis y={10},
tick align=outside,
tick pos=left,
title={Error Range of Networks Trained from 100 Random Initializations},
x grid style={darkgray176},
xlabel={Time (s)},
xmin=0.025, xmax=10,
xtick style={color=black},
y grid style={darkgray176},
ylabel={Width of Error Band},
ymin=0.00593249175890049, ymax=0.0487007965311798,
ymode=log,
ytick style={color=black},
ytick={0.0001,0.001,0.01,0.1,1},
yticklabels={
  \(\displaystyle {10^{-4}}\),
  \(\displaystyle {10^{-3}}\),
  \(\displaystyle {10^{-2}}\),
  \(\displaystyle {10^{-1}}\),
  \(\displaystyle {10^{0}}\)
}
]
\addplot [semithick, steelblue31119180]
table {%
0.025 0.0196255282912785
0.05 0.0184087033297243
0.075 0.0182053293380956
0.1 0.0180751177637308
0.125 0.017852702544095
0.15 0.0173922799398236
0.175 0.0167138409770459
0.2 0.0159600423980238
0.225 0.0152475117212104
0.25 0.0146111845600725
0.275 0.0140433967576879
0.3 0.013528171681388
0.325 0.013053701672297
0.35 0.0126134873330968
0.375 0.0122039169213542
0.4 0.01182178121597
0.425 0.0114628065994614
0.45 0.0111215557881937
0.475 0.0107924086109431
0.5 0.0104707434191875
0.525 0.0101536033537042
0.55 0.00983973269139444
0.575 0.00952918692162793
0.6 0.0092227548718164
0.625 0.00892137605257495
0.65 0.0086256620321761
0.675 0.0083459464083911
0.7 0.00819129905032819
0.725 0.00805102930975612
0.75 0.00792386380745498
0.775 0.00780851274350053
0.8 0.00770425460437403
0.825 0.00761104205718507
0.85 0.00752909195554514
0.875 0.00745816511782754
0.9 0.0073969745460389
0.925 0.007343217707252
0.95 0.00729416147257703
0.975 0.00724706257808794
1 0.00719910487264919
1.025 0.00714735649585148
1.05 0.00708918959468
1.075 0.00702289299136087
1.1 0.00694790120651452
1.125 0.00686443708931806
1.15 0.00681994372967516
1.175 0.00676392805794229
1.2 0.00669472613229035
1.225 0.00661529365664493
1.25 0.00652824110162257
1.275 0.00656672332390109
1.3 0.00664043373134374
1.325 0.0067109177025154
1.35 0.00677981087372275
1.375 0.0068774688344393
1.4 0.00702927722954041
1.425 0.00715092965820921
1.45 0.00724264316233575
1.475 0.00731588249681106
1.5 0.00738759461397599
1.525 0.00747762641477487
1.55 0.00755594431988047
1.575 0.00762698029905675
1.6 0.00768972648103035
1.625 0.00774065412995723
1.65 0.00777583259045711
1.675 0.00779163582516866
1.7 0.00778537663713323
1.725 0.0078290590646195
1.75 0.00789461380941967
1.775 0.00794756731151439
1.8 0.00805127326106489
1.825 0.00816695011161291
1.85 0.00832001658884456
1.875 0.0084873328303034
1.9 0.0086276480946418
1.925 0.00873346085475832
1.95 0.00881204361052645
1.975 0.00887286271151068
2 0.00893881483307697
2.025 0.00906955642309631
2.05 0.00927995107680304
2.075 0.00949333209672568
2.1 0.00964243358729439
2.125 0.00973989837315588
2.15 0.00980947182500581
2.175 0.00986127108928939
2.2 0.00989912120140932
2.225 0.00992430431003064
2.25 0.00993689094205158
2.275 0.00993661982989878
2.3 0.00992389086751441
2.325 0.0099009714450761
2.35 0.0101925037767775
2.375 0.0106008447902878
2.4 0.0111041770974214
2.425 0.0115994762000198
2.45 0.0120011338542681
2.475 0.0122488654204085
2.5 0.012368686099956
2.525 0.0124262745456451
2.55 0.012460331164325
2.575 0.0124924601232136
2.6 0.01254117277955
2.625 0.0126290728530187
2.65 0.0127789247928348
2.675 0.01300007993274
2.7 0.0132811107296034
2.725 0.0135931308946128
2.75 0.0139080802012983
2.775 0.0142163552535942
2.8 0.0147439143738875
2.825 0.0153245378554062
2.85 0.0158600049896435
2.875 0.0163619028577489
2.9 0.0168377993089914
2.925 0.0172717848751963
2.95 0.0176619757763956
2.975 0.0180144845995686
3 0.0183210669220276
3.025 0.018564547644612
3.05 0.0187317016428733
3.075 0.0188231988096289
3.1 0.0188559480857825
3.125 0.0188567981675908
3.15 0.0188522530892727
3.175 0.0188583964870133
3.2 0.0188746070748991
3.225 0.0188862964982786
3.25 0.0188748605053228
3.275 0.0188270239855886
3.3 0.0187397363241572
3.325 0.0186211355197197
3.35 0.0184889042822949
3.375 0.0183645803751385
3.4 0.0182617568622734
3.425 0.0181768703819107
3.45 0.0180941341833098
3.475 0.0179979034299508
3.5 0.0178810197812192
3.525 0.0177454769324619
3.55 0.0175980731271045
3.575 0.0174466247068534
3.6 0.0172987397749005
3.625 0.0171623644199051
3.65 0.0170469905565544
3.675 0.0169637978980277
3.7 0.0169223083219393
3.725 0.0169234702515982
3.75 0.0169552665919266
3.775 0.0169967768440101
3.8 0.0170287918177175
3.825 0.0170431081318972
3.85 0.0170408933965332
3.875 0.0170236222919892
3.9 0.0169892879404828
3.925 0.016935728803422
3.95 0.0168633191728242
3.975 0.0167739671410967
4 0.0166695389372569
4.025 0.0165513672073067
4.05 0.016420373373662
4.075 0.0162776833514257
4.1 0.0162750794359317
4.125 0.0162620339352923
4.15 0.0162387103327624
4.175 0.0162054430605495
4.2 0.0161631099569605
4.225 0.0161135535096843
4.25 0.0160595418042749
4.275 0.0160037512790596
4.3 0.0159471257897463
4.325 0.0158878718927445
4.35 0.0158218536116487
4.375 0.0157439300149256
4.4 0.0156495497791988
4.425 0.0155363328696644
4.45 0.0154052140731355
4.475 0.0152604515309425
4.5 0.0151083868214775
4.525 0.0149555164653368
4.55 0.0148066918238658
4.575 0.0146647287048162
4.6 0.0145923391450479
4.625 0.014719423390028
4.65 0.0148494481322729
4.675 0.0149803782013143
4.7 0.0151099266523666
4.725 0.0152359713179042
4.75 0.015356645729957
4.775 0.0154702250071583
4.8 0.0155754288144647
4.825 0.015672164158913
4.85 0.0157600975335703
4.875 0.0158338943703767
4.9 0.015883419008023
4.925 0.0159077932196928
4.95 0.0159229073638147
4.975 0.01594275717929
5 0.0159681598843652
5.025 0.015992861049876
5.05 0.0160101029464573
5.075 0.0160149088225762
5.1 0.0160043291747095
5.125 0.0159772529911941
5.15 0.0159340849614343
5.175 0.0158758295086511
5.2 0.0159007888280406
5.225 0.0162019468339082
5.25 0.0165479209222829
5.275 0.0168944004333256
5.3 0.0172402541030991
5.325 0.0175855074947543
5.35 0.0179314599147266
5.375 0.0182798046153811
5.4 0.0186314968952362
5.425 0.0189855203825716
5.45 0.019337609206021
5.475 0.0196798964548668
5.5 0.0200027939877489
5.525 0.0202988832508287
5.55 0.0205662560642495
5.575 0.0208087282551877
5.6 0.0210333018248052
5.625 0.0212473113003576
5.65 0.0214567861294884
5.675 0.0216659558274676
5.7 0.0218772455614881
5.725 0.0220911837488139
5.75 0.0223056533068359
5.775 0.022513542710091
5.8 0.0226980977153215
5.825 0.0228354786183002
5.85 0.0229338984107161
5.875 0.0230542168759516
5.9 0.0232383708240082
5.925 0.0234802119716443
5.95 0.0237588059941494
5.975 0.0240587307598244
6 0.0243712313358757
6.025 0.02469145885682
6.05 0.0250164738961102
6.075 0.0253441243914871
6.1 0.0256724855538099
6.125 0.025999863203234
6.15 0.0263254339481549
6.175 0.0266498938062497
6.2 0.0269750056005446
6.225 0.0273023443276609
6.25 0.0276326034396478
6.275 0.0279657264904477
6.3 0.0283012750696153
6.325 0.0286386772194589
6.35 0.0289773291435886
6.375 0.0293166134051928
6.4 0.0296558838626956
6.425 0.0299944467306377
6.45 0.0303315590606314
6.475 0.0306664680508066
6.5 0.0309985182983107
6.525 0.0313273416894417
6.55 0.0316530896800419
6.575 0.0319765721747642
6.6 0.0322991317129579
6.625 0.0326222501313346
6.65 0.032983045224844
6.675 0.0334085771233217
6.7 0.0338407255315432
6.725 0.0342798375432816
6.75 0.0347252555230832
6.775 0.0351753448277525
6.8 0.0356278837967538
6.825 0.0360807085754874
6.85 0.0365322437631172
6.875 0.0369816703717226
6.9 0.0374287852276141
6.925 0.037873739519448
6.95 0.0383167907269229
6.975 0.0387581121236719
7 0.0391976542817943
7.025 0.0396350395555413
7.05 0.0400694719661629
7.075 0.0404996488662459
7.1 0.0409236635702939
7.125 0.0413388898804621
7.15 0.0417418410368986
7.175 0.0421279978742584
7.2 0.0424916044558594
7.225 0.0428254347626732
7.25 0.0431205418659763
7.275 0.0434117925608077
7.3 0.0437426390782487
7.325 0.0440059387076276
7.35 0.04418386318131
7.375 0.0442564956740499
7.4 0.0442025143228746
7.425 0.0440003654374762
7.45 0.0436299733063959
7.475 0.043074916835107
7.5 0.0423248317134597
7.525 0.0413776146045245
7.55 0.040240897206618
7.575 0.0389323139542477
7.6 0.0374783423985486
7.625 0.0359118775420488
7.65 0.034269048446479
7.675 0.0325859420037621
7.7 0.0308958195626252
7.725 0.0292271773262811
7.75 0.0276027395667697
7.775 0.0260392773384475
7.8 0.0247088552102242
7.825 0.0239062619772554
7.85 0.0231402472828146
7.875 0.02241149846717
7.9 0.0217197419621923
7.925 0.0210642597566889
7.95 0.0204442780174403
7.975 0.0198591951510092
8 0.0193440770970462
8.025 0.0189854656627167
8.05 0.0186584350995753
8.075 0.0183609221431658
8.1 0.0180908616084984
8.125 0.0178461892521012
8.15 0.0176248489938356
8.175 0.0174248065376862
8.2 0.0172440681171331
8.225 0.0170807014020363
8.25 0.0169328554494313
8.275 0.0167987774713869
8.3 0.0166768254932702
8.325 0.0165654771284921
8.35 0.0164633353527146
8.375 0.0163855916001419
8.4 0.0163147805963242
8.425 0.0162477946937415
8.45 0.0161834336767202
8.475 0.0161207729713633
8.5 0.0160592275617746
8.525 0.0159985802131286
8.55 0.0159389595218825
8.575 0.0158807685616212
8.6 0.0158245800183587
8.625 0.0157710214307221
8.65 0.0157206725819176
8.675 0.0156739894832325
8.7 0.0156312604867606
8.725 0.0155925931299686
8.75 0.0155579263848101
8.775 0.0155270615184199
8.8 0.0154997045787977
8.825 0.0154755135143689
8.85 0.0154541426801559
8.875 0.0154352774036546
8.9 0.0154186524035428
8.925 0.0154040509744497
8.95 0.0153912866342197
8.975 0.0153801737274492
9 0.0153704963054196
9.025 0.0153619845860981
9.05 0.0153543063180311
9.075 0.0153470782672633
9.1 0.0153399019301387
9.125 0.0153324262708436
9.15 0.0153244348581853
9.175 0.0153159407584597
9.2 0.0153072518108908
9.225 0.0152989581728342
9.25 0.0152918189266576
9.275 0.0152865859851821
9.3 0.0152838515875641
9.325 0.0152839858983889
9.35 0.0152871618777713
9.375 0.0152934165724227
9.4 0.0153027028480097
9.425 0.0153149156997481
9.45 0.01532989920809
9.475 0.0153474457457098
9.5 0.0153672953983011
9.525 0.0153891383317106
9.55 0.0154126194572586
9.575 0.0154373437073396
9.6 0.0154628808349886
9.625 0.0154887699772124
9.65 0.0155145255153696
9.675 0.0155396464815082
9.7 0.0155636315097955
9.725 0.015585999922502
9.75 0.0156063171742596
9.775 0.0156242203733657
9.8 0.015639438242725
9.825 0.0156518007285579
9.85 0.015661236447513
9.875 0.0156677598857253
9.9 0.0156714528554579
9.925 0.0156724450948541
9.95 0.0156708973522513
9.975 0.0156669880281036
10 0.015660902661432
};
\addlegendentry{Width of Error Band}
\end{axis}

\end{tikzpicture}